\documentclass[final]{cvpr}

\usepackage{times}
\usepackage{epsfig}
\usepackage{graphicx}
\usepackage{amsmath}
\usepackage{amssymb}
\usepackage{subcaption}
\usepackage{multirow}
\usepackage{makecell}
\usepackage{gensymb}

\newcommand{\zc}[1]{\textcolor{black}{{#1}}}
\newcommand{\beforecaption}{\vspace{-1.2em}}

\newcommand\blfootnote[1]{%
  \begingroup
  \renewcommand\thefootnote{}\footnote{#1}%
  \addtocounter{footnote}{-1}%
  \endgroup
}

\usepackage[colorlinks=true,allcolors=black,urlcolor=blue,pagebackref=true,breaklinks=true,bookmarks=false]{hyperref}
\let\orgautoref\autoref
\providecommand{\Autoref}[1]{\def\figureautorefname{Figure}\orgautoref{#1}}
\renewcommand{\autoref}[1]{\def\figureautorefname{Fig.}\orgautoref{#1}}

\def\cvprPaperID{****} % *** Enter the CVPR Paper ID here
\def\confYear{CVPR 2021}
\begin{document}

%%%%%%%%% TITLE
% \title{Holistic Scene Understanding via Implicit Contextural Constraints}
% \title{Holistic Scene Understanding with Deep Structured Implicit Representation}
% \title{Exploiting Deep Structured Implicit Representation for Holistic Scene Understanding}
% \title{When Deep Structured Implicit Representation Meets Holistic Scene Understanding}
%\title{IM4SUN: Empowering Holistic Scene Understanding \\ with Local  Implicit Representation}
\title{Holistic 3D Scene Understanding from a Single Image \\ with Implicit Representation }
% \vspace{-3em}
\author{
    Cheng Zhang$^{2*}$\ \ \ Zhaopeng Cui$^{1*}$\ \ \ Yinda Zhang$^{3}$\thanks{Equal contribution}\ \ \ Bing Zeng$^{2}$\ \ \ Marc Pollefeys$^{4}$ \ \ \ Shuaicheng Liu$^{2}$\thanks{Corresponding author}\\
% 	\\
	$^{1}$ State Key Lab of CAD \& CG, Zhejiang University \\
    $^{2}$ University of Electronic Science and Technology of China \quad
    $^{3}$ Google \quad $^{4}$ETH Z\"{u}rich \\
}
% \vspace{-2.5em}

\maketitle

%%%%%%%%% ABSTRACT
\begin{abstract}
We present a new pipeline for holistic 3D scene understanding from a single image, which could predict object shapes, object poses, and scene layout. As it is a highly ill-posed problem, existing methods usually suffer from inaccurate estimation of both shapes and layout especially for the cluttered scene due to the heavy occlusion between objects. We propose to utilize the latest deep implicit representation to solve this challenge. 
We not only propose an image-based local structured implicit network to improve the object shape estimation, but also refine the 3D object pose and scene layout via a novel implicit scene graph neural network that exploits the implicit local object features.
% On one hand, we proposed an image-based local structured implicit network to improve the object shape estimation. On the other hand, we proposed to refine the estimated 3D pose and scene layout via an novel implicit scene graph neural network that well exploits the implicit local object features. 
A novel physical violation loss is also proposed to avoid incorrect context between objects. Extensive experiments demonstrate that our method outperforms the state-of-the-art methods in terms of object shape, scene layout estimation, and 3D object detection.\blfootnote{Project webpage:
\href{https://chengzhag.github.io/publication/im3d/}{https://chengzhag.github.io/publication/im3d/}}
   
\end{abstract}

\section{Introduction}
3D indoor scene understanding is a long-lasting computer vision problem and has tremendous impact on several applications, e.g., robotics, virtual reality. 
Given a single color image, the goal is to reconstruct the room layout as well as each individual object and estimate its semantic type in the 3D space. 
Over decades, there are plenty of works consistently improving the performance of such a task over two focal points of the competition.
One is the \textbf{3D shape representation} preserving fine-grained geometry details, evolving from the 3D bounding box, 3D volume, point cloud, to the recent triangulation mesh.
The other is the joint inference of multiple objects and layout in the scene leveraging \textbf{contextual information}, such as co-occurring or relative locations among objects of multiple categories.
However, the cluttered scene is a double-blade sword, which unfortunately increases the complexity of 3D scene understanding by introducing large variations in object poses and scales, and heavy occlusion.
Therefore, the overall performance is still far from satisfactory.

\begin{figure}[t]
	\centering
	\begin{subfigure}[t]{0.23\textwidth}
		\includegraphics[width=\textwidth]  
		{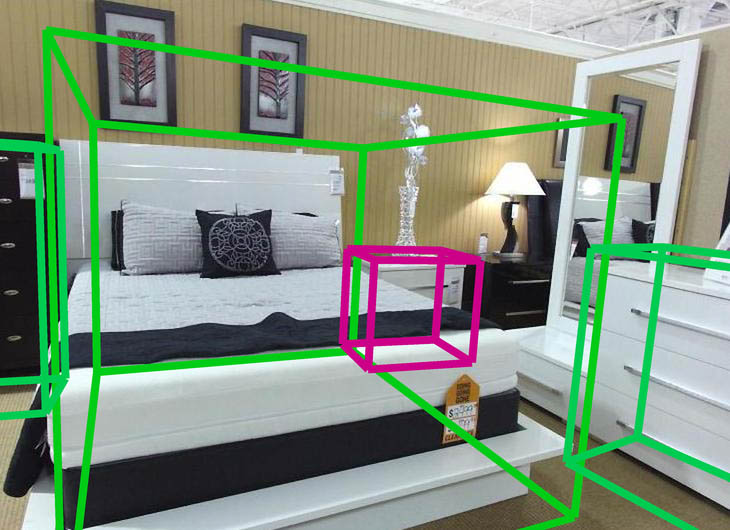}
	\end{subfigure}
	\begin{subfigure}[t]{0.23\textwidth}
		\includegraphics[width=\textwidth]  
		{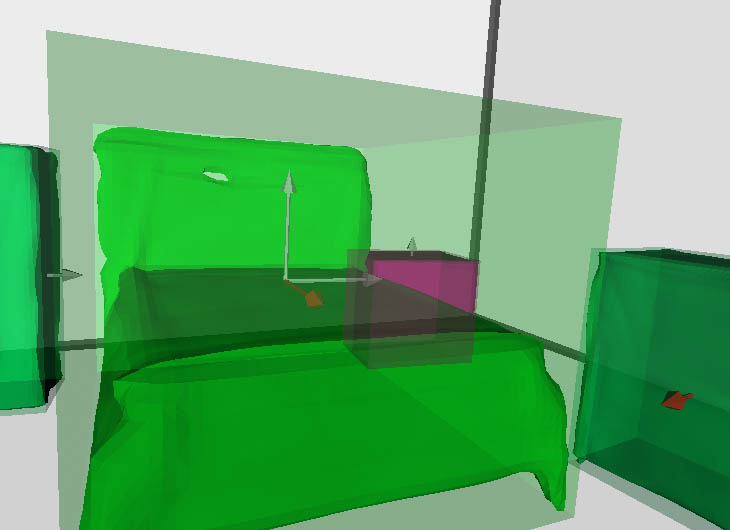}
	\end{subfigure}
	\vspace{0.5em}
	\beforecaption
	\caption{Our proposed pipeline takes a single image as input, estimates layout and object poses, then reconstructs the scene with Signed Distance Function (SDF) representation.}
	\label{fig:teaser}
\end{figure}

In this work, we propose a deep learning system for holistic 3D scene understanding, which predicts and refines object shapes, object poses, and scene layout jointly with \textbf{deep implicit representation}.
%On one hand, 
At first, similar to previous methods, we exploit standard Convolutional Neural Networks (CNN) to learn an initial estimation of 3D object poses, scene layout as well as 3D shapes. 
Different from previous methods using explicit 3D representation like volume or mesh, we utilize the local structured implicit representation of shapes motivated by~\cite{genova2020local}. Instead of taking depth images as input like \cite{genova2020local}, we design a new local implicit shape embedding network to learn the latent shape code directly from images, which can be further decoded to generate the implicit function for 3D shapes. Due to the power of implicit representation, the 3D shape of each object can be reconstructed with higher accuracy and finer surface details compared to other representations.

%Our model first detects room layout and object arrangement in 3D, and jointly estimates the fine shape of each object.
%Different from previous methods using explicit 3D representation like volume or mesh, we represent objects in implicit signed distance functions, which are decoded from network from a latent vector and sampled 3D locations.
%An immediate benefit is the capability of encoding 3D shape of each object with higher accuracy and finer surface details, as shown in our experiments and also observed in other works.

%not only for each individual object, the deep implicit representation also benefits the learning of scene context. Specifically speaking, 
Then, we propose a novel graph-based scene context network to gather information from local objects, i.e., bottom-up features extracted from the initial predictions, and learn to refine the initial 3D pose and scene layout via scene context information with the implicit representation.
Being one of the core topics studied in scene understanding, the context has been achieved in the era of deep learning mainly from two aspects - the model architecture and the loss function.
From the perspective of model design, we exploit the graph-based convolutional neural network (GCN) to learn context since it has shown competitive performance to learn context~\cite{yang2018graph}. %\yd{missing citation}.
%From the perspective of model design, context-aware architectures gathers information from local objects, e.g. bottom-up feature extracted from the input image or initial predictions, and then build pair-wise or high order connections to provide the possibility of learning context.
%Recently, graph based convolutional neural network (GCN) has shown competitive performance to learn context.
%While the design of GCN naturally fits the context learning, we argue that the input feature are equally important. 
With the deep implicit representation, the learned local shape latent vectors are naturally a compact and informative feature measuring of the object geometries, which results in more effective context models compared to features extracted from other representations such as mesh.
% From the perspective of model design, context-aware architectures have been proposed to build pair-wise or high-order connections among objects in the scene, through a pre-defined template or more recently a graph based convolutional neural network (GCN).
% In a typical GCN for indoor scene understanding, each object is represented by a node, and edges are created between every pair of nodes (i.e. a complete graph).
% The forward pass of GCN imitates the message propagation among objects, which provides possibility of learning context.
% To facilitate the learning, the input to the GCN, which are usually features extracted from the input image or 
% While the GCN is efficient in exchanging local information from each object with others, the input feature to the graph, i.e. from the bottom-up analysis, are important to guarantee the 
% Both nodes and edges are assigned features from bottom-up analysis, and then feed forward through GCN for optimal solution respecting top-down context.
% The compact latent representation for deep implicit representation is naturall

Not only the architecture, but the deep implicit representation also benefits the context learning on the loss function.
One of the most basic contextual information yet still missing in many previous works - objects should not intersect with each other, could be easily applied as supervision by penalizing the existence of 3D locations with negative predicted SDF in more than one objects\footnote{The object interior is with negative SDF, and thus no location should be inside of two objects.}.
We define this constraint as {a novel physical violation loss} and find it particularly helpful in preventing intersecting objects and producing reasonable object layouts.

Overall, our contributions are mainly in four aspects.
First, we design a two-stage single image-based holistic 3D scene understanding system that could predict object shapes, object poses, and scene layout with deep implicit representation, then optimize the later two.
%leveraging deep implicit representation, which leads to superior geometry accuracy.
%Second, we proposed a new GCN based scene context network to refine the object arrangement taking the latent implicit code as the feature, and find it especially effective.
Second, a new image-based local implicit shape embedding network is proposed to extract latent shape information which leads to superior geometry accuracy.
Third, we propose a novel GCN-based scene context network to refine the object arrangement which well exploits the latent and implicit features from the initial estimation. 
Last but not least, we design a physical violation loss, thanks to the implicit representation, to effectively prevent the object intersection.
Extensive experiments show that our model achieves the state-of-the-art performance on the standard benchmark.

\section{Related works}

\noindent\textbf{Single Image Scene Reconstruction.}
As a highly ill-posed problem, single image scene reconstruction sets a high bar for learning-based algorithms, especially in a cluttered scene with heavy occlusion. 
The problem can be divided into layout estimation, object detection and pose estimation, and 3D object reconstruction. 
A simple version of the first problem is to simplify the room layout as a bounding box \cite{hedau2009recovering, lee2009geometric, mallya2015learning, dasgupta2016delay, ren2016coarse}. 
To detect objects and estimate poses in 3D space, Recent works \cite{du2018learning, huang2018cooperative, chen2019holistic++} try to infer 3D bounding boxes from 2D detection by exploiting relationships among objects with a graph or physical simulation.
At the same time, other works \cite{izadinia2017im2cad, hueting2017seethrough, huang2018holistic} further extend the idea to align a CAD model with similar style to each detected object.
Still, the results are limited by the size of the CAD model database which results in an inaccurate representation of the scene.
To tackle the above limitations of previous works, Total3D \cite{nie2020total3dunderstanding} is proposed as an end-to-end solution to jointly estimate the layout box and object poses while reconstructing each object from the detection and utilizing the reconstruction to supervise the pose estimation learning.
However, they only exploit relationships among objects with features based on appearance and 2D geometry.

\noindent\textbf{Shape Representation.}
In the field of computer graphics, traditional shape representation methods include mesh, voxel, and point cloud. 
Some of the learning-based works try to encode the shape prior into a feature vector but stick to the traditional representations by decoding the vector into mesh \cite{groueix2018papier, wang2018pixel2mesh, pan2019deep, smith2019geometrics, gkioxari2019mesh}, voxel \cite{wu20153d, choy20163d, brock2016generative, wu2016learning, stutz2018learning} or point cloud \cite{lin2017learning, achlioptas2018learning, yang2019pointflow}. 
Others try to learn structured representations which decompose the shape into simple shapes \cite{li2017grass, gao2019sdm, paschalidou2019superquadrics}. 
Recently, implicit surface function \cite{mescheder2019occupancy, park2019deepsdf, xu2019disn, saito2019pifu, peng2020convolutional, saito2020pifuhd} has been widely used as a new representation method to overcome the disadvantages of traditional methods (i.e. unfriendly data structure to neural network of mesh and point cloud, low resolution and large memory consumption of voxel). 
Most recent works \cite{genova2019learning, genova2020local, wu2020pq} try to combine the structured and implicit representation which provides a physically meaningful feature vector while introducing significant improvement on the details of the decoded shape.

\noindent\textbf{Graph Convolutional Networks.}
Proposed by \cite{gori2005new}, graph neural networks or GCNs have been widely used to learn from graph-structured data. 
Inspired by convolutional neural networks, convolutional operation has been introduced to graph either on spectral domain \cite{bruna2013spectral, defferrard2016convolutional, kipf2016semi} or non-spectral domain \cite{hamilton2017inductive} which performs convolution with a message passing neural network to gather information from the neighboring nodes. 
Attention mechanism has also been introduced to GCN and has been proved to be efficient on tasks like node classification \cite{velivckovic2017graph}, scene graph generation \cite{yang2018graph} and feature matching \cite{sarlin2020superglue}. 
Recently, GCN has been even used on super-resolution \cite{zhou2020cross} which is usually the territory of CNN. In the 3D world which interests us most, GCN has been used on classification \cite{wang2019dynamic} and segmentation \cite{te2018rgcnn, wang2019graph, wang2019dynamic} on point cloud, which is usually an enemy representation to traditional neural networks. 
The most related application scenario of GCN with us is 3D object detection on points cloud. 
Recent work shows the ability of GCN to predict relationship \cite{avetisyan2020scenecad} or 3D object detections \cite{najibi2020dops} from point cloud data.

\begin{figure*}
	\centering
	\vspace{-1.5em}
	\includegraphics[width=\textwidth]  
		{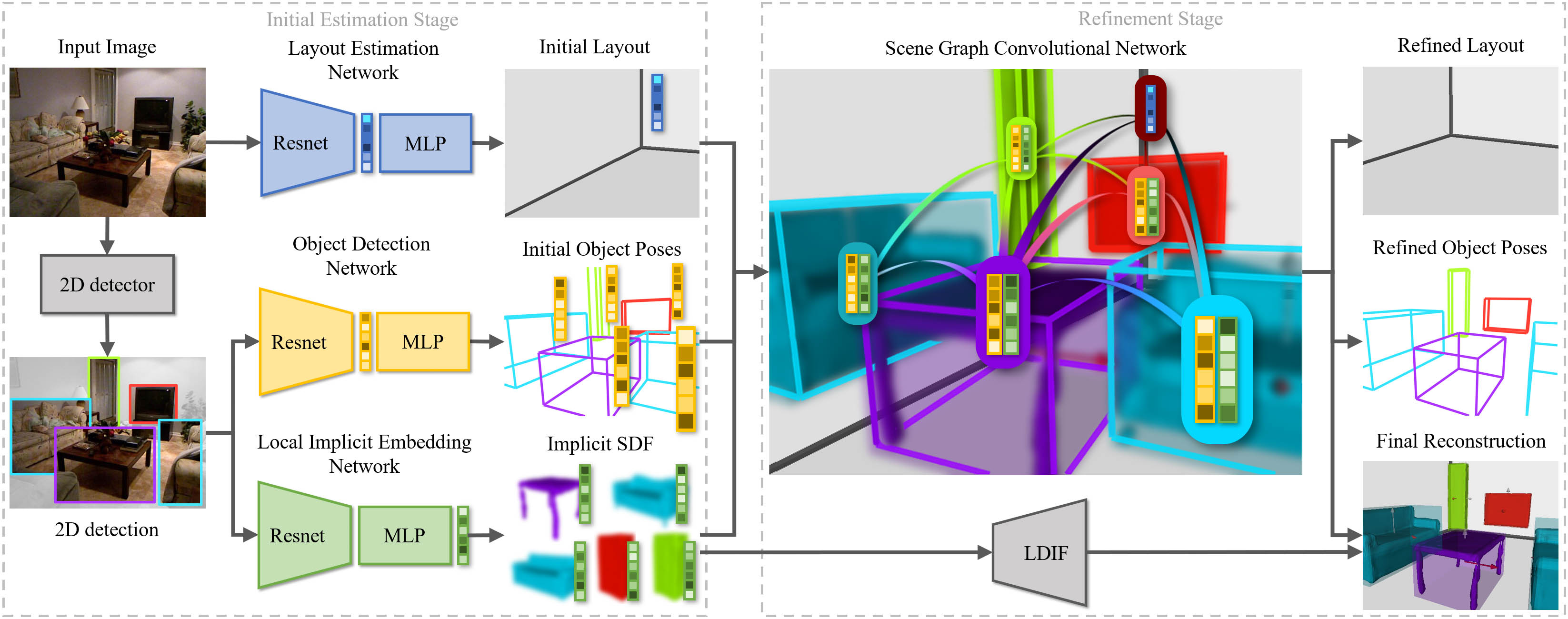}
	\vspace{-0.8em}
	\beforecaption
	\caption{Our proposed pipeline. We initialize the layout estimation and 3D object poses with LEN and ODN from prior work \cite{nie2020total3dunderstanding}, then refine them with Scene Graph Convolutional Network (SGCN). We utilize a Local Implicit Embedding Network (LIEN) to encode latent code for LDIF decoder \cite{genova2020local} and to extract implicit features for SGCN. With the help of LDIF and marching cube algorithm, object meshes are extracted then rotated, scaled, and put into places to construct the scene.}
	\vspace{-1em}
	\label{fig:pipeline}
\end{figure*}

\section{Our method}

%As shown in \autoref{fig:pipeline}, our proposed pipeline consists of five modules: a 2D detector, a Layout Estimation Network (LEN) that estimates the initial room layout (including 3D layout bounding box and camera pose) from a single image, a 3D Object Detection Network (ODN) that predicts initial object poses as 3D bounding boxes from the 2D detection, a Local Implicit Embedding Network (LIEN) that work with Local Deep Implicit Functions (LDIF) to infers 3D geometry from the object images, and a Graph Convolutional Network (GCN) that models the scene as a graph to refine layout and object poses predictions.
As shown in \autoref{fig:pipeline}, the proposed system consists of two stages, i.e., the initial estimation stage, and the refinement stage. In the initial estimation stage, similar to \cite{huang2018cooperative, nie2020total3dunderstanding}, a 2D detector is first adopted to extract the 2D bounding box from the input image, followed by an Object Detection Network (ODN) to recover the object poses as 3D bounding boxes and a new Local Implicit Embedding Network (LIEN) to extract the implicit local shape information from the image directly, which can further be decoded to infer 3D geometry.
The input image is also fed into a Layout Estimation Network (LEN) to produce a 3D layout bounding box and relative camera pose.
% the initial room layout including 3D layout bounding box and camera pose is first estimated from a single image via the  Layout Estimation Network (LEN), and initial object poses is estimated as 3D bounding boxes via the Object Detection Network (ODN). A new Local Implicit Embedding Network (LIEN) is designed to extract the implicit local shape information from the image directly, which can further decoded to infer 3D geometry.
In the refinement stage, a novel Scene Graph Convolutional Network (SGCN) is designed to refine the initial predictions via the scene context information.
As 2D detector, LEN, ODN has the standard architecture similar to prior works \cite{huang2018cooperative, nie2020total3dunderstanding}, %\textcolor{red}{[cite one more paper]}
in this section, we will describe the details of the novel SGCN and LIEN in detail. Please refer to our supplementary materials for the details of our 2D detector, LEN, ODN.

%\subsection{Image-based Local Implicit Embedding Network}
%The local deep implicit representation \cite{ldif} have shown superior performance for 3D shape reconstruction of a single object. 
%Motivated by Genova \etal \cite{genova2020local}, we propose to utilize the local deep implicit representation for the scene understanding due to its superior performance for single object reconstruction compared to other representation. However, it is nontrivial to adapt \cite{genova2020local} to 

\subsection{Scene Graph Convolutional Network}

%We adopt the basic structure of attentional GCN from Graph R-CNN \cite{yang2018graph} but without attentions. 
%Graph R-CNN constructs scene objects and their relationships as a graph with their features in each node, effectively capturing contextual information from the objects and relationships to get a more accurate relation prediction. 
%Our proposed GCN tries to do the same, but in 3D space and to get accurate layout estimation and object detection by refining the output of LEN and ODN.
As shown in \autoref{fig:pipeline}, motivated by Graph R-CNN \cite{yang2018graph}, we model the whole 3D scene as a graph $G$, in which the nodes represent the objects, the scene layout, and their relationships.
The graph is constructed starting from a complete graph with undirected edges between all objects and layout nodes, which allows information to flow among objects and the scene layout. %as skip connection \yd{what does this skip connection refer to?}
Then, we add relation nodes to each pair of neighboring object/layout nodes.
Considering the nature of directional relation \cite{krishna2017visual}, %\cite{cite here: https://visualgenome.org/}, 
we add two relation nodes between each pair of neighbors in different directions. %\yd{should there be a figure showing this, if this is not copied from somewhere else?}

% Considering that the information flow needed from relation node by its connected two nodes might be different, we use directed edges to connect relation nodes with others. 
% Also considering that the relation from source node to destination node might be different in reverse, each pair of nodes is connected with two relation nodes with different directions. 

It is well known that the input features are the key to an effective GCN \cite{wang2018pixel2mesh}.
%Finally, the features from initial estimation are extracted and embedded with a two-layer MLP into an representation $z_i \in \mathbb{R}^d$ for each node $i$. 
For different types of nodes, we design features carefully from different sources as follows.
For each node, features from different sources are flattened and concatenated into a vector, then embedded into a node representation vector with the same length using MLP. %yd{How? Flatten? Same below}
% \yd{I found the description below are very vague and probably not replicable.}

\noindent\textbf{Layout Node.} 
We use the feature from the image encoder of LEN, which encodes the appearance of layout, and the parameterized output of layout bounding box and camera pose from LEN, as layout node features.
We also concatenate the camera intrinsic parameters normalized by the image height into the feature to add camera priors. %yd{How? Flatten? Same below}

\noindent\textbf{Object Node.} 
We collect the appearance-relationship feature \cite{nie2020total3dunderstanding} %\yd{what is this feature?}
from ODN, and the parameterized output of object bounding box from ODN, %\yd{should this be ODN?} 
along with the element centers in the world coordinate and analytic code from LIEN (which we will further describe in the next section). 
We also use the one-hot category label from the 2D detector to introduce semantic information to SGCN.

\noindent\textbf{Relationship Node.} 
For nodes connecting two different objects, the geometry feature \cite{hu2018relation, vaswani2017attention} of 2D object bounding boxes and the box corner coordinates of both connected objects normalized by the image height and width are used as features. %\yd{how?} of the box corners
The coordinates are flattened and concatenated in the order of source-destination, which differentiate the relationships of different directions.
% \yd{how is this related with the direction?}
For nodes connecting objects and layouts, since the relationship is presumably different from object-object relationship, we initialize the representations with constant values, leaving the job of inferring reasonable relationship representation to SGCN.

For a graph with $N$ objects and $1$ layout, object-layout nodes and relationship nodes can then be put into two matrices $Z^o\in\mathbb{R}^{d{\times}(N+1)}$ and $Z^r\in\mathbb{R}^{d{\times}(N+1)^2}$.

% Since the graph is modeled with connections and nodes with different types, we define the message passing method in a asymmetric manner following Graph R-CNN \yd{I don't understand the causality in this sentence}. 
Since the graph is modeled with different types of nodes, which makes a difference in the information needed from different sources to destinations, we define independent message passing weights for each of the source-destination types.
We denote the linear transformation and the adjacent matrix from source node to destination node with type $a$ and $b$ as $W^{ab}$ and $\alpha^{ab}$, in which node types can be source object (or layout) $s$, destination object (or layout) $d$, and relationships $r$. 
Thus, 
% the i-th update for object and layout nodes can be defined as
the representation of object and layout nodes can be updated as
\begin{equation}
% {Z}_{i}^o = \sigma (
% {Z}_{i-1}^o + 
% \overbrace{W^{sd} {Z}_{i-1}^o}^{\substack{\text{Message from}\\ \text{Layout or Objects}}} + 
% \overbrace{W^{rs} {Z}_{i-1}^r{\alpha}^{rs} + W^{rd} {Z}_{i-1}^r{\alpha}^{rd}}^{\substack{\text{Messages from}\\ \text{Neighboring Relationships}}}
{{Z}^o}^{\prime} = \sigma (
{Z}^o + 
\overbrace{W^{sd} {Z}^o}^{\substack{\text{Message from}\\ \text{Layout or Objects}}} + 
\overbrace{W^{rs} {Z}^r{\alpha}^{rs} + W^{rd} {Z}^r{\alpha}^{rd}}^{\substack{\text{Messages from}\\ \text{Neighboring Relationships}}}
),
\end{equation}
% and the i-th update for relationship node can be defined as
and the relationship node representations can be updated as
\begin{equation}
% {Z}_{i}^r = \sigma (
% {Z}_{i-1}^r + 
% \underbrace{W^{sr} {Z}_{i-1}^{o} {\alpha}^{sr} +  W^{dr} {Z}_{i-1}^{o} {\alpha}^{dr}}_{\text{Messages from Layout or Neighboring Objects}}
{{Z}^r}^{\prime} = \sigma (
{Z}^r + 
\underbrace{W^{sr} {Z}^{o} {\alpha}^{sr} +  W^{dr} {Z}^{o} {\alpha}^{dr}}_{\text{Messages from Layout or Neighboring Objects}}
).
\end{equation}
% After four steps of message passing, two-layer MLP is used to predict a residual which is later added into each output from LEN and ODN.
% \zpc{Please assign some mathematical symbols for each output of LEN and ODN.}
After four steps of message passing, independent MLPs are used to decode object node representations into residuals for corresponding object bounding box parameters $\left({\delta},d,{s},\theta\right)$, and layout node representation into residuals for initial layout box $\left({C},{s}^{l},\theta^{l}\right)$ and camera pose $\mathbf{R}\left(\beta,\gamma\right)$. Please refer to our supplementary or
% \zpc{our supplementary or} % \zc{Our supplementary dose not have definition of these parameters}
\cite{nie2020total3dunderstanding} for the details of the definition.
The shape codes can be also refined in the scene graph, while we find that it doesn't improve empirically as much as for the layout and object poses in our pipeline because our local implicit embedding network, which will be introduced in the following, is powerful enough to learn accurate shapes.

\subsection{Local Implicit Embedding Network}
\label{sec:lien}
With a graph constructed for each scene, we naturally ask what features help SGCN effectively capture contextual information among objects. 
Intuitively, we expect features that well describe 3D object geometry and their relationship in 3D space. 
%Since MGN from Total3D only provides appearance feature without explicit geometry information, we turn to structured representation with physically meaningful parameters. 
%We adopted the representation method of LDIF\cite{genova2020local}, a 3D shape representation combining structured representation and implicit function, and propose an encoder to accomplish the object reconstruction task. 
Motivated by Genova \etal \cite{genova2020local}, we propose to utilize the local deep implicit representation as the features embedding object shapes due to its superior performance for single object reconstruction.
In their model, 
% Genova \etal proposed a function as a decoder to represent 3D geometry by predicting the inside-outside label from latent code and 3D point coordinate. 
the function is a combination of 32 3D elements (16 with symmetry constraints), with each element described with 10 Gaussian function parameters analytic code %\yd{Do they call this analytic code? Sounds a bit weird} 
and 32-dim latent variables (latent code). %\yd{what are these numbers? dimension? If so, write like xxx-dim at least}
The Gaussian parameters describe the scale constant, center point, radii, and Euler angle of every Gaussian function, which contains structured information of the 3D geometry.
We use analytic code as a feature for object nodes in SGCN, %\yd{which node}
% which provides a physically meaningful \yd{what does ``physically meaningful'' mean} description of the local object structure.
which should provide information on the local object structure.
Furthermore, since the centers of the Gaussian functions presumably correspond to centers of different parts of an object, %substructures \yd{what does substructure mean?} of a object, 
we also transform them from the object coordinate system to the world coordinate system as a feature for every object node in SGCN.
The transformation provides global information about the scene, which makes SGCN easier to infer relationships between objects.
The above two features make up the implicit features of LIEN.

%To use the LDIF representation for 3D object reconstruction task, we replaced its depth 
%encoder with our proposed LIEN, 
As LDIF \cite{genova2020local} is designed for 3D object reconstruction from one or multiple depth images, we design a new image-based Local Implicit Embedding Network (LIEN) to learn the 3D latent shape representation %\yd{3D latent representation?} 
from the image which is obviously a more challenging problem.
Our LIEN consists of a Resnet-18 as image encoder, along with a three-layer MLP to get the analytic and latent code. 
Additionally, in order to learn the latent features effectively, we concatenate the category code with the image feature from the encoder to introduce shape priors to the LIEN, which improves the performance greatly. Please refer to our supplementary material for the detailed architecture of the proposed LIEN. 

\subsection{Loss Function}

% \yd{Since the 2nd loss is for refinement, call the first one for intialization stage?}
\noindent\textbf{Losses for Initialization Modules.} 
When training LIEN along with LDIF decoder individually, we follow \cite{genova2020local} to use the shape element center %\yd{element center seems never defined} 
loss $\mathcal{L}_{c}$ with weight $\lambda_{c}$ and point sample loss,
\begin{equation}
\begin{aligned}
\mathcal{L}_{p} = \lambda_{ns}\mathcal{L}_{ns} + \lambda_{us}\mathcal{L}_{us},
\end{aligned}
\end{equation}
where $\mathcal{L}_{ns}$ and $\mathcal{L}_{us}$ evaluates %\yd{evaluate or weight?} 
$L_2$ losses %\yd{what L2 loss? never defined} 
for near-surface samples and uniformly sampled points.
When training LEN and ODN, we follow \cite{huang2018cooperative,nie2020total3dunderstanding} to use classification and regression loss for every output parameter of LEN and ODN,
\begin{equation}
\begin{aligned}
\mathcal{L}_{LEN} = & \sum_{y \in \{ \beta,\gamma, {C}, {s}^{l}, \theta^{l} \}} \lambda_{y}\mathcal{L}_{y},
% + \lambda _{co}\mathcal{L}_{co},
\end{aligned}
\end{equation}

\begin{equation}
\begin{aligned}
\mathcal{L}_{ODN} = & \sum_{x \in \{  {\delta}, d, {s}, \theta  \} }\lambda_{x}\mathcal{L}_{x}.
\end{aligned}
\end{equation}

\begin{figure}[t]
	\centering
	\vspace{-1em}
	\includegraphics[width=0.4\textwidth]
		{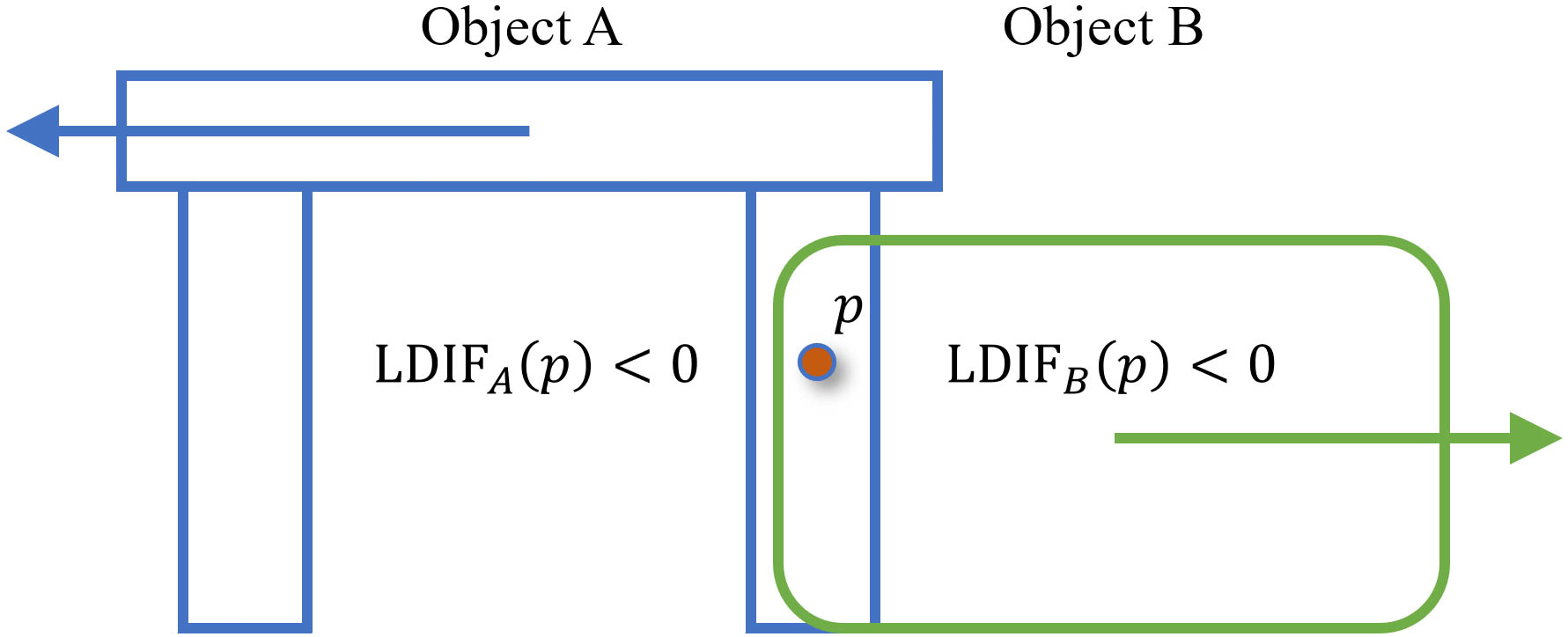}
	\vspace{0.5em}
	\beforecaption
	\caption{Object physical violation loss. Based on the insight that objects should not intersect, we punish points inside neighboring objects (demonstrated as $p$, which has negative LDIF values in both object A and object B). With error back-propagated through the LDIF decoder, intersected objects should be pushed back from each other, reducing intersection resulting from bad object pose estimation.}
	\label{fig:lphy}
\end{figure}

\noindent\textbf{Joint Refinement with Object Physical Violation Loss.} 
%Prior work \cite{huang2018cooperative} proposed a physical violation loss as a part of cooperative loss to punish objects outside layout bounding box. 
%We further extend the idea of physical violation to object-object intersection. 
For the refinement stage, we aim to optimize the scene layout and object poses using the scene context information by minimizing the following loss function,
\begin{equation}
\begin{aligned}
\mathcal{L}_{j} = \mathcal{L}_{LEN} + \mathcal{L}_{ODN} + \lambda _{co}\mathcal{L}_{co} + \lambda _{phy} \mathcal{L}_{phy}.
\end{aligned}
\end{equation}
Besides $\mathcal{L}_{LEN}$, $\mathcal{L}_{ODN}$ and cooperative loss $\mathcal{L}_{co}$ \cite{nie2020total3dunderstanding}, we propose a novel physical violation loss as a part of joint loss for the scene graph convolutional network to make sure that objects will not intersect with each other.
The neural SDF representation used by local implicit representation gives us a convenient way to propagate gradient from undesired geometry intersection back to the object pose estimation. %from error of output SDF values back to input point coordinates \yd{from undesired geometry intersection back to the object and scene layout?}. %\yd{Don't understand. Why implicit representation make point sampling easier?}. 
% To utilize this idea, we first randomly sample points inside the bounding box of each object, adding to the already known center points of Gaussian elements \yd{how?}. 
%To utilize this idea, 
To achieve this,
we first sample points inside objects.
For each object $i$, we randomly sample points inside the bounding box of each object, along with the center points of Gaussian elements as point candidates.
% To get more samples near object surface, we do a few steps of optimization on the point coordinates with the objective of minimizing the error between the SDF output and the iso surface value \yd{Don't understand how and why?}. 
We then queue these candidates into LDIF decoder of the object and filter out points outside object surfaces to get inside point samples $\mathbb{S}_{i}$. 
% Finally, we transform the sampled inside points of each object $i$ to the object-coordinate of its k-nearest neighbors $\mathbb{N}_{i}$ to verify if they are bellow the iso surfaces of the neighboring objects. 
Finally, we queue $\mathbb{S}_{i}$ into the LDIF decoder of the k-nearest objects %\yd{what does its refer to? and k-nearest of what? object?} 
$\mathbb{N}_{i}$ to verify if they have intersection with other objects (if the predicted label is "inside"). %\yd{inside? cuz you are checking intersection as mentioned in prev sentence}
We follow \cite{genova2020local} to compute a $L_2$ loss between the predicted labels of intersected points with the ground truth surface label (where we use $1$, $0$, $0.5$ for "outside", "inside", "surface" labels). %iso surface values 
The object physical violation loss can be defined as:
\begin{equation}
\begin{aligned}
    \mathcal{L}_{phy} = \frac{1}{N}\sum_{i=1}^{N}\frac{1}{|\mathbb{S}_{i}|}\sum_{\mathbf{x} \in \mathbb{S}_{i}} \|\mathtt{relu}(0.5 - \mathtt{sig}(\alpha \mathrm{LDIF}_{i}(\mathbf{x})))\|,
\end{aligned}
\end{equation}
where $\mathrm{LDIF}_{i}(\mathbf{x})$ is the LDIF for object $i$ to decode a world coordinate point $\mathbf{x}$ into LDIF value.
A sigmoid is applied on the LDIF value (scaled by $\alpha$) to get the predicted labels, and a ReLU is applied to consider only the intersected points.
As shown in \autoref{fig:lphy}, the loss punishes intersected sample points thus push both objects away from each other to prevent intersections.
%inside with intersection along the gradient descent direction of the signed distance field of other objects, thus reducing intersection between objects.
% \yd{this paragraph is not easy to follow.}

\section{Experiments}
In this section, we compare our method with state-of-the-art 3D scene understanding methods in various aspects and provide an ablation study to highlight the effectiveness of major components. %, including object reconstruction, 3D object detection, layout estimation, camera pose prediction, and scene mesh reconstruction. We also provide ablation study to highlight the effectiveness of major components.

\subsection{Experiment Setup}

\begin{figure}[!ht]
	\centering
	\vspace{-1em}
	\begin{subfigure}[t]{0.1\textwidth}
		\includegraphics[width=\textwidth]
		{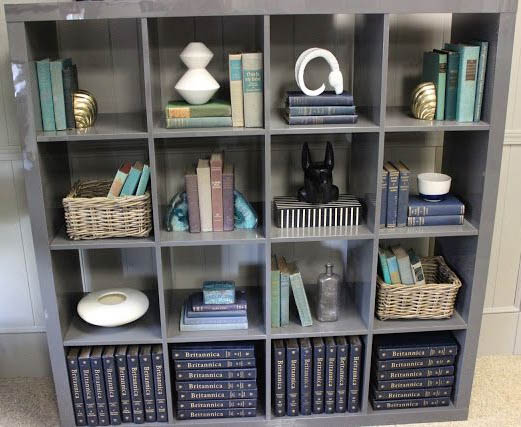}
		\includegraphics[width=\textwidth]  
		{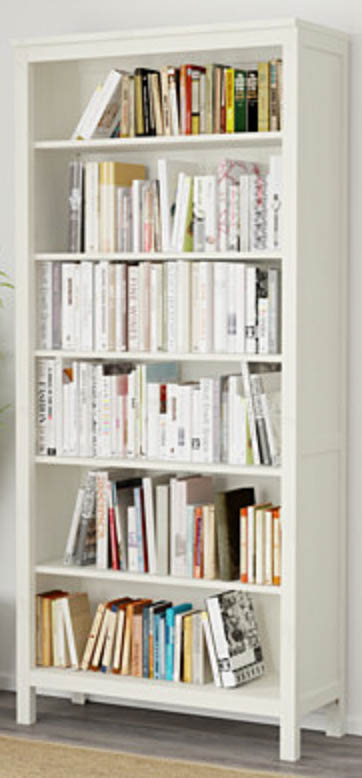}
		\includegraphics[width=\textwidth]
		{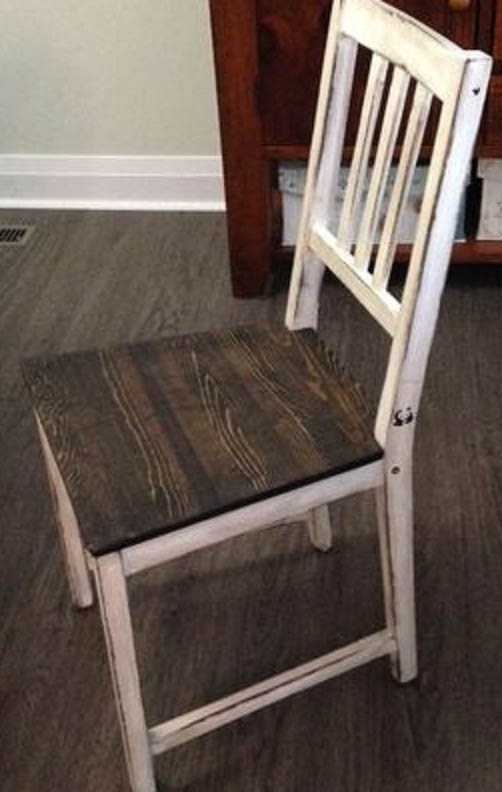}
		\includegraphics[width=\textwidth]  
		{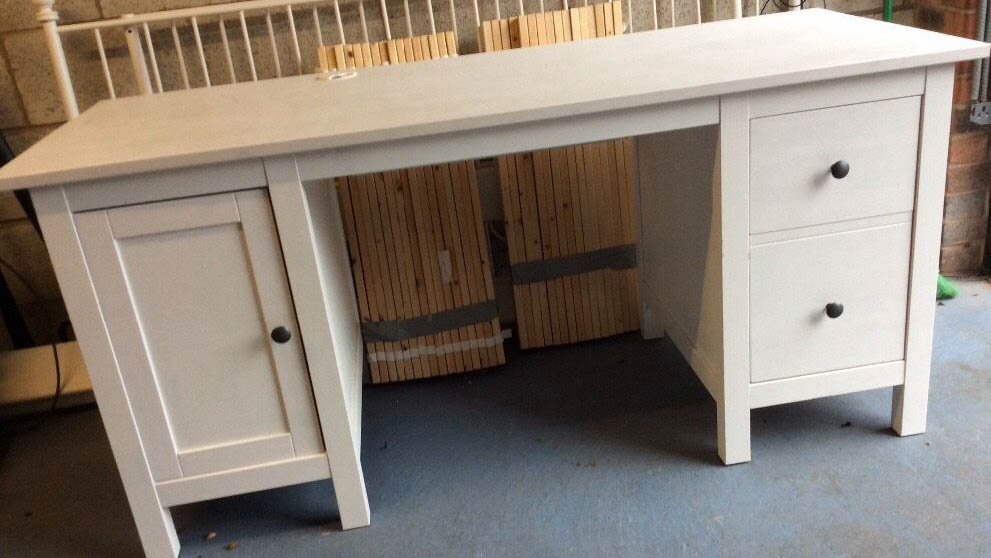}
		\includegraphics[width=\textwidth]  
		{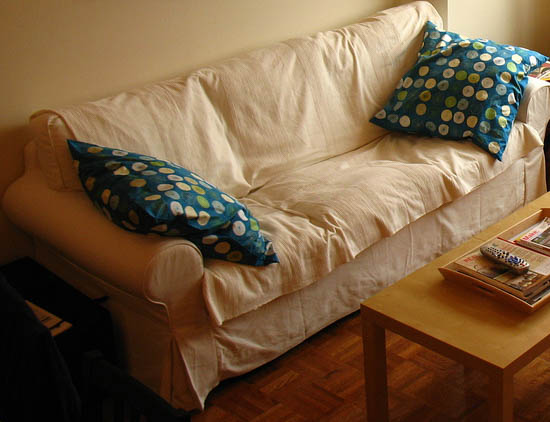}
		\includegraphics[width=\textwidth]  
		{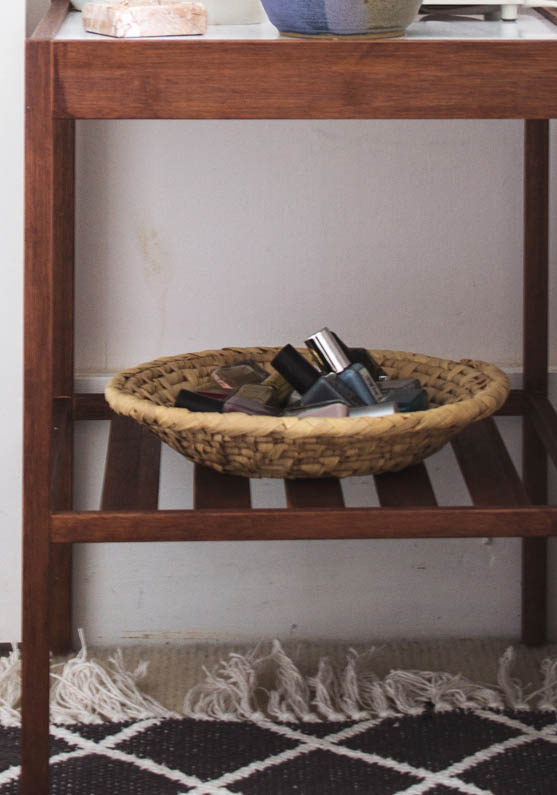}
		\vspace{-6mm}
		\caption{Input}
	\end{subfigure}
		\begin{subfigure}[t]{0.1\textwidth}
		\includegraphics[width=\textwidth]
		{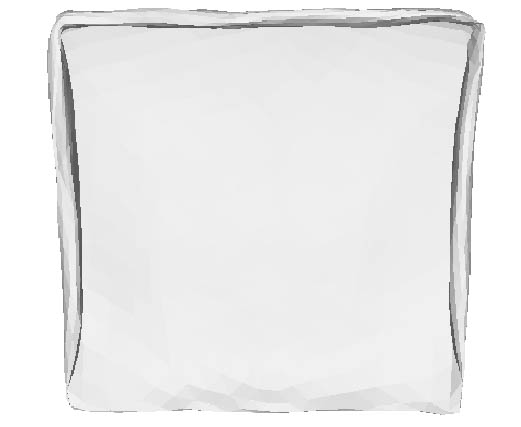}
		\includegraphics[width=\textwidth]  
		{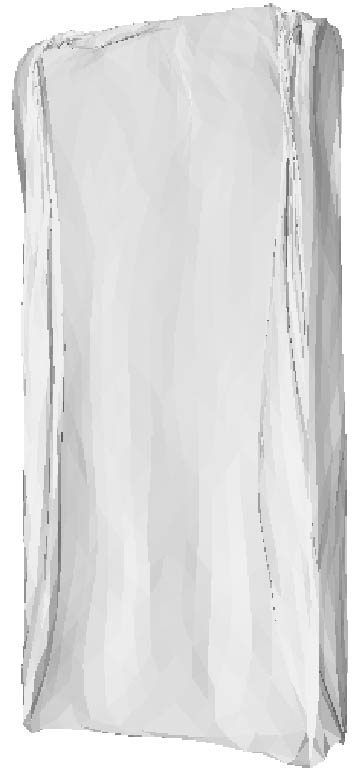}
		\includegraphics[width=\textwidth]
		{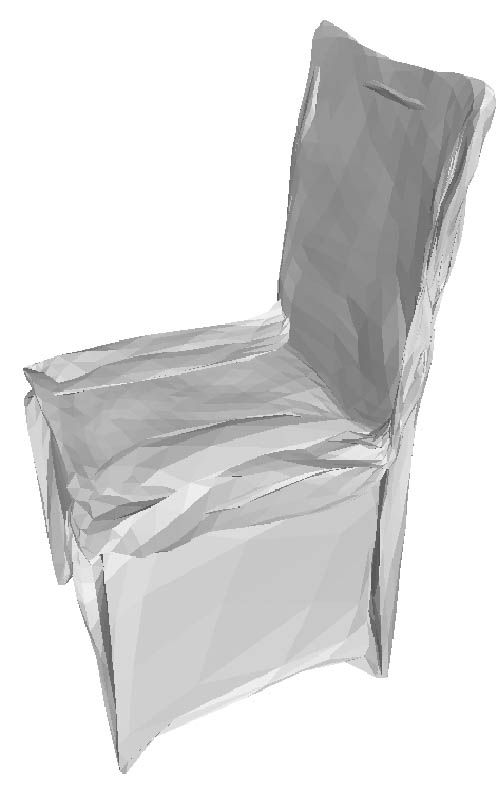}
		\includegraphics[width=\textwidth]  
		{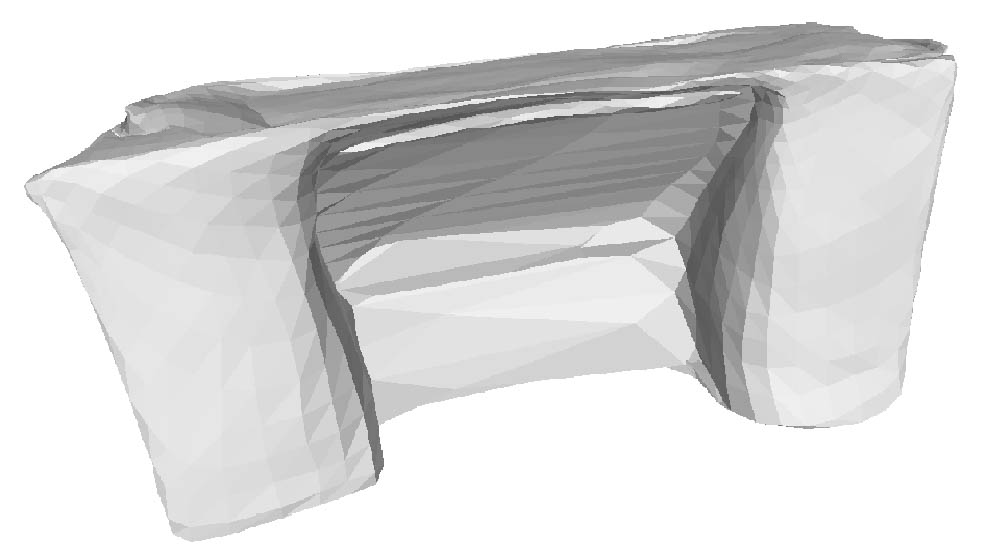}
		\includegraphics[width=\textwidth]  
		{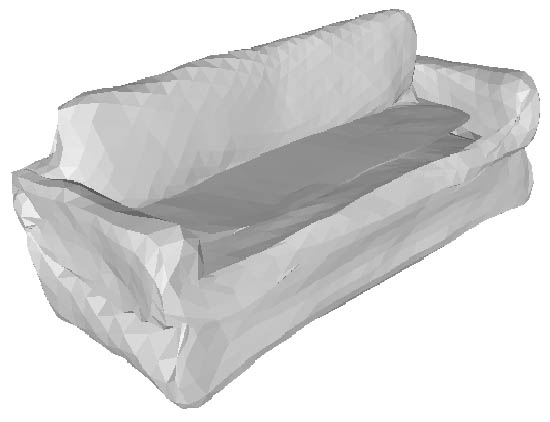}
		\includegraphics[width=\textwidth]  
		{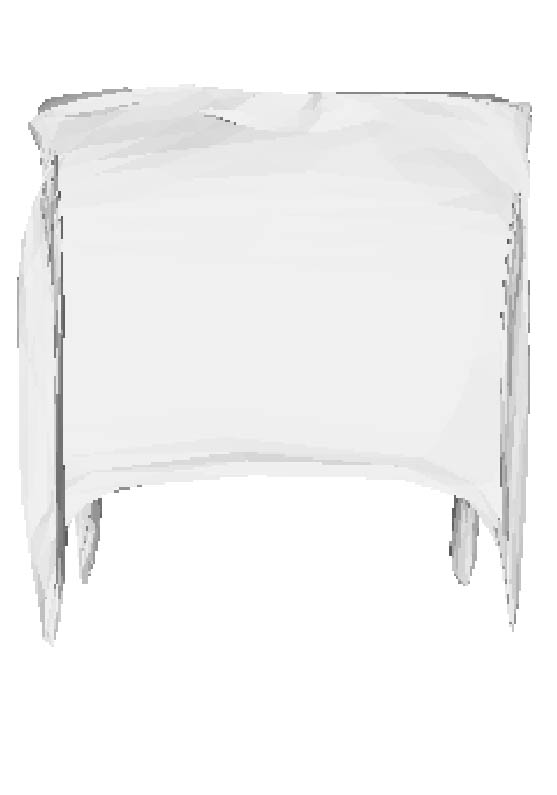}
		\vspace{-6mm}
		\caption{AtlasNet}
		\label{fig:objrecon_atalasnet}
	\end{subfigure}
	\begin{subfigure}[t]{0.1\textwidth}
		\includegraphics[width=\textwidth]
		{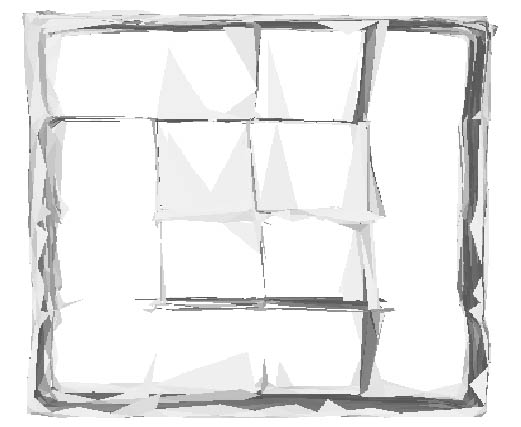}
		\includegraphics[width=\textwidth]  
		{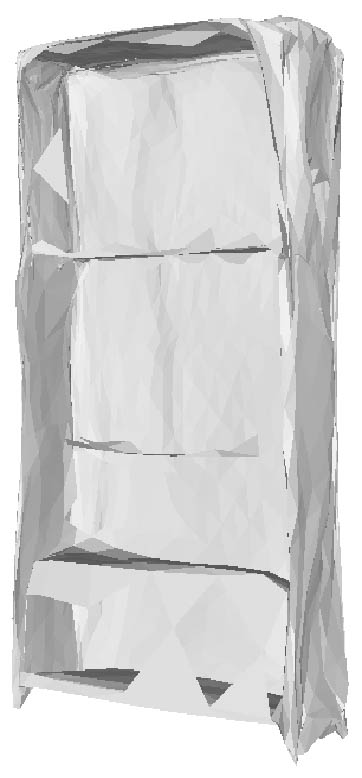}
		\includegraphics[width=\textwidth]
		{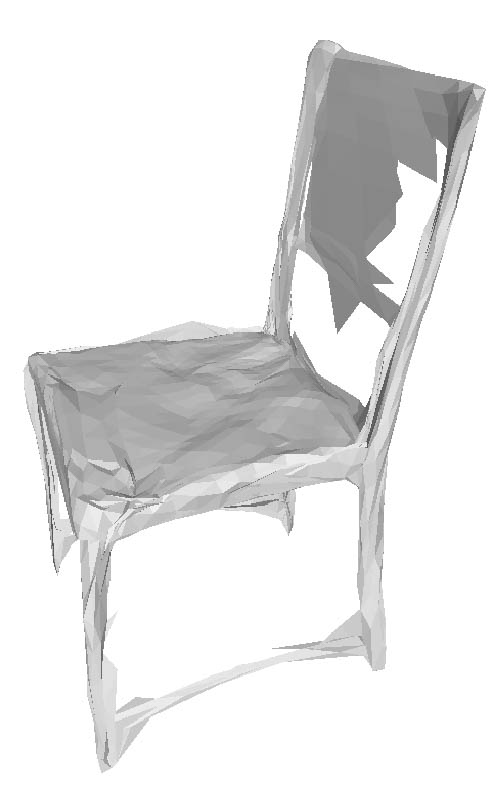}
		\includegraphics[width=\textwidth]  
		{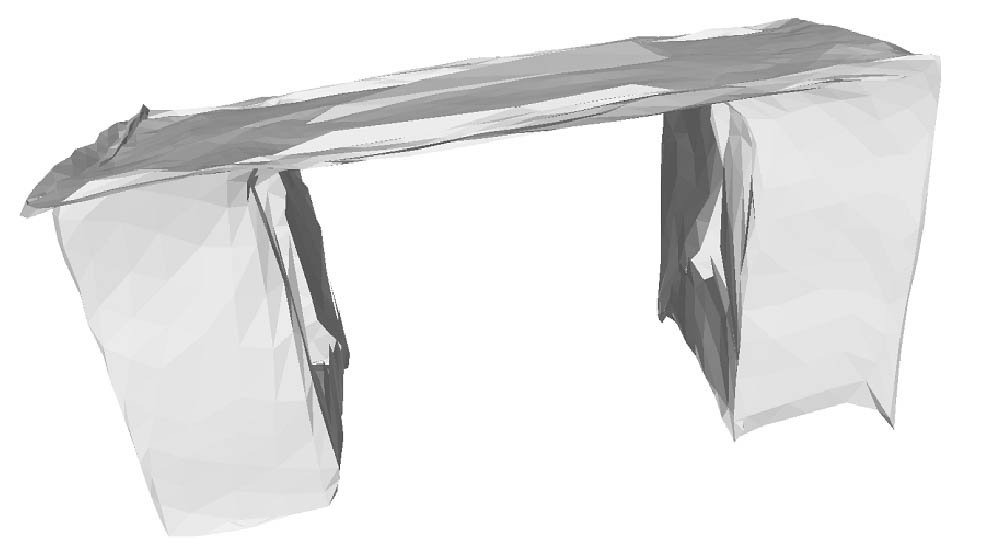}
		\includegraphics[width=\textwidth]  
		{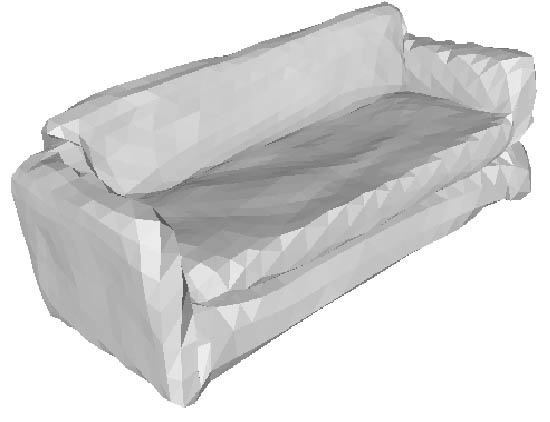}
		\includegraphics[width=\textwidth]  
		{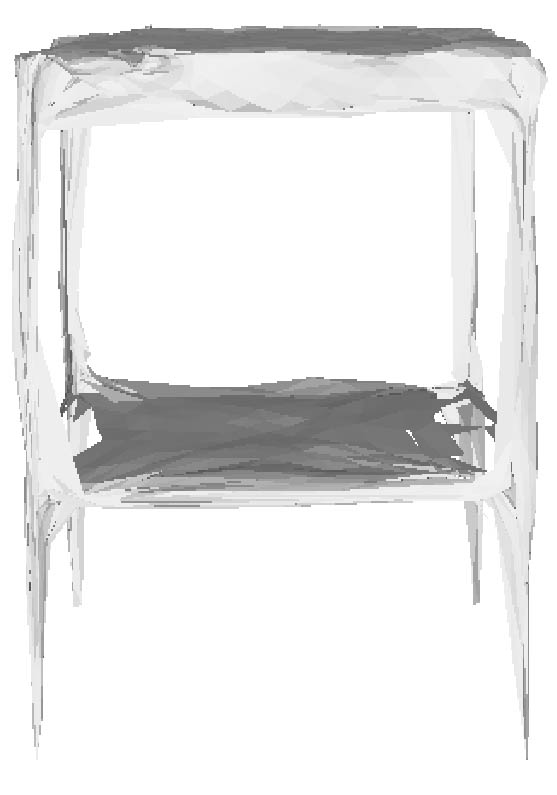}
		\vspace{-6mm}
		\caption{MGN}
		\label{fig:objrecon_mgn}
	\end{subfigure}
	\begin{subfigure}[t]{0.1\textwidth}
		\includegraphics[width=\textwidth]
		{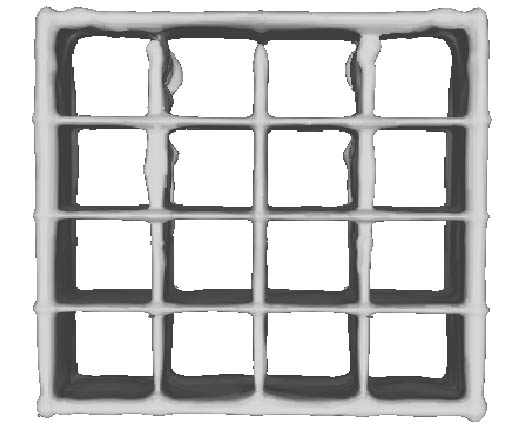}
		\includegraphics[width=\textwidth]  
		{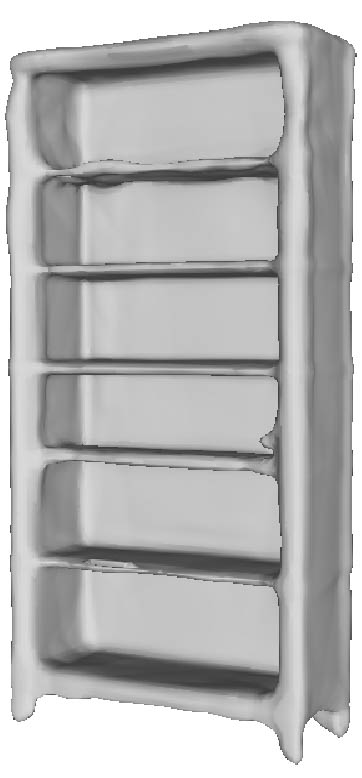}
		\includegraphics[width=\textwidth]
		{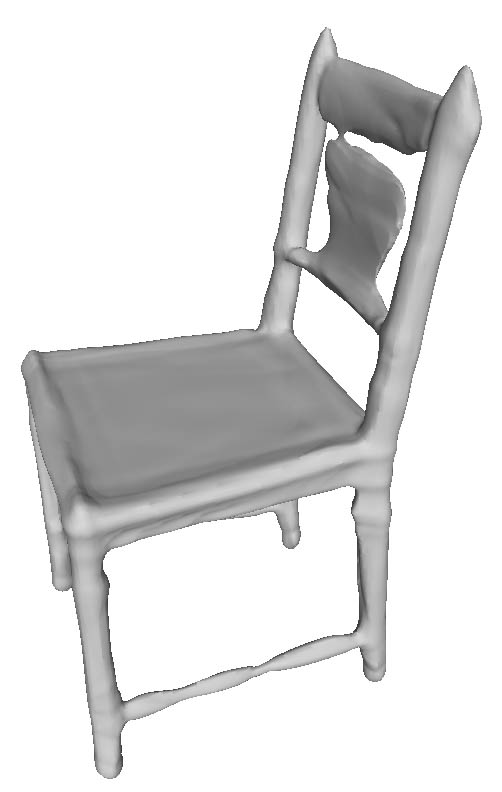}
		\includegraphics[width=\textwidth]  
		{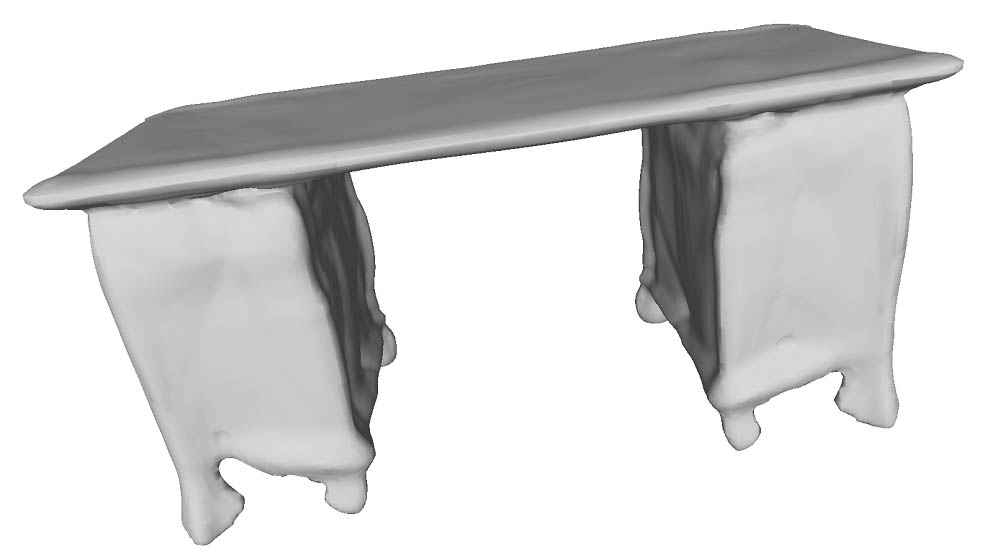}
		\includegraphics[width=\textwidth]  
		{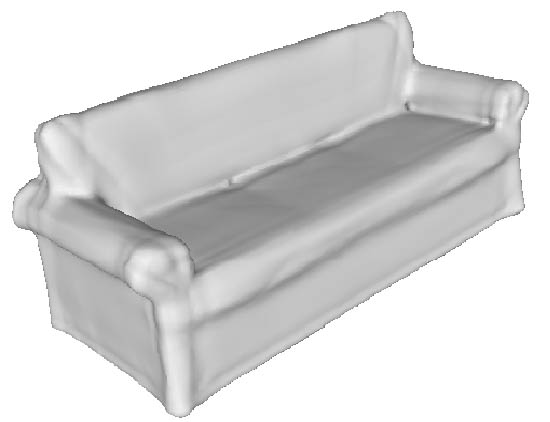}
		\includegraphics[width=\textwidth]  
		{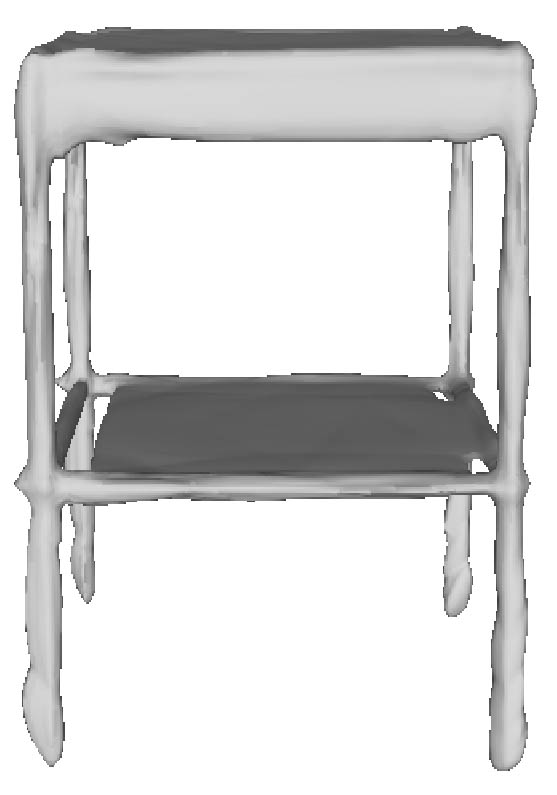}
		\vspace{-6mm}
		\caption{Ours}
		\label{fig:objrecon_ours}
	\end{subfigure}
	\vspace{0.5em}
	\beforecaption
	\caption{Object reconstruction qualitative comparison. We use the implementation from \cite{nie2020total3dunderstanding} for AtlasNet \cite{groueix2018}. Our results contain finer details and have more smooth surfaces.}
	\vspace{-0.2em}
	\label{fig:objrecon}
\end{figure}

\begin{table*}[!h]
    \vspace{-1.5em}
	\begin{center}
		\begin{tabular}{|l|c c c c c c c c c|c|}
			\hline
			Category & bed & bookcase & chair & desk & sofa & table & tool & wardrobe & misc & mean\\
			\hline\hline
			AtlasNet \cite{groueix2018} & 9.03 & 6.91 & 8.37 & 8.59 & 6.24 & 19.46 & 6.95 & 4.78 & 40.05 & 12.26\\
			TMN \cite{pan2019deep} & 7.78 & 5.93 & 6.86 & 7.08 & 4.25 & 17.42 & 4.13 & 4.09 & \textbf{23.68} & 9.03\\
			MGN \cite{nie2020total3dunderstanding} & 5.99 & 6.56 & \textbf{5.32} & \textbf{5.93} & \textbf{3.36} & 14.19 & 3.12 & \textbf{3.83} & 26.93 & 8.36\\
			Ours & \textbf{4.11} & \textbf{3.96} & 5.45 & 7.85 & 5.61 & \textbf{11.73} & \textbf{2.39} & 4.31 & 24.65 & \textbf{6.72}\\
			\hline
		\end{tabular}
	\end{center}
	\vspace{-0.5em}
	\beforecaption
	\caption{Object reconstruction comparison. We report the Chamfer distance scaled with the factor of $10^3$. We follow \cite{nie2020total3dunderstanding} to align the reconstructed mesh to ground-truth with ICP then sample 10K points from the output and the ground-truth meshes. Although trained on watertight meshes with more noise, our results still shows better results.}
	\vspace{-0.5em}
	\label{tbl:objrecon}
\end{table*}

\begin{table*}[!h]
    \vspace{-0.5em}
	\begin{center}
		\begin{tabular}{|l|c|c|c|c|c|c|c|c|c|c|c|}
			\hline
			Method & bed & chair & sofa & table & desk & dresser & nightstand & sink & cabinet & lamp & mAP\\
			\hline\hline
			3DGP \cite{choi2013understanding} & 5.62 & 2.31 & 3.24 & 1.23 & - & - & - & - & - & - & - \\
			HoPR \cite{huang2018holistic} & 58.29 & 13.56 & 28.37 & 12.12 & 4.79 & 13.71 & 8.80 & 2.18 & 0.48 & 2.41 & 14.47 \\
			CooP \cite{huang2018cooperative} & 57.71 & 15.21 & 36.67 & 31.16 & 19.90 & 15.98 & 11.36 & 15.95 & 10.47 & 3.28 & 21.77\\
			Total3D \cite{nie2020total3dunderstanding} & 60.65 & 17.55 & 44.90 & 36.48 & 27.93 & 21.19 & 17.01 & 18.50 & 14.51& 5.04& 26.38\\
			Ours & \textbf{89.32} & \textbf{35.14} & \textbf{69.10} & \textbf{57.37} & \textbf{49.03} & \textbf{29.27} & \textbf{41.34} & \textbf{33.81} & \textbf{33.93} & \textbf{11.90} & \textbf{45.21}\\
			\hline
		\end{tabular}
	\end{center}
	\vspace{-0.5em}
	\beforecaption
	\caption{3D object detection comparison. For CooP, we report the better results from \cite{nie2020total3dunderstanding} trained on NYU-37 object labels. Our method outperforms SOTA, benefiting from a better understanding of the object relationships and the scene context.}
	\vspace{-1em}
	\label{tbl:3ddetection}
\end{table*}

\noindent \textbf{Datasets.} 
We follow \cite{nie2020total3dunderstanding} to use two datasets to train each module individually and jointly. % \yd{what does ``complementarily train'' mean?}.
We use two datasets for training and evaluation.
1) \textbf{Pix3D} dataset \cite{sun2018pix3d} is presented as a benchmark for shape-related tasks including reconstruction, providing 9 categories of 395 furniture models and 10,069 images with precise alignment.
% When training LIEN, watertight meshes must be used to retrieve the ground truth value of inside-outside labels. 
% However, models of Pix3D is quite dirty with occasionally inverted surface normals and large holes, which causes failure with traditional flood fill algorithm. 
% To get more robust results, we use the mesh fusion pipline from Occupancy Network \cite{mescheder2019occupancy}, which generate watertight meshes by fusing signed distance field from several virtual cameras and applying marching cube algorithm on it.
% Although the mesh fusion pipeline makes the model thicker and introduces noise to the ground truth sample points, we evaluate it on the original mesh to directly compare with previous works.
% \zc{removed for limited space}
We use the mesh fusion pipeline from Occupancy Network \cite{mescheder2019occupancy} to get watertight meshes for LIEN training and evaluate LIEN on original meshes.
%\yd{Is this the same fusion you mention below?}
2) \textbf{SUN RGB-D} dataset \cite{song2015sun} contains 10K RGB-D indoor images captured by four different sensors and is densely annotated with 2D segmentation, semantic labels, 3D room layout, and 3D bounding boxes with object orientations.
Follow Total3D \cite{nie2020total3dunderstanding}, we use the train/test split from \cite{gkioxari2019mesh} on the Pix3D dataset and the official train/test split on the SUN RGB-D dataset. 
The object labels are mapped from NYU-37 to Pix3D as presented by \cite{nie2020total3dunderstanding}.

\noindent \textbf{Metrics.} 
We adopt the same evaluation metrics with \cite{huang2018cooperative,nie2020total3dunderstanding}, including average 3D Intersection over Union (IoU) for layout estimation; mean absolute error for camera pose; average precision (AP) for object detection; and chamfer distance for single-object mesh generation from single image. 

\noindent \textbf{Implementation.} 
We use the outputs of the 2D detector from Total3D as the input of our model. 
We also adopted the same structure of ODN and LEN from Total3D. 
LIEN is trained with LDIF decoder on Pix3D with watertight mesh, using Adam optimizer with a batch size of 24 and learning rate decaying from 2e-4 (scaled by 0.5 if the test loss stops decreasing for 50 epochs, 400 epochs in total) and evaluated on the original non-watertight mesh. 
SGCN is trained on SUN RGB-D, using Adam optimizer with a batch size of 2 and learning rate decaying from 1e-4 (scaled by 0.5 every 5 epochs after epoch 18, 30 epochs in total). 
We follow \cite{nie2020total3dunderstanding} to train each module individually then jointly.
When training SGCN individually, we use $\mathcal{L}_{j}$ without $\mathcal{L}_{phy}$, and put it into the full model %and load other modules (i.e. ODN, LEN, LIEN, LDIF decoder) with pretrained weights and fix them. 
with pre-trained weights of other modules.
In joint training, we adopt the observation from \cite{nie2020total3dunderstanding} that object reconstruction depends on clean mesh for supervision, to fix the weights of LIEN and LDIF decoder.
% \yd{I would remove all the network details as we already refer to supp or mentioned in paper. Only details about network training is needed. Besides LR, you can also talk about optimizer, regularization, batch size.}

\subsection{Comparison to State-of-the-art}
\label{sec:comp}
In this section, we compare to the state-of-the-art methods for holistic scene understand from aspects including object reconstruction, 3D object detection, layout estimation, camera pose prediction, and scene mesh reconstruction.
% \yd{order these tasks in the same order as presented below}

% \noindent \textbf{3D Object Reconstruction:} 
% When training LIEN, watertight meshes must be used to retrieve the ground truth value of inside-outside labels. 
% However, models of Pix3D is quite dirty with occasionally inverted surface normals and large holes, which causes failure with traditional flood fill algorithm. 
% To get more robust results, we utilize the mesh fusion pipeline \cite{mescheder2019occupancy} which generate watertight meshes by fusing signed distance field from several virtual cameras and applying marching cube algorithm on it.
% Although the mesh fusion pipeline makes the model thicker and introduces noise to the ground truth sample points, we evaluate it on the original mesh to directly compare with previous works.
% \yd{If this is the fusion you mentioned when you introduce the dataset, I would remove all of them.}

\noindent \textbf{3D Object Reconstruction.} 
We first compare the performance of LIEN with previous methods, including AtlasNet \cite{groueix2018}, TMN \cite{pan2019deep}, and Total3D \cite{nie2020total3dunderstanding}, for the accuracy of the predicted geometry on Pix3D dataset.
All the methods take as input a crop of image of the object and produce 3D geometry.
To make a fair comparison, the one-hot object category code is also concatenated with the appearance feature for AtlasNet \cite{groueix2018} and TMN \cite{pan2019deep}.
For our method, we run a marching cube algorithm on 256 resolution to reconstruct the mesh.
The quantitative comparison is shown in \autoref{tbl:objrecon}.
Our method produces the most accurate 3D shape compared to other methods, yielding the lowest mean Chamfer Distance across all categories.
Qualitative results are shown in \autoref{fig:objrecon}.
AtlasNet produces results in limited topology and thus generates many undesired surfaces.
MGN mitigates the issue with the capability of topology modification, which improves the results but still leaves obvious artifacts and unsmooth surface due to the limited representation capacity of the triangular mesh.
In contrast, our method produces 3D shape with correct topology, smooth surface, and fine-grained details, which clearly shows the advantage of the deep implicit representation.

\begin{table}[t]
    \vspace{-0.25em}
	\begin{center}
	    \resizebox{0.8\columnwidth}{!}{
    		\begin{tabular}{|l|c|c|c|}
    			\hline
    			Method & Layout IoU & Cam pitch & Cam roll \\
    			\hline\hline
    			3DGP \cite{choi2013understanding} & 19.2 & - & - \\
    			Hedau \cite{hedau2009recovering} & - & 33.85 & 3.45\\
    			HoPR \cite{huang2018holistic} & 54.9 & 7.60 & 3.12 \\
    			CooP \cite{huang2018cooperative} & 56.9 & 3.28 & 2.19\\
    			Total3D \cite{nie2020total3dunderstanding} & 59.2 & 3.15 & \textbf{2.09}\\
    			Ours & \textbf{64.4} & \textbf{2.98} & 2.11 \\
    			\hline
    		\end{tabular}
    	}
	\end{center}
	\vspace{-0.5em}
	\beforecaption
	\caption{3D layout and camera pose estimation comparison. Our method outperforms SOTA by 5.2\% in layout estimation while on par with SOTA on camera pose estimation.}
	\label{tbl:layout}
\end{table}

\begin{figure*}[!ht]
	\centering
	\scriptsize
	\newcommand{\rgb}[1]{\raisebox{-0.5\height}{\includegraphics[width=.15\linewidth]{#1}}}
	\newcommand{\oblique}[1]{\raisebox{-0.5\height}{\includegraphics[width=.15\linewidth,clip,trim=48 12 48 12]{#1}}}
	\newcommand{\rot}[1]{\rotatebox[origin=c]{90}{#1}}
	\def\arraystretch{0.5}%
	\begin{tabular}{c|c*{6}{c@{\hspace{1px}}}}
    	\multicolumn{8}{l}{\hspace{34px}\includegraphics[width=.91\linewidth]{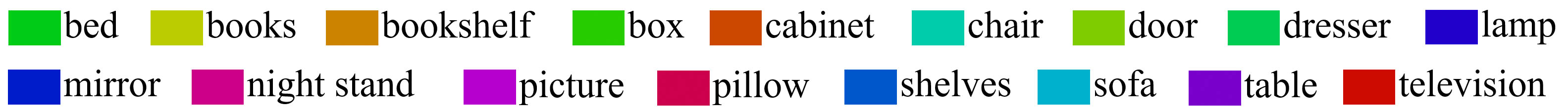}\vspace{5px}} \\
        \rule{0pt}{5px}&\rot{Input}& 
    	    \rgb{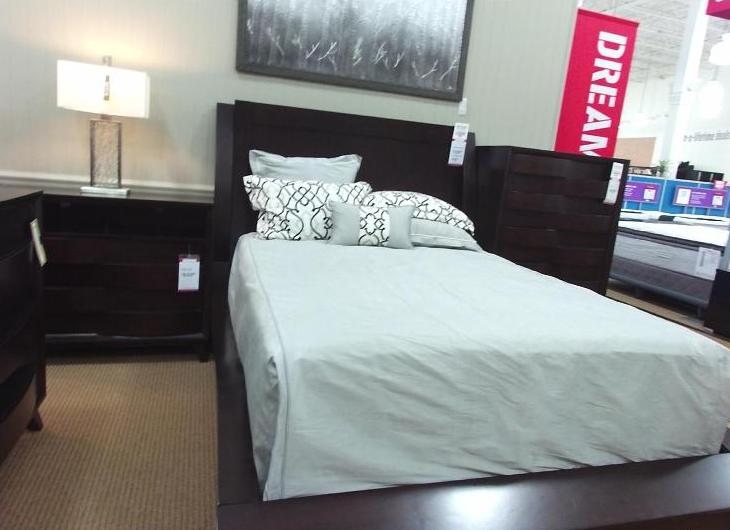} &
    	    \rgb{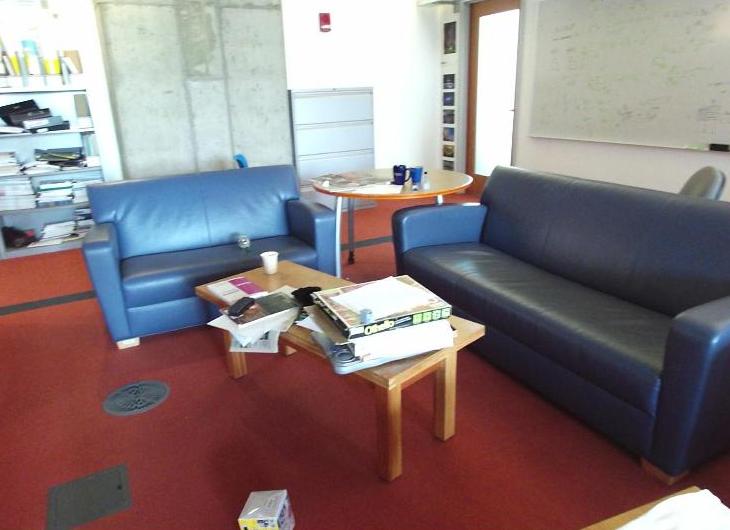} &
    	    \rgb{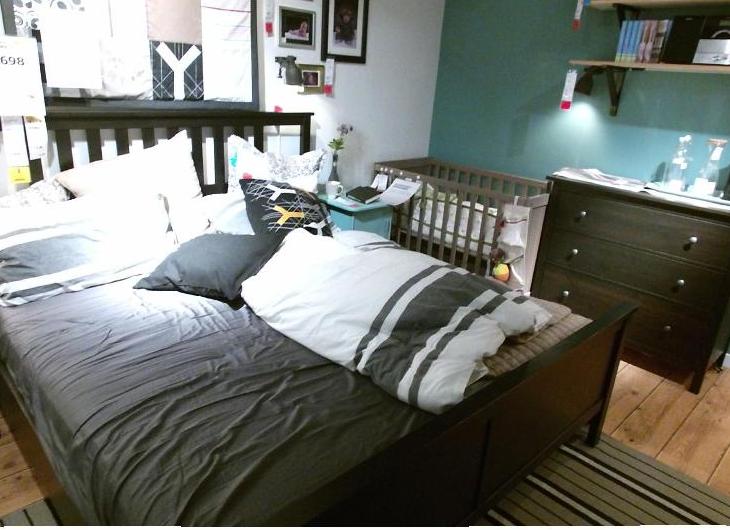} &
    	    \rgb{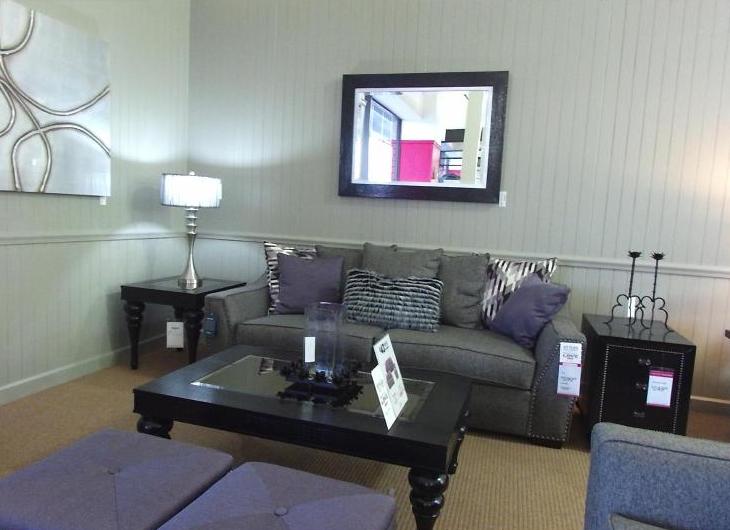} &
    	   % \rgb{figure/scnrecon/input/000845} &
    	    \rgb{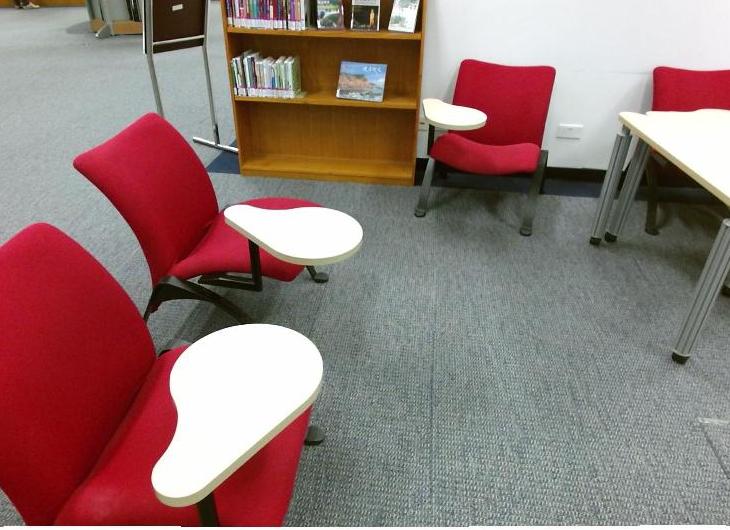} &
    	   % \rgb{figure/scnrecon/input/002435} &
    	    \rgb{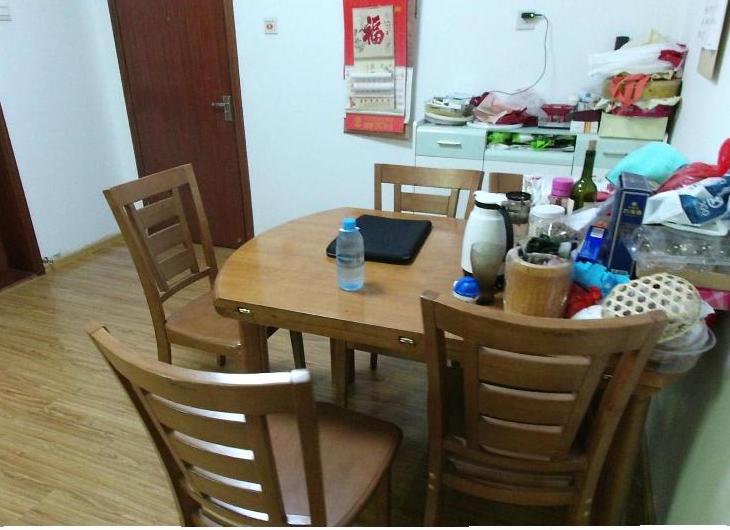} \\
    	\hline
    	    &\rot{Total3D}& 
        	    \oblique{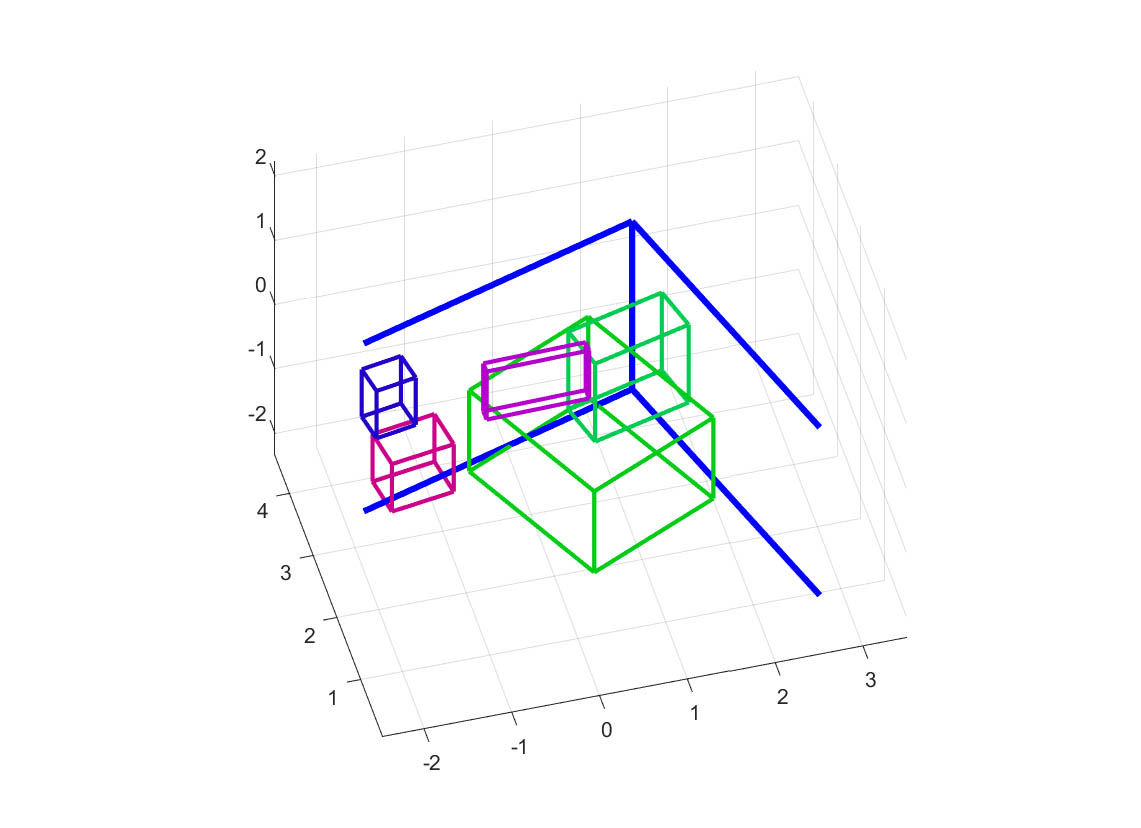} &
        	    \oblique{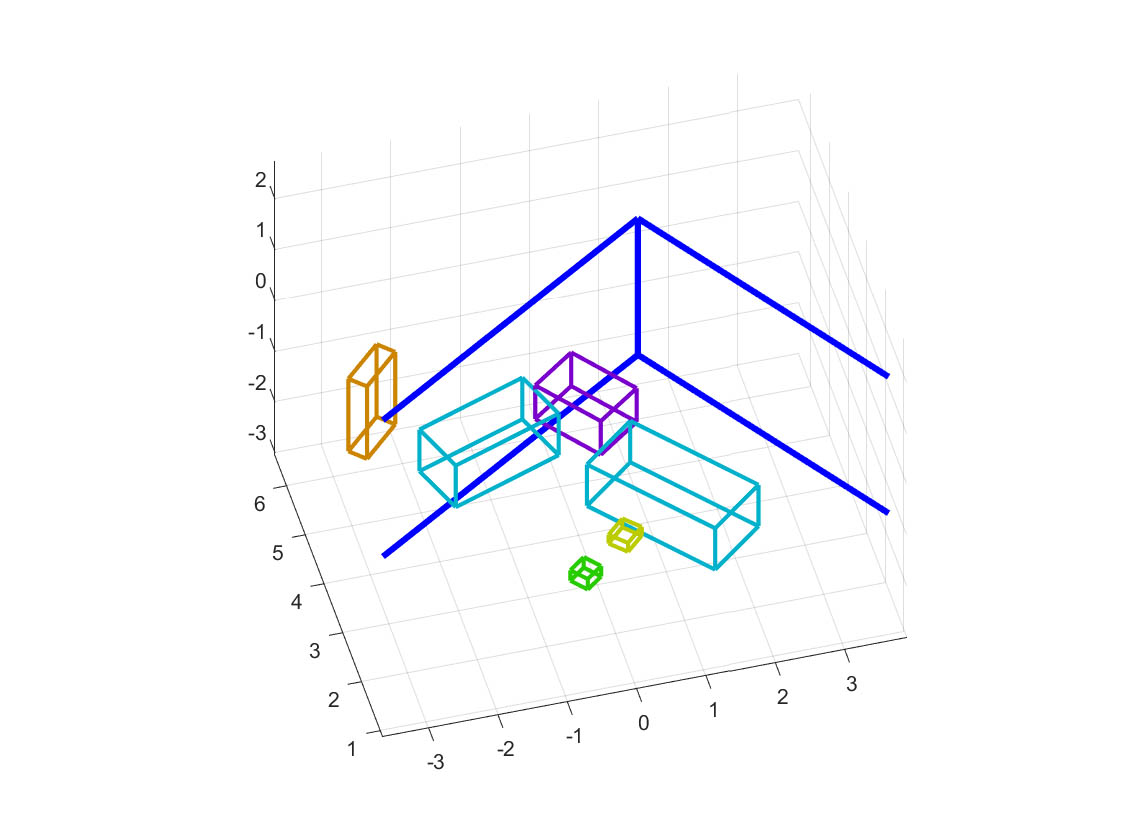} &
        	    \oblique{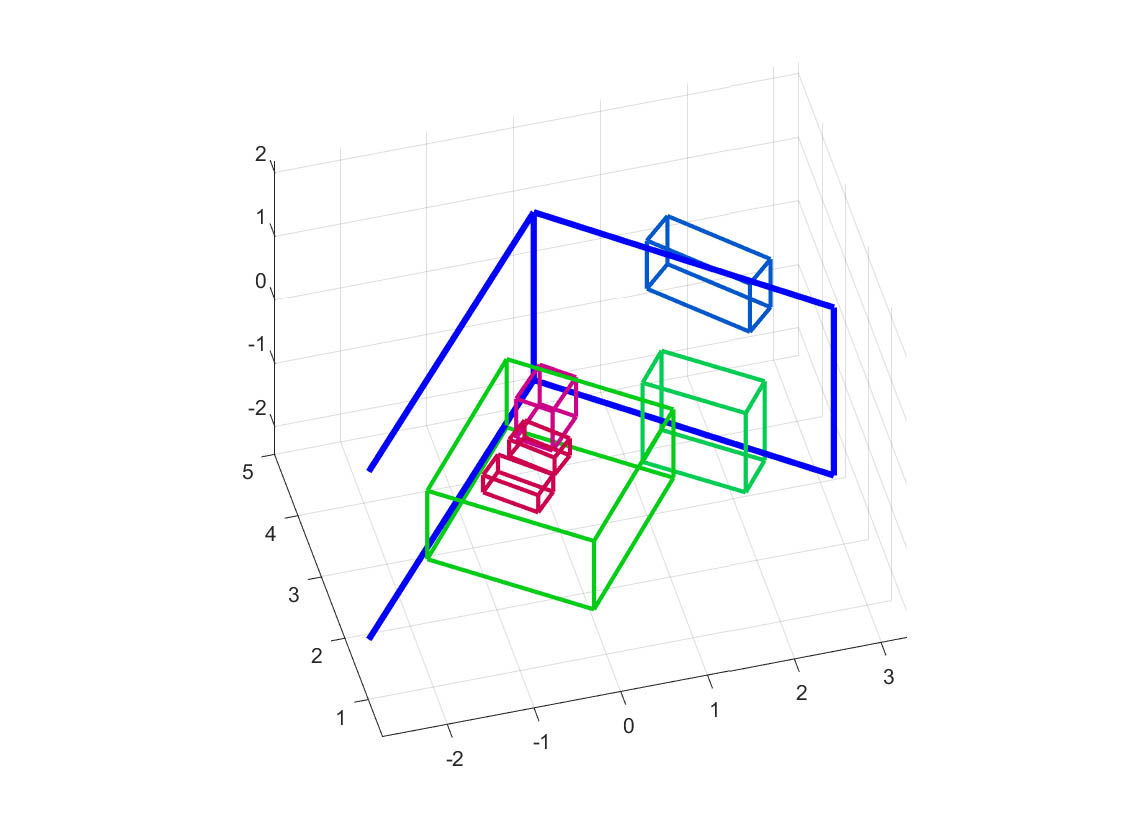} &
        	    \oblique{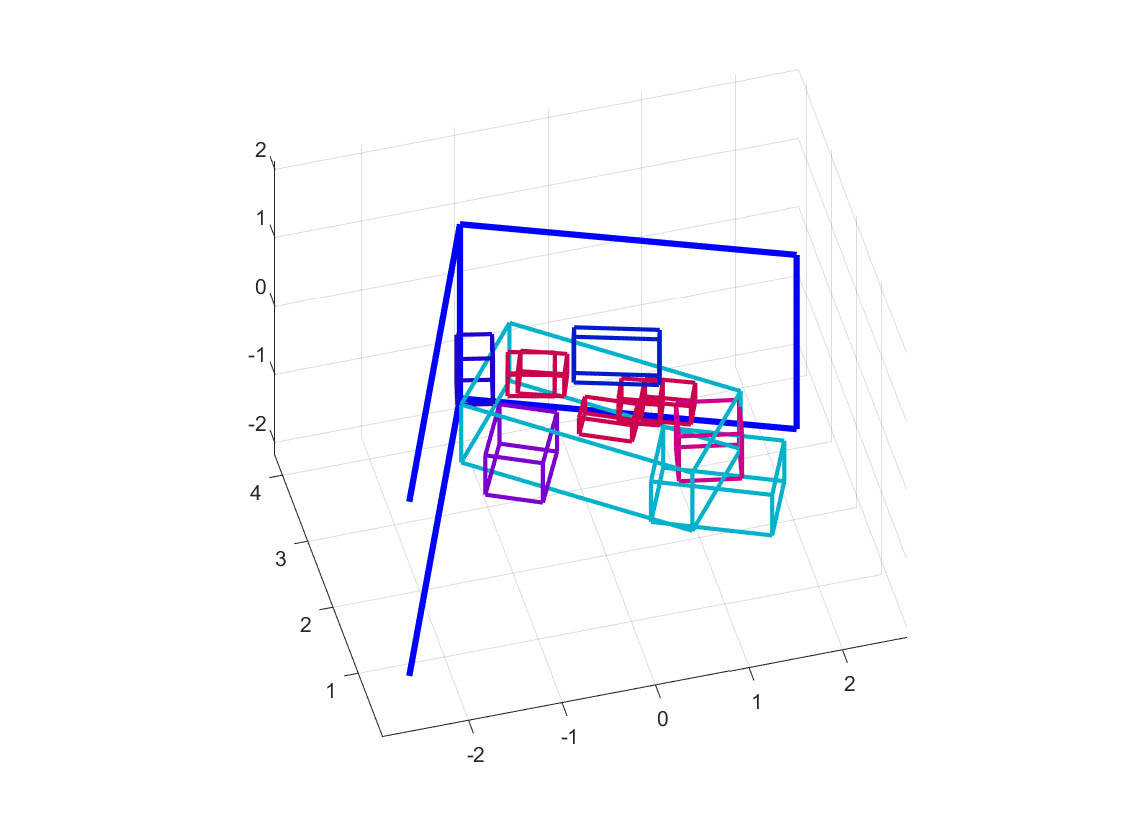} &
        	   % \oblique{figure/scnrecon/Total3D/845_oblique_3d} &
        	    \oblique{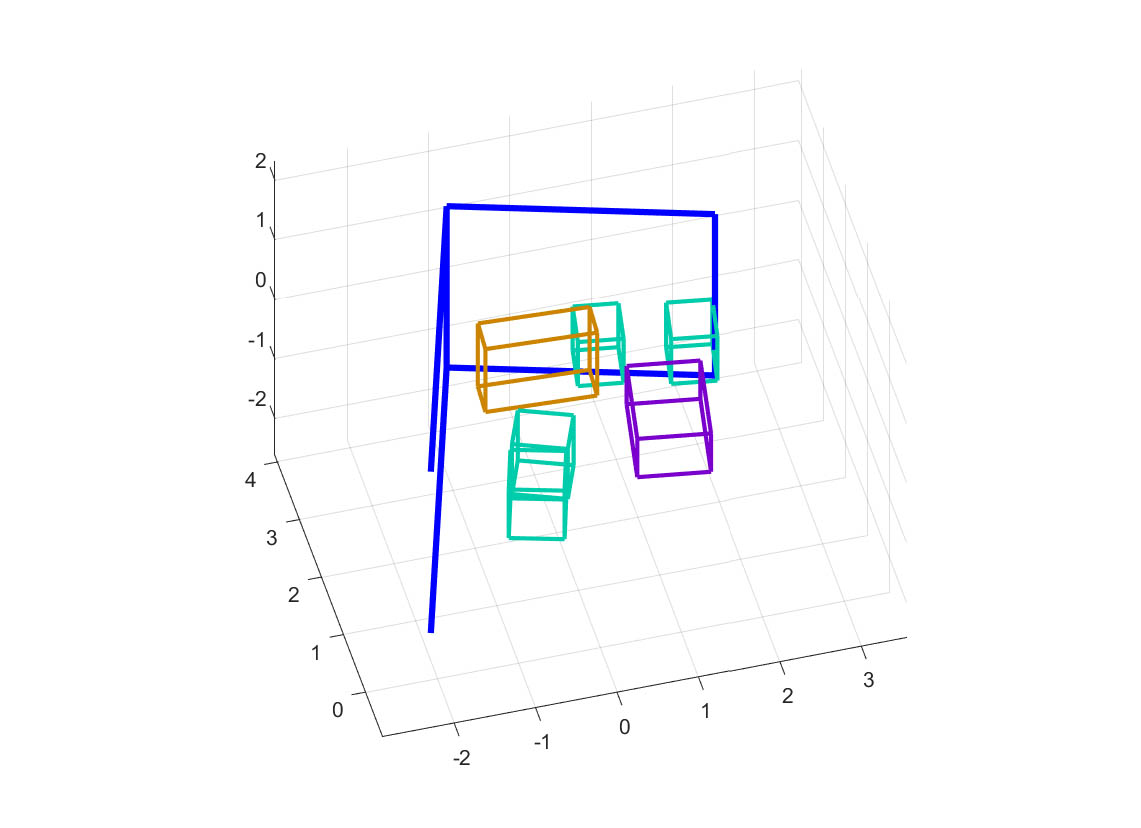} &
        	   % \oblique{figure/scnrecon/Total3D/2435_oblique_3d} &
        	    \oblique{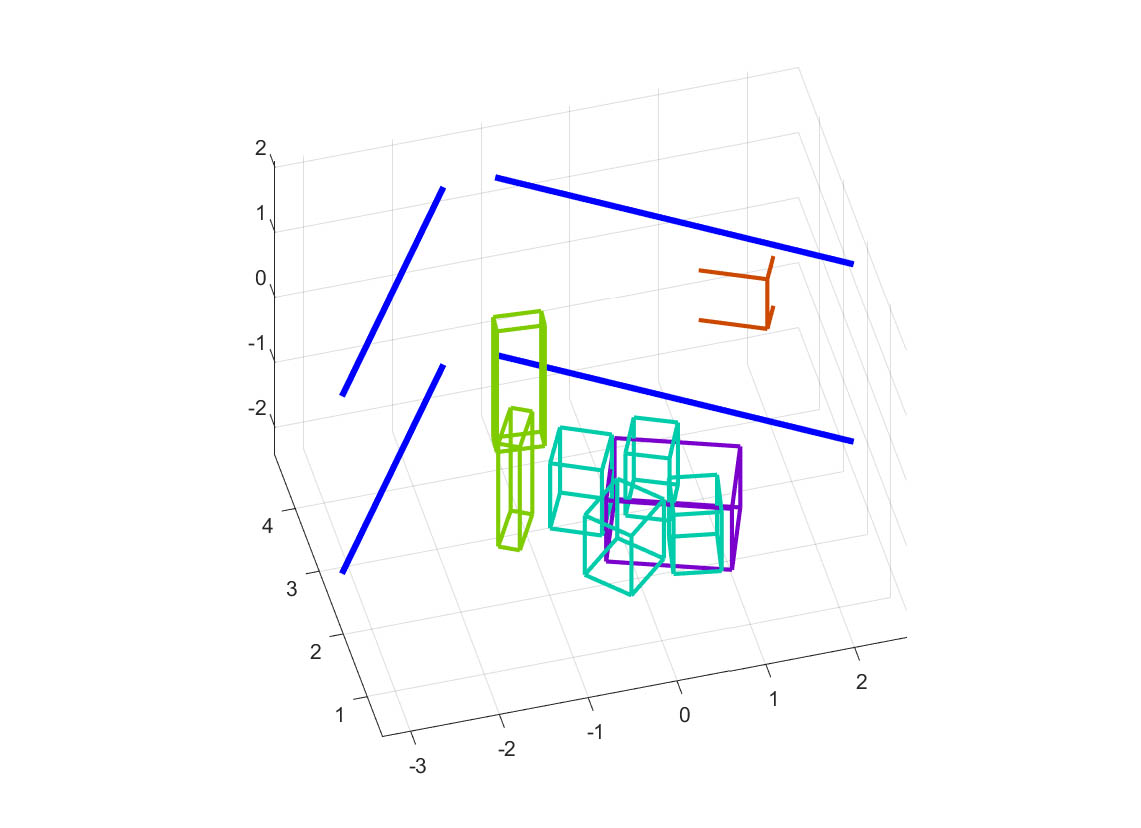} \\
        	\rot{Oblique View}& \rot{Ours}& 
        	    \oblique{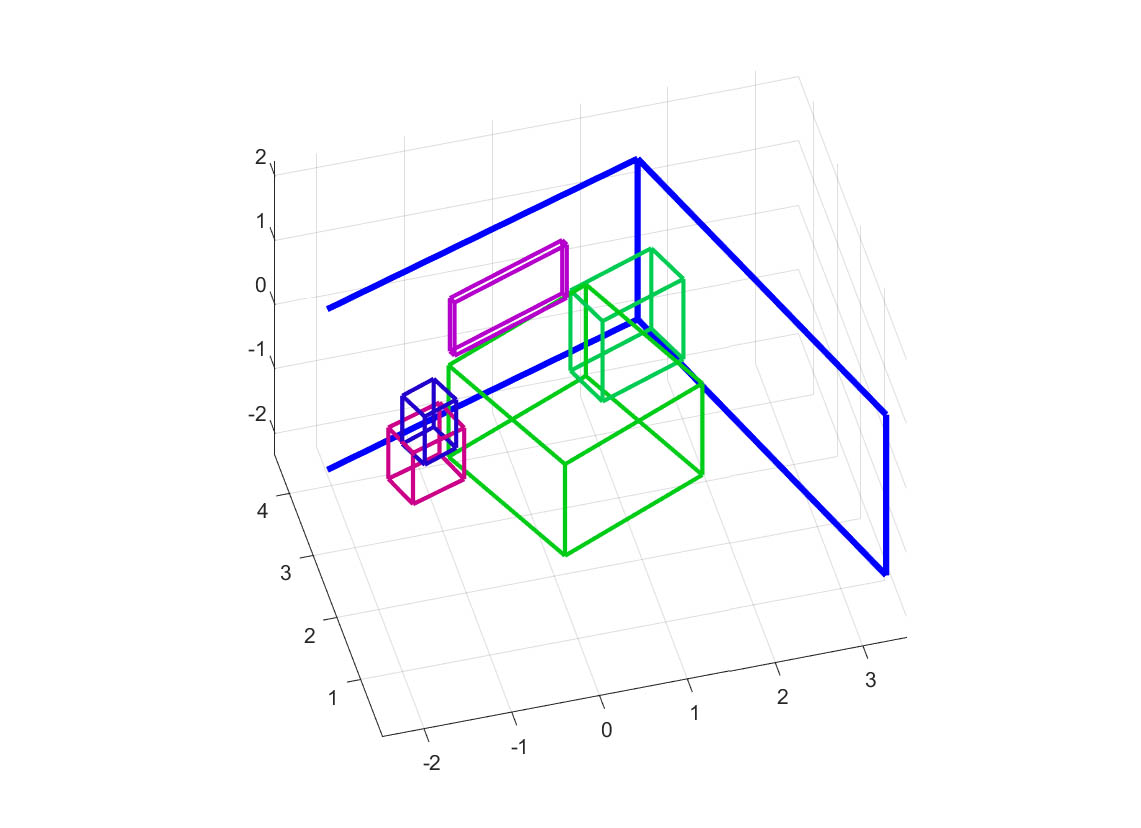} &
        	    \oblique{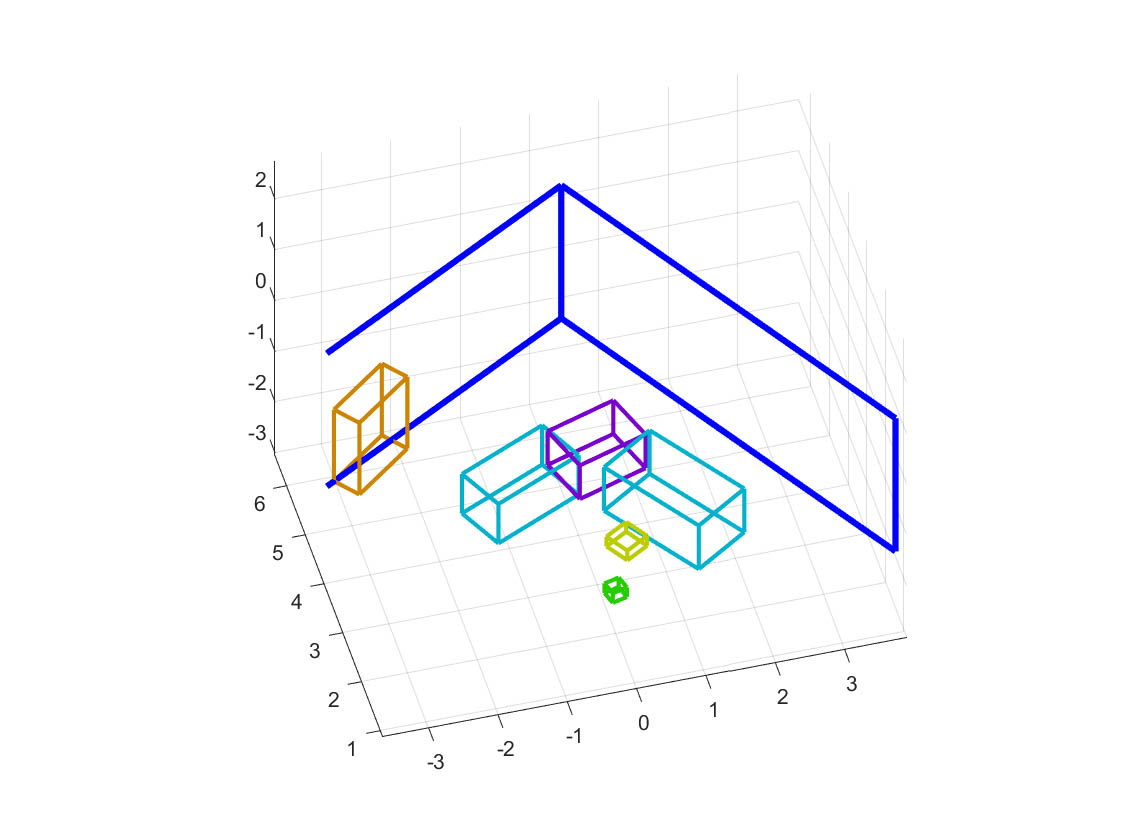} &
        	    \oblique{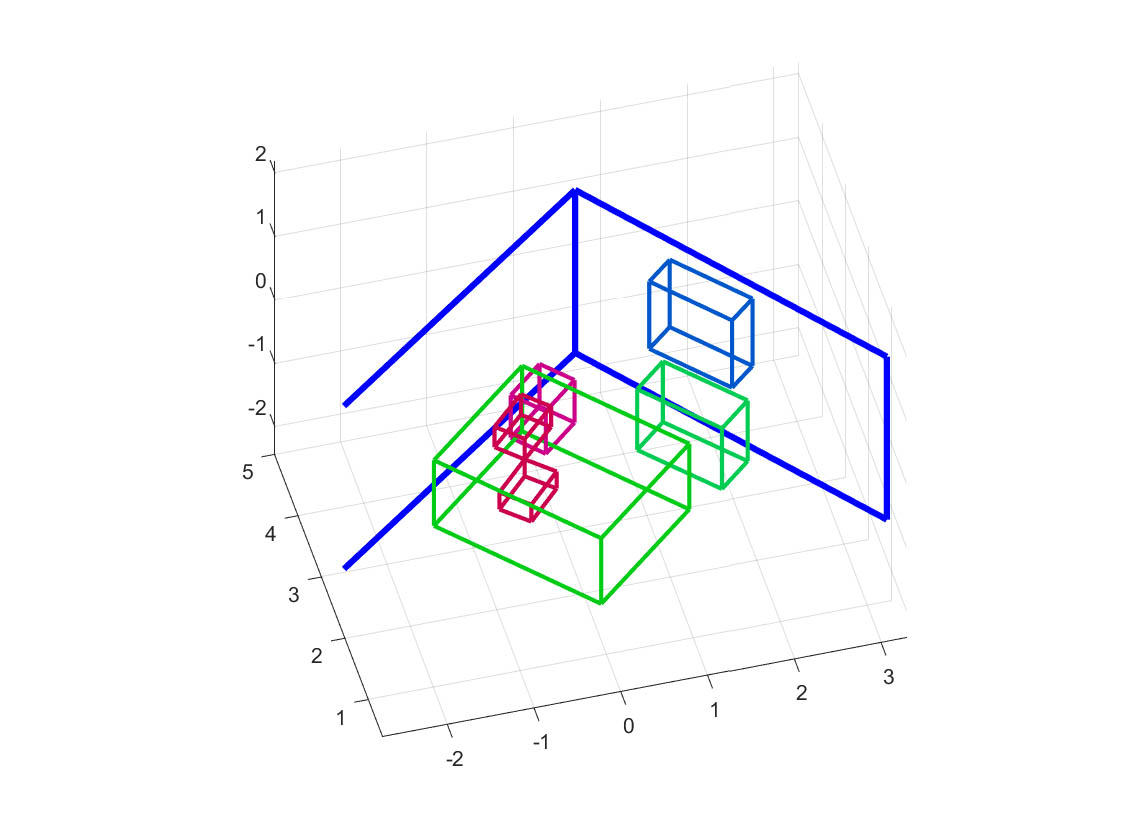} &
        	    \oblique{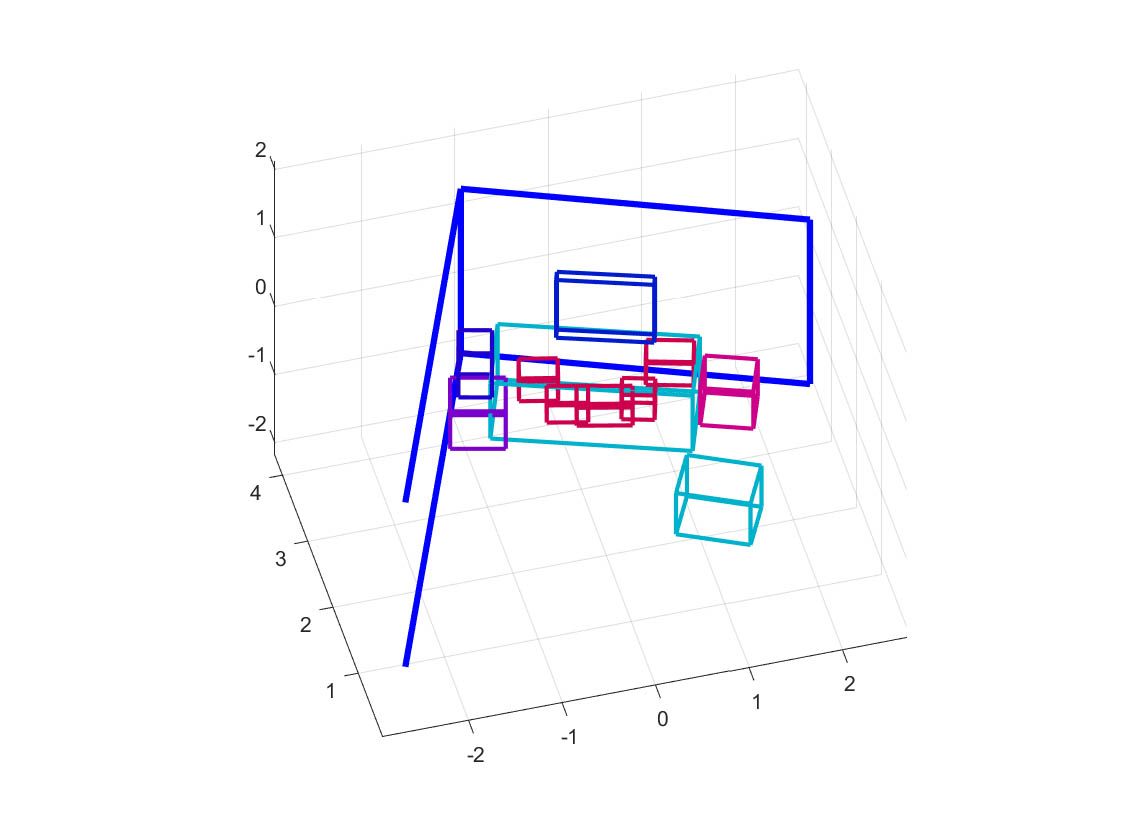} &
        	   % \oblique{figure/scnrecon/ours/845_oblique_3d} &
        	    \oblique{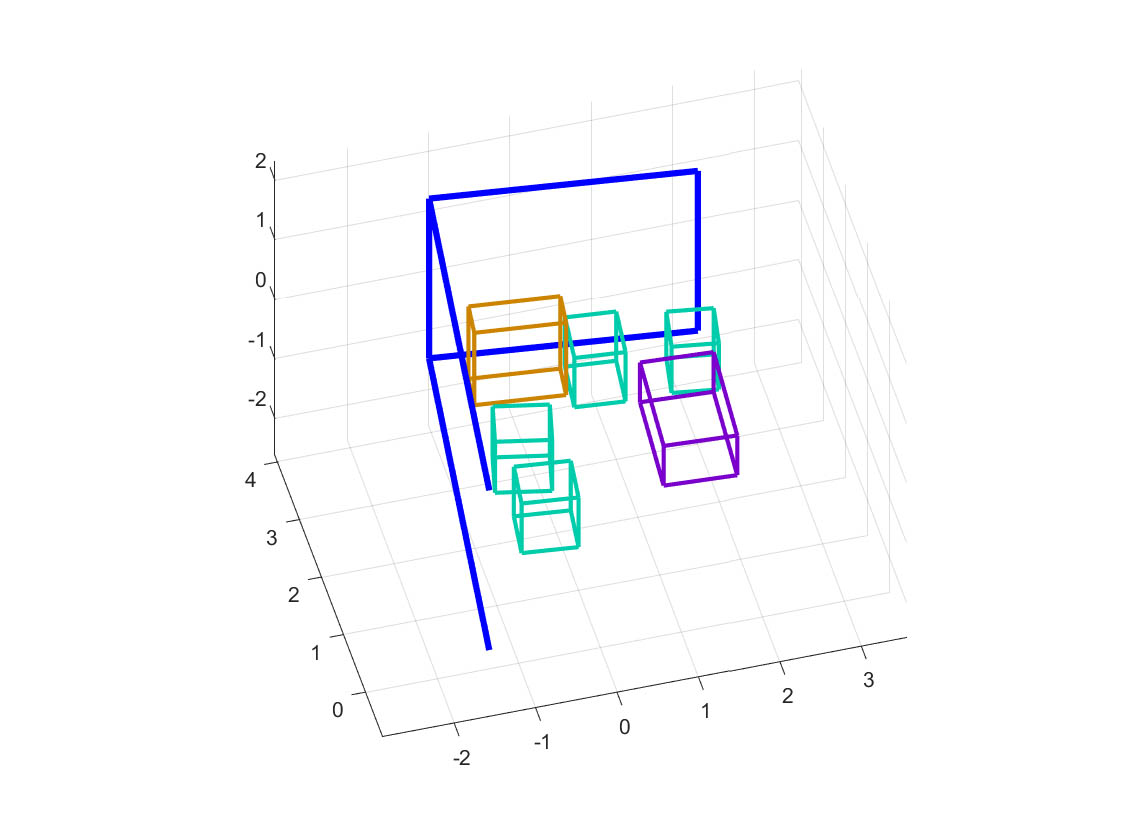} &
        	   % \oblique{figure/scnrecon/ours/2435_oblique_3d} &
        	    \oblique{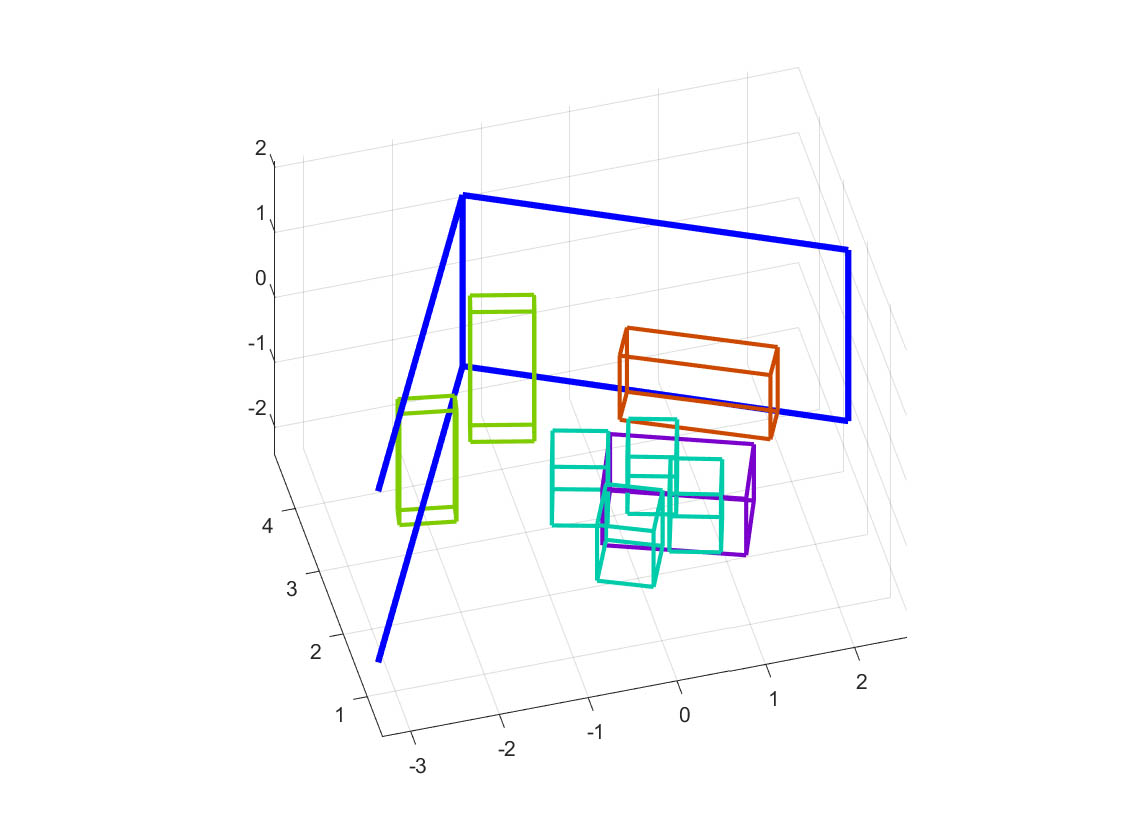} \\
        	&\rot{GT}& 
        	    \oblique{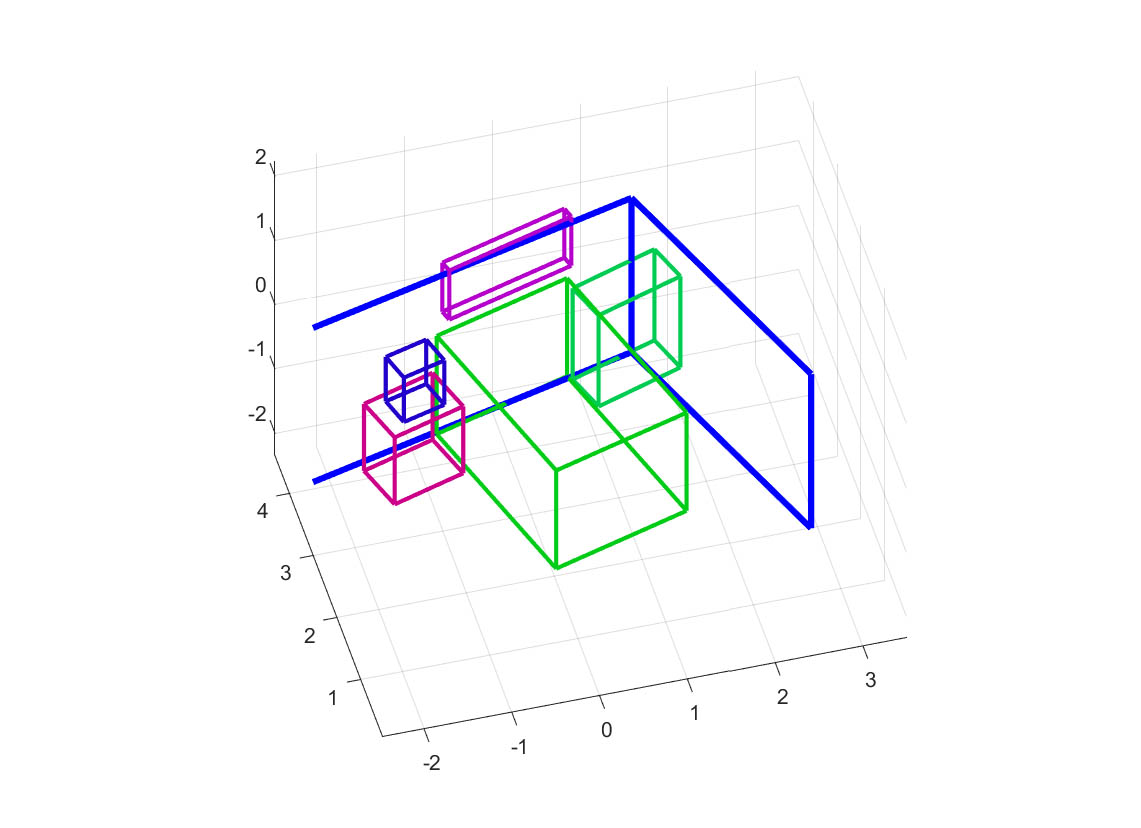} &
        	    \oblique{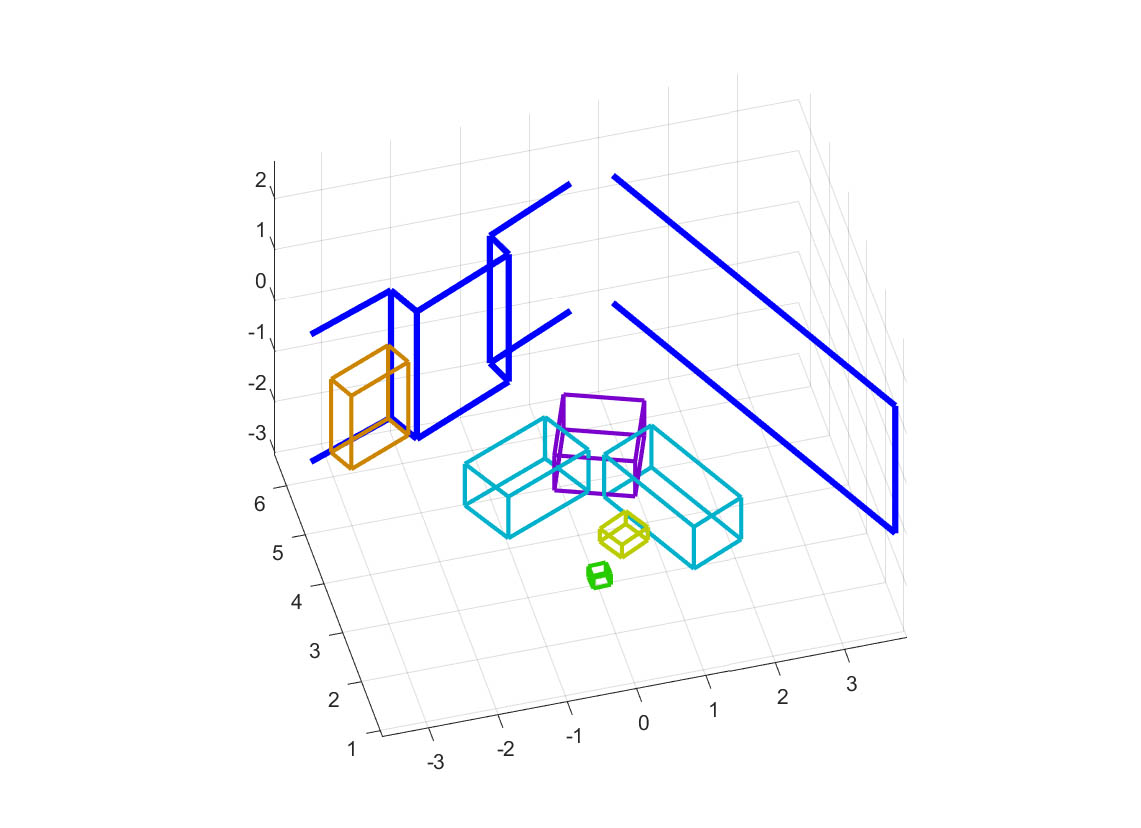} &
        	    \oblique{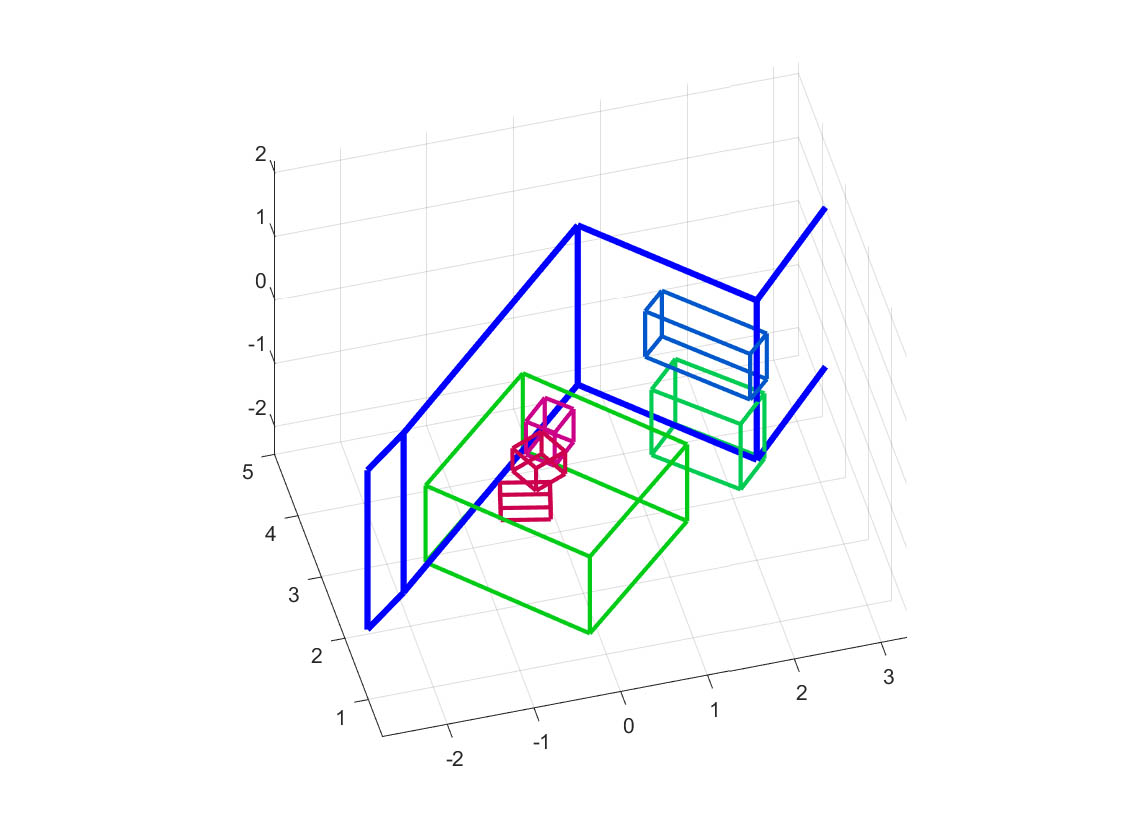} &
        	    \oblique{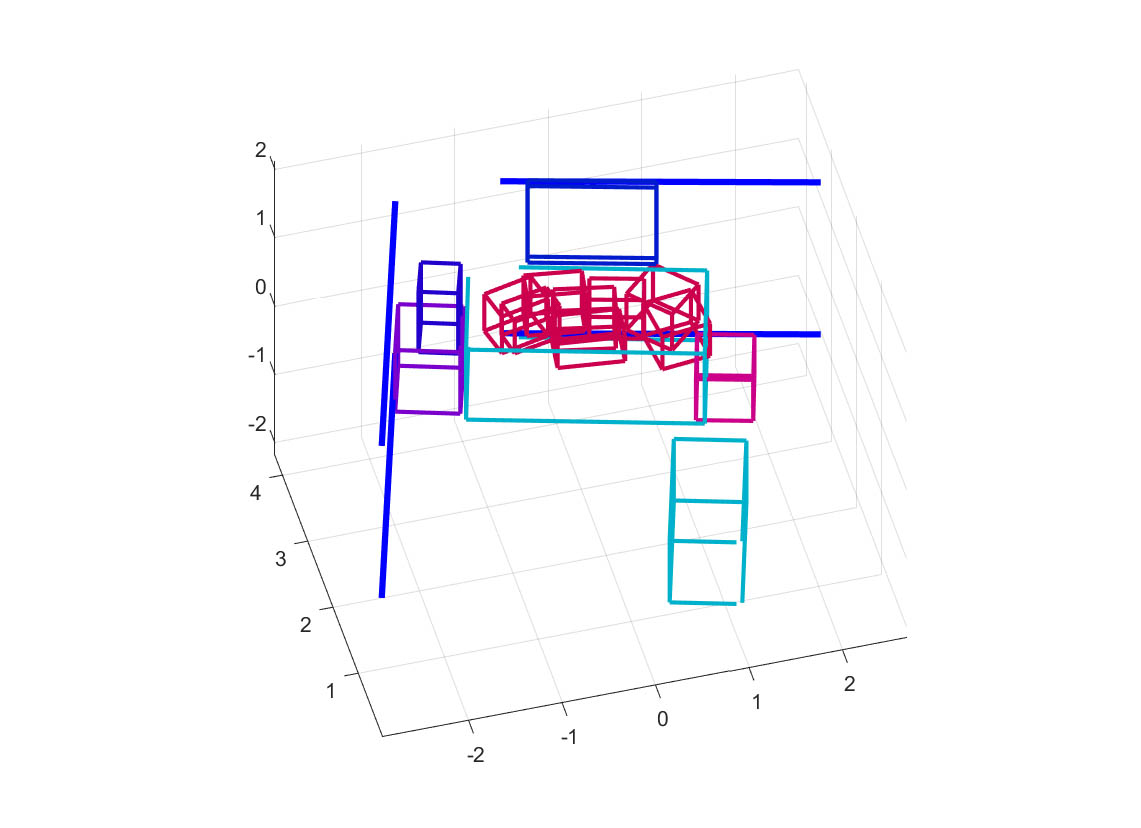} &
        	   % \oblique{figure/scnrecon/gt/845_oblique_3d} &
        	    \oblique{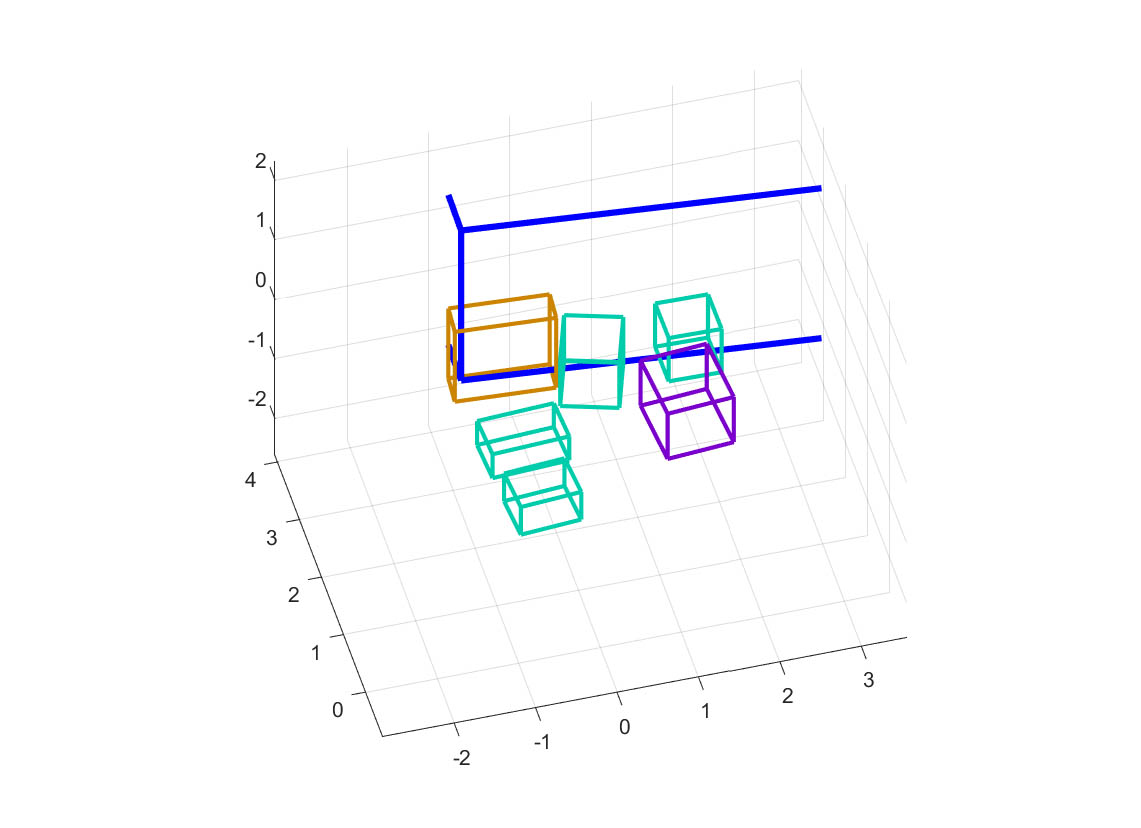} &
        	   % \oblique{figure/scnrecon/gt/2435_oblique_3d} &
        	    \oblique{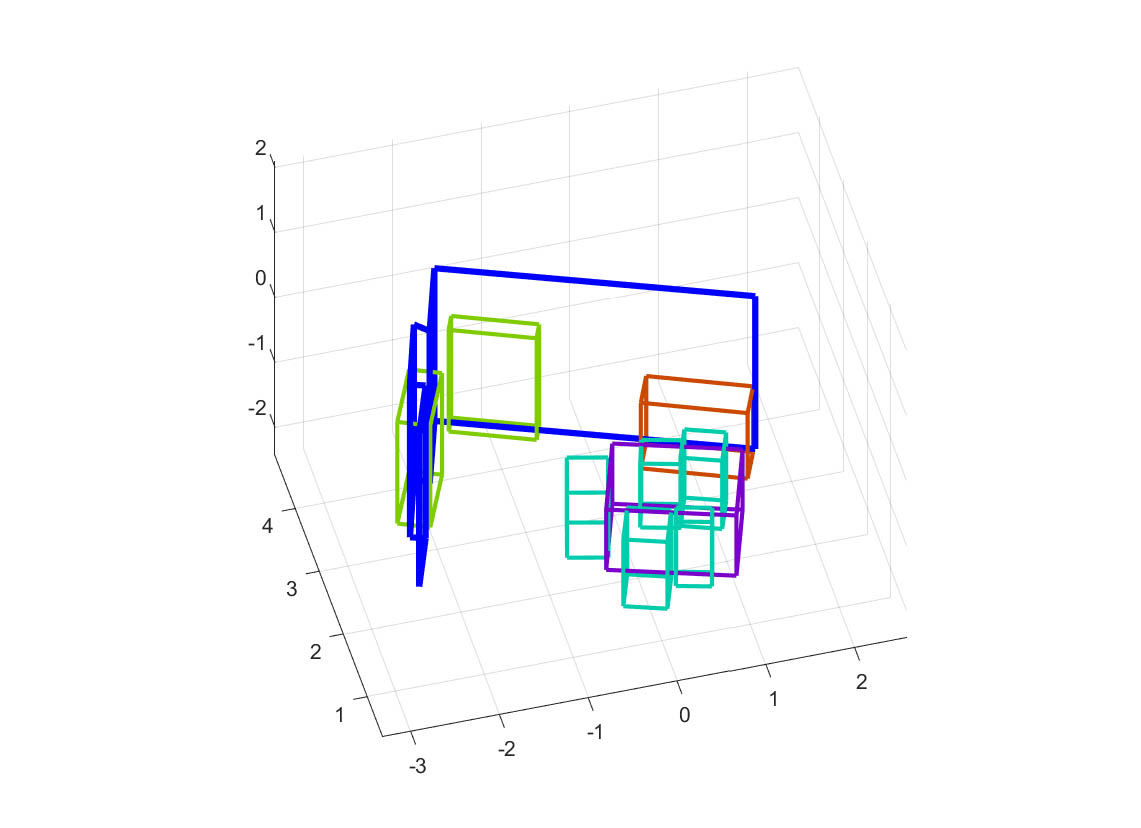} \\
    	\hline
    	    &\rot{Total3D}& 
        	    \rgb{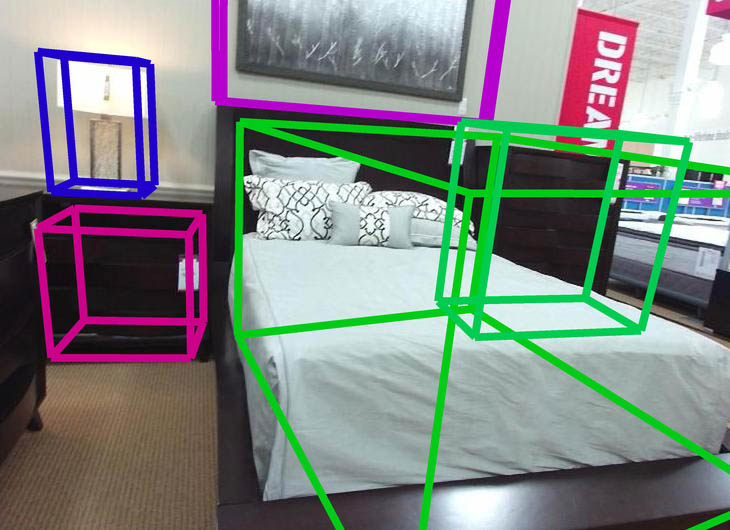} &
        	    \rgb{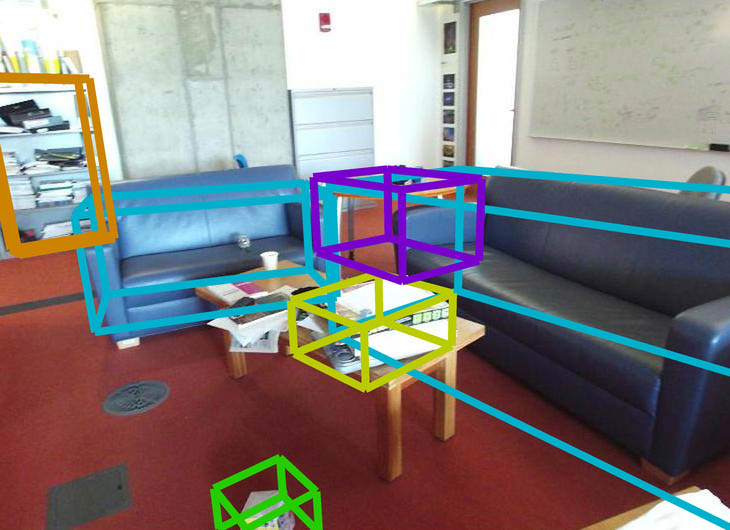} &
        	    \rgb{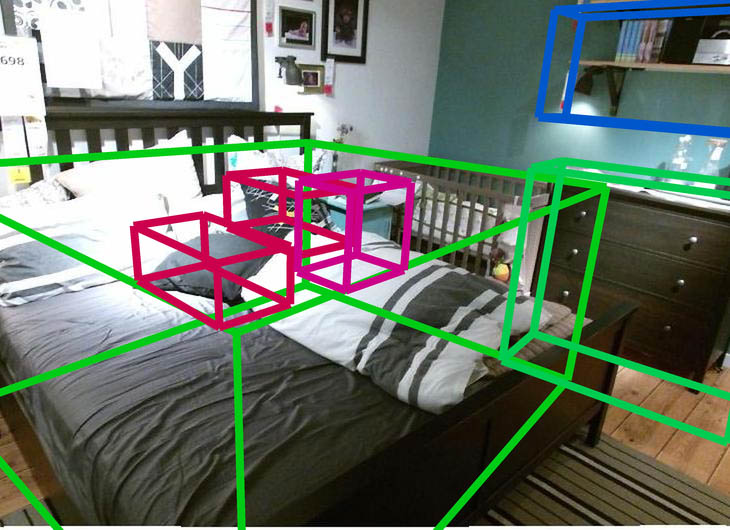} &
        	    \rgb{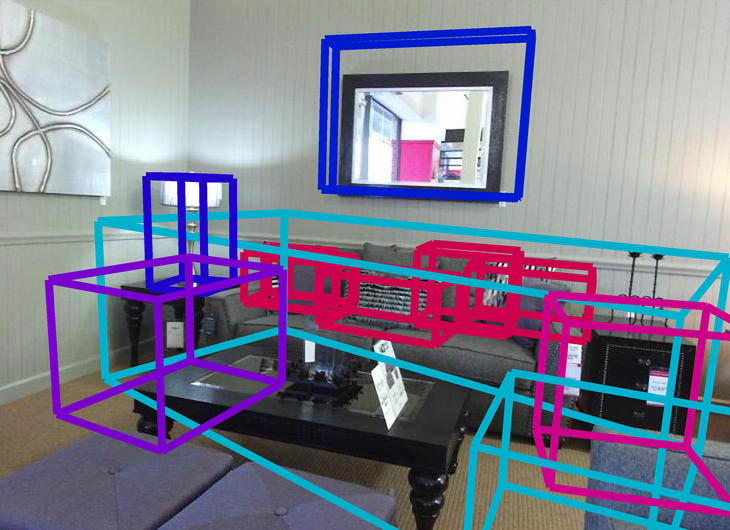} &
        	   % \rgb{figure/scnrecon/Total3D/845_bbox} &
        	    \rgb{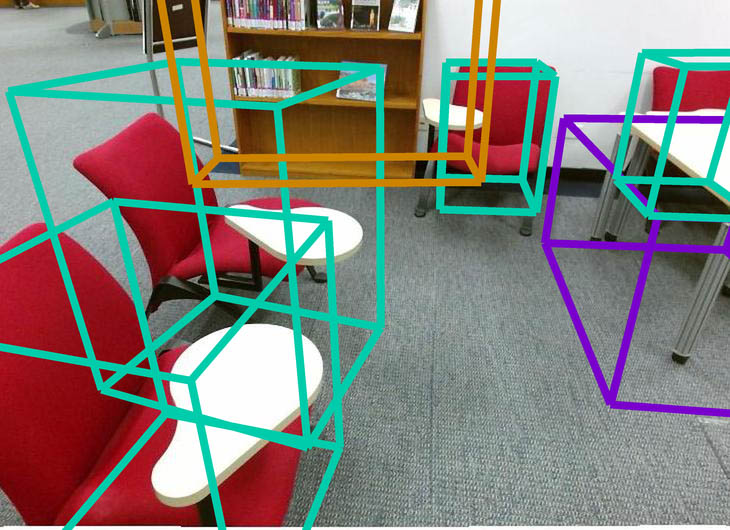} &
        	   % \rgb{figure/scnrecon/Total3D/2435_bbox} &
        	    \rgb{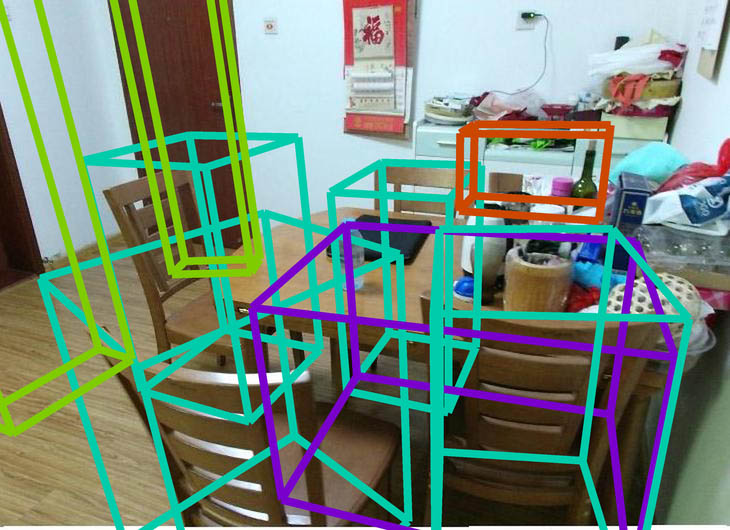} \\
            \rot{Camera View}&\rot{Ours}& 
        	    \rgb{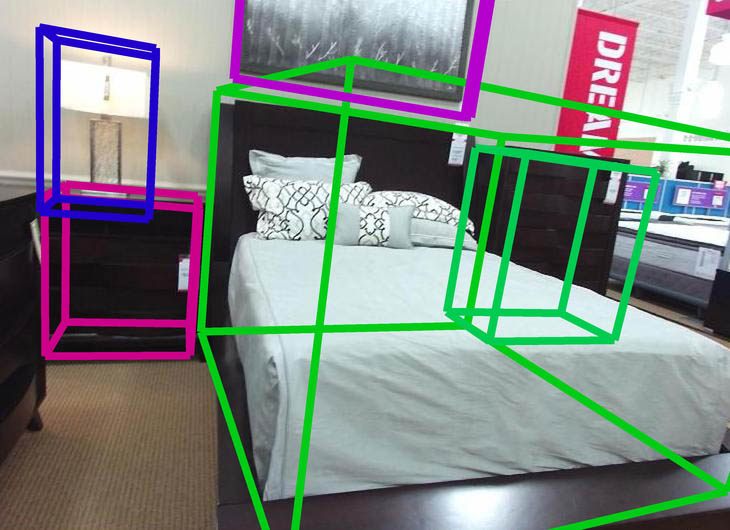} &
        	    \rgb{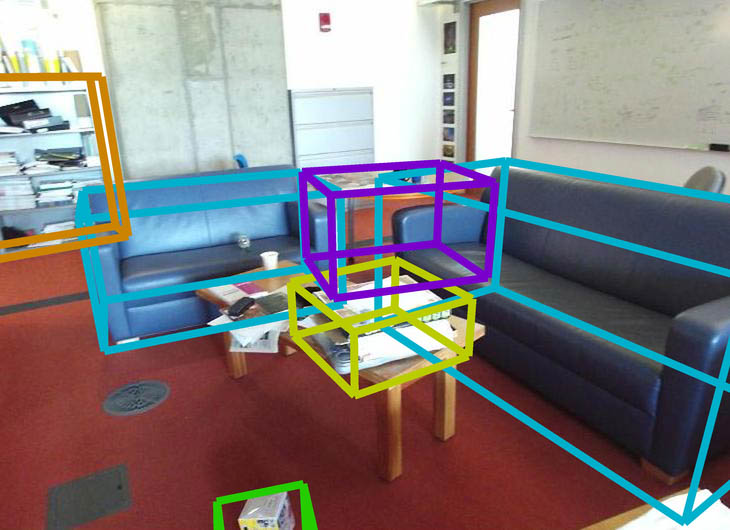} &
        	    \rgb{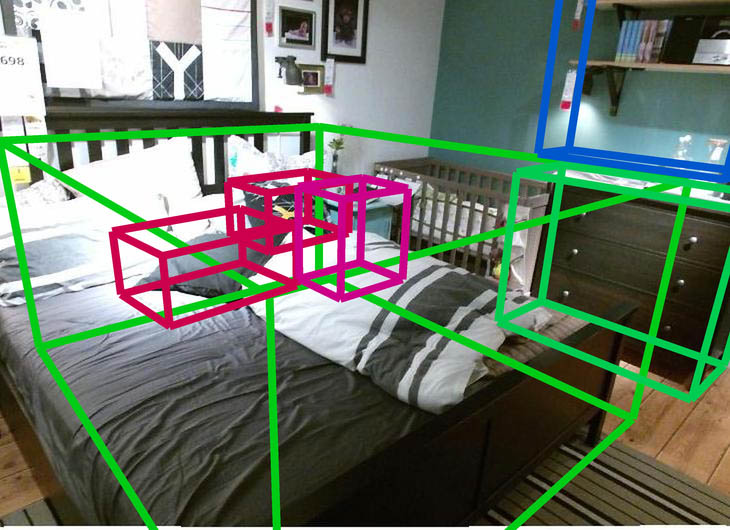} &
        	    \rgb{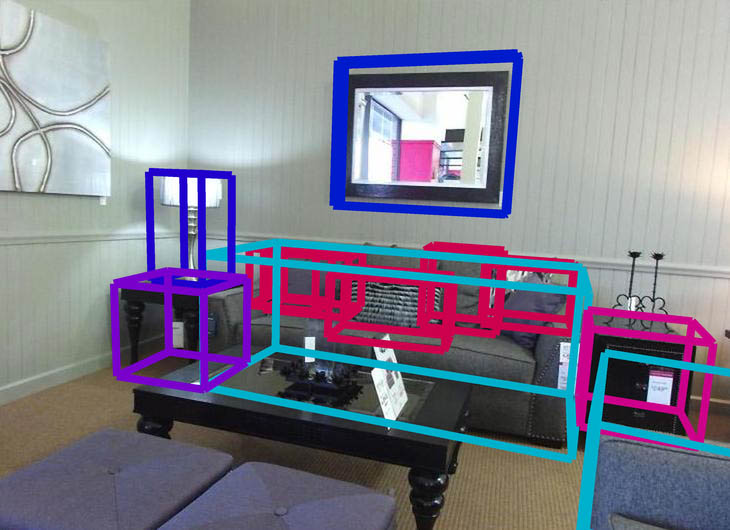} &
        	   % \rgb{figure/scnrecon/ours/845_bbox} &
        	    \rgb{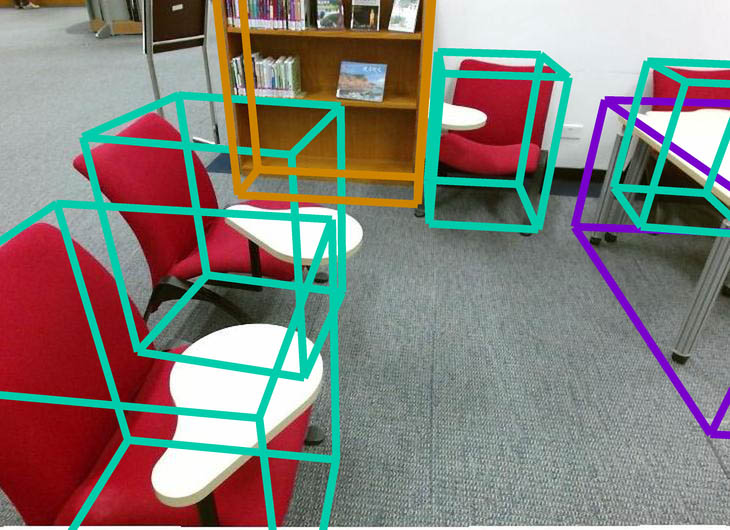} &
        	   % \rgb{figure/scnrecon/ours/2435_bbox} &
        	    \rgb{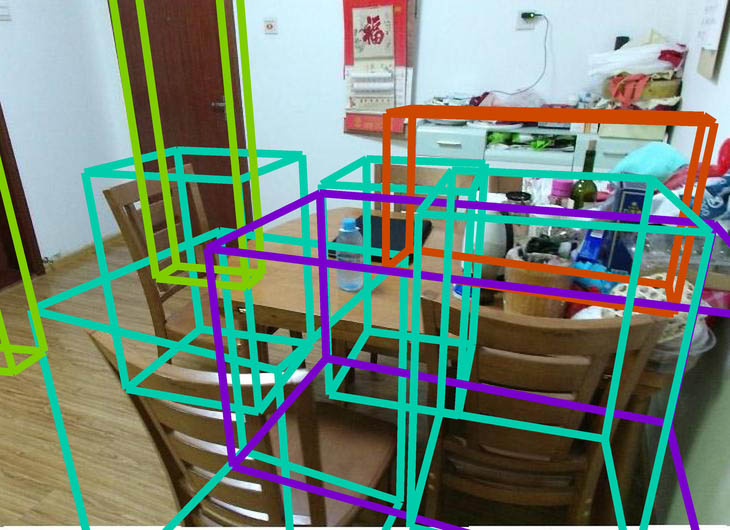} \\
        	&\rot{GT}& 
        	    \rgb{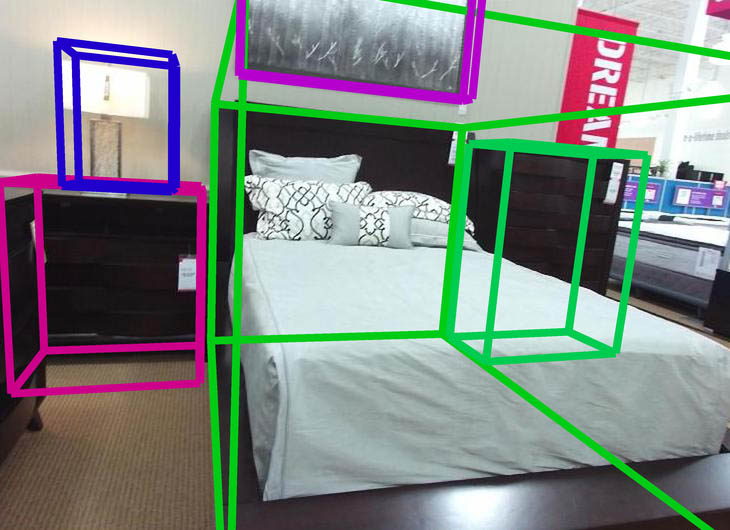} &
        	    \rgb{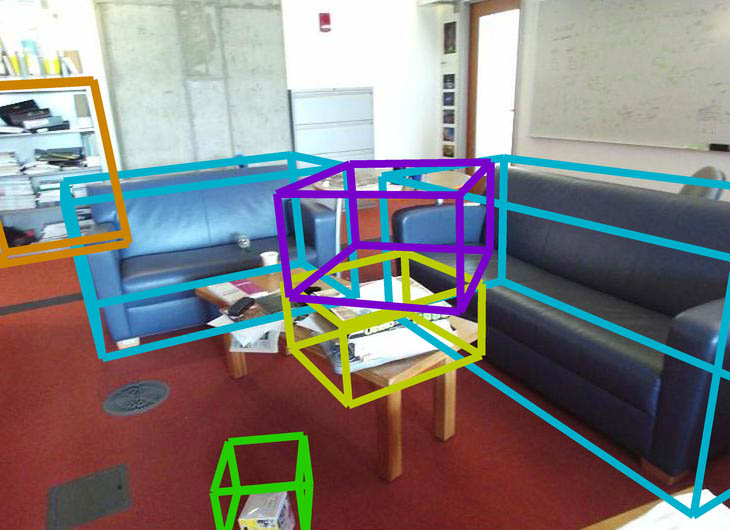} &
        	    \rgb{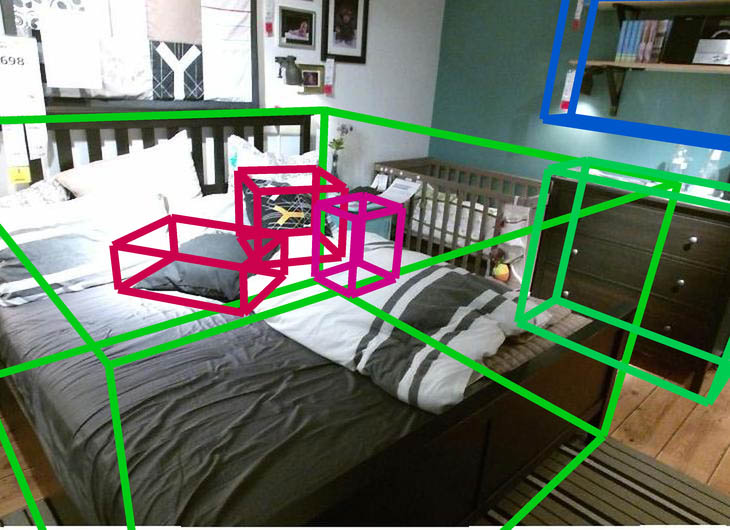} &
        	    \rgb{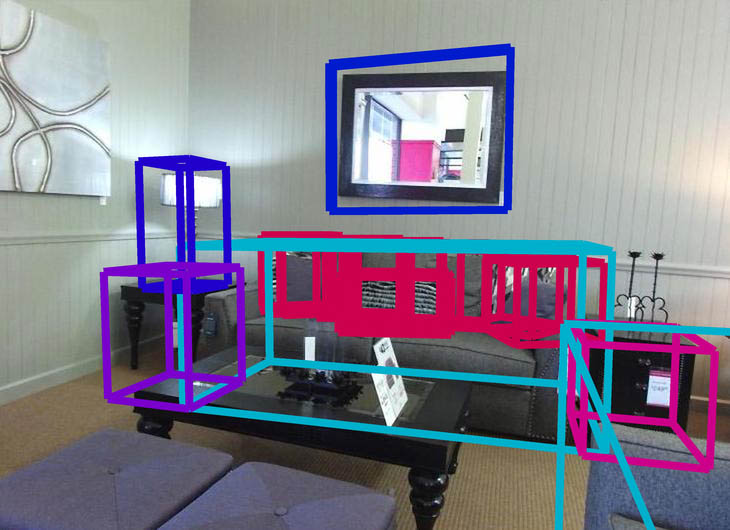} &
        	   % \rgb{figure/scnrecon/gt/845_bbox} &
        	    \rgb{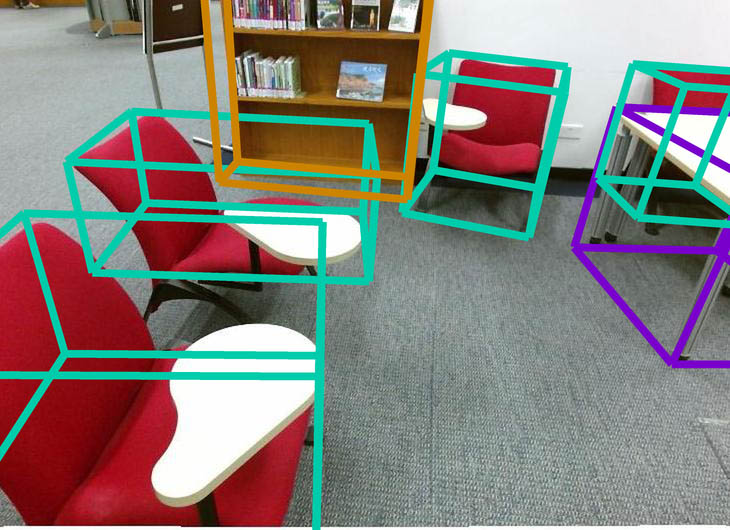} &
        	   % \rgb{figure/scnrecon/gt/2435_bbox} &
        	    \rgb{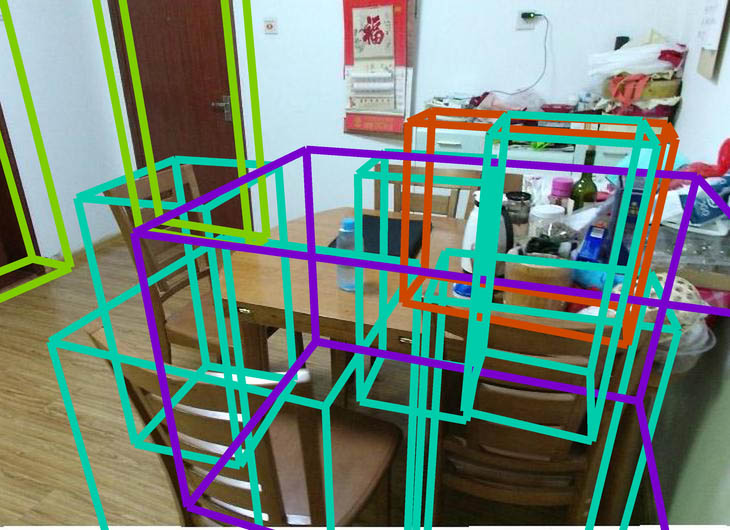} \\
    	\hline
    	\multirow{2}{*}{\rot{Scene Reconstruction}}
        	&\rot{Total3D}& 
        	    \rgb{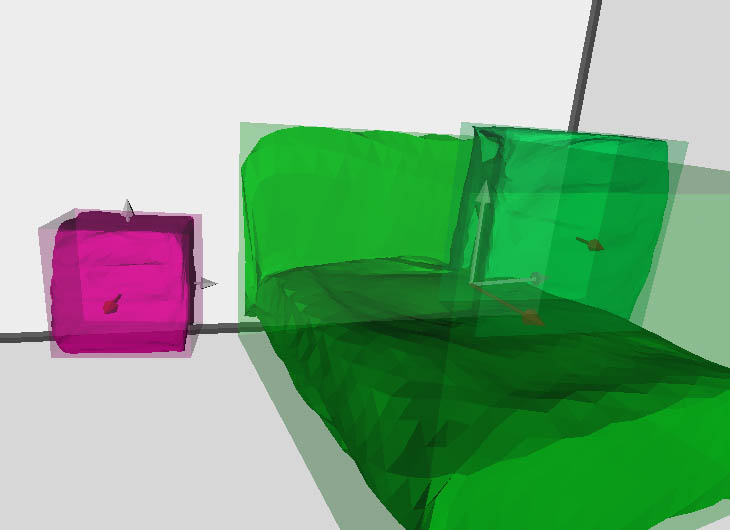} &
        	    \rgb{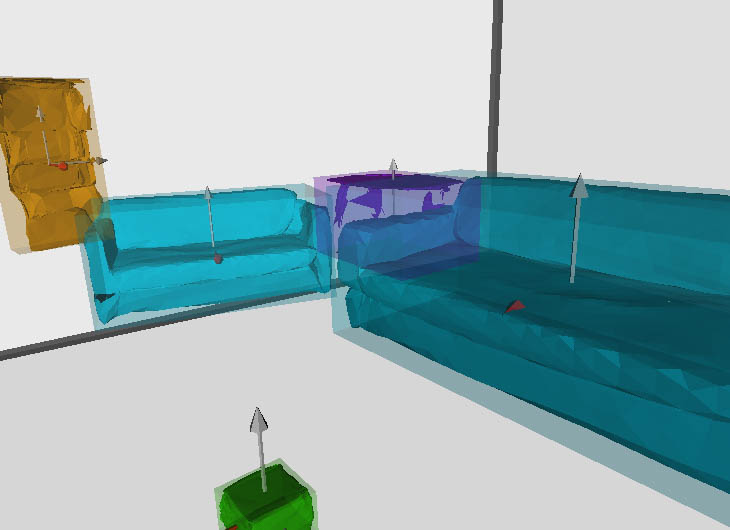} &
        	    \rgb{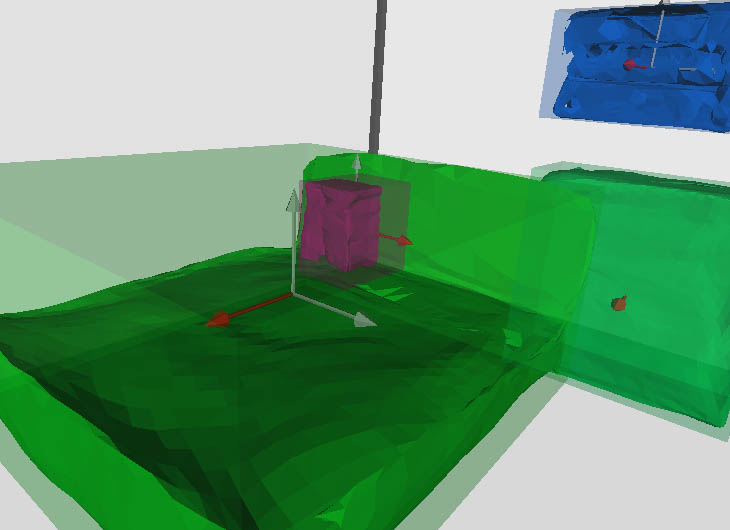} &
        	    \rgb{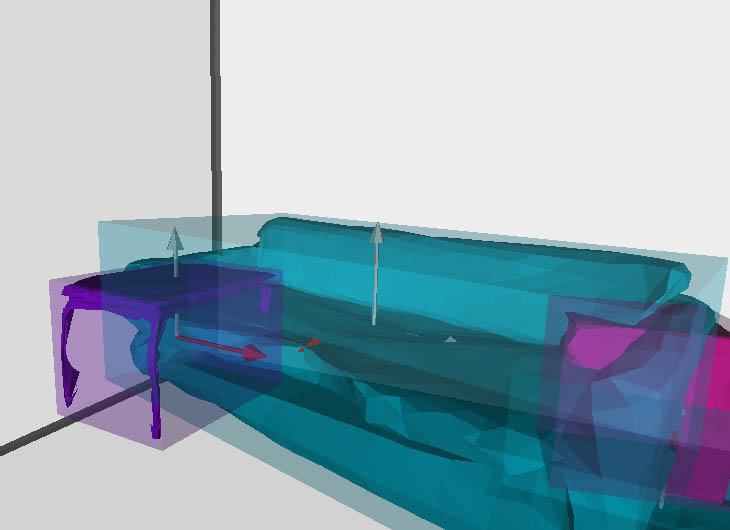} &
        	   % \rgb{figure/scnrecon/Total3D/845_recon} &
        	    \rgb{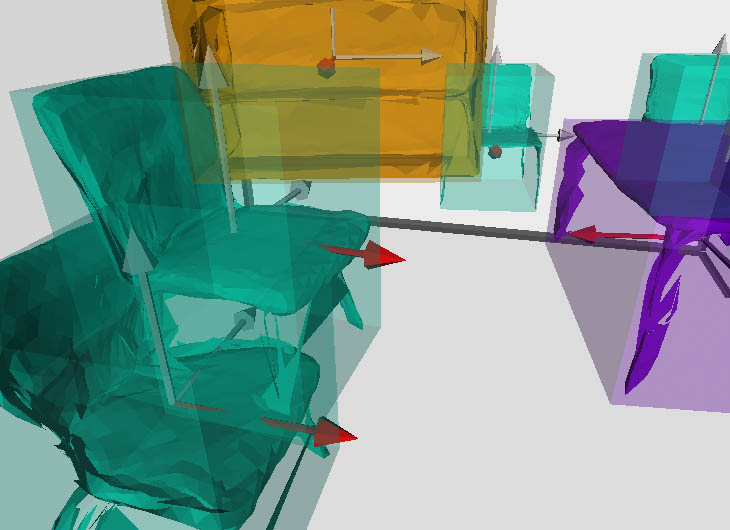} &
        	   % \rgb{figure/scnrecon/Total3D/2435_recon} &
        	    \rgb{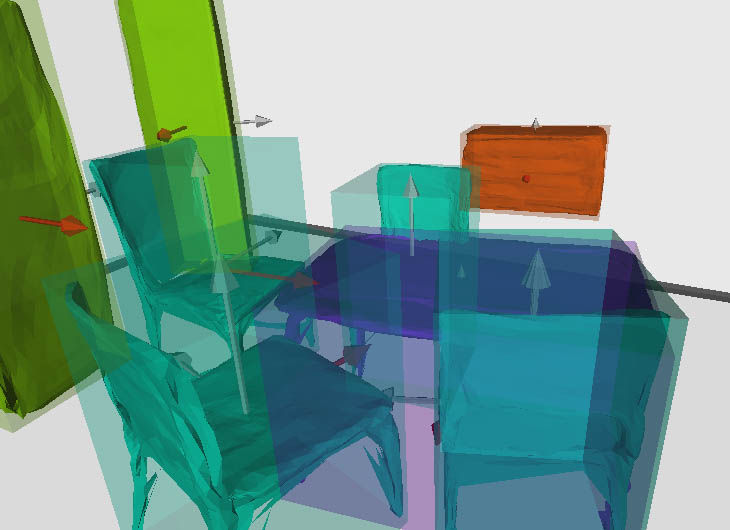} \\
        	&\rot{Ours}& 
        	    \rgb{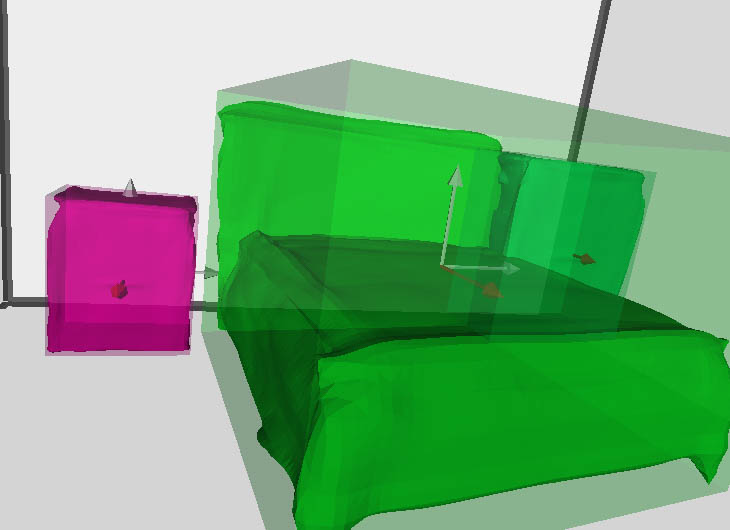} &
        	    \rgb{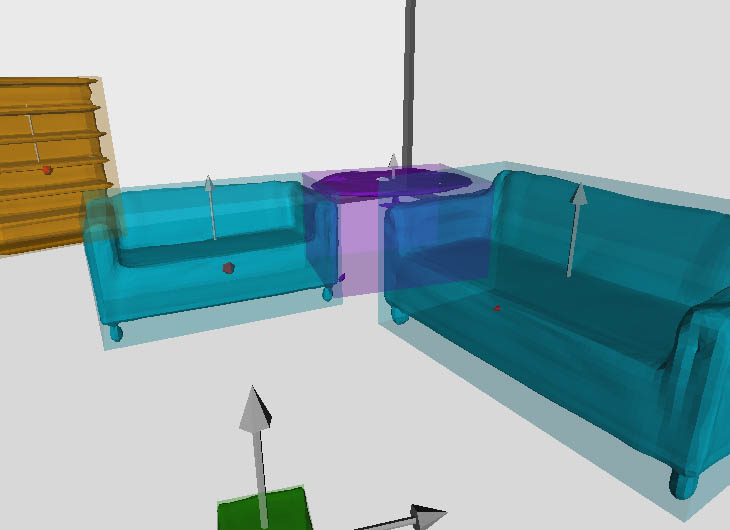} &
        	    \rgb{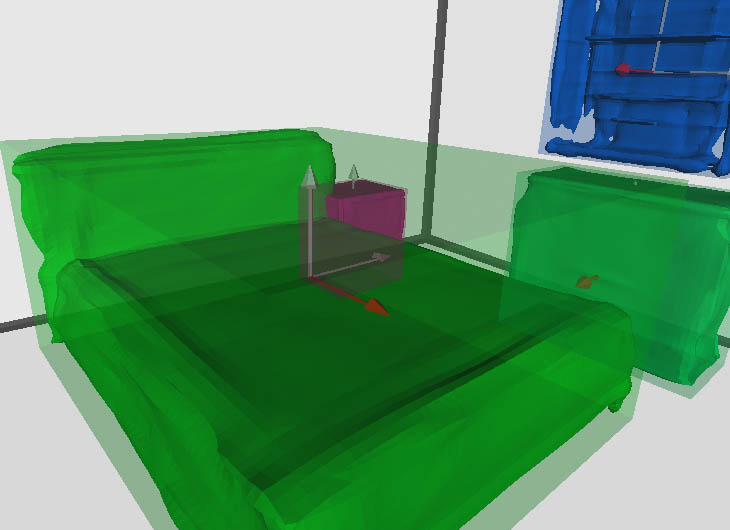} &
        	    \rgb{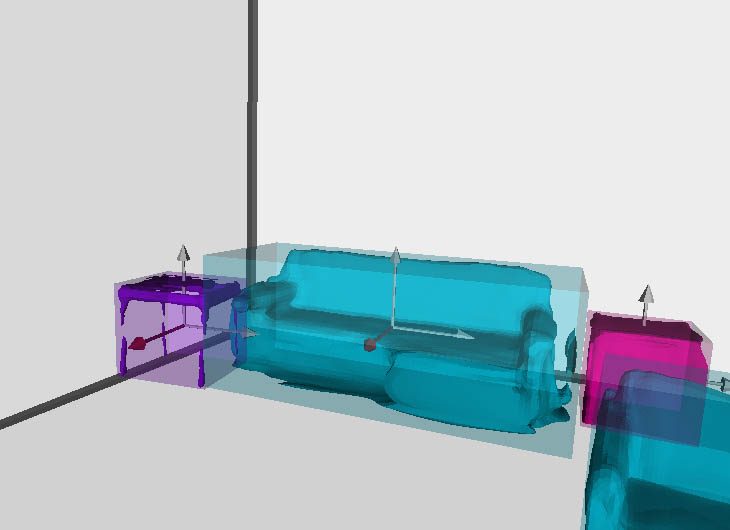} &
        	   % \rgb{figure/scnrecon/ours/845_recon} &
        	    \rgb{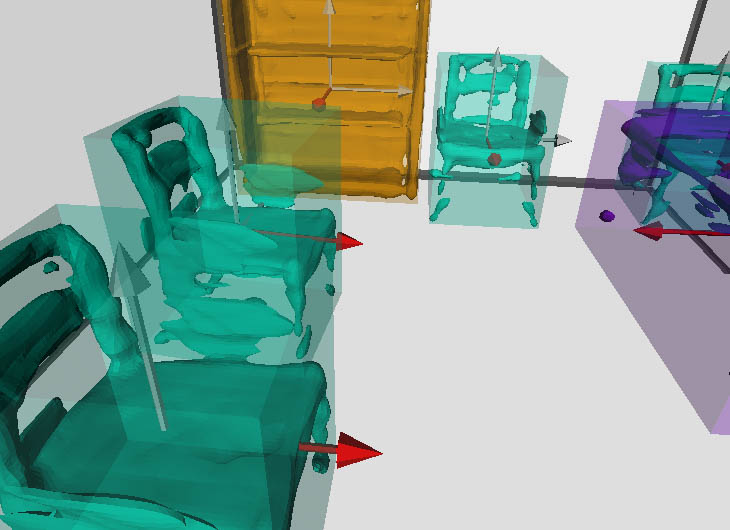} &
        	   % \rgb{figure/scnrecon/ours/2435_recon} &
        	    \rgb{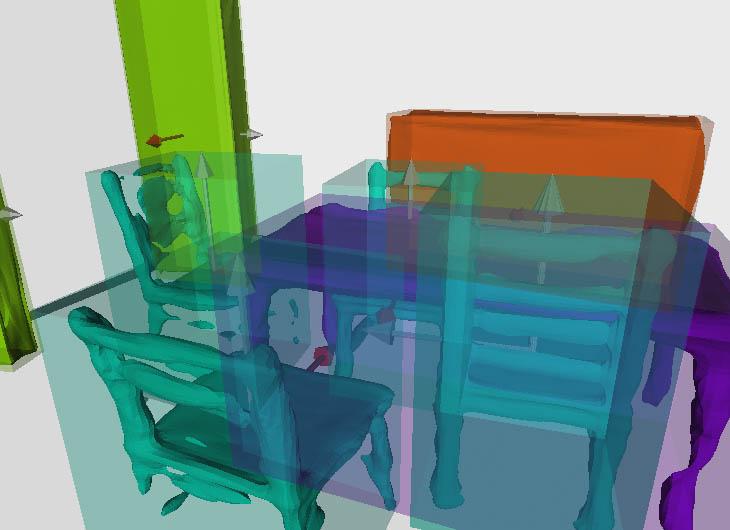} \\
    \end{tabular}
    % \vspace{-1.0em}
	\caption{Qualitative comparison on object detection and scene reconstruction. We compare object detection results with Total3D \cite{nie2020total3dunderstanding} and ground truth in both oblique view and camera view. The results show that our method gives more accurate bounding box estimation and with less intersection. We compare scene reconstruction results with Total3D in camera view and observe more reasonable object poses.}
	\label{fig:scnrecon}
\end{figure*}

\noindent \textbf{3D Object Detection.} 
We then evaluate the 3D object detection performance of our model.
Follow \cite{nie2020total3dunderstanding,huang2018cooperative}, we use mean average precision (mAP) with the threshold of 3D bounding box IoU set at 0.15 as the evaluation metric.
The quantitative comparison to state-of-the-art methods \cite{choi2013understanding,huang2018holistic,huang2018cooperative,nie2020total3dunderstanding} is shown in \autoref{tbl:3ddetection}.
Our method performs consistently the best over all semantic categories and significantly outperforms the state-of-the-art (i.e. improving AP by 18.83\%).
\Autoref{fig:scnrecon} shows some qualitative comparison.
Note how our method produces object layout not only more accurate but also in reasonable context compared to Total3D, e.g. objects are parallel to wall direction.

% We follow \cite{nie2020total3dunderstanding,huang2018cooperative} to use mean average precision (mAP) defined in \cite{song2015sun}. \cite{huang2018cooperative} proposed to set the threshold of IoU from 0.25 to 0.15 considering no depth map is used in this task. 
% We adopt their evaluation setting for a fair comparison.
% We make a quantitative comparison in \autoref{tbl:3ddetection} with \cite{choi2013understanding,huang2018holistic,huang2018cooperative,nie2020total3dunderstanding} on the same object categories shared by them. 
% The result shows that our method not only outperforms previous work on average precision of every single category, but also outperforms the state-of-the-art\cite{nie2020total3dunderstanding} by 18.83\% on mAP. 
% We also make a qualitative comparison in \autoref{fig:scnrecon}. 
% By comparing ours with Total3D and ground truth in both oblique and camera view, we observe a significant improvement on position accuracy and size estimation.

\noindent \textbf{Layout Estimation.} 
We also compare the 3D room layout estimation with Total3D \cite{nie2020total3dunderstanding} and other state-of-the-arts \cite{choi2013understanding,huang2018holistic,huang2018cooperative}.
The quantitative evaluation is shown in \autoref{tbl:layout} (Layout IoU).
Overall, our method outperforms all the baseline methods.
This indicates that the GCN is effective in measuring the relation between layout and objects and thus benefits the layout prediction.

% Our model is compared with Total3D \cite{nie2020total3dunderstanding} and existing layout estimation works \cite{choi2013understanding,huang2018holistic,huang2018cooperative} quantitatively in \autoref{tbl:layout}. 
% The results show that our model outperforms Total3D by 5.2\%. 
% Although we model layout node in the same graph with other objects and update their representation and relationship with the same method, which simplifies the problem by ignoring the difference between layout and objects, we still get considerable improvement. 
% On the one hand, the improvement might comes from better understanding of object relationships and scene context, which contributes to a better estimation of object bounding boxes, thus improves the understanding of the layout. 
% On the other hand, this shows the flexibility of the proposed GCN on handling different relationships of inclusion or proximity.

\noindent \textbf{Camera Pose Estimation.} 
\autoref{tbl:layout} also shows the comparison over camera pose prediction, following the evaluation protocol of Total3D.
Our method achieves 5\% better camera pitch %(i.e. improve by 5\%) \zc{removed for limited space}
and slightly worse camera roll.
% The overall performance is in general on par with Total3D

% We compare pitch $\beta$ and roll $\gamma$ of camera pose $\mathbf{R}\left(\beta,\gamma\right)$ with the mean absolute error in degrees in \autoref{tbl:layout}. 
% Our model outperforms Total3D on camera pitch estimation by 0.17 degrees while on par with Total3D on camera roll estimation.

\noindent \textbf{Holistic Scene Reconstruction.}
To our best knowledge, Total3D \cite{nie2020total3dunderstanding} is the only work achieving holistic scene reconstruction from a single RGB, and thus we compare to it.
% by combining scene understanding and mesh generation.
Since no ground truth is presented in SUN RGB-D dataset, we mainly show qualitative comparison in \autoref{fig:scnrecon}. 
Compared to Total3D, our model has less intersection and estimates more reasonable object layout and direction.
We consider this as a benefit from a better understanding of scene context by GCN. Our proposed physical violation loss $\mathcal{L}_{phy}$ also contributes to less intersection.
% We will discuss the contributions made by several aspects quantitatively in the next section.

\subsection{Ablation Study}

In this section, we verify the effectiveness of the proposed components for holistic scene understanding. 
As shown in \autoref{tbl:ablation}, we disable certain components and evaluate the model for 3D layout estimation and 3D object detection, % and the results are shown in \autoref{tbl:ablation}. \zc{removed for limited space}
We do not evaluate the 3D object reconstruction since it is highly related to the usage of deep implicit representation, which has been already evaluated in \autoref{sec:comp}.

\noindent \textbf{Does GCN Matter?}
To show the effectiveness of GCN, we first attach the GCN to the original Total3D to improve the object and scene layout (\autoref{tbl:ablation}, Total3D+GCN).
% We use xxxx as the node feature.
For the difference between MGN of Total3D and LIEN of ours, we replace deep implicit features with the feature from image encoder of MGN and use their proposed partial Chamfer loss $\mathcal{L}_{g}$ instead of $\mathcal{L}_{phy}$.
Both object bounding box and scene layout are improved.
We also train a version of our model without the GCN (Ours-GCN), and the performance drops significantly.
Both experiments show that GCN is effective in capturing scene context.

\noindent \textbf{Does Deep Implicit Feature Matter?}
As introduced in \autoref{sec:lien}, the LDIF representation provides informative node features for the GCN.
Here we demonstrate the contribution from each component of the latent representation.
Particularly, we remove either element centers or analytic code from the GCN node feature (Ours-element, Ours-analytic), and find both hurts the performance.
This indicates that the complete latent representation is helpful in pursuing better scene understanding performance.

\noindent \textbf{Does Physical Violation Loss Matter?}
\zc{
%Last but not least, 
Additionally, 
we evaluate the effectiveness of the physical violation loss.
We train our model without it (Ours-$\mathcal{L}_{phy}$), and also observe performance drop 
for both scene layout and object 3D bounding box 
% on most of the metrics
in \autoref{tbl:ablation}.
We refer to supplementary material for qualitative comparison.
}

\noindent \textbf{Evaluating on Other Metrics.}
\zc{
We also test our method in other aspects including supporting relation, geometry accuracy, and room layout as shown in \autoref{tbl:more_ablation}.
1) We calculate the mean distance between the predicted bottom of on-floor objects and the ground truth floor to measure the supporting relationship.
As ground truth, an object is considered to be on-floor if its bottom surface is within 15cm to the floor.
While GCN significantly improves the metric, $\mathcal{L}_{phy}$ slightly hurts possibly because it tends to push objects away.
Further qualitative results are shown in the supplementary material.
Besides, we also measure the average volume of the collision per scene between objects (Coll Vol), and our full model effectively prevent collision.
2) We follow Total3D \cite{nie2020total3dunderstanding} to evaluate the alignment between scene reconstruction and ground truth depth map with global loss $\mathcal{L}_{g}$, and our full model performs the best.
% To evaluate collision, we propose to use with average volume of the collision per scene. 
3) We also project the predicted layout onto the image and evaluate with image based metrics \cite{dasgupta2016delay, ren2016coarse}.
Our full model achieves the best on both corner and pixel errors.
Overall, the GCN and $\mathcal{L}_{phy}$ benefit on all these aspects.
}

\begin{table}[!t]
    \vspace{-0.2em}
	\begin{center}
	    \resizebox{0.8\columnwidth}{!}{
    		\begin{tabular}{|l|c|c|c|}
    			\hline
    			Setting & Layout IoU $(\uparrow)$ & Detection mAP $(\uparrow)$\\
    			\hline\hline
    			Total3D & 59.25 & 26.38\\
    			Total3D+GCN & 62.49 & 37.04 \\
    			\hline
    			Ours-GCN & 60.04 & 27.47 \\
    			Ours-element & 64.22 & 42.05 \\
    			Ours-analytic & 63.76 & 43.10 \\
    			Ours-$\mathcal{L}_{phy}$ & 63.52 & 43.33 \\
    			\textbf{Full} & \textbf{64.41} & \textbf{45.21}\\
    			\hline
    		\end{tabular}
    	}
	\end{center}
	\vspace{-0.7em}
	\beforecaption
	\caption{Ablation study. We evaluate layout estimation with layout IoU and 3D object detection with mAP.}
	\label{tbl:ablation}
\end{table}

\begin{table}[!t]
    \vspace{-1em}
	\begin{center}
    	\resizebox{1\columnwidth}{!}{
    		\begin{tabular}{|l|c|c|c|c|c|}
    			\hline
    			Setting & \makecell[c]{Sup Err \\ $(cm)$} & $\mathcal{L}_{g}$ & \makecell[c]{Coll Vol \\ $(dm^3/scene)$} & \makecell[c]{Corner Err \\ (\%)} & \makecell[c]{Pixel Err \\ (\%)} \\
    			\hline\hline
    			Total3D & 26.72 & 1.43 & - & 13.29 & 20.51\\
    % 			Total3D+GCN & 19.12 & - & - & 12.58 & 19.00 \\
    			\hline
    % 			Ours-GCN-$\mathcal{L}_{phy}$ & 25.35 & 1.36 & 16.27 & 13.30 & 20.40 \\
    			Ours-GCN & 24.18 & 1.41 & 16.64 & 13.17 & 20.05 \\
    			Ours-$\mathcal{L}_{phy}$ & \textbf{13.35} & 1.14 & 13.65 & 11.60 & 17.91 \\
    			\textbf{Full} & 14.71 & \textbf{1.11} & \textbf{13.55} & \textbf{11.45} & \textbf{17.60}\\
    			\hline
    		\end{tabular}
    	}
	\end{center}
	\vspace{-0.7em}
	\beforecaption
	\caption{\zc{Ablation study on other metrics. We compare on supporting error, $\mathcal{L}_{g}$ (in units of ${10}^{-2}$), average collision volume, corner error, and pixel error.}}
	\label{tbl:more_ablation}
\end{table}

\begin{figure}[!t]
	\centering
    \vspace{-1em}
	\begin{subfigure}[t]{0.15\textwidth}
		\includegraphics[width=\textwidth]  
		{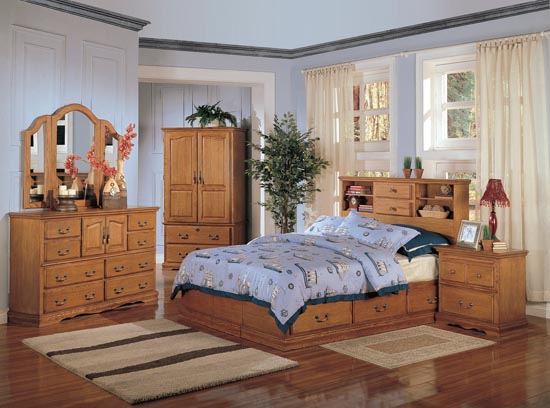}
		\includegraphics[width=\textwidth]  
		{figure/generalizeRebuttal/IMG_0991/IMG_0991}
		\vspace{-6mm}
		\caption{Input}
	\end{subfigure}
	\begin{subfigure}[t]{0.15\textwidth}
		\includegraphics[width=\textwidth]  
		{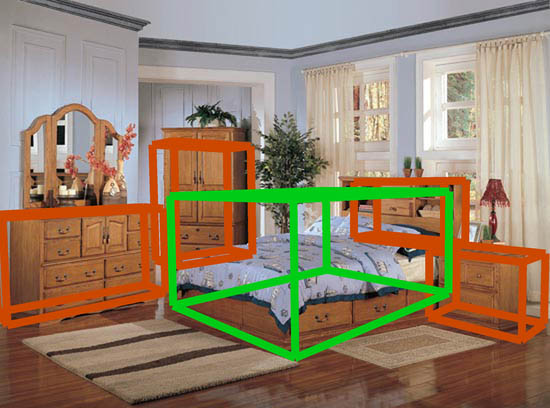}
		\includegraphics[width=\textwidth]  
		{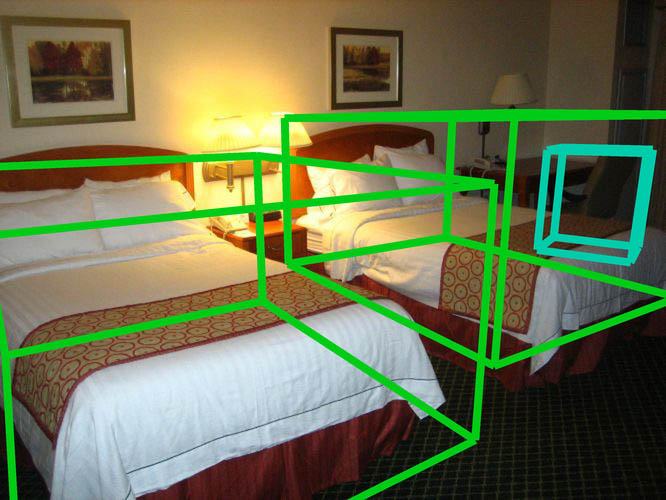}
		\vspace{-6mm}
		\caption{3D Detection}
	\end{subfigure}
	\begin{subfigure}[t]{0.15\textwidth}
		\includegraphics[width=\textwidth]  
		{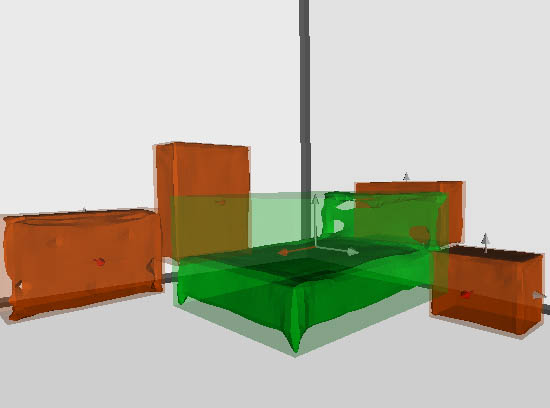}
		\includegraphics[width=\textwidth]  
		{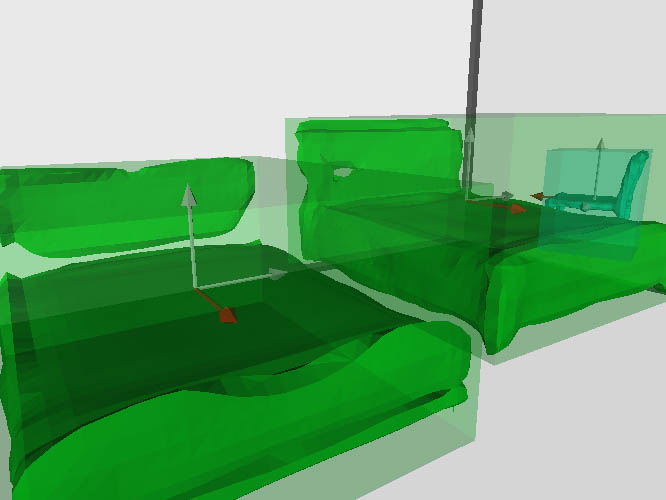}
		\vspace{-6mm}
		\caption{Reconstruction}
	\end{subfigure}
	\vspace{0.2em}
	\beforecaption
	\caption{Qualitative results on ObjectNet3D dataset \cite{xiang2016objectnet3d} (row 1) and the layout estimation dataset in \cite{hedau2009recovering} (row 2).}
    % \vspace{-1em}
	\label{fig:rebuttal_genralize}
\end{figure}

\subsection{Generalization to other datasets}

\zc{
We also show qualitative results of our method tested on the 3D detection dataset ObjectNet3D \cite{xiang2016objectnet3d} and the layout estimation dataset in \cite{hedau2009recovering} without fine-tuning in \autoref{fig:rebuttal_genralize}.
Our method shows good generalization capability and performs reasonably well on these unseen datasets.
}

\section{Conclusion}
We have presented a deep learning model for holistic scene understanding by leveraging deep implicit representation.
Our model not only reconstructs accurate 3D object geometry, but also learns better scene context using GCN and a novel physical violation loss, which can deliver accurate scene and object layout.
Extensive experiments show that our model improves various tasks in holistic scene understanding over existing methods.
A promising future direction could be exploiting %high-level context information or 
object functionalities for better 3D scene understanding.

\vspace{0.5em}
\noindent \textbf{Acknowledgement:} This research was supported by National Natural Science Foundation of China (NSFC) under grants No.61872067 and No.61720106004.

%xxxx \yd{In case you want to back up something}.

% With graph convolutional network as the core idea, we cooperate it with LDIF, a 3D representation combining structured and implicit representation, to better understand the relationship between objects and layout. 
% We further propose a feature aiming at representing the geometric information for each object, and an object physical violation loss to punish intersection between objects. 
% The experiment results show our method introduces significant improvements on several tasks of scene understanding.

{\small
\bibliographystyle{ieee_fullname}
\bibliography{egbib}
}

% \end{document}

\clearpage

% \begin{document}
\newpage\appendix
% \noindent{\textbf{\large{Supplementary Material}}}
\begin{center}
{\Large \textbf{Supplementary Material}}
\end{center}
%%%%%%%%% TITLE
% \title{Holistic 3D Scene Understanding from a Single Image \\ with Implicit Representation: Supplementary Material}

\maketitle

In this supplementary material, we provide the detailed network architecture, implementation details, 3D detection on all categories, more qualitative results, and discussion of failure cases. 
% \vspace{-1em}

\section{Architecture of Our Pipeline}
\begin{figure*}
	\centering
	\includegraphics[width=.9\textwidth]  
		{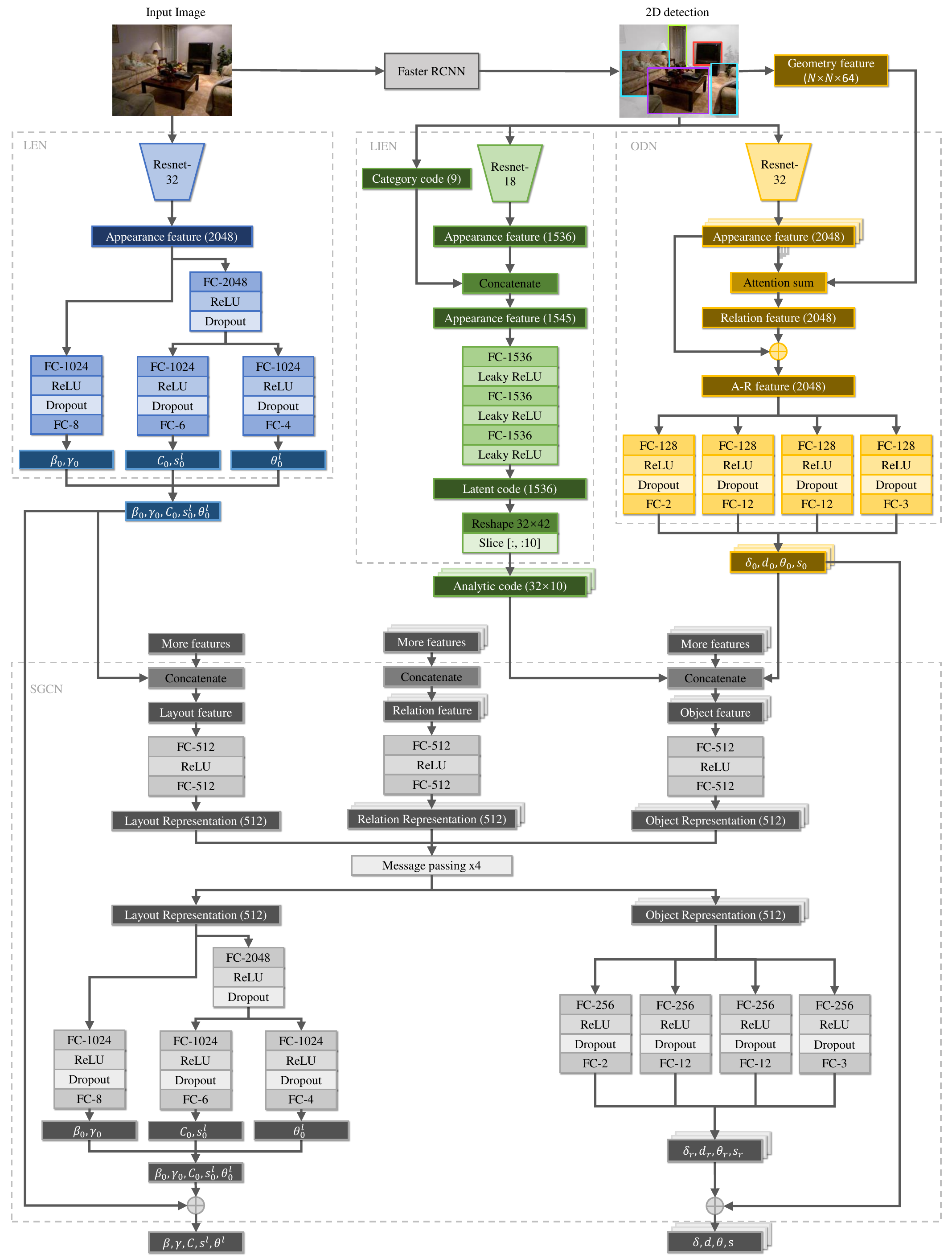}
	\caption{Architecture of LEN, ODN, LIEN, and SGCN. Our pipeline takes features from LEN, ODN, LIEN and other sources and embeds them into node representations. The parameter $x_0$ initialized by LEN and ODN is then refined with residual $x_r$ decoded from updated node representations, $\forall x \in \{\beta, \gamma, C, s^l, \theta^l, \delta, d, \theta, s\}$. Variables $\beta, \gamma, \theta^{l}, d, \theta$ are parameterized following \cite{nie2020total3dunderstanding, huang2018holistic}. We set dropout rate to 0.5 for all dropout blocks.}
	\label{fig:architecture}
\end{figure*}

We show the architecture of LEN, ODN, LIEN, and SCGN in \autoref{fig:architecture}.

\noindent\textbf{2D Detector, LEN, ODN.} 
Following \cite{nie2020total3dunderstanding, huang2018cooperative}, we use Faster RCNN \cite{ren2016faster} trained on COCO dataset \cite{lin2014microsoft} and fine-tuned on SUN RGB-D~\cite{song2015sun} as 2D detector.
The 2D detection results on SUN RGB-D are filtered and matched with the ground-truth 3D object bounding box during the data preparation procedure provided by \cite{nie2020total3dunderstanding}.
During the initialization stage, we use LEN and ODN architecture shown in \autoref{fig:architecture} similar with \cite{nie2020total3dunderstanding}. 

\noindent\textbf{LIEN.} 
Our proposed LIEN consists of an image encoder followed by a three-layer MLP to embed a single image into a code. 
When evaluating on SUN RGB-D, the category labels are mapped to the ones used by Pix3D and concatenated to the image feature following \cite{nie2020total3dunderstanding}.
To construct the shape elements for LDIF decoder, we follow \cite{genova2020local} to reshape the 1344-dim vector into a 32x42 array, which corresponds to 42-dim (10 for analytic code and 32 for latent code) codes of the 32 shape elements.

\noindent\textbf{SGCN.}
Before being fed into each node, the features from different sources are flattened, concatenated, and embedded into a 512-dim representation using FC layers.
The weights of the embedding network for layout, object, and relationship nodes are independent of each other.
After updated with four steps of message passing, the representations of layout and object nodes are decoded into parameters with the networks specially designed for each of the node types.
The decoding networks follow the design of LEN and ODN, and refine the parameterized initial outputs of them.

\section{Implementation details}
\noindent\textbf{Data Processing.} 
For the training of LIEN, watertight meshes~\cite{genova2020local} must be used to retrieve the ground-truth values of inside-outside labels. 
However, the models of Pix3D~\cite{sun2018pix3d} are not that clean with inverted surface normals and holes occasionally, which causes failure with the traditional flood fill algorithm. 
To get more robust results, we utilize the mesh fusion pipeline \cite{mescheder2019occupancy} which generates watertight meshes by fusing signed distance fields from several virtual cameras and applying the marching cube algorithm on it.
Although the mesh fusion pipeline makes the model thicker and introduces noise to the ground-truth sample points, we evaluate it on the original mesh to directly compare with previous works.

\noindent\textbf{SGCN Outputs.}\zc{
As mentioned in main paper Section 3.1, our SGCN predicts residuals to refine the parameters of object bounding boxs, layout box, and camera pose.
We follow \cite{nie2020total3dunderstanding} to set the origin of the world coordinate frame at the camera center, with the y-axis up and perpendicular to the floor, and the x-axis aligned to the orientation of the camera forward.
Thus the camera pose can be parameterized as $\mathbf{R}\left(\beta,\gamma\right)$, where $\beta$ is the camera pitch and $\gamma$ is the camera roll.
Also, a bounding box can be parameterized as 3D center ${C} \in \mathbb{R}^{3}$, size ${s} \in \mathbb{R}^{3}$ and orientation $\theta \in [-\pi, \pi)$.
Specifically, a layout box can be represented as $\left({C},{s}^{l},\theta^{l}\right)$, and a object box can be represented as $\left({\delta},d,{s},\theta\right)$, where ${\delta} \in \mathbb{R}^{2}$ is the offset between 2D projection of 3D center and detected 2D object bounding box center, and $d$ is the distance between 3D center and camera center.
}

\noindent\textbf{Hyper Parameters.}
When training LIEN, we use 1024 near-surface samples and 1024 uniformly samples, and set their loss weight $\lambda_{ns}=0.1$ and $\lambda_{us}=1$.
For shape element center loss, we let $\lambda_{c}=0.2$.
Following \cite{nie2020total3dunderstanding, huang2018holistic}, classification and regression loss is used for parameters of both LEN and ODN, which we denote as $\mathcal{L}^{cls,reg}_{x} = \mathcal{L}^{cls}_{x} + \lambda^{reg}_{x}\mathcal{L}^{reg}_{x}, \forall x \in \{\beta, \gamma, \theta^{l}, d, \theta\}$. Other parameters of LEN and ODN are using only regression loss.
For camera parameters, we set $\lambda_{\beta}=0.25, \lambda^{reg}_{\beta}=40, \lambda_{\gamma}=0.25$, and $\lambda^{reg}_{\gamma}=20$.
For layout box parameters, we set $\lambda_{C}=10, \lambda_{{s}^{l}}=10, \lambda_{\theta^{l}}=0.25$, and $\lambda^{reg}_{\theta^{l}}=30$.
For object box parameters, we set $\lambda_{\delta}=1, \lambda_{d}=0.75, \lambda^{reg}_{d}=6.7, \lambda_{s}=10, \lambda_{\theta}=0.33$, and $\lambda^{reg}_{\theta}=30$.
When training with cooperative loss and object physical violation loss, we set $\lambda_{co}=150, \lambda_{phy}=20, \alpha=100, k=4$.

\noindent\textbf{Scene Mesh Reconstruction.}
Since our LIEN is trained on Pix3D with only 9 categories like MGN of Total3D, we suffer from the same problem with them when testing on SUN RGB-D, that our LIEN can not generalize to some of the categories.
For the accuracy of the scene reconstruction, we follow \cite{nie2020total3dunderstanding} to only consider certain categories of objects (i.e. cabinet, bed, chair, sofa, table, door, bookshelf, desk, shelves, dresser, refrigerator, television, box, whiteboard, nightstand).
As a result, the reconstructed scene mesh has fewer objects than 3D detections.

\section{3D Detection on all categories}
%Following \cite{nie2020total3dunderstanding}, 
In this section, we report the average precision of 3D object detection on all categories of SUN RGB-D for a full comparison in \autoref{tbl:3ddetectionSupp}.
Our method achieves the best performance for 27 over 33 categories and a significantly better mean average precision.

\begin{table}[!ht]
	\begin{center}
		\begin{tabular}{l|c|c|c}
			Method & CooP \cite{huang2018cooperative} & Total3D \cite{nie2020total3dunderstanding} & Ours \\
			\hline
			cabinet & 10.47 & 14.51 & \textbf{33.93} \\
			bed & 57.71 & 60.65 & \textbf{89.34} \\
			chair & 15.21 & 17.55 & \textbf{35.14} \\
			sofa & 36.67 & 44.90 & \textbf{69.10} \\
			table & 31.16 & 36.48 & \textbf{57.37} \\
			door & 0.14 & 0.69 & \textbf{5.82} \\
			window & 0.00 & \textbf{0.62} & 0.00 \\
			bookshelf & 3.81 & 4.93 & \textbf{18.33} \\
			picture & 0.00 & 0.37 & \textbf{1.04} \\
			counter & 27.67 & 32.08 & \textbf{57.02} \\
            blinds & \textbf{2.27} & 0.00 & 1.69 \\
            desk & 19.90 & 27.93 & \textbf{49.03} \\
            shelves & 2.96 & 3.70 & \textbf{16.68} \\
            curtain & 1.35 & 3.04 & \textbf{7.38} \\
            dresser & 15.98 & 21.19 & \textbf{29.27} \\
            pillow & 2.53 & 4.46 & \textbf{11.41} \\
            mirror & 0.47 & 0.29 & \textbf{0.87} \\
            clothes & 0.00 & 0.00 & 0.00 \\
            books & 3.19 & 2.02 & \textbf{5.44} \\
			fridge & 21.50 & 24.42 & \textbf{39.12} \\
			tv & 5.20 & 5.60 & \textbf{11.17} \\
			paper & 0.20 & \textbf{0.97} & 0.03 \\
			towel & 2.14 & 2.07 & \textbf{7.73} \\
			shower curtain & \textbf{20.00} & \textbf{20.00} & 0.00 \\
			box & 2.59 & 2.46 & \textbf{6.71} \\
			whiteboard & 0.16 & 0.61 & \textbf{2.39} \\
			person & 20.96 & \textbf{31.29} & 20.82 \\
			nightstand & 11.36 & 17.01 & \textbf{41.34} \\
			toilet & 42.53 & 44.24 & \textbf{70.81} \\
		    sink & 15.95 & 18.50 & \textbf{33.81} \\
		    lamp & 3.28 & 5.04 & \textbf{11.90} \\
		    bathtub & 24.71 & 21.15 & \textbf{53.64} \\
		    bag & 1.53 & 2.47 & \textbf{6.82} \\
		    \hline
		    mAP & 12.23 & 14.28 & \textbf{24.10} \\
		\end{tabular}
	\end{center}
	\vspace{-1em}
	\caption{Average precision of 3D object detection on all categories. For CooP, we report the better results from \cite{nie2020total3dunderstanding} trained on NYU-37 object labels.}
	\label{tbl:3ddetectionSupp}
\end{table}

\section{More Qualitatively Comparison with MGN on Object Mesh Reconstruction}
\begin{figure*}[!ht]
	\centering
	\scriptsize
	\newcommand{\verti}[1]{\raisebox{-.5\height}{\includegraphics[width=.14\linewidth]{#1}}}
	\newcommand{\squa}[1]{\raisebox{-.5\height}{\includegraphics[width=.1\linewidth]{#1}}}
	\newcommand{\hori}[1]{\raisebox{-.5\height}{\includegraphics[width=.05\linewidth]{#1}}}
	\begin{tabular}{*{9}{c@{\hspace{5px}}}}
	    \verti{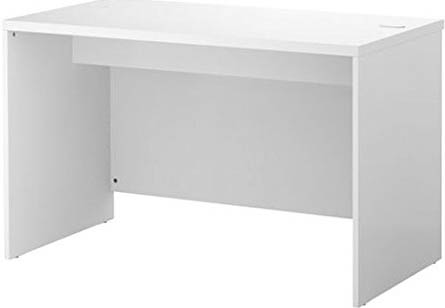} &
	    \verti{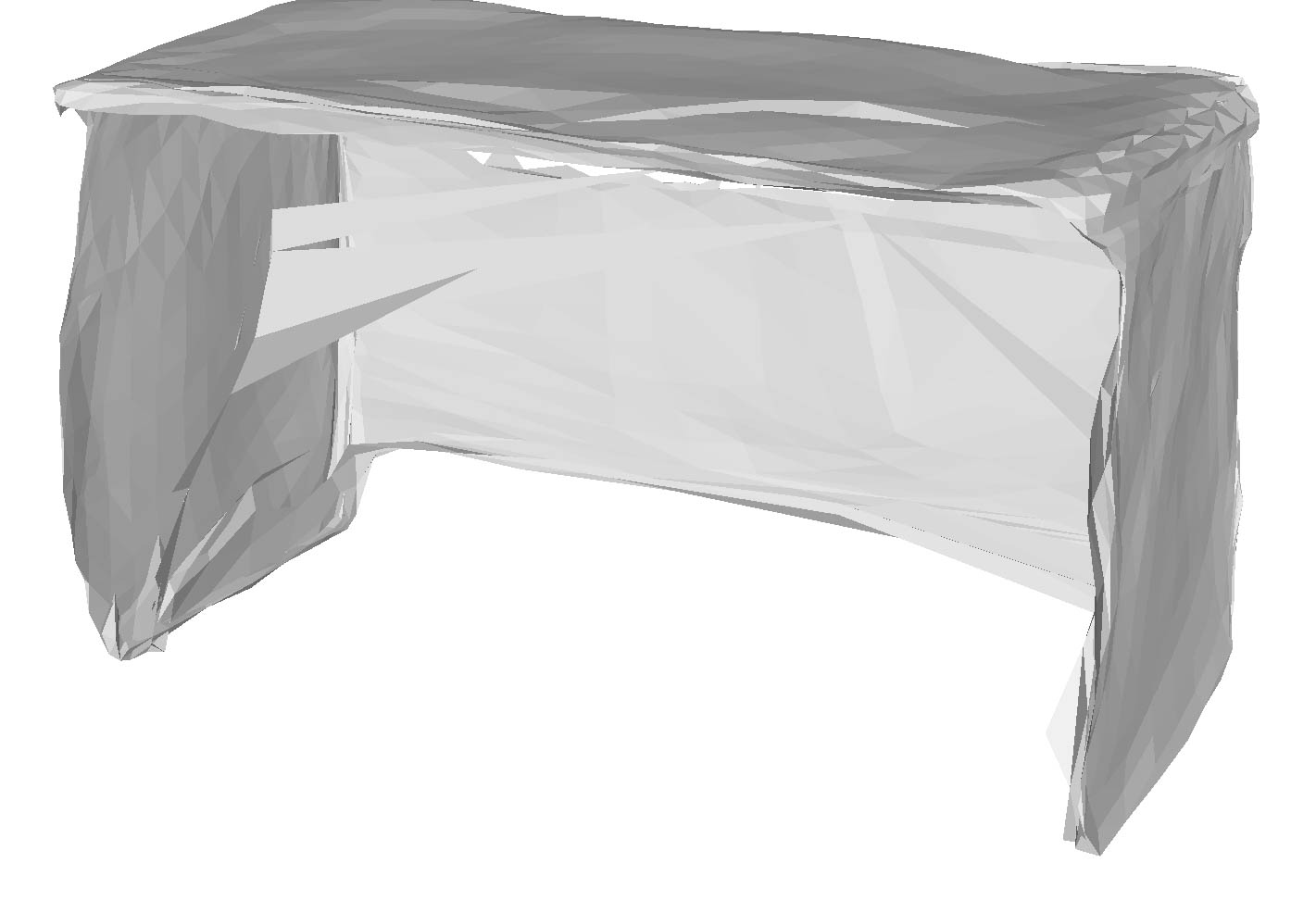} &
	    \verti{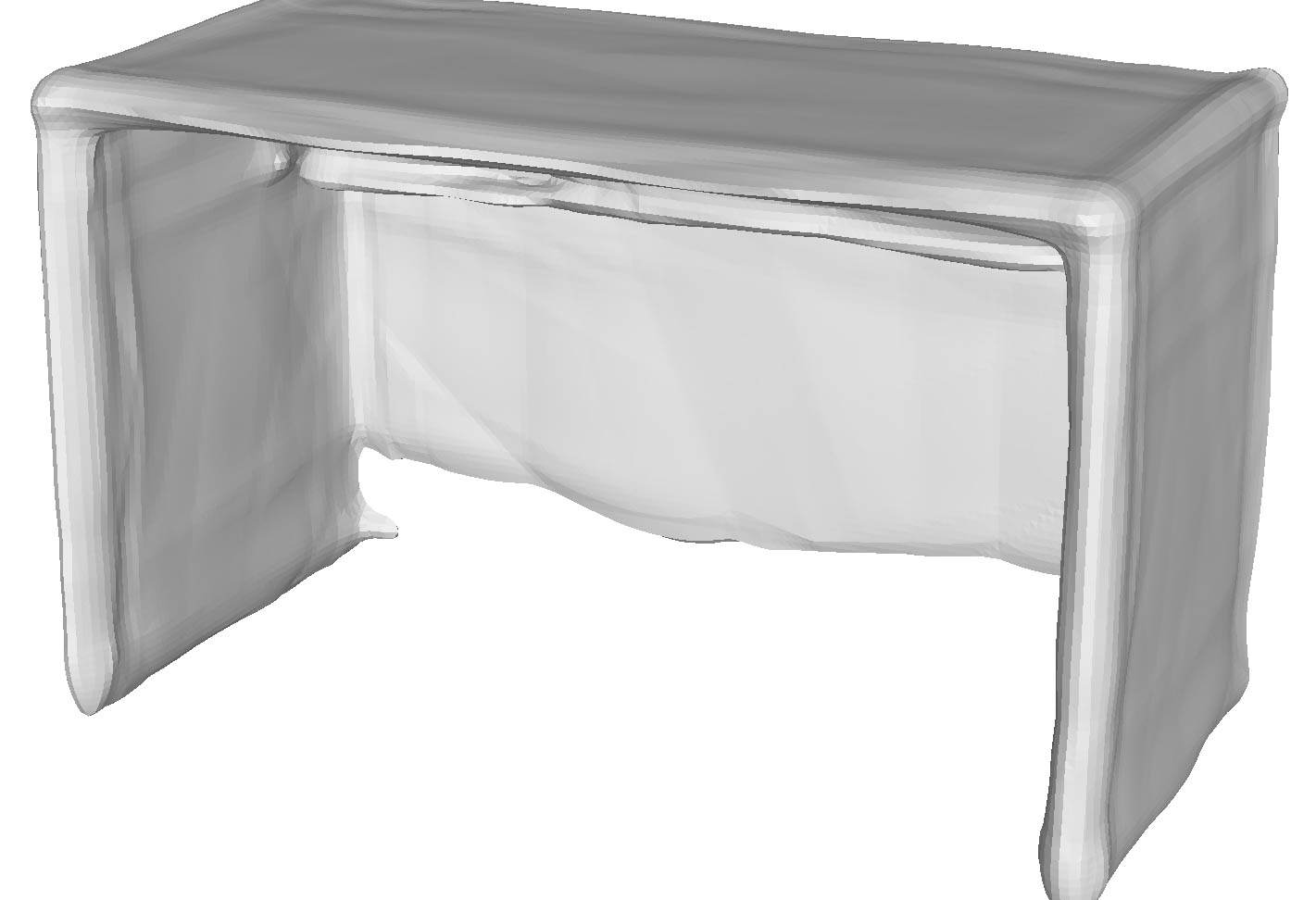} &
	    \squa{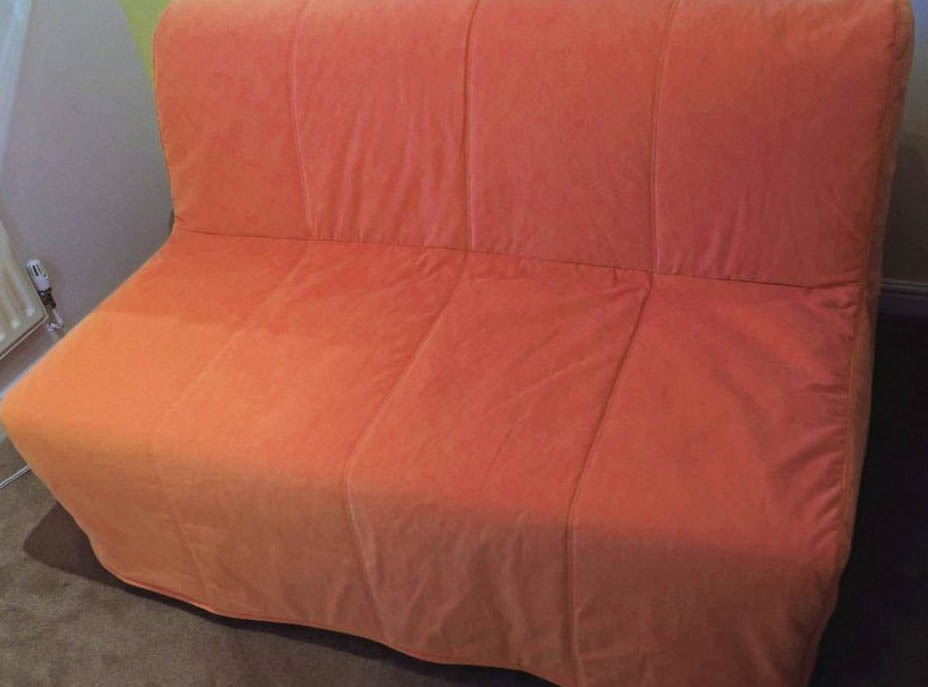} &
	    \squa{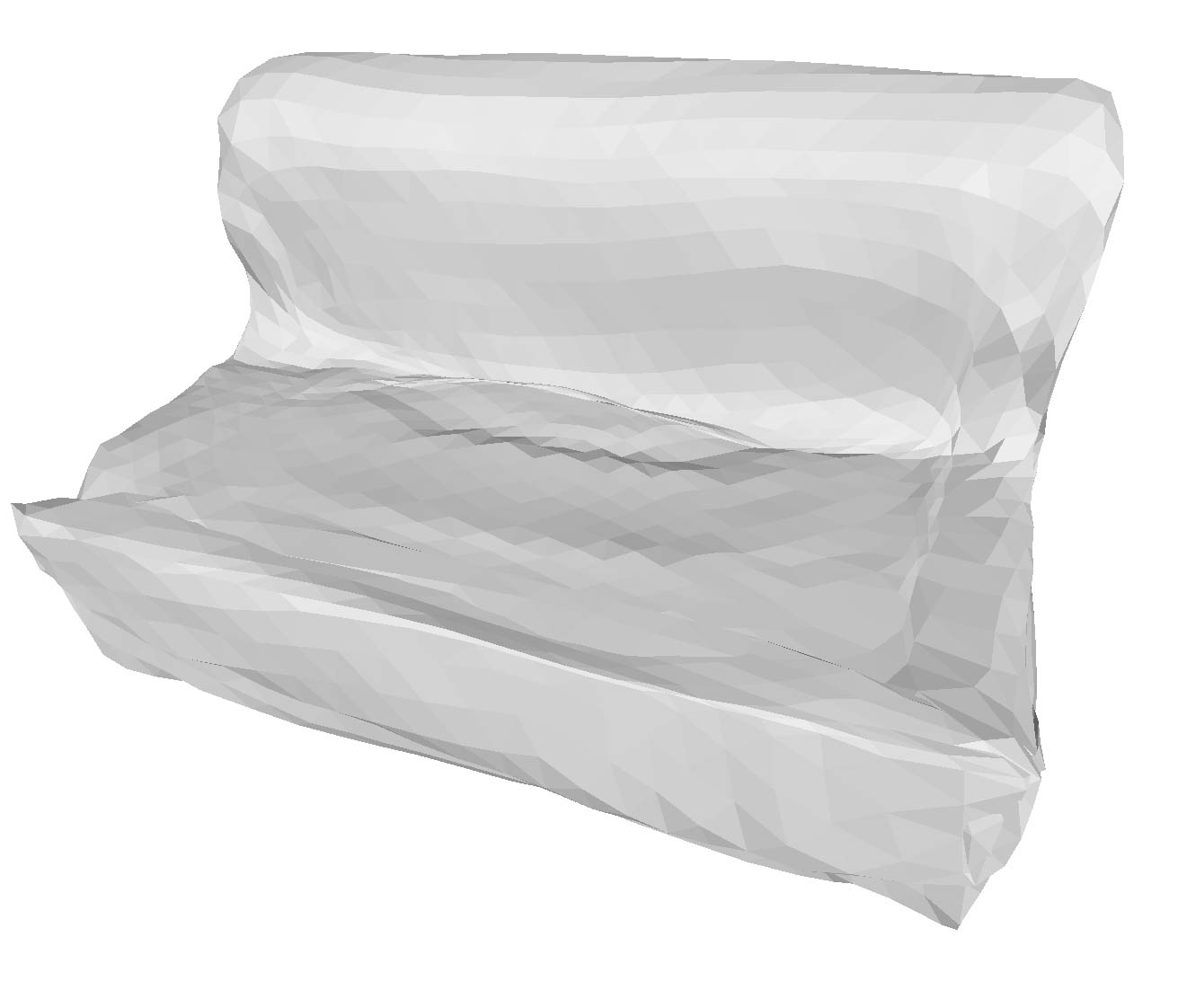} &
	    \squa{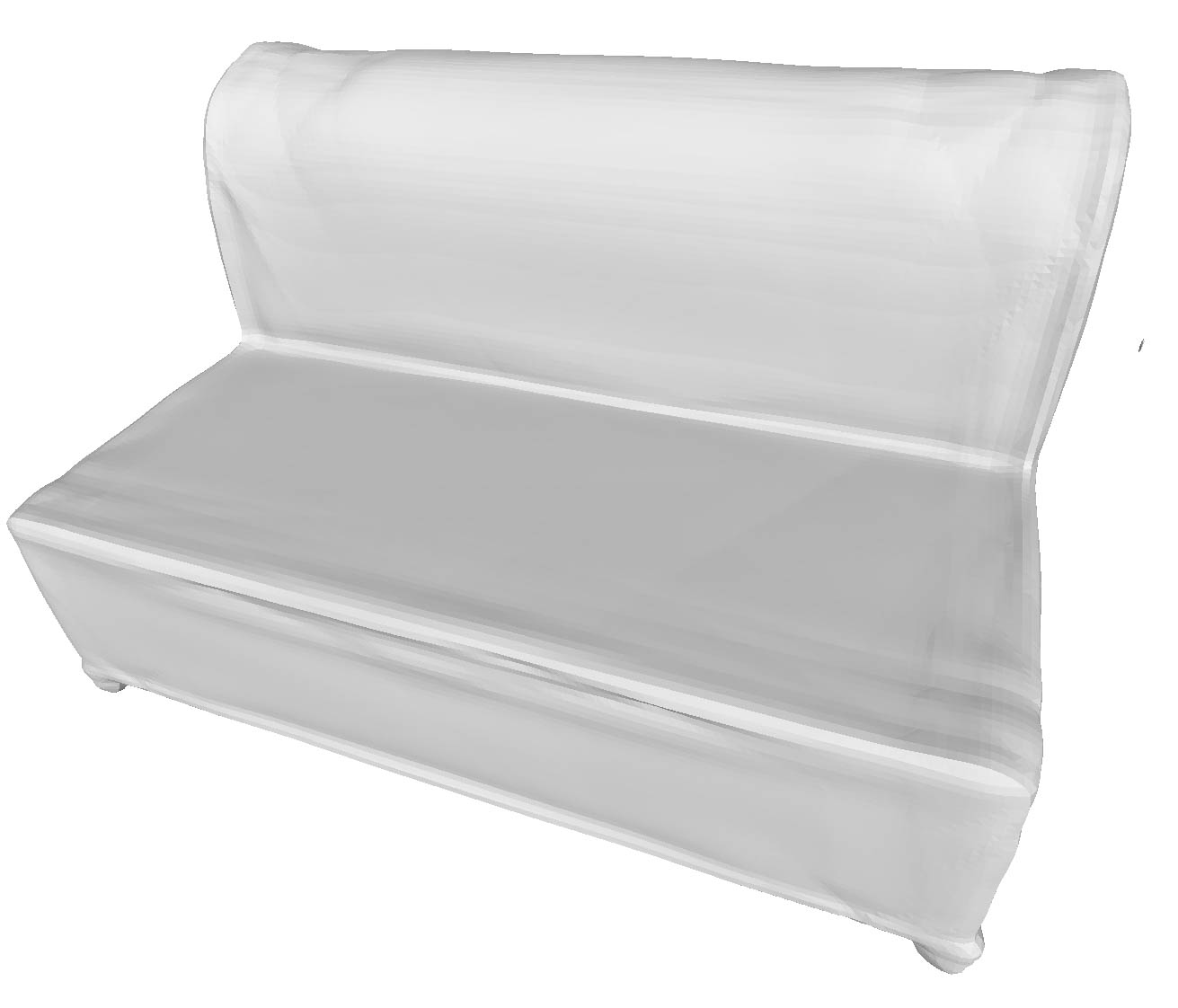} &
	    \hori{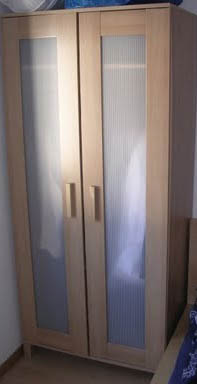} &
	    \hori{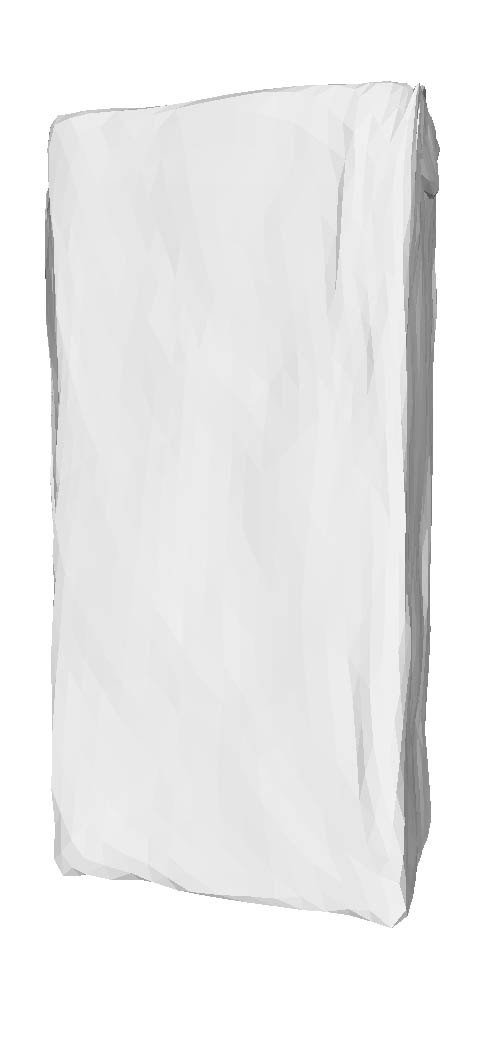} &
	    \hori{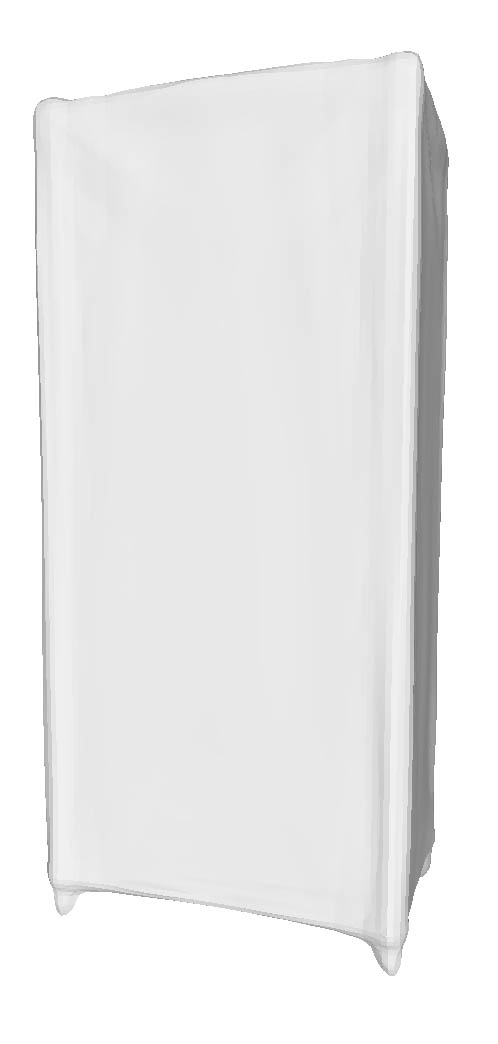} \\
	    
	    \verti{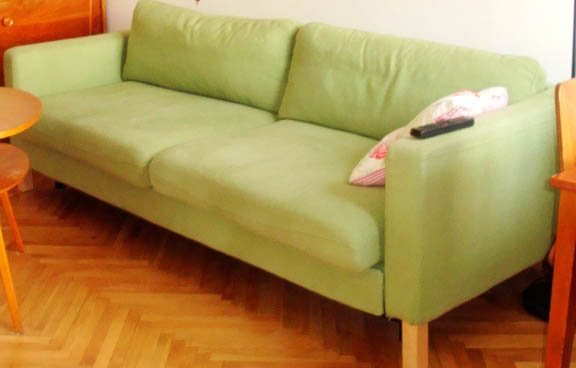} &
	    \verti{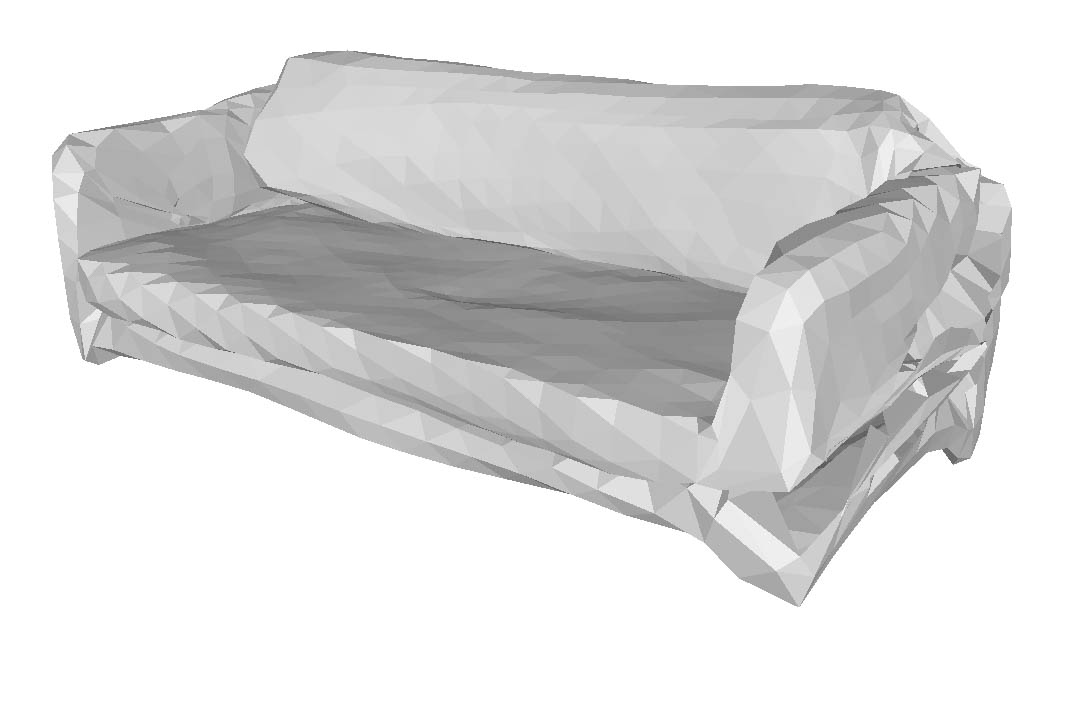} &
	    \verti{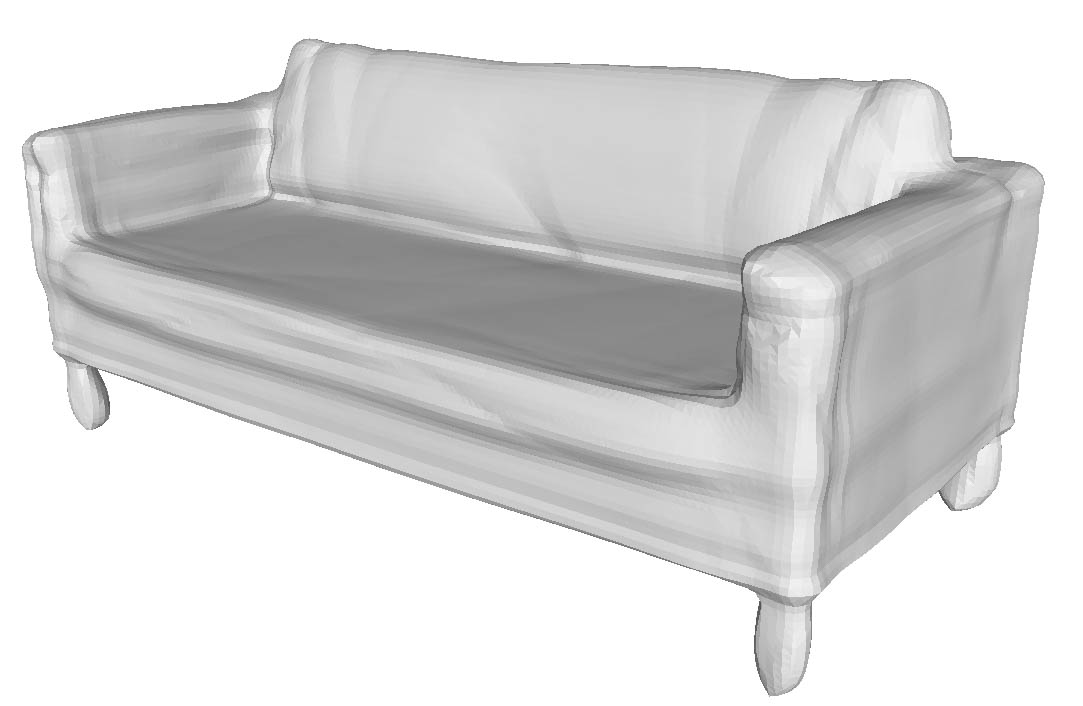} &
	    \squa{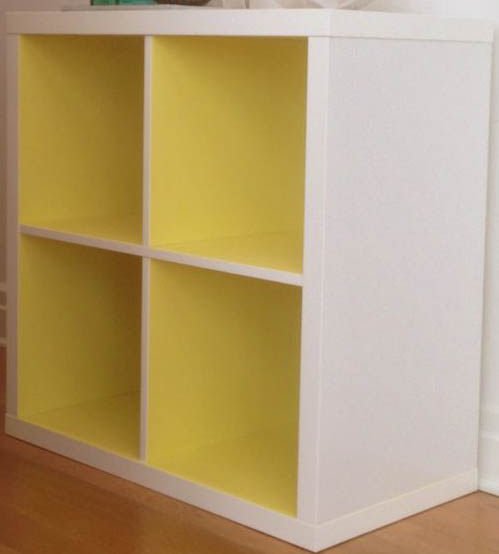} &
	    \squa{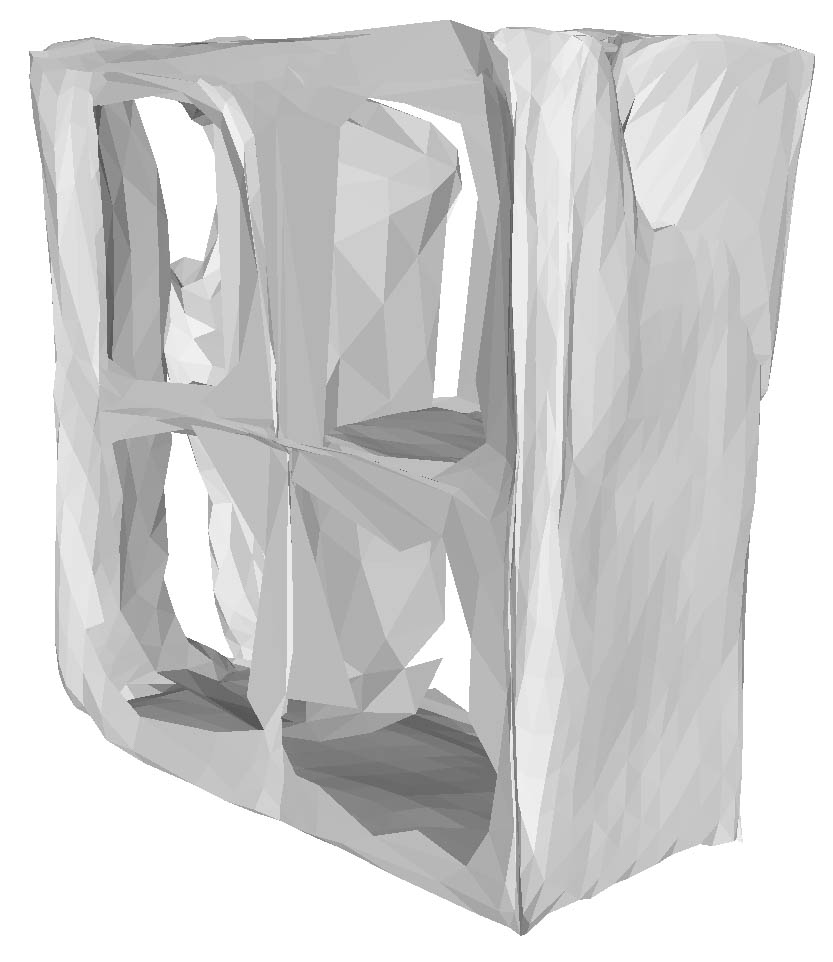} &
	    \squa{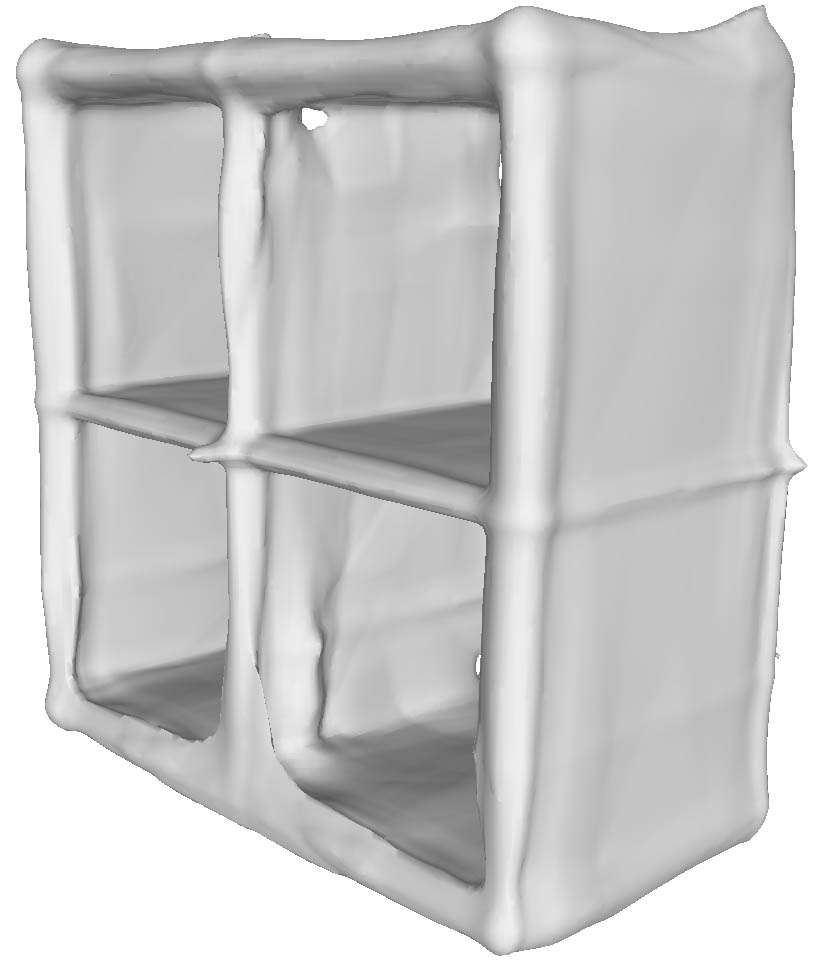} &
	    \hori{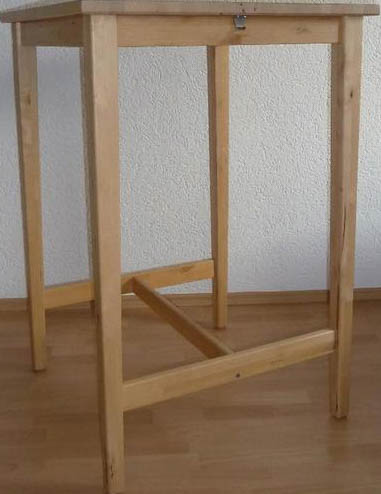} &
	    \hori{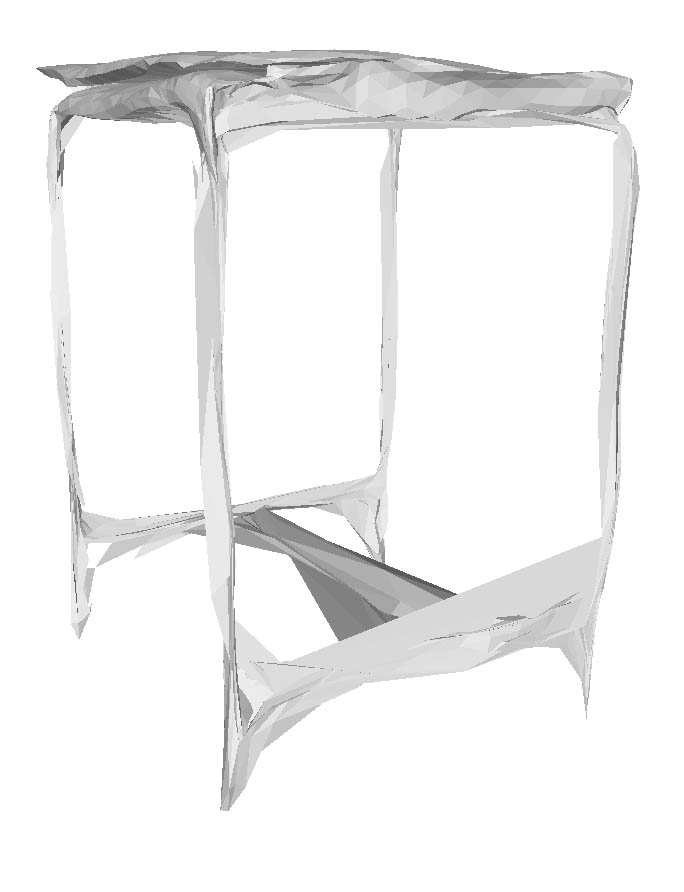} &
	    \hori{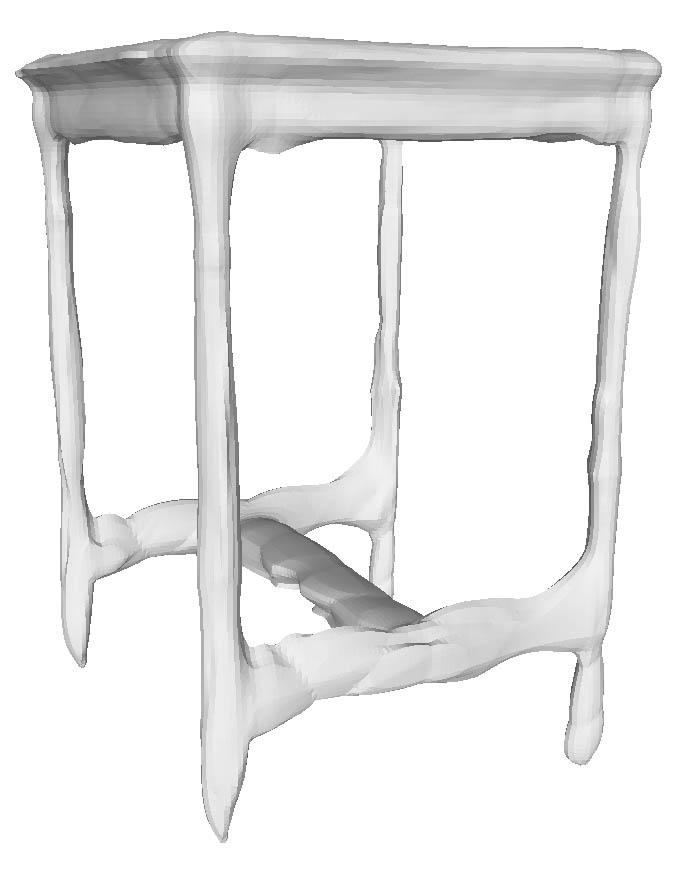} \\
	    
	    \verti{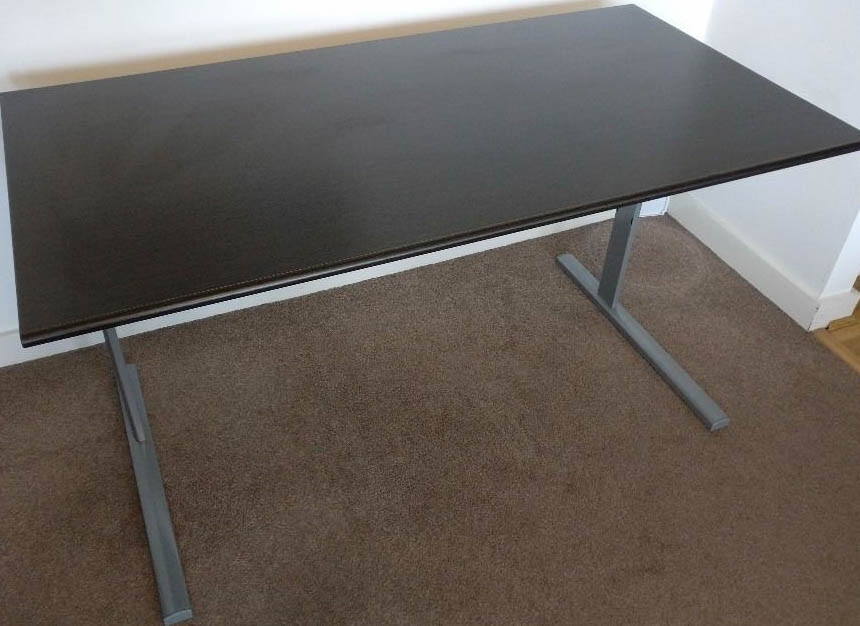} &
	    \verti{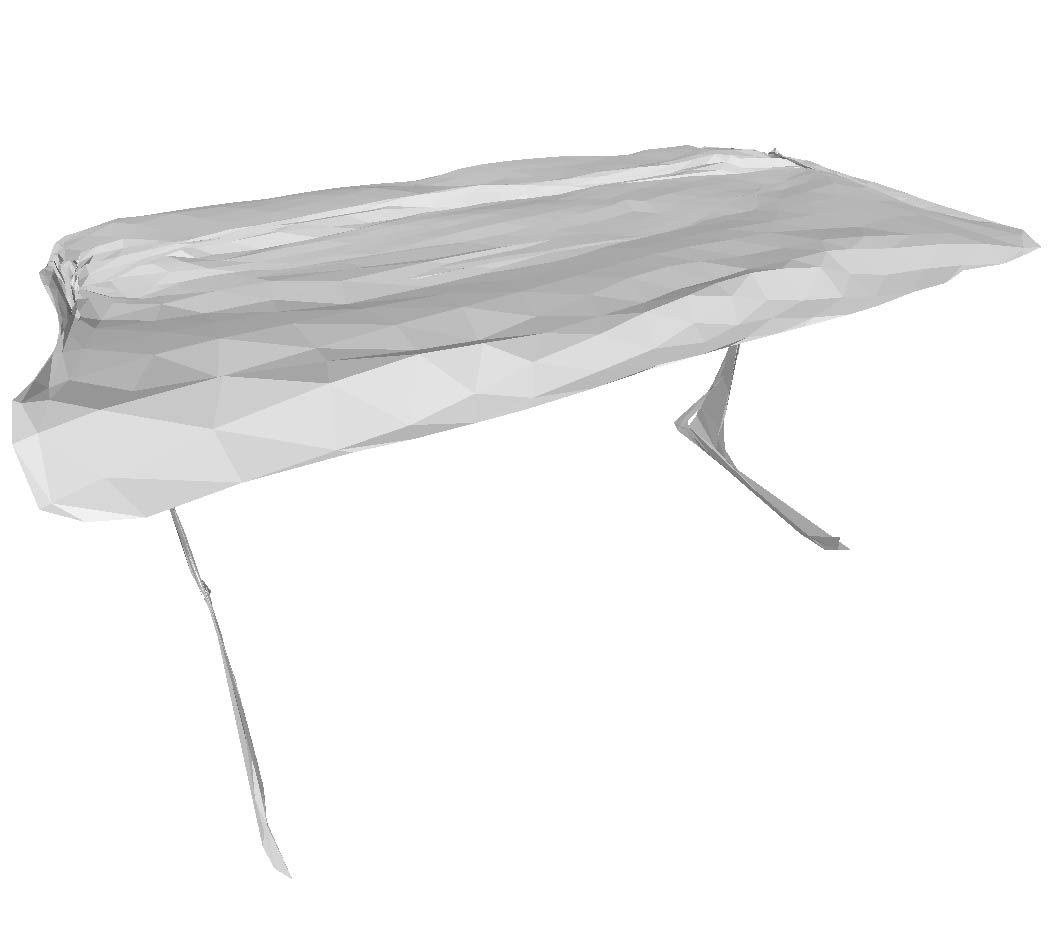} &
	    \verti{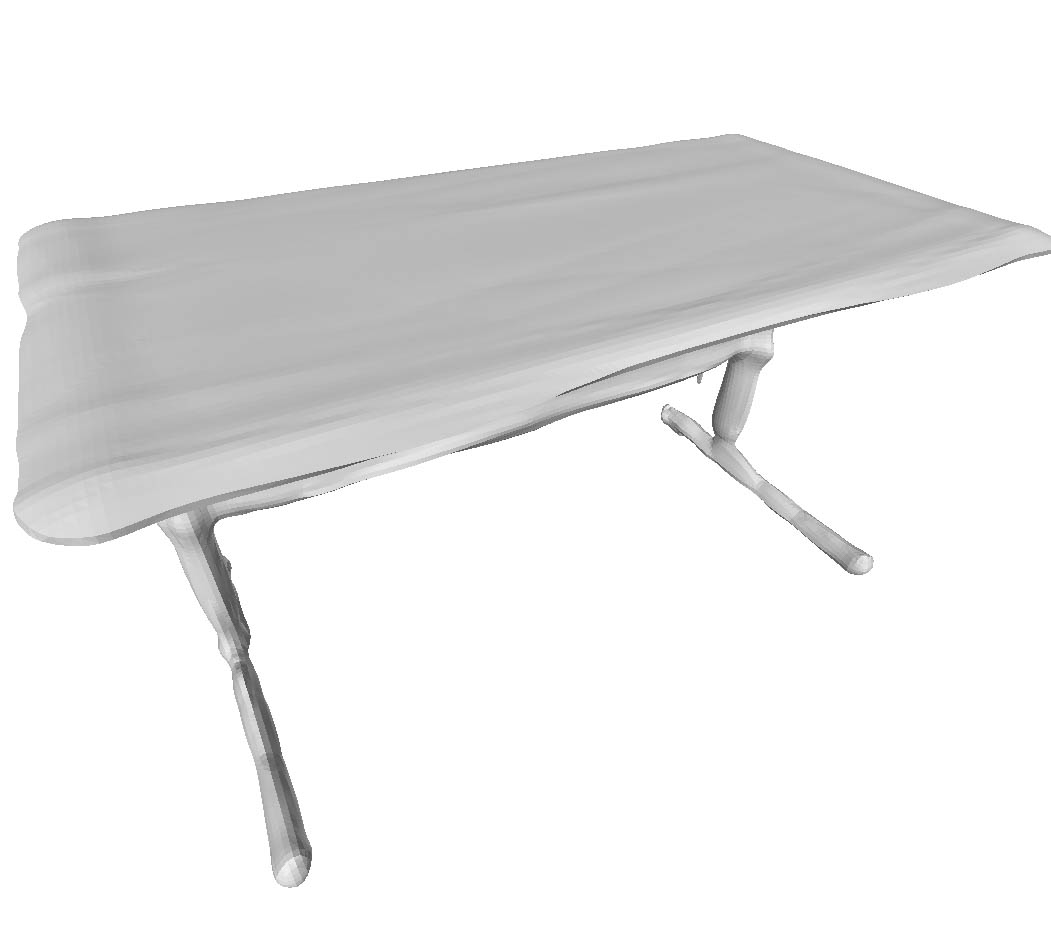} &
	    \squa{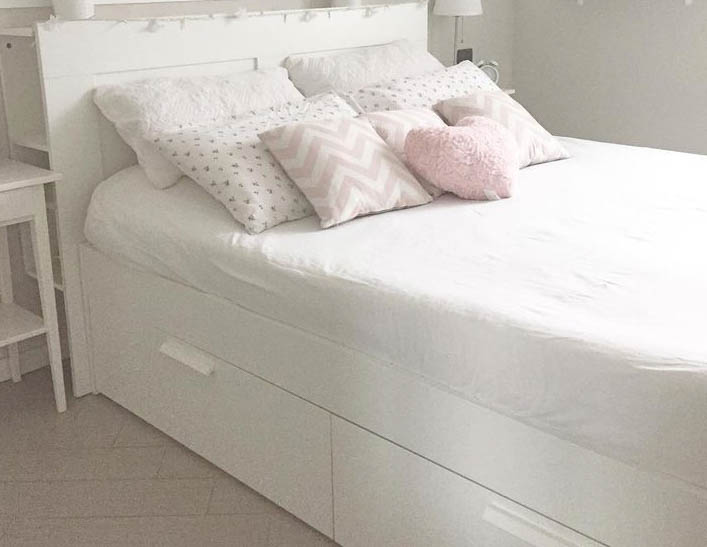} &
	    \squa{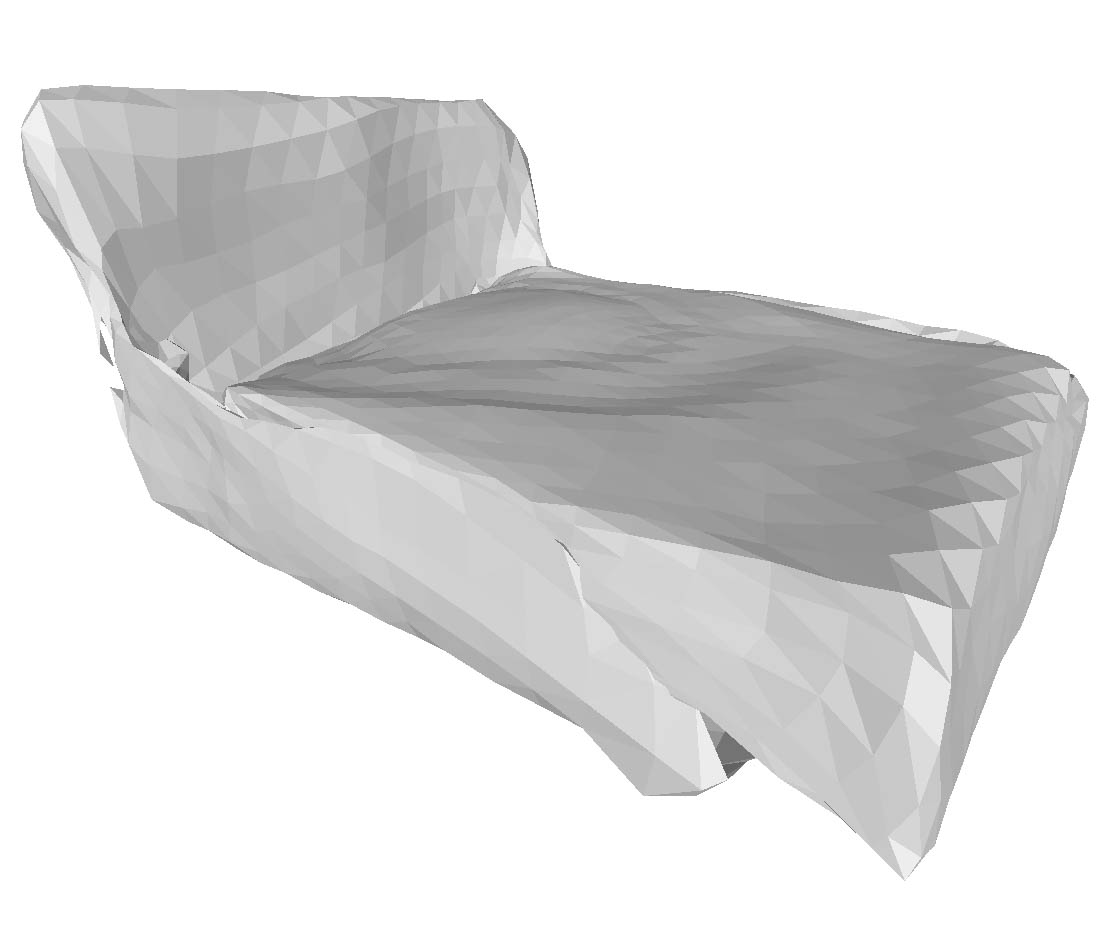} &
	    \squa{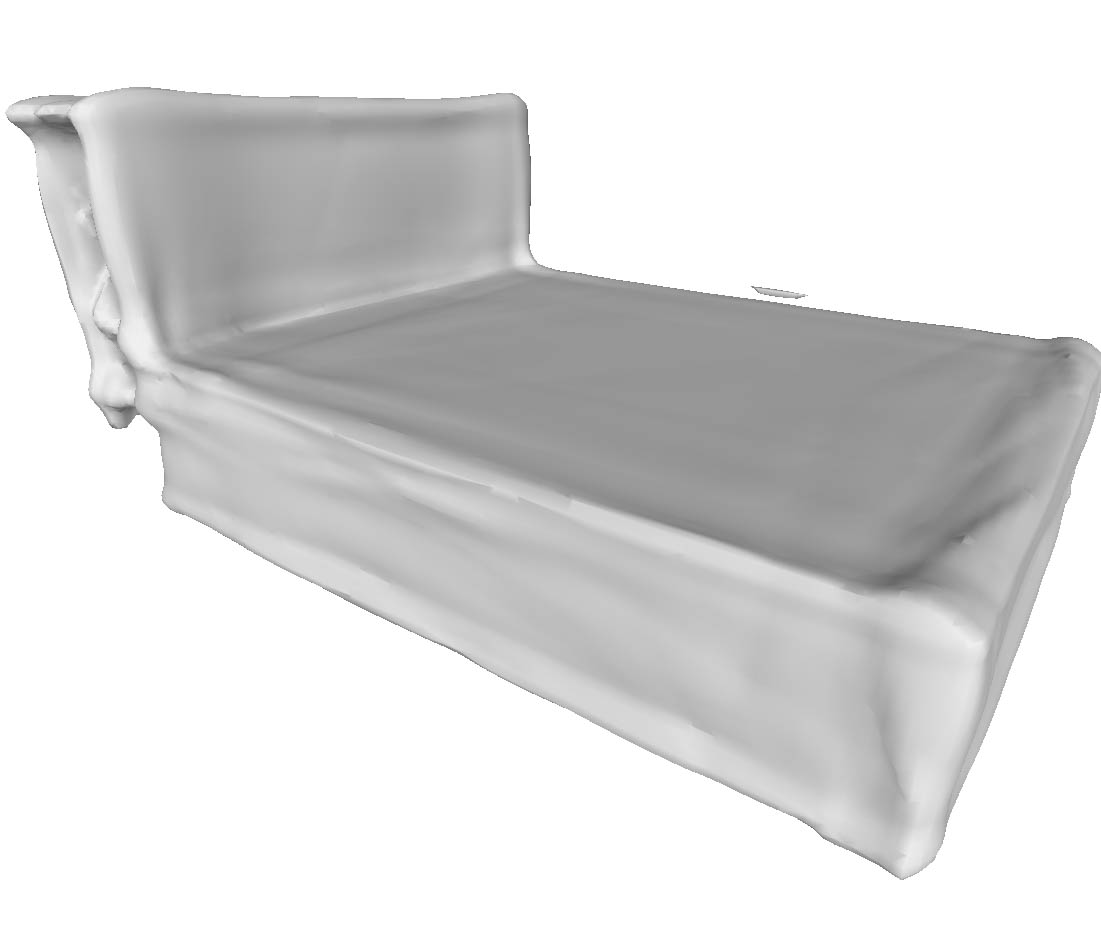} &
	    \hori{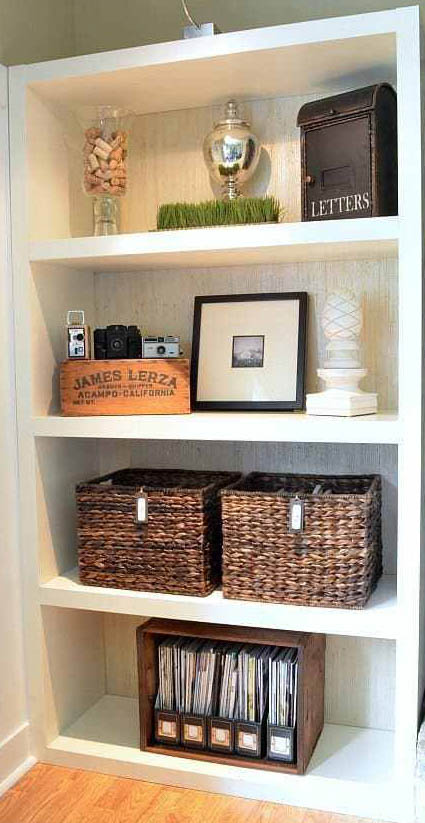} &
	    \hori{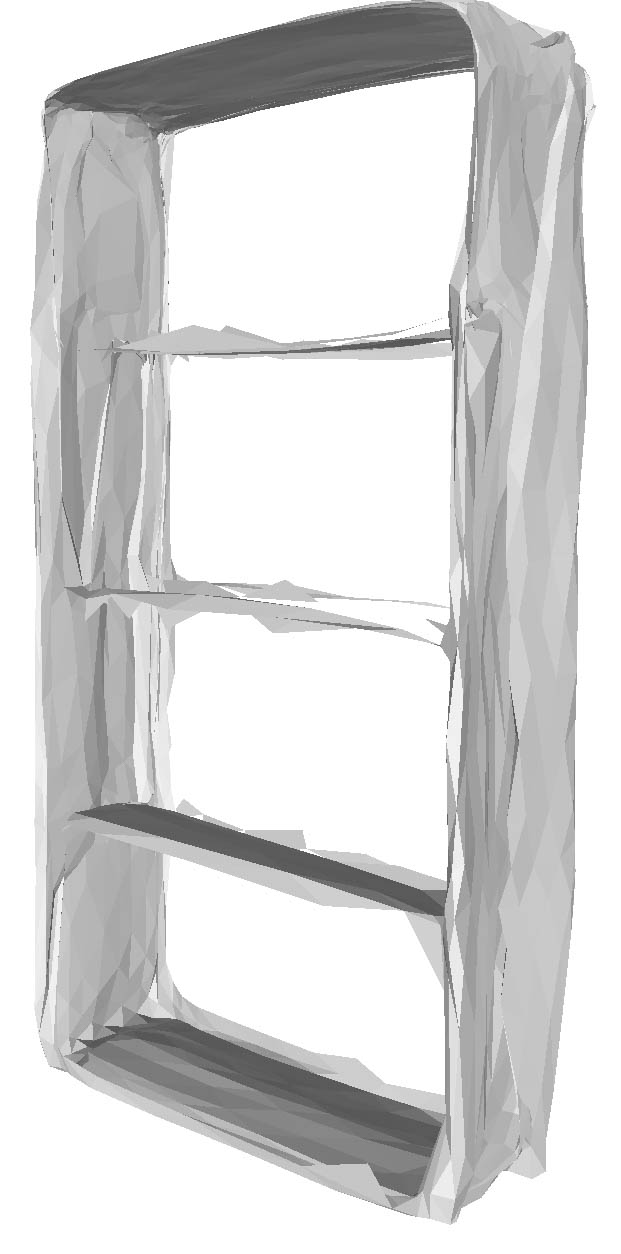} &
	    \hori{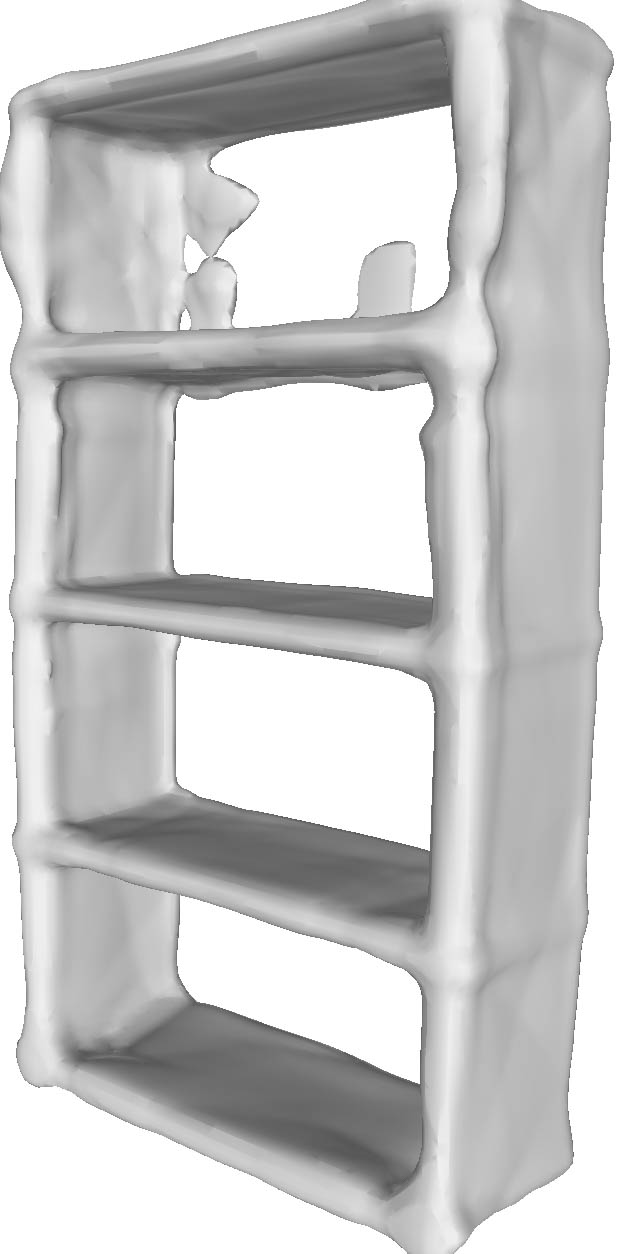} \\
	    
	    \verti{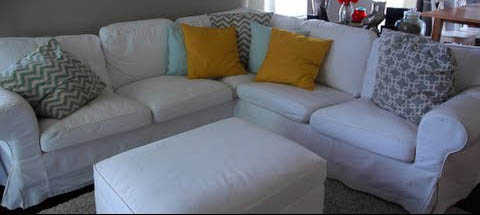} &
	    \verti{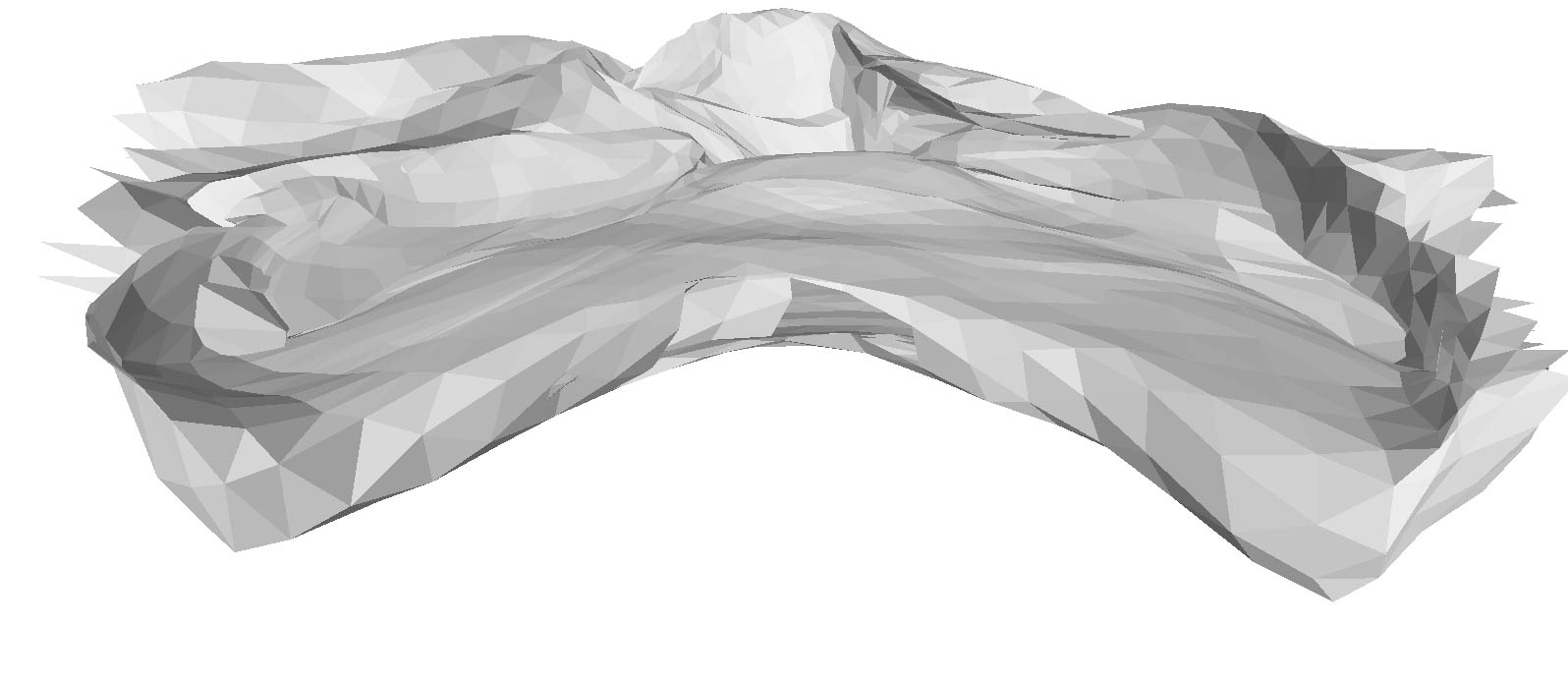} &
	    \verti{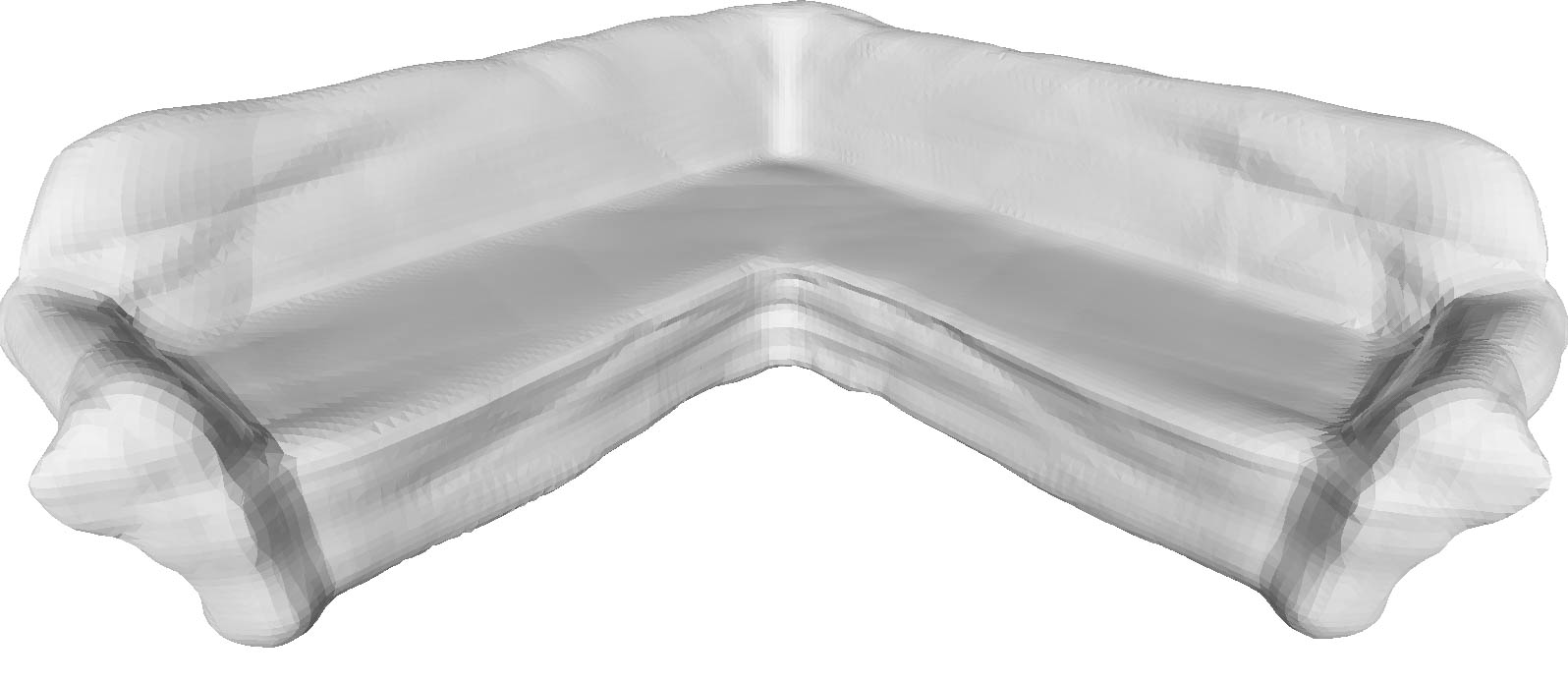} &
	    \squa{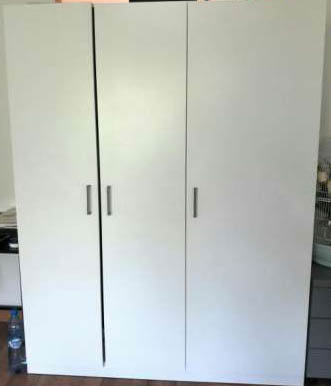} &
	    \squa{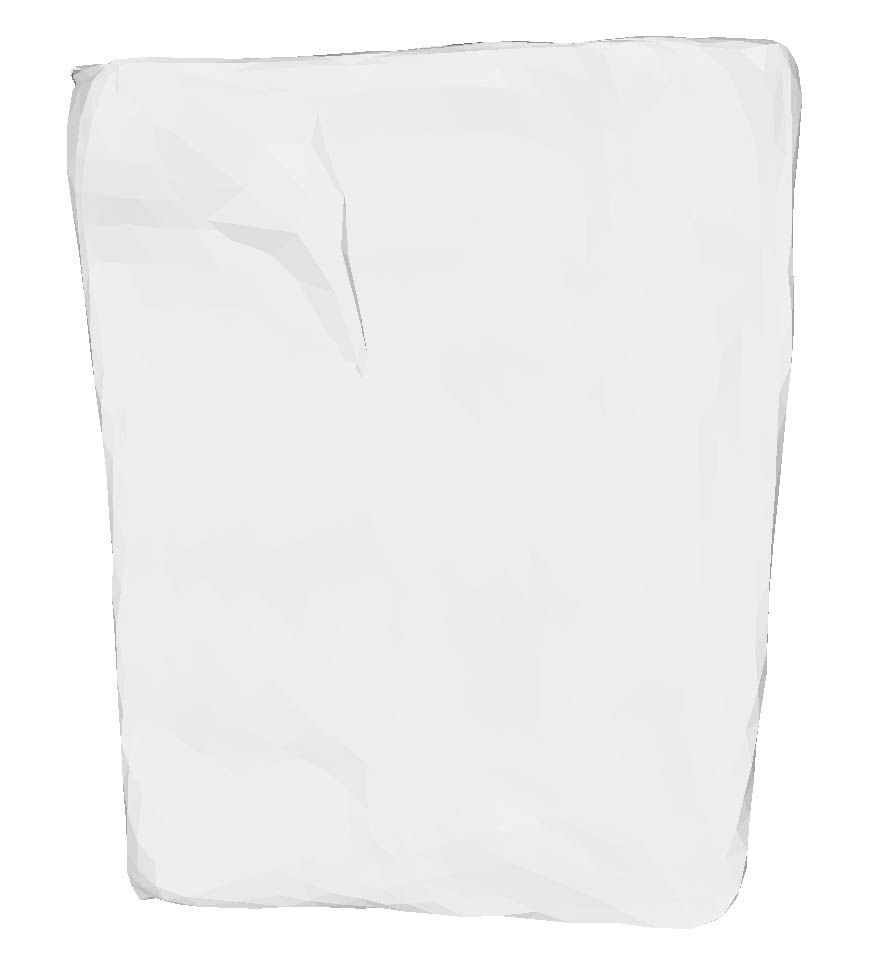} &
	    \squa{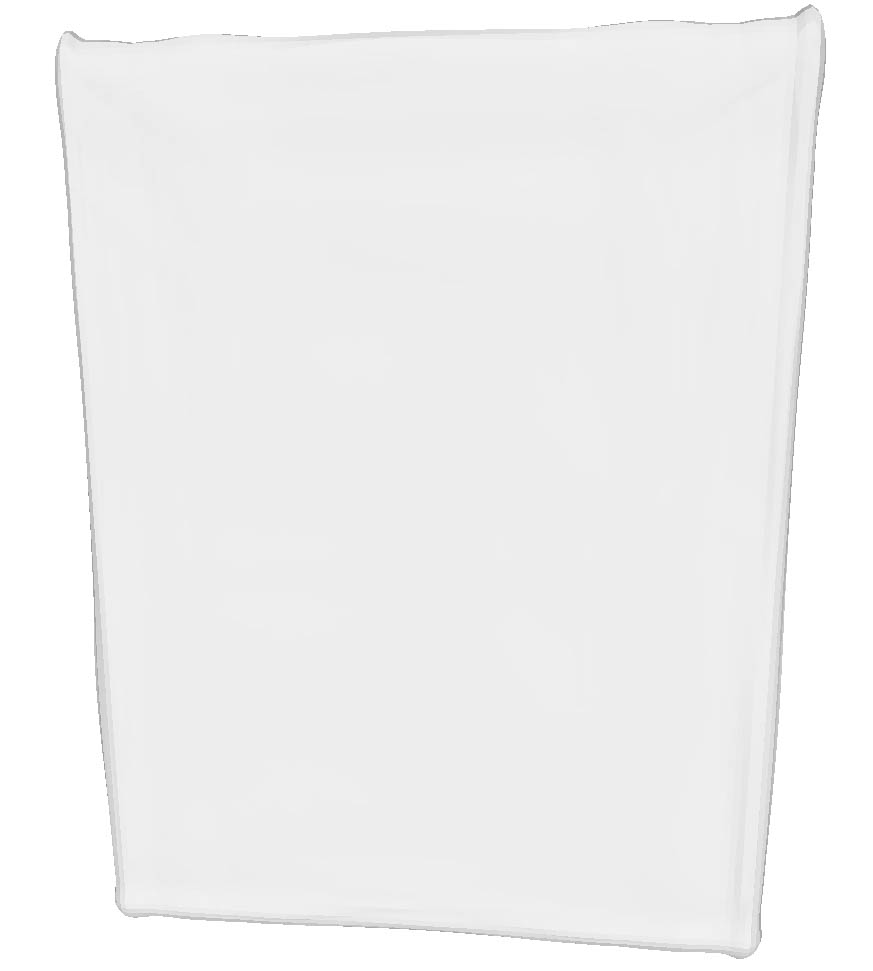} &
	    \hori{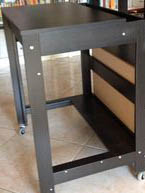} &
	    \hori{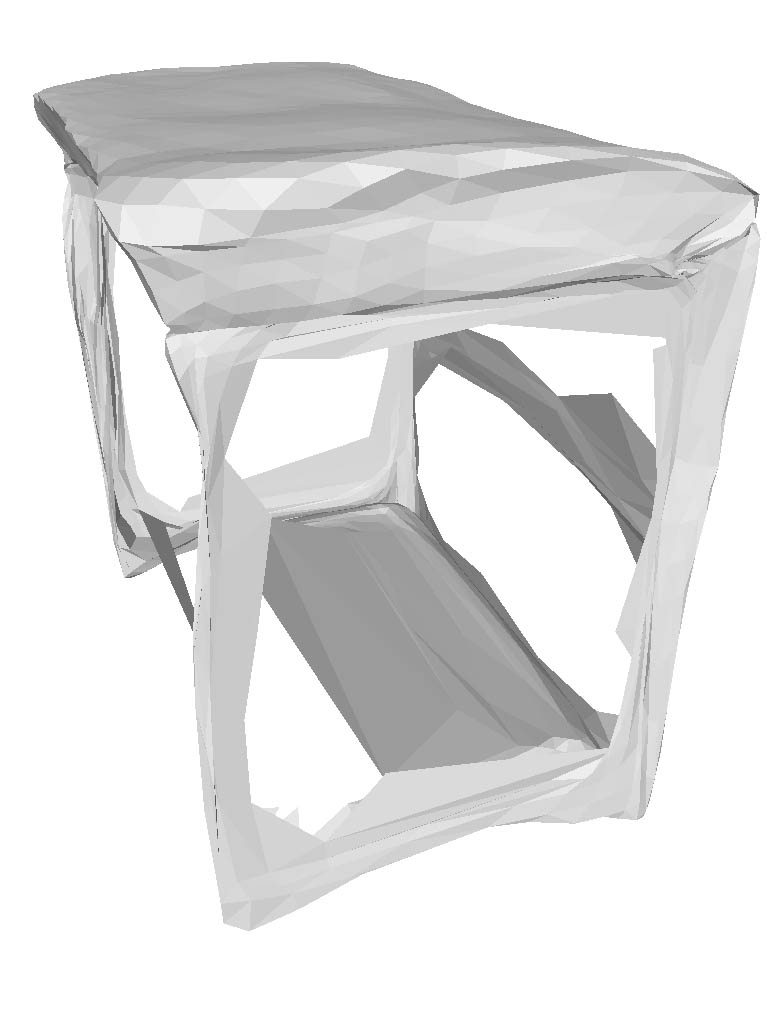} &
	    \hori{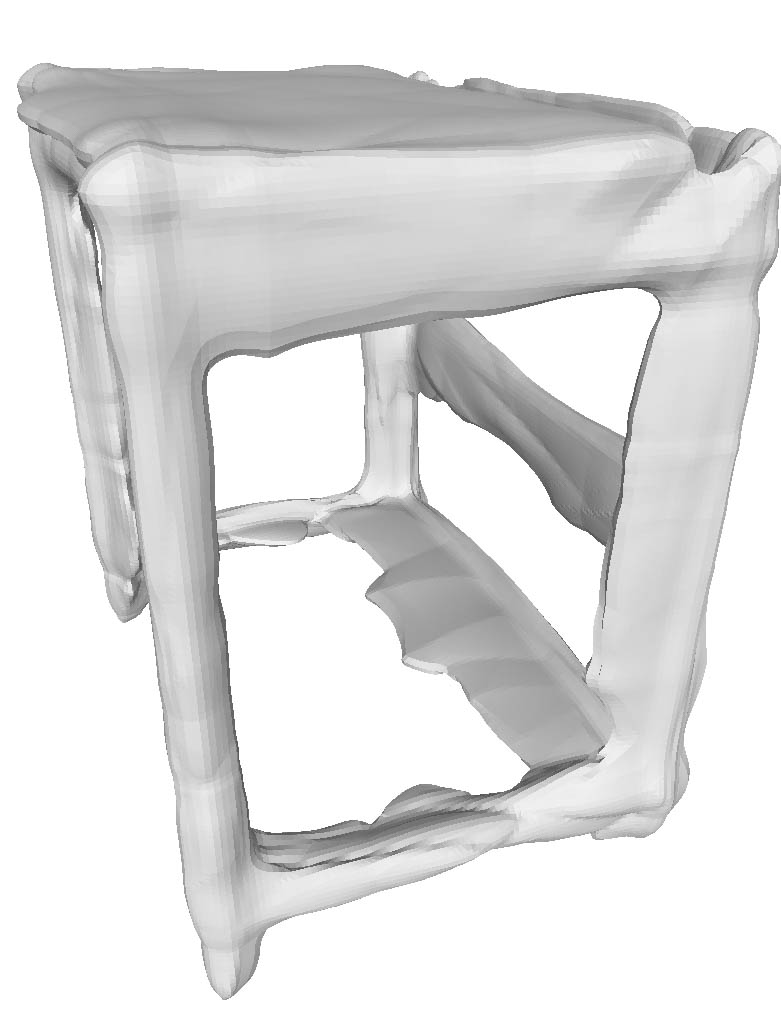} \\
	    
	    \verti{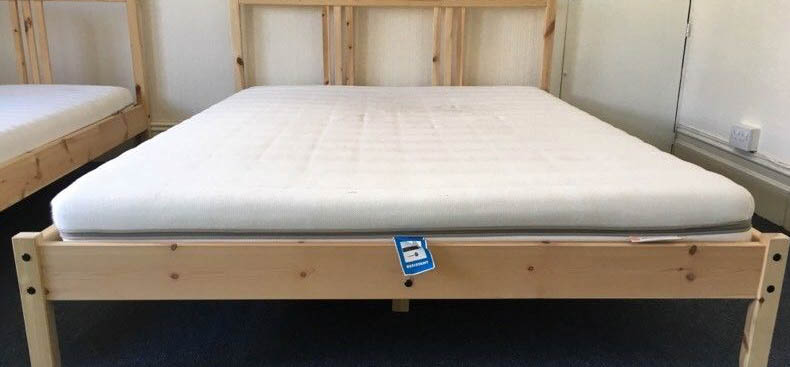} &
	    \verti{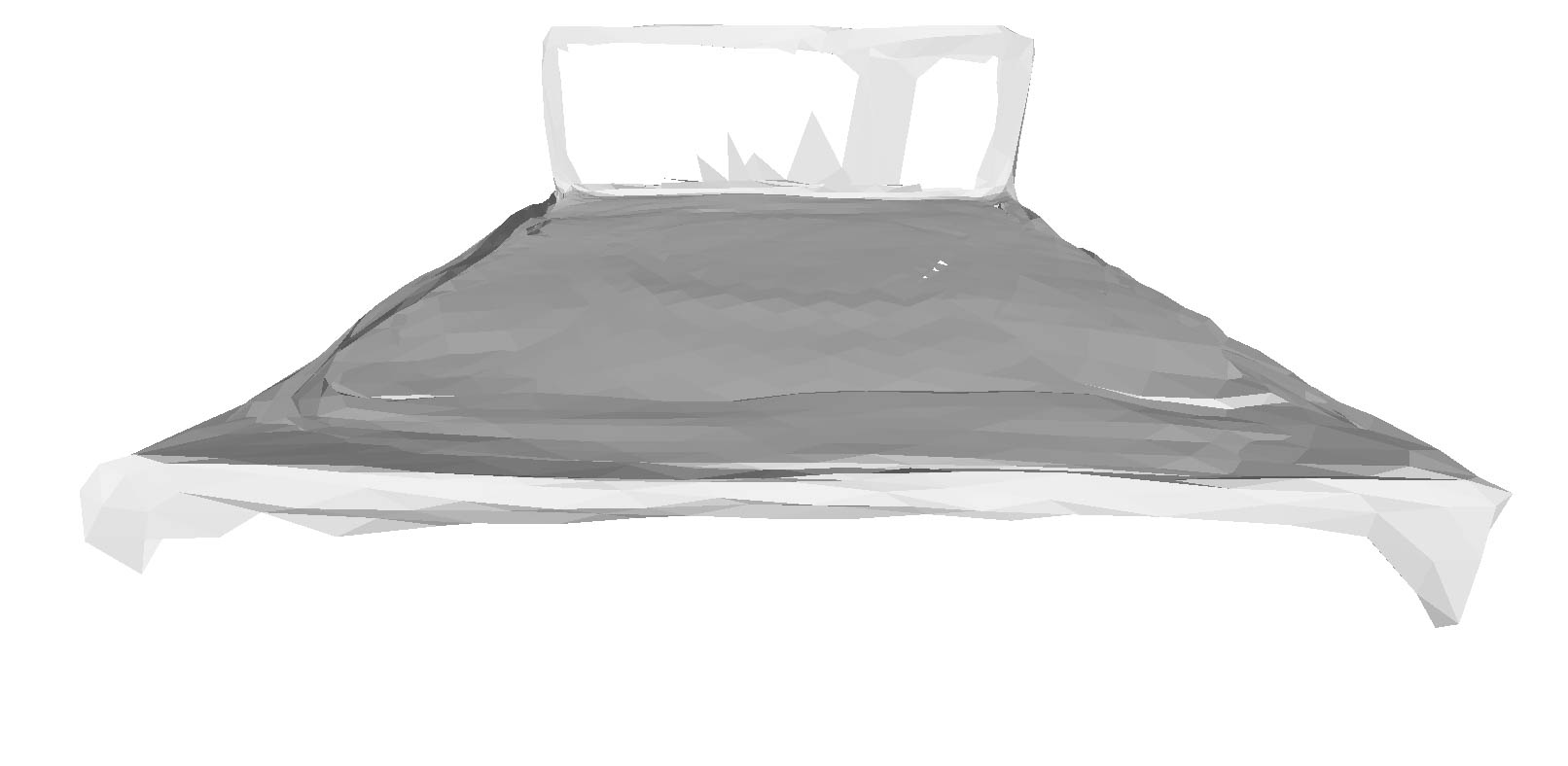} &
	    \verti{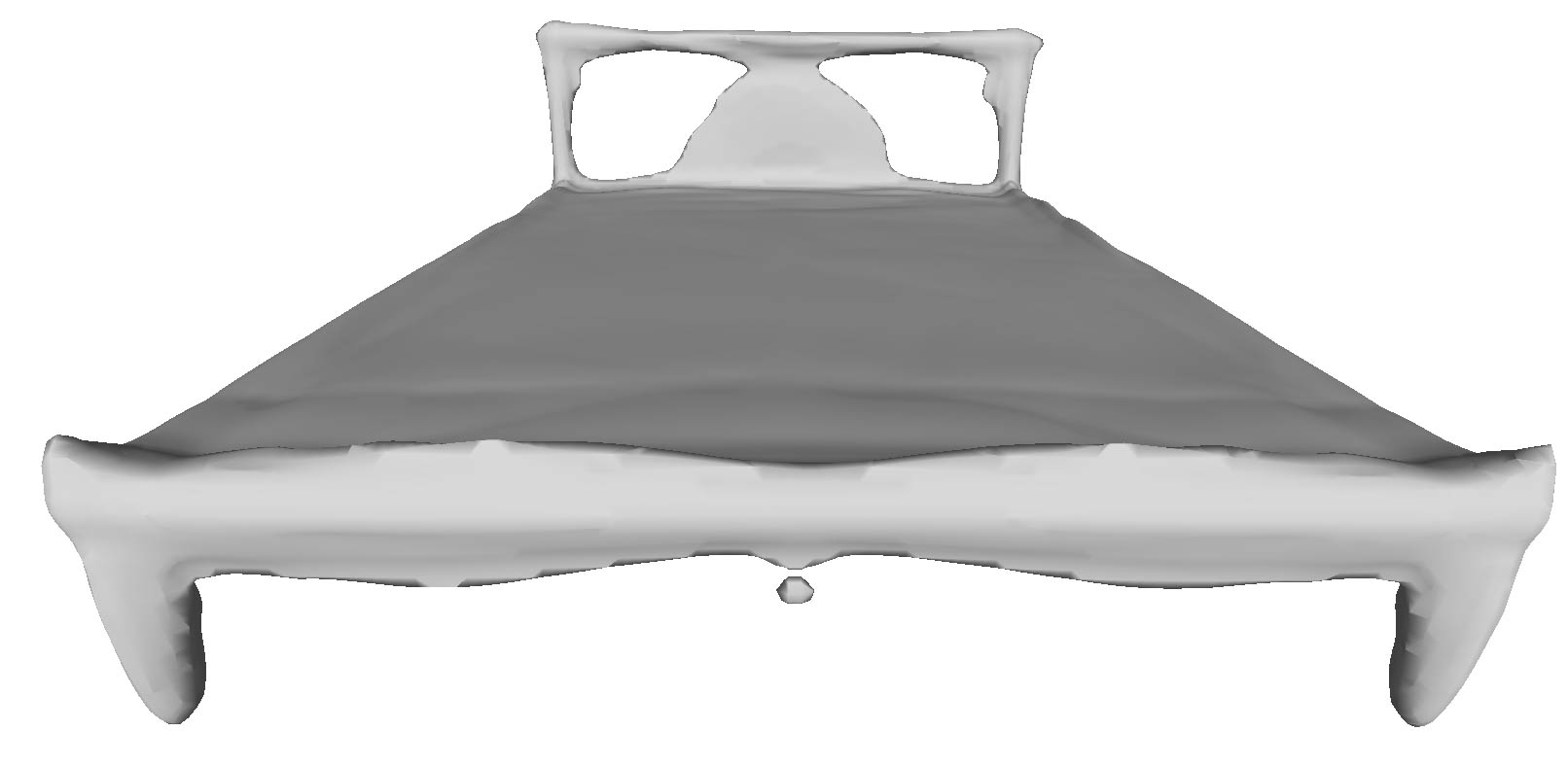} &
	    \squa{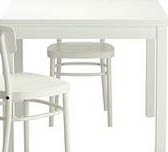} &
	    \squa{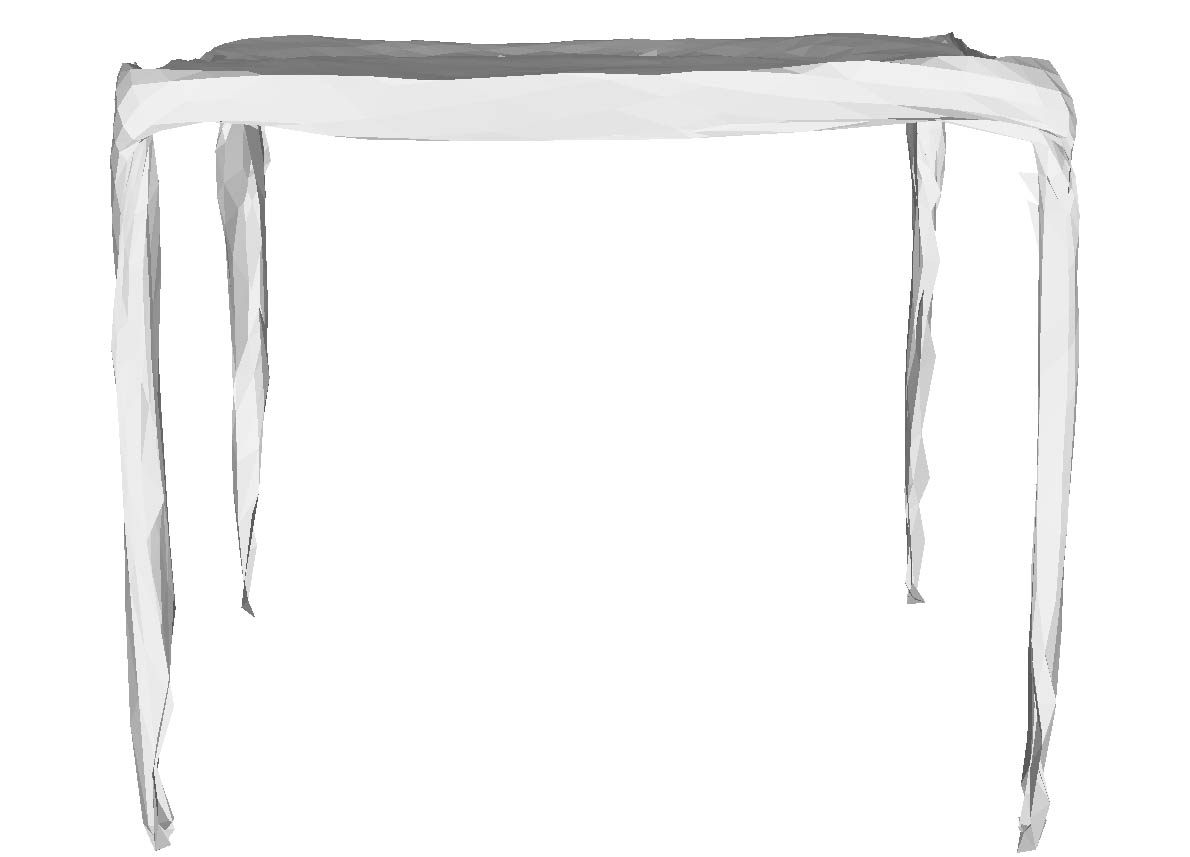} &
	    \squa{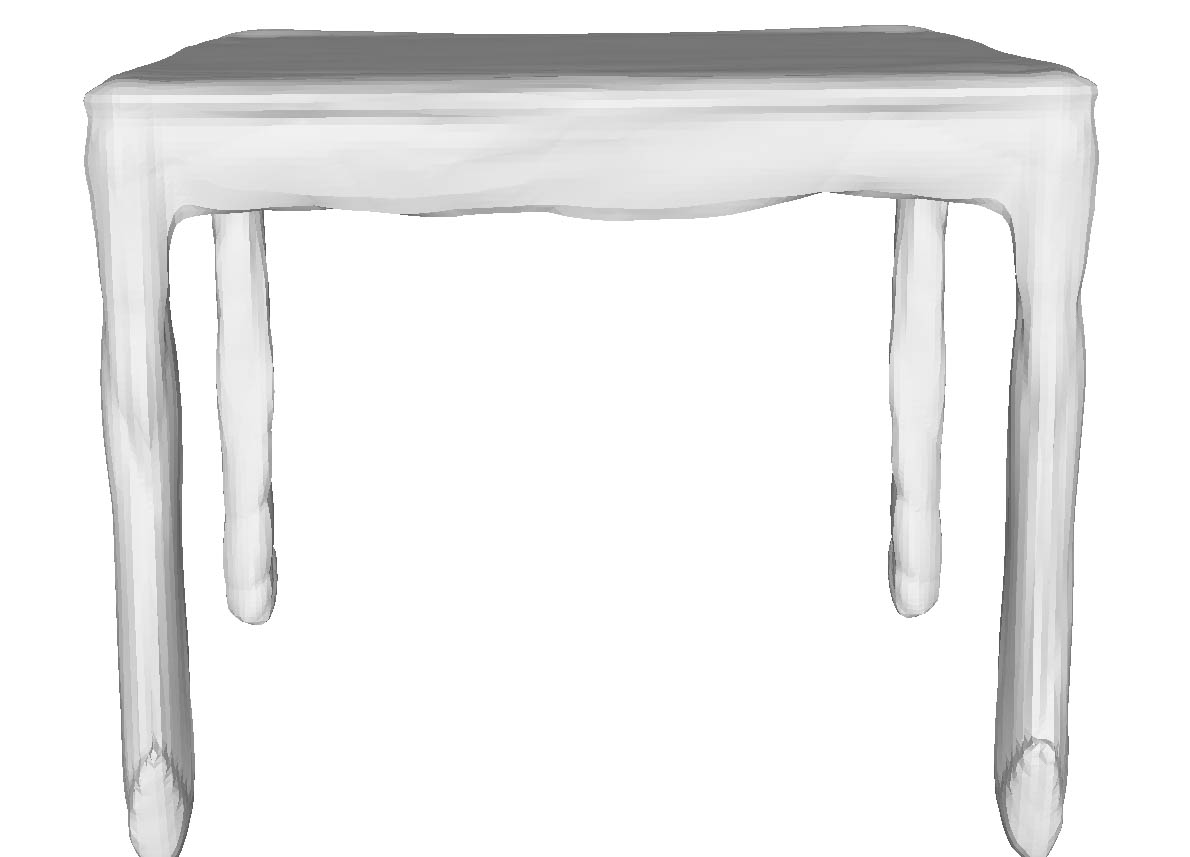} &
	    \hori{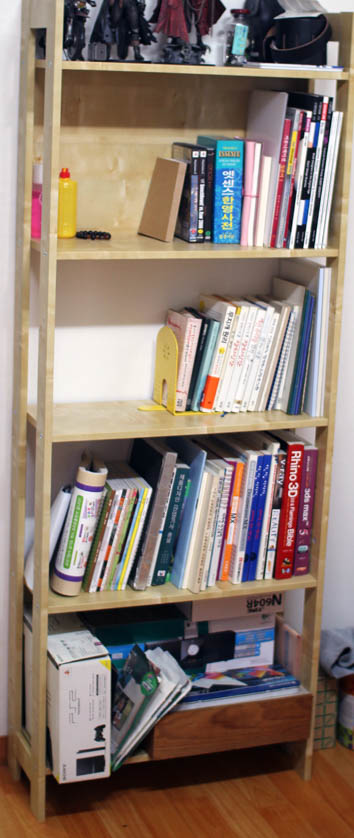} &
	    \hori{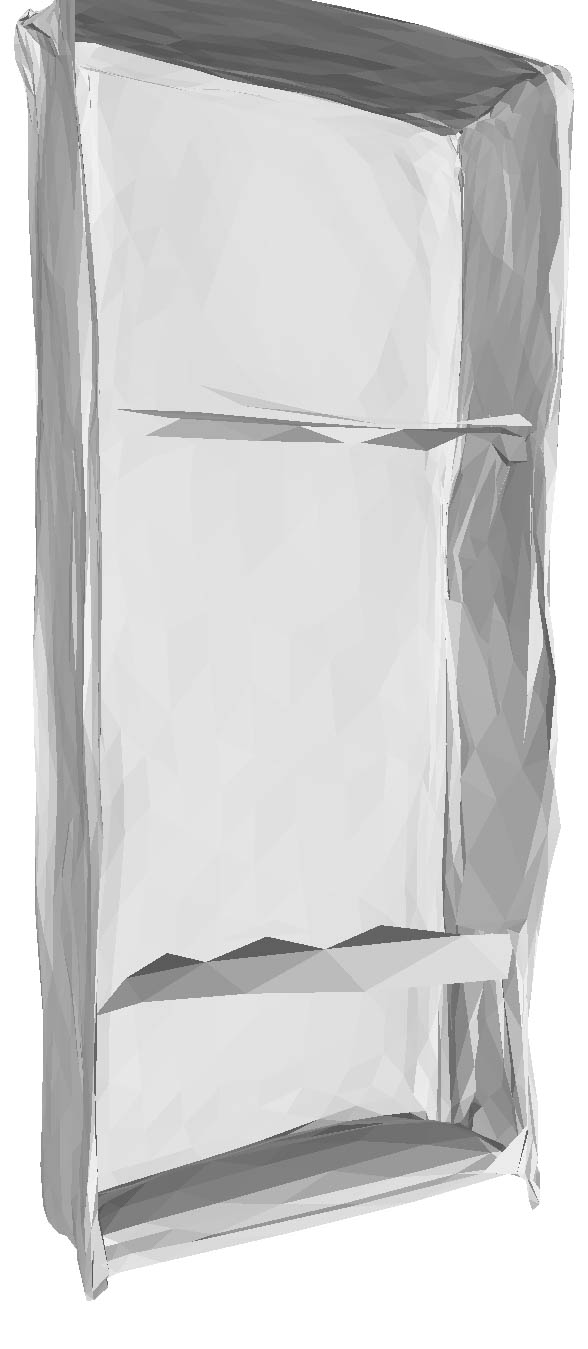} &
	    \hori{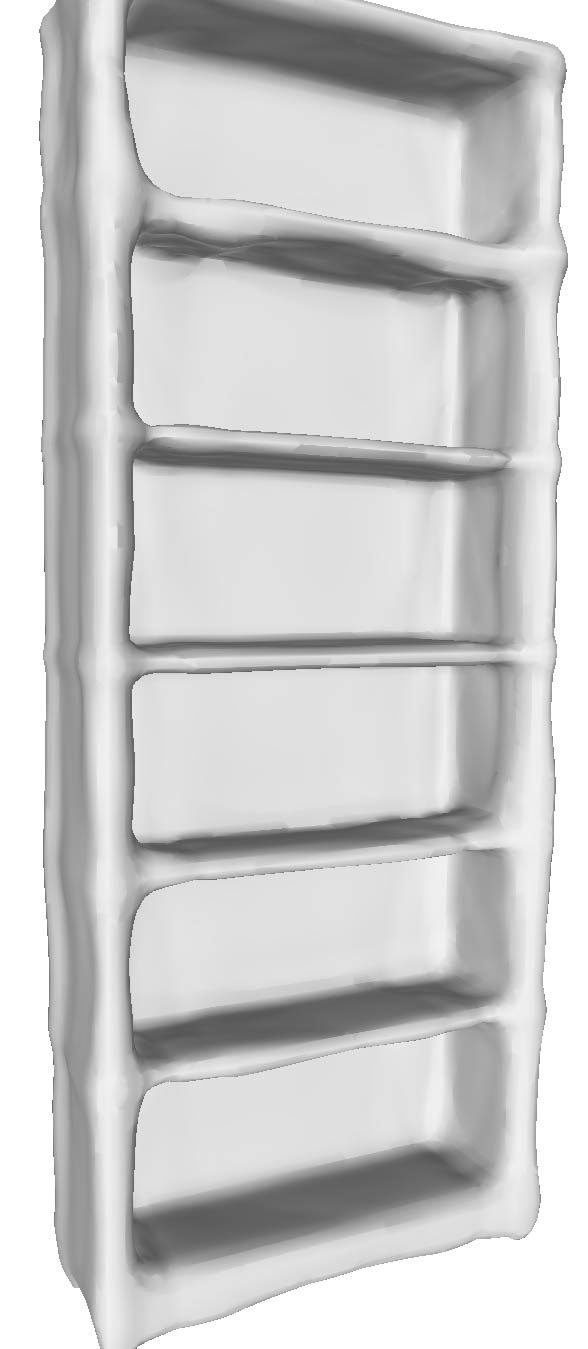} \\
        
	    \verti{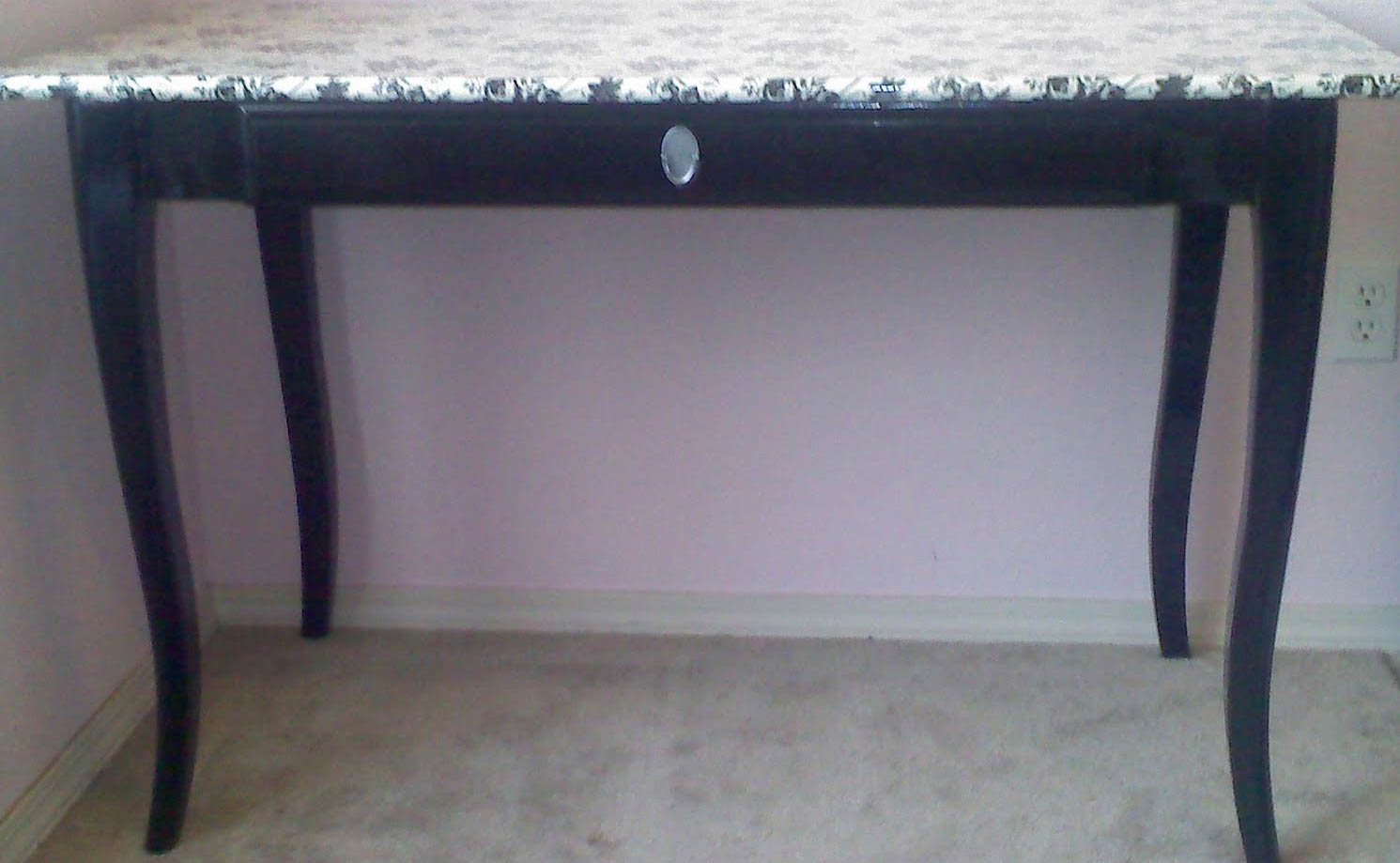} &
	    \verti{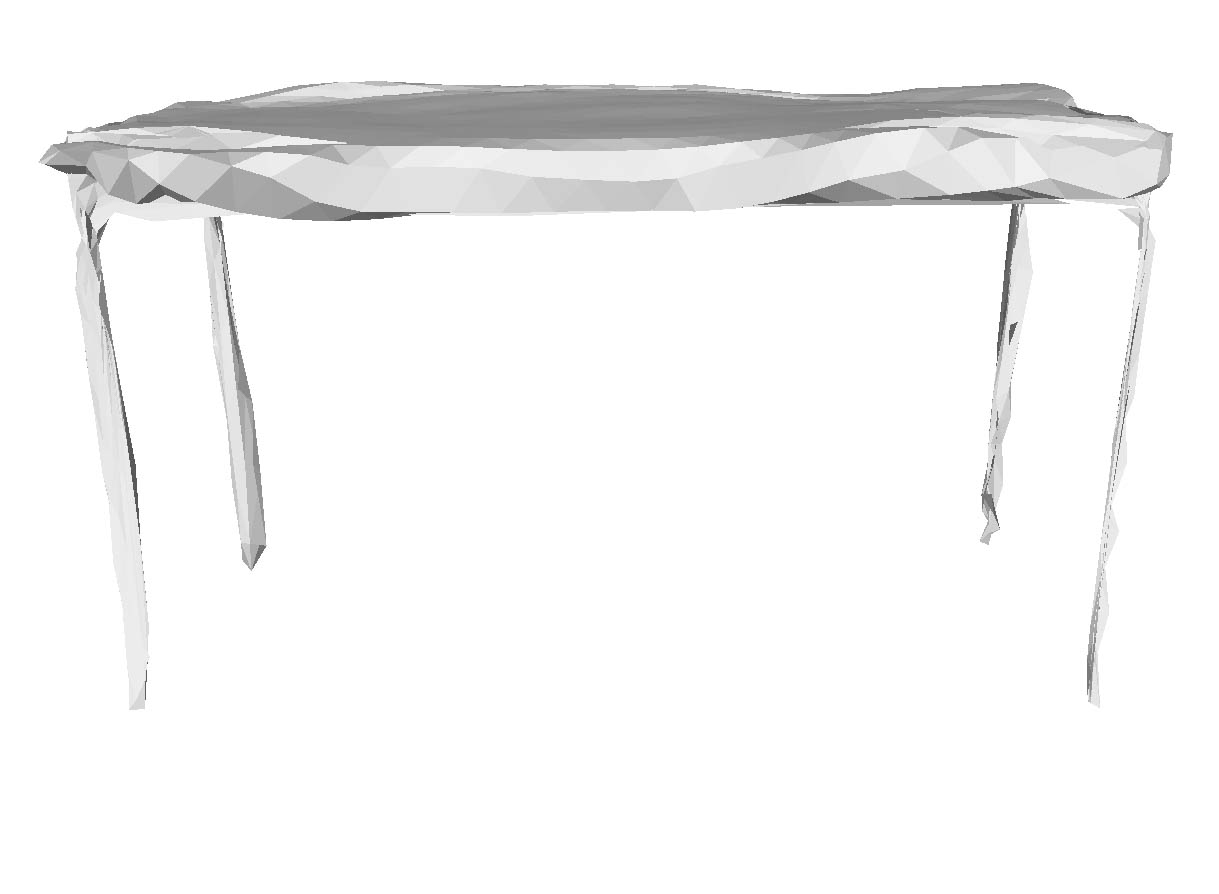} &
	    \verti{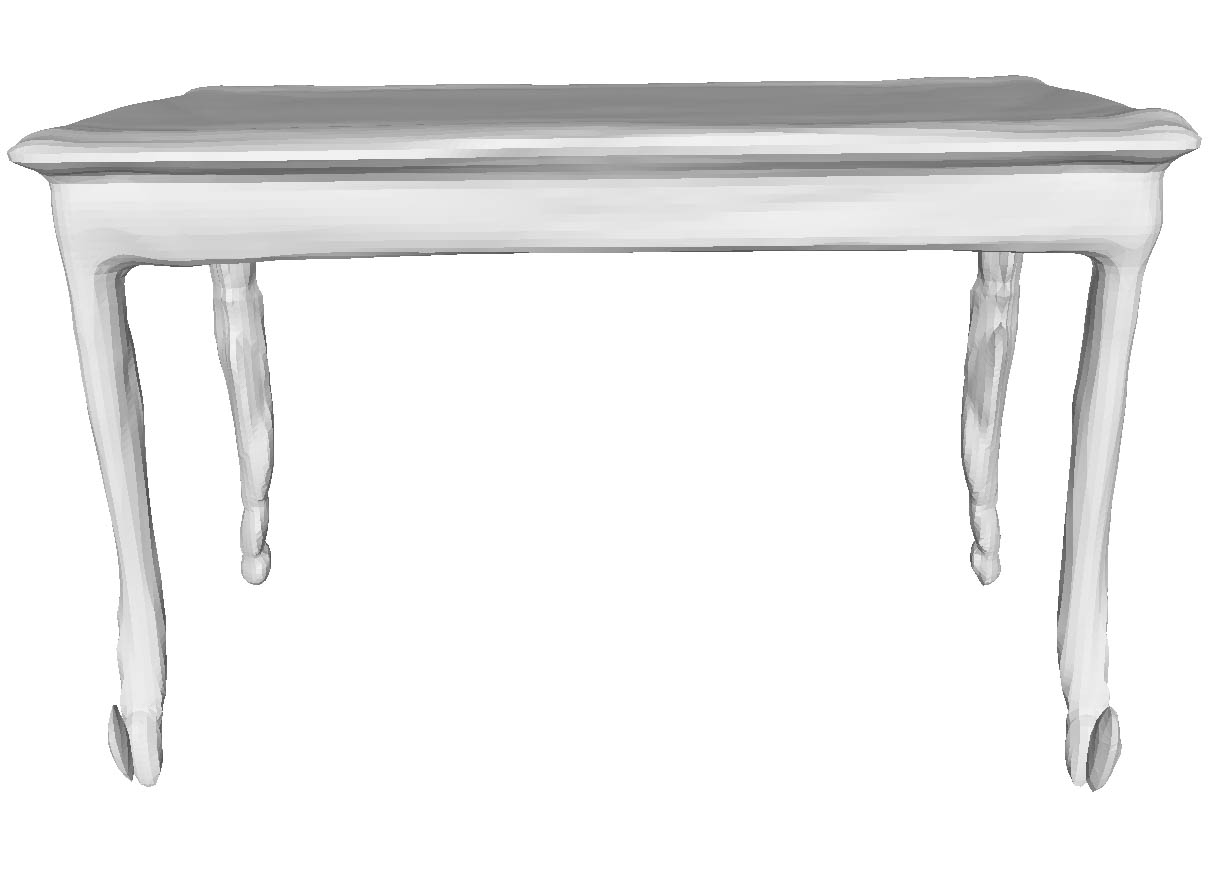} &
	    \squa{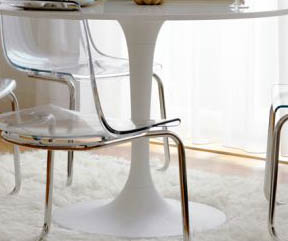} &
	    \squa{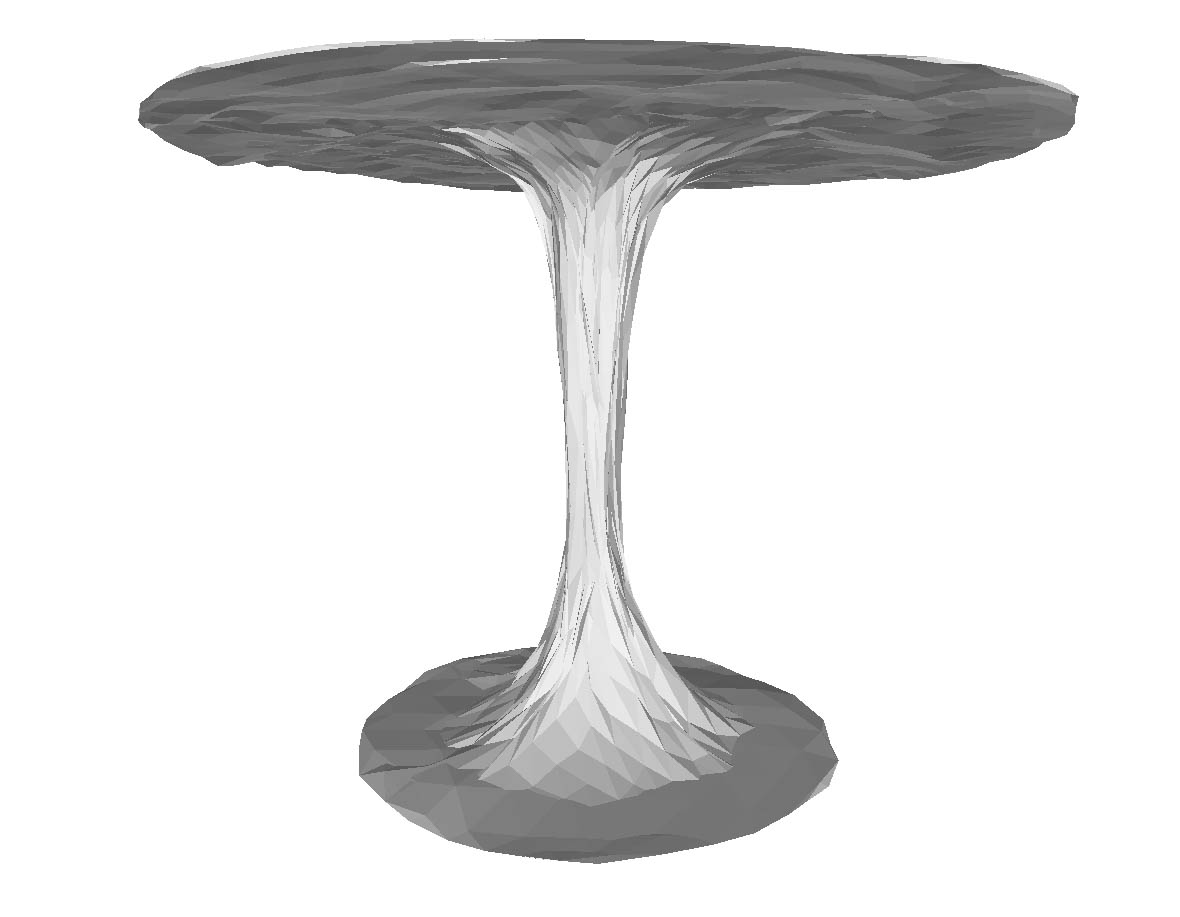} &
	    \squa{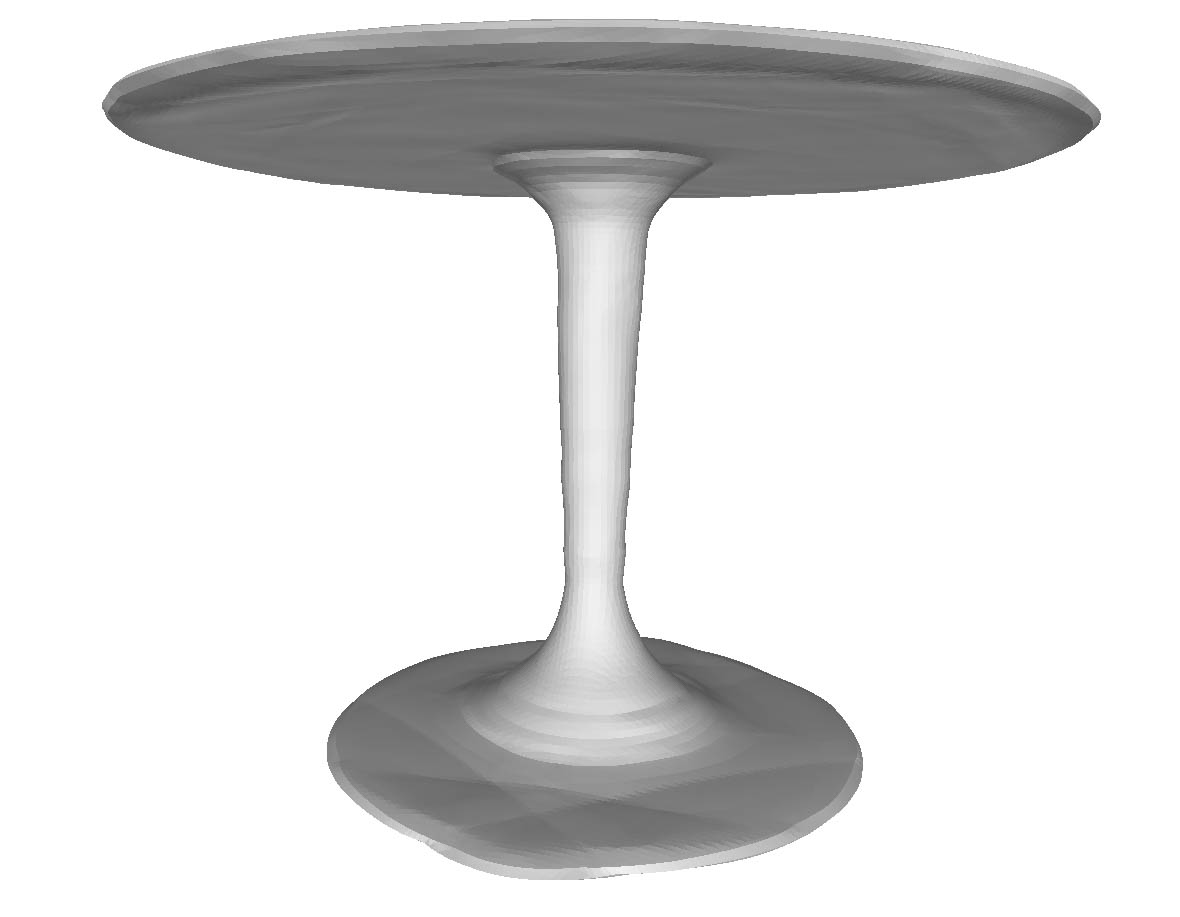} &
	    \hori{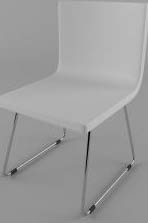} &
	    \hori{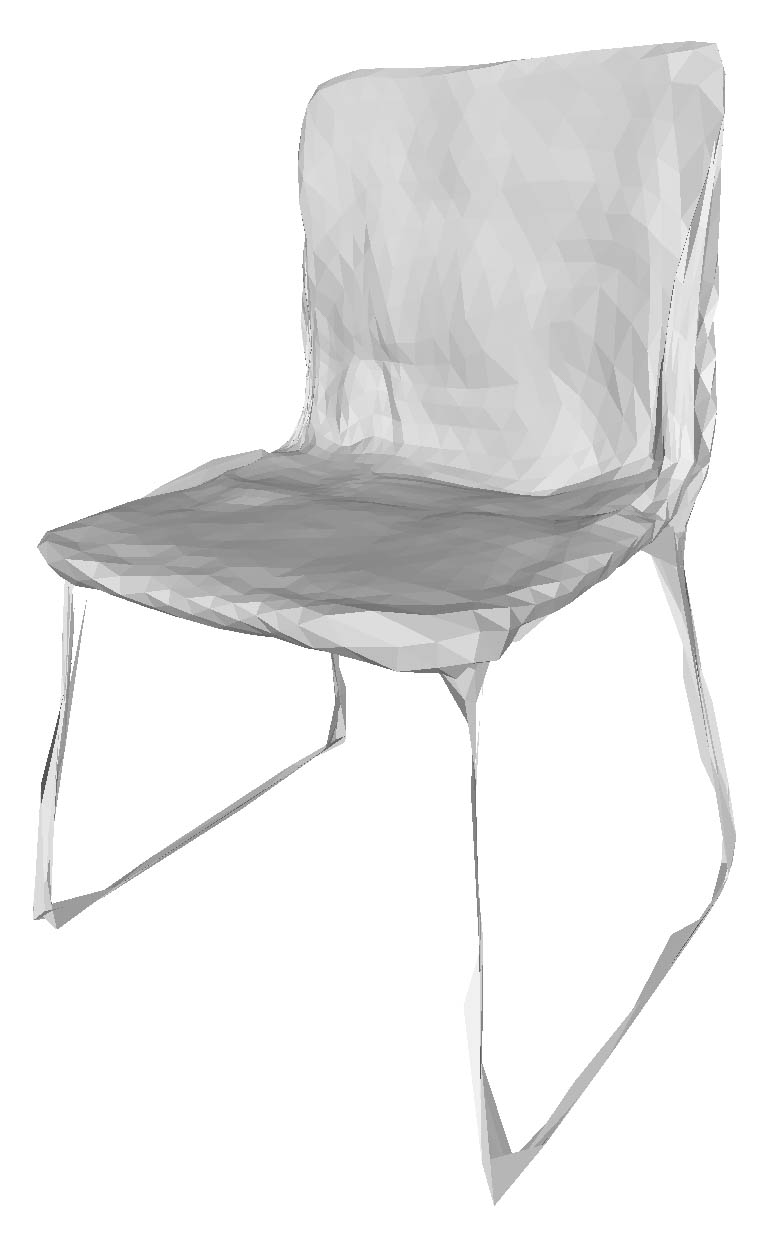} &
	    \hori{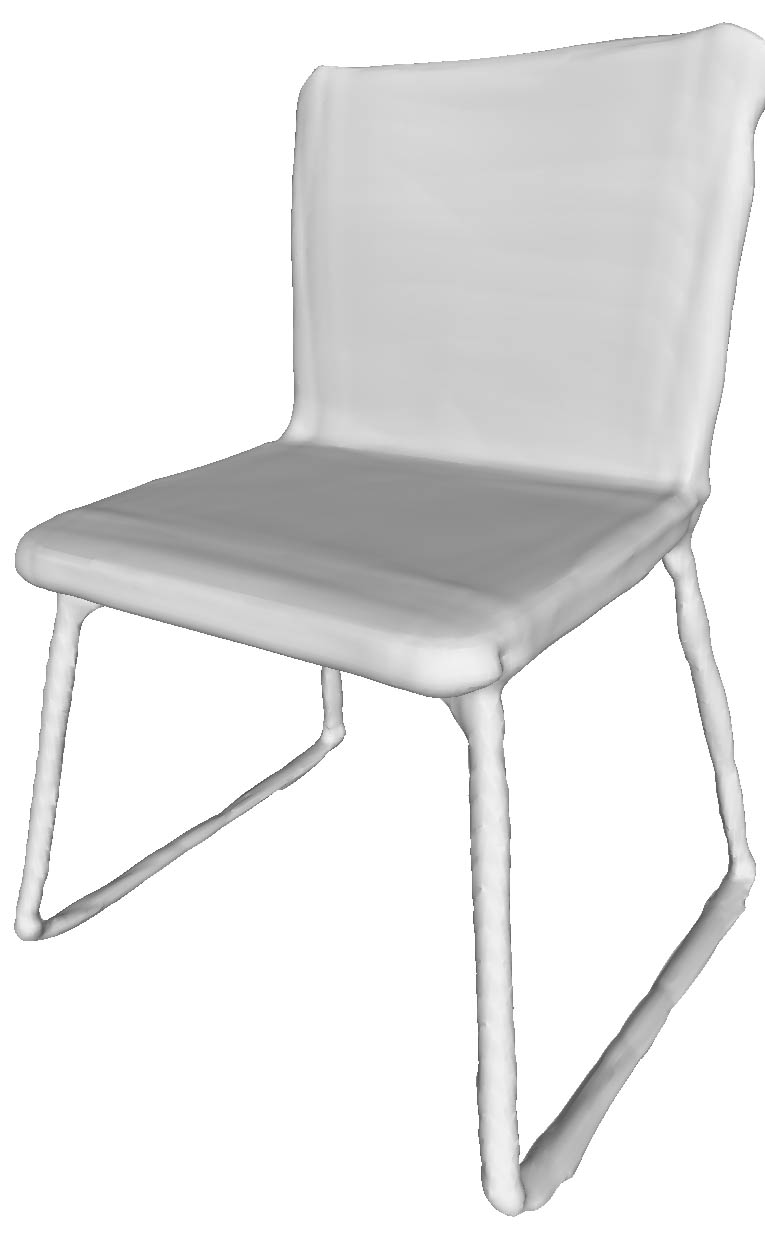} \\
	    \vspace{5px}
	    \verti{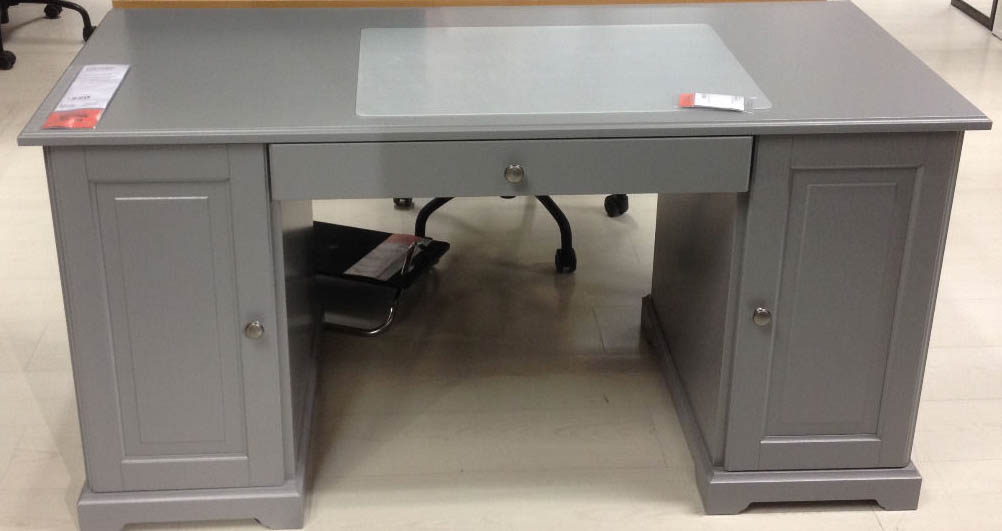} &
	    \verti{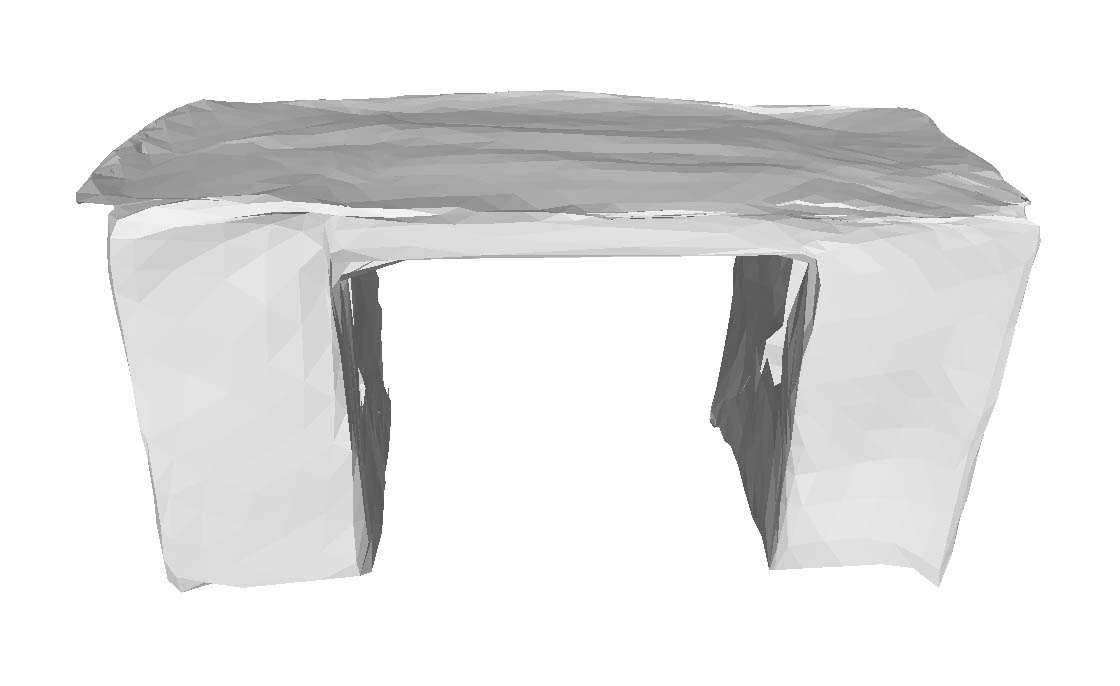} &
	    \verti{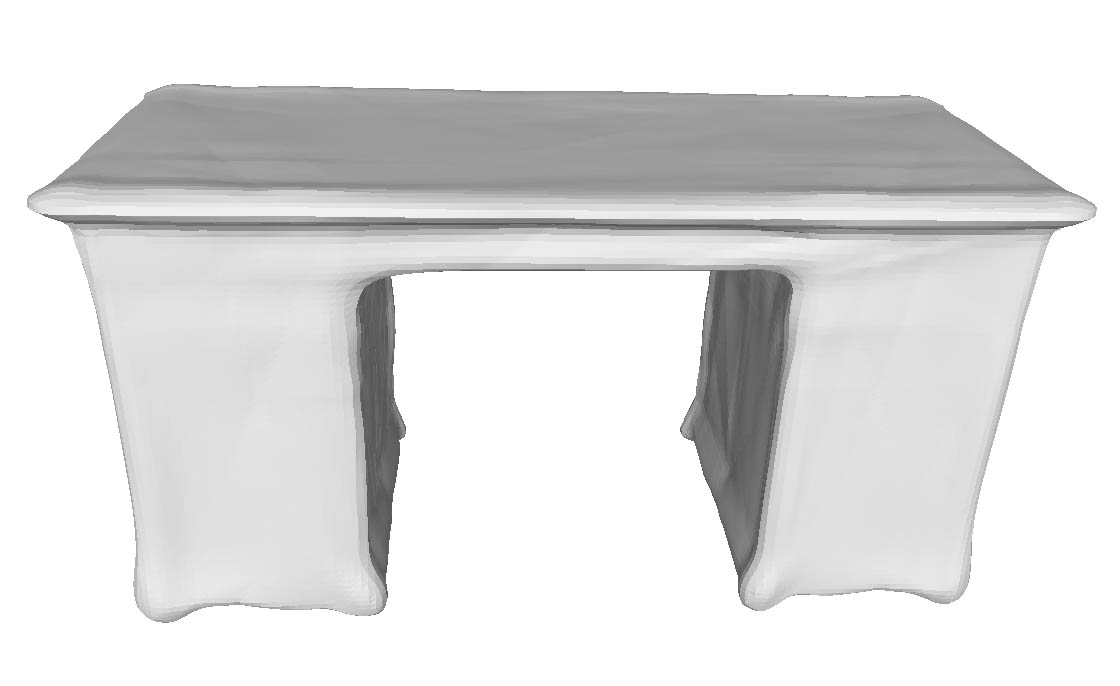} &
	    \squa{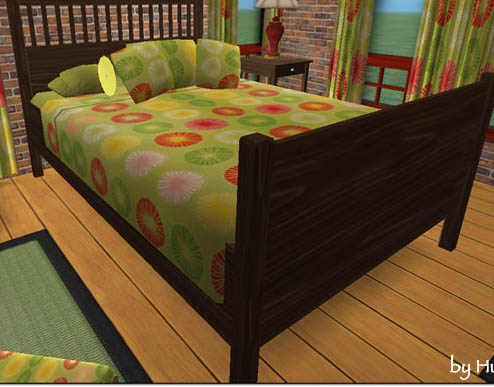} &
	    \squa{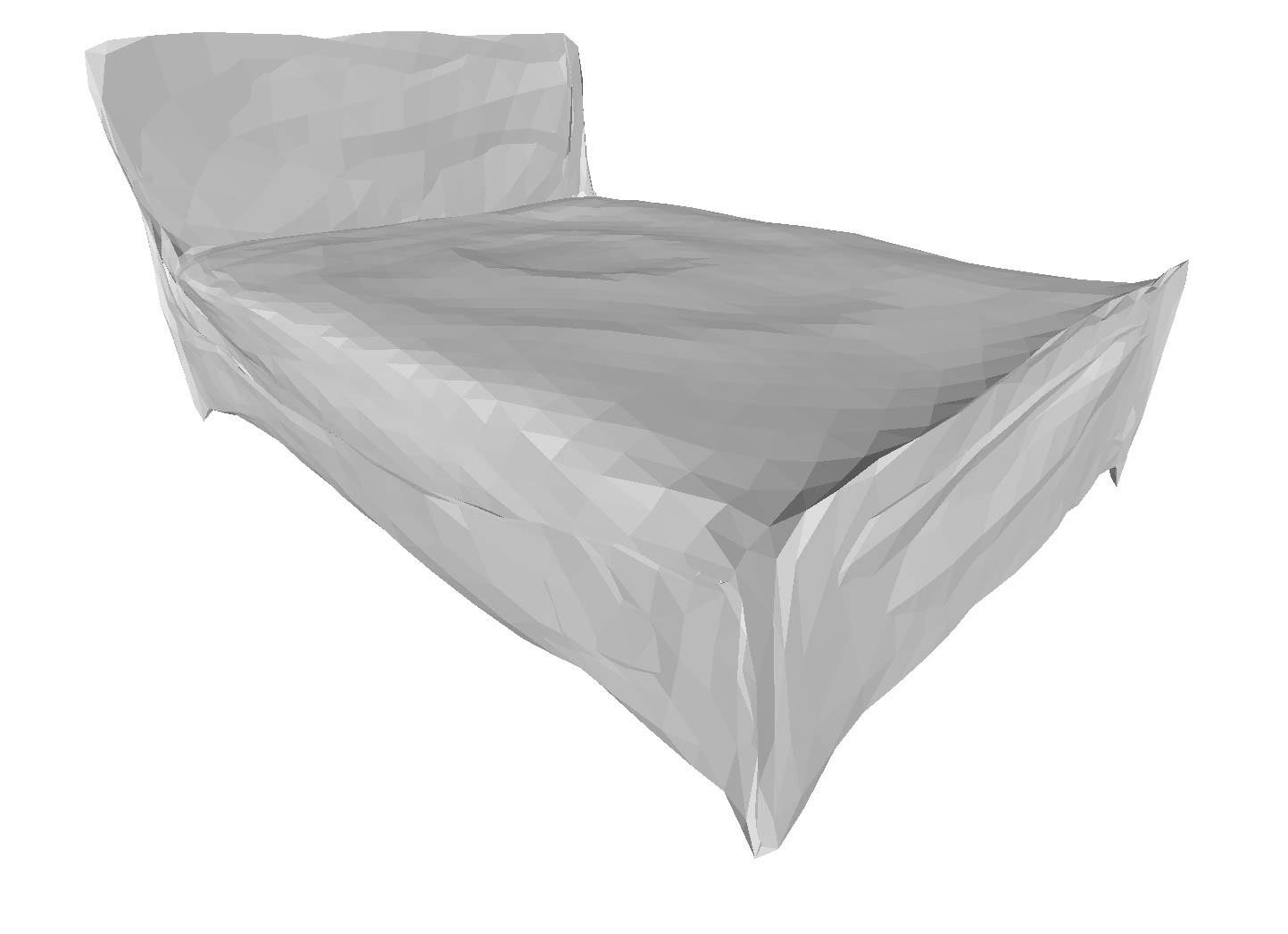} &
	    \squa{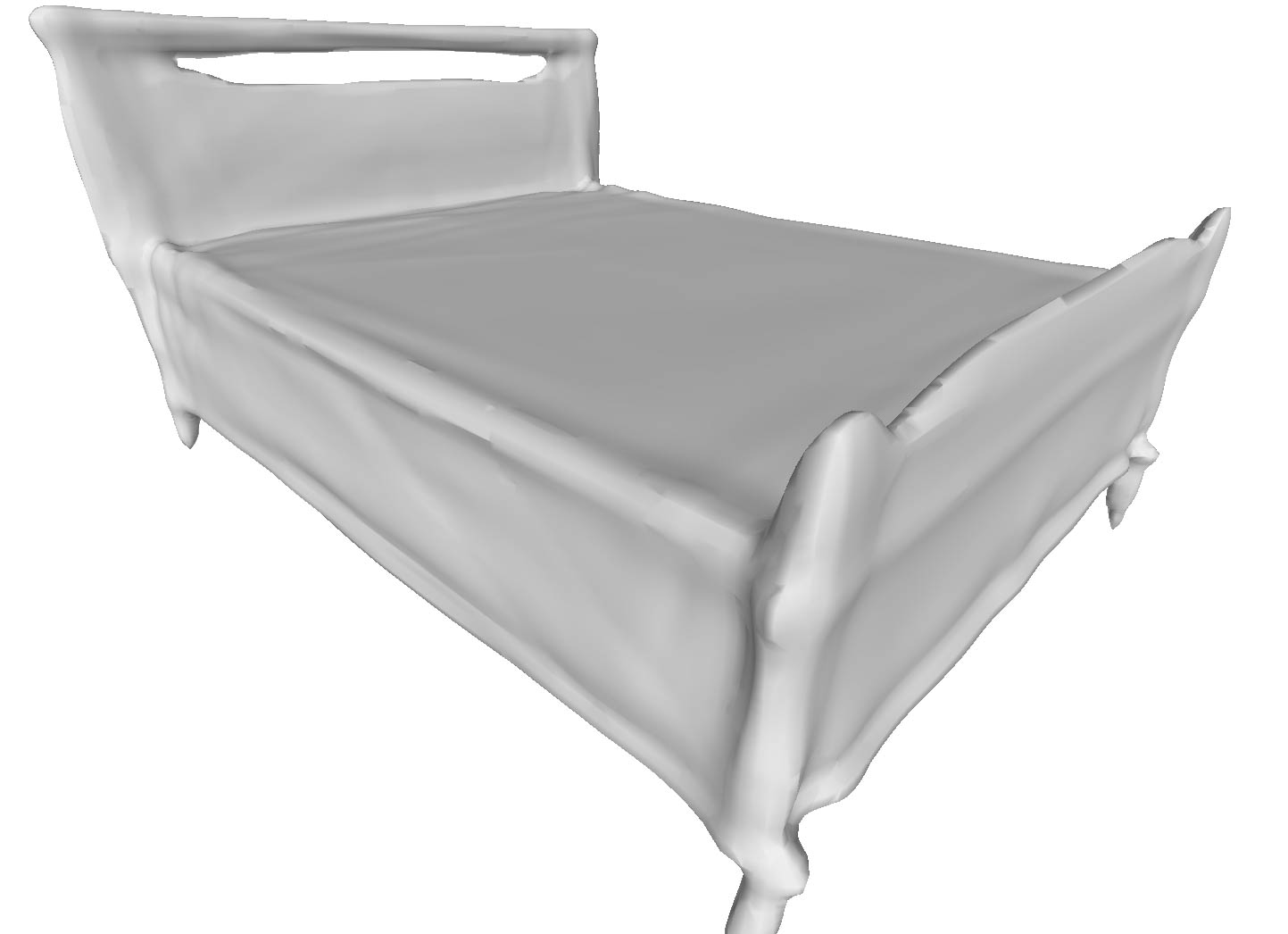} &
	    \hori{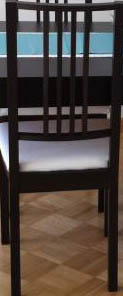} &
	    \hori{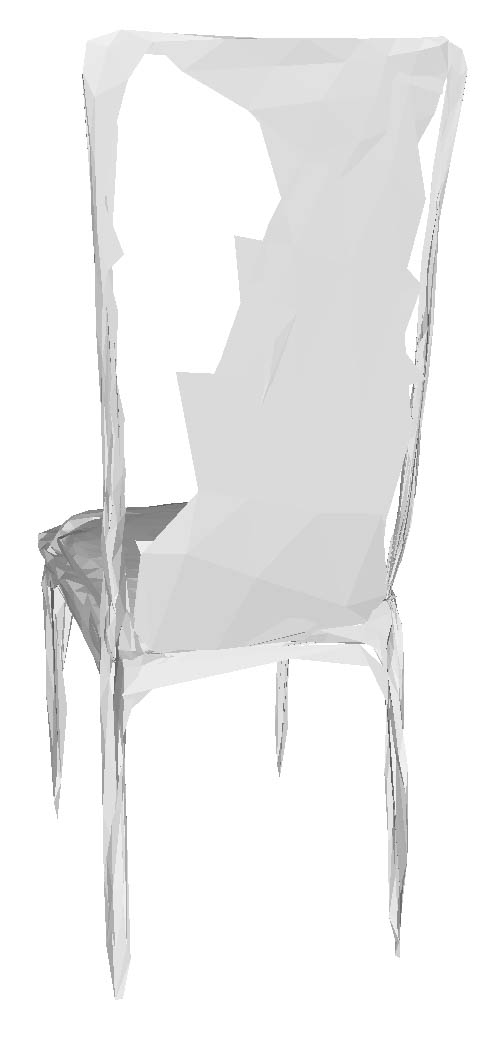} &
	    \hori{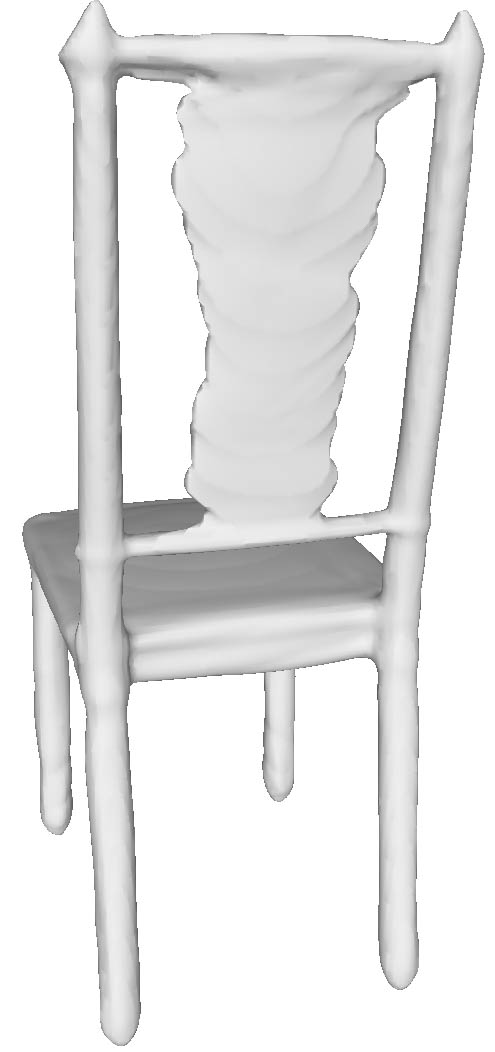} \\
	    \vspace{5px}
	    \verti{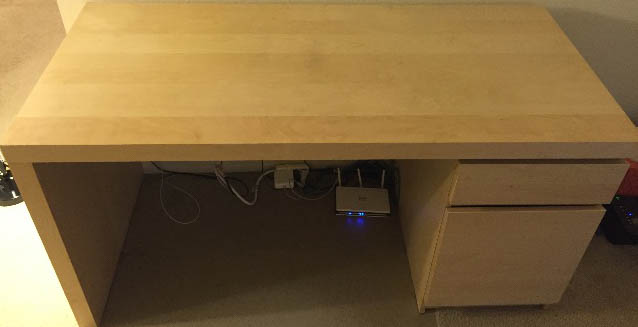} &
	    \verti{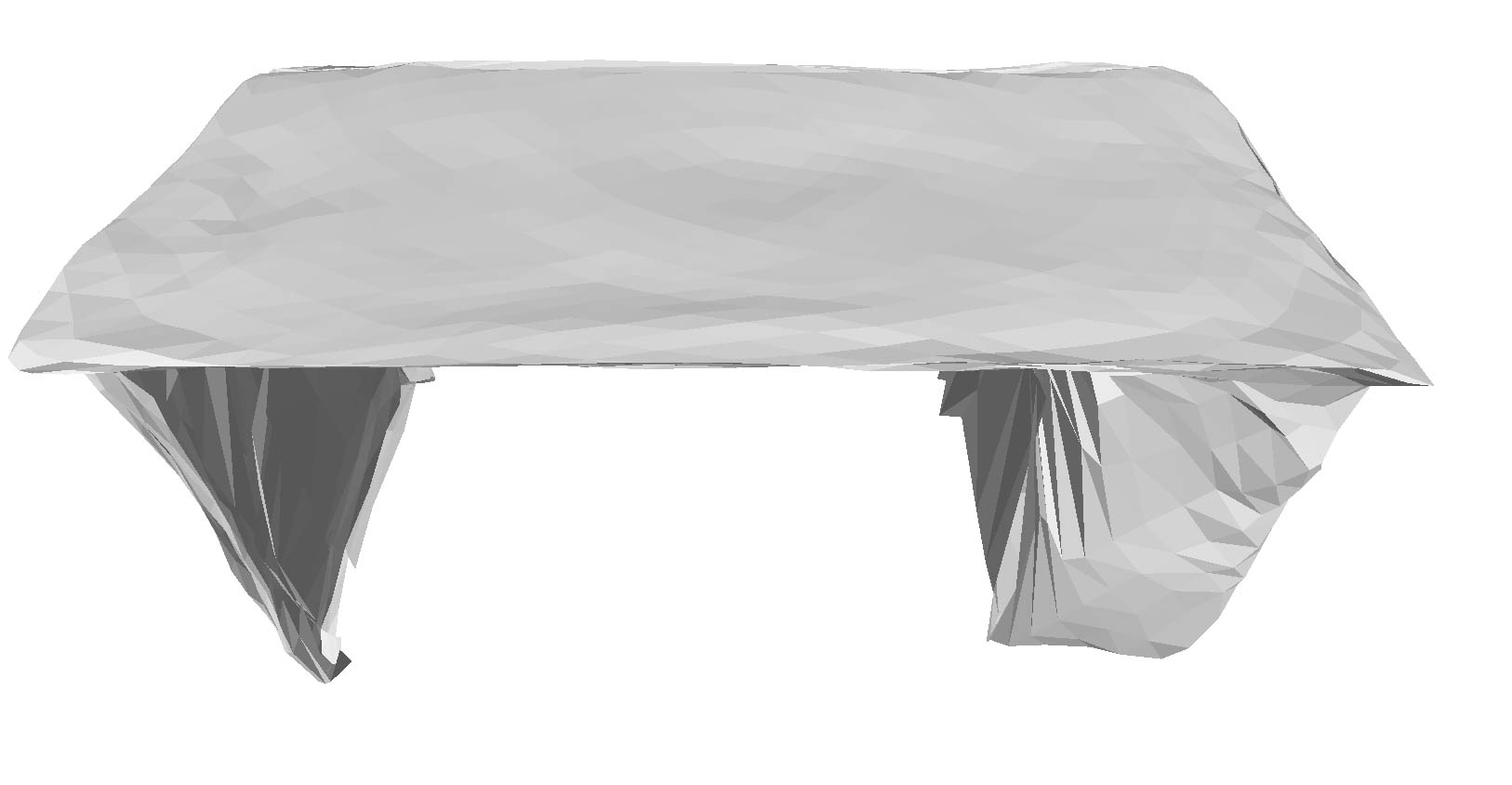} &
	    \verti{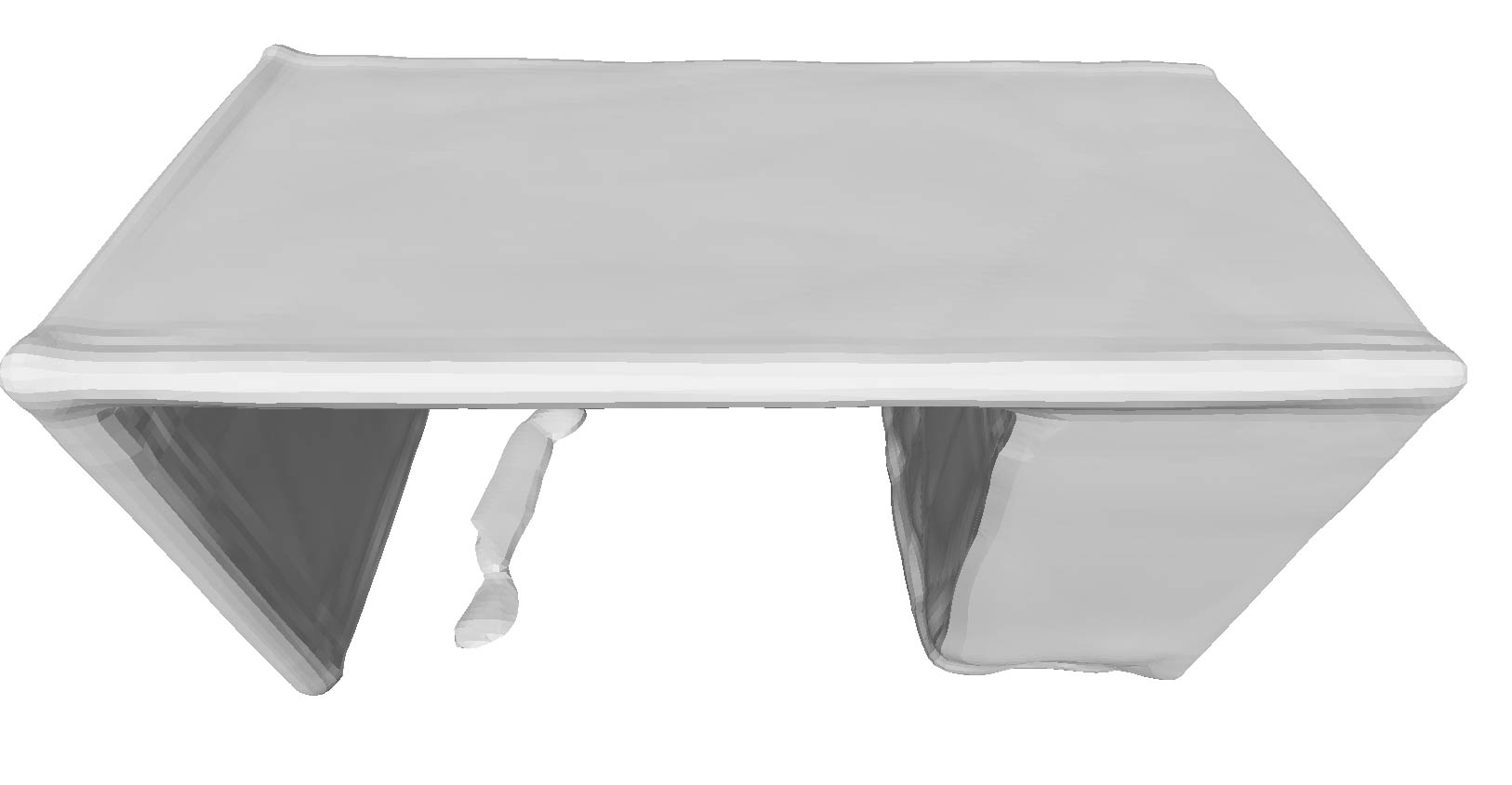} &
	    \squa{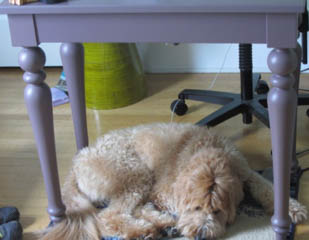} &
	    \squa{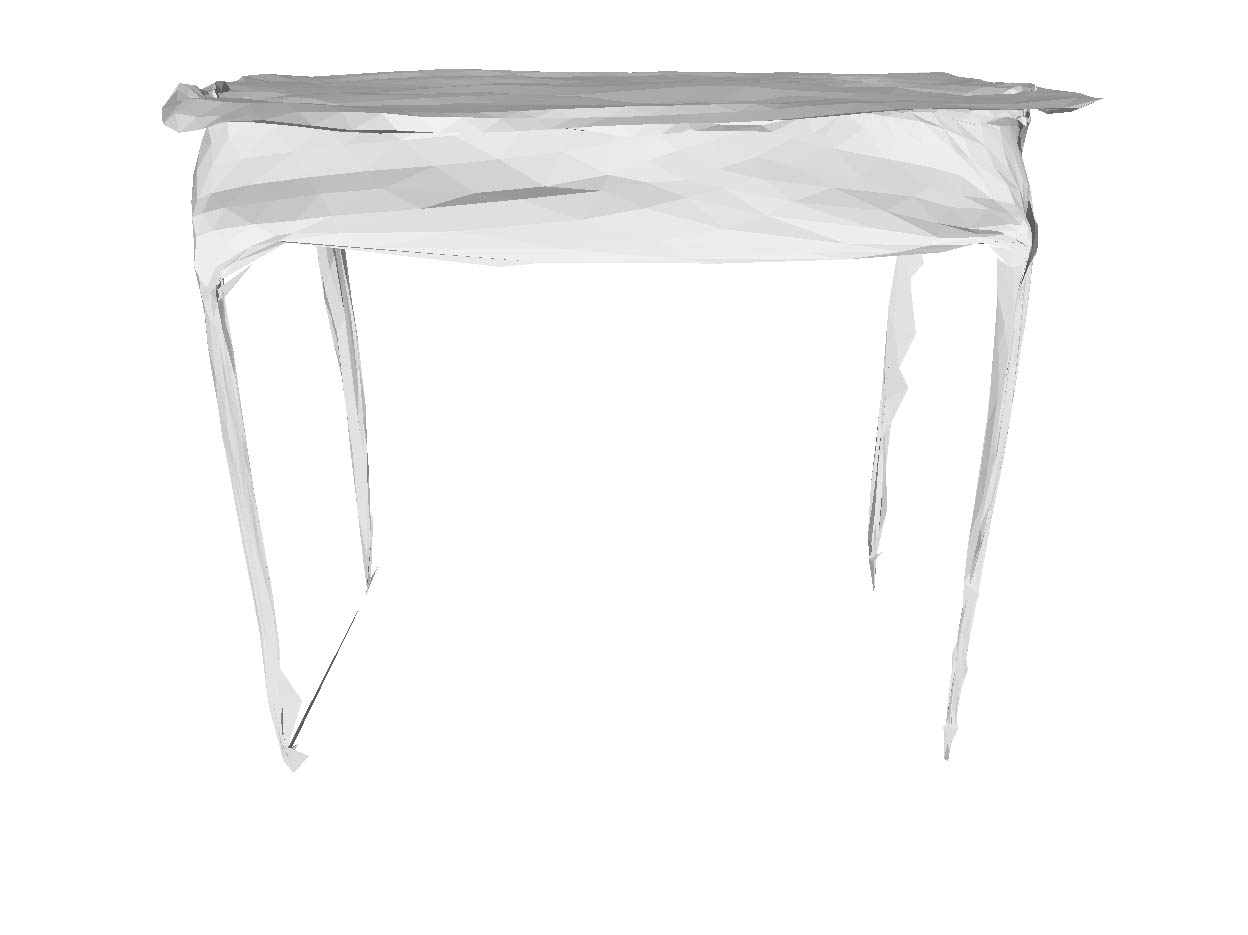} &
	    \squa{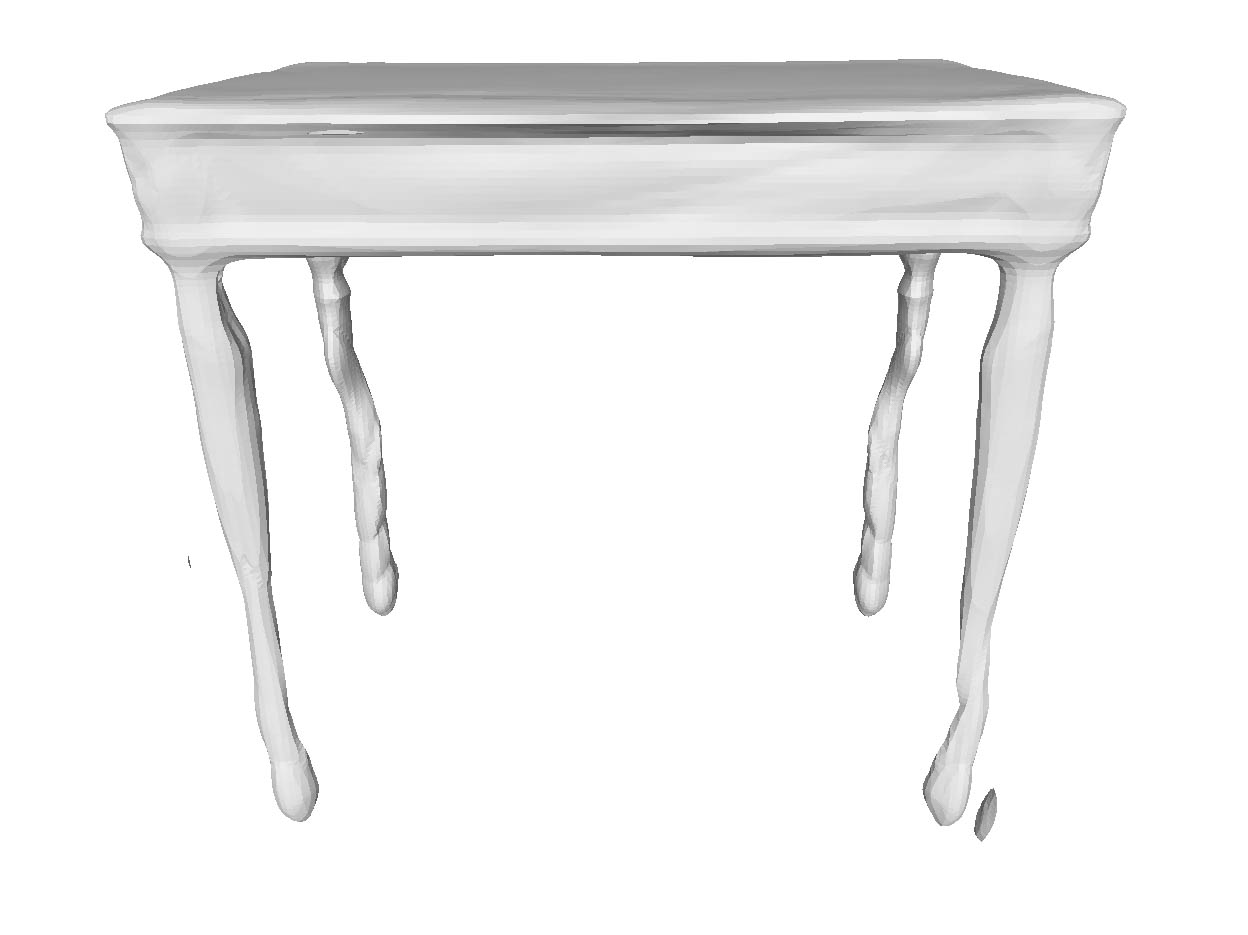} &
	    \hori{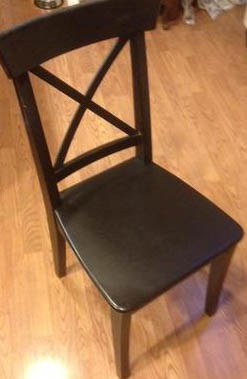} &
	    \hori{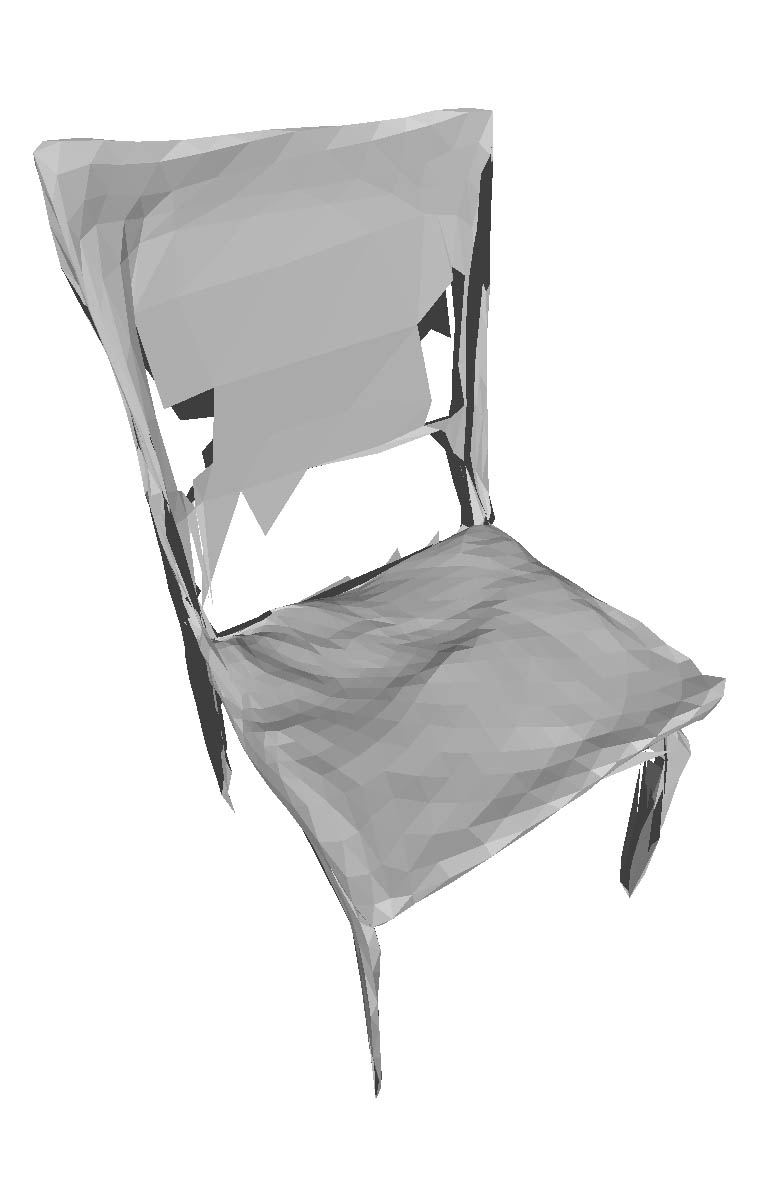} &
	    \hori{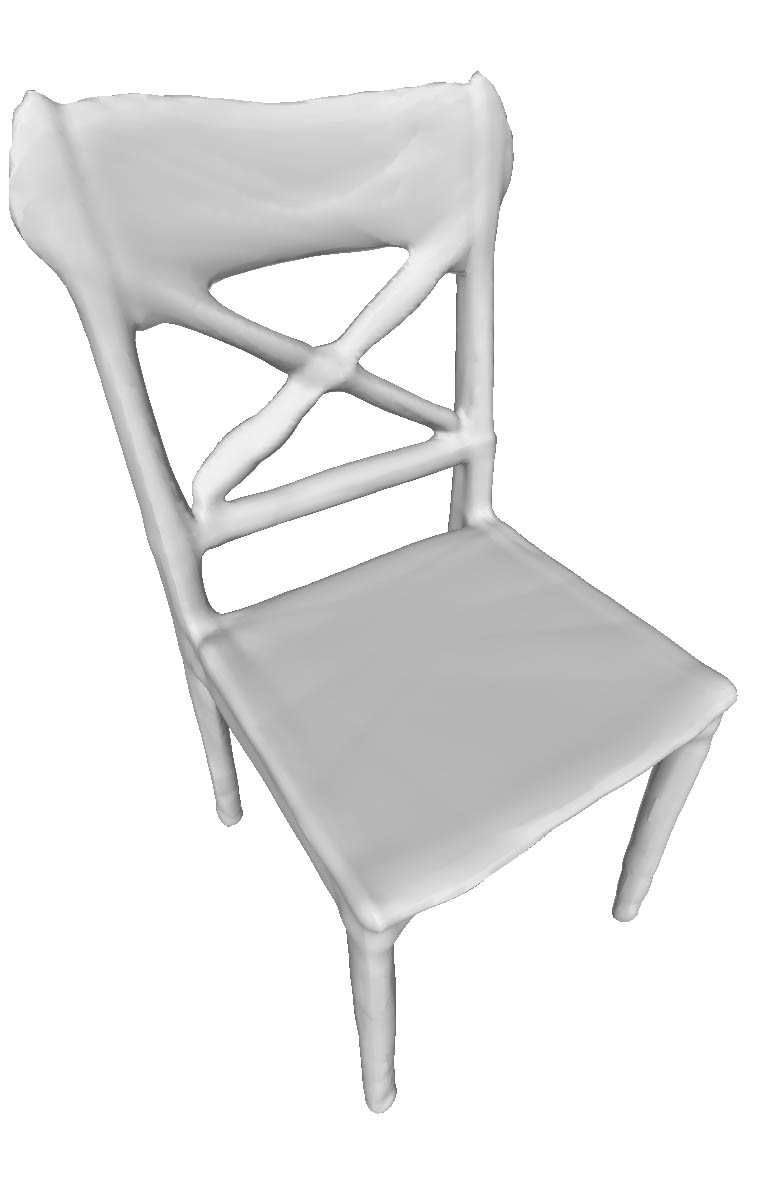} \\
	    
	    \verti{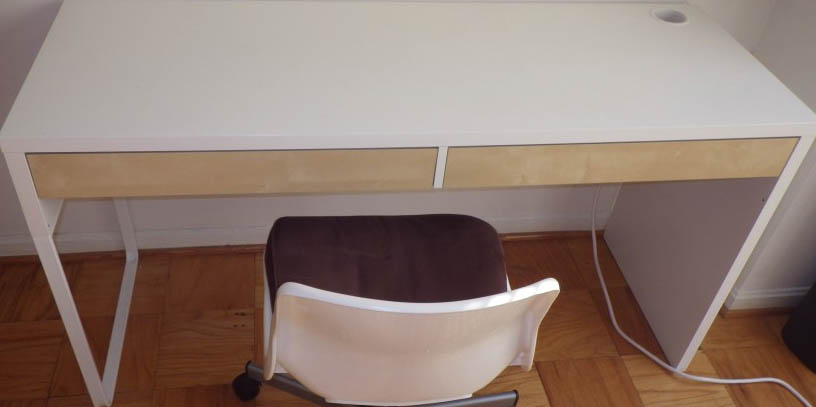} &
	    \verti{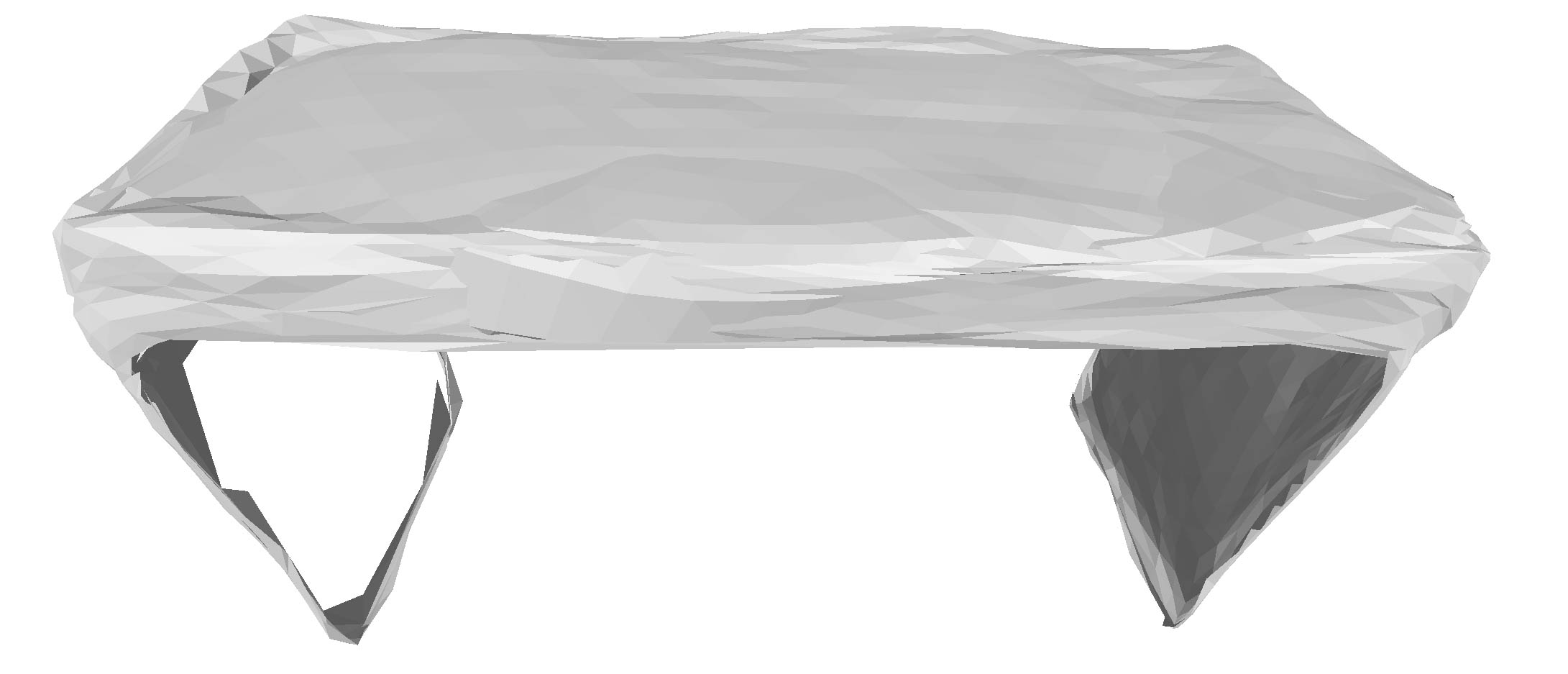} &
	    \verti{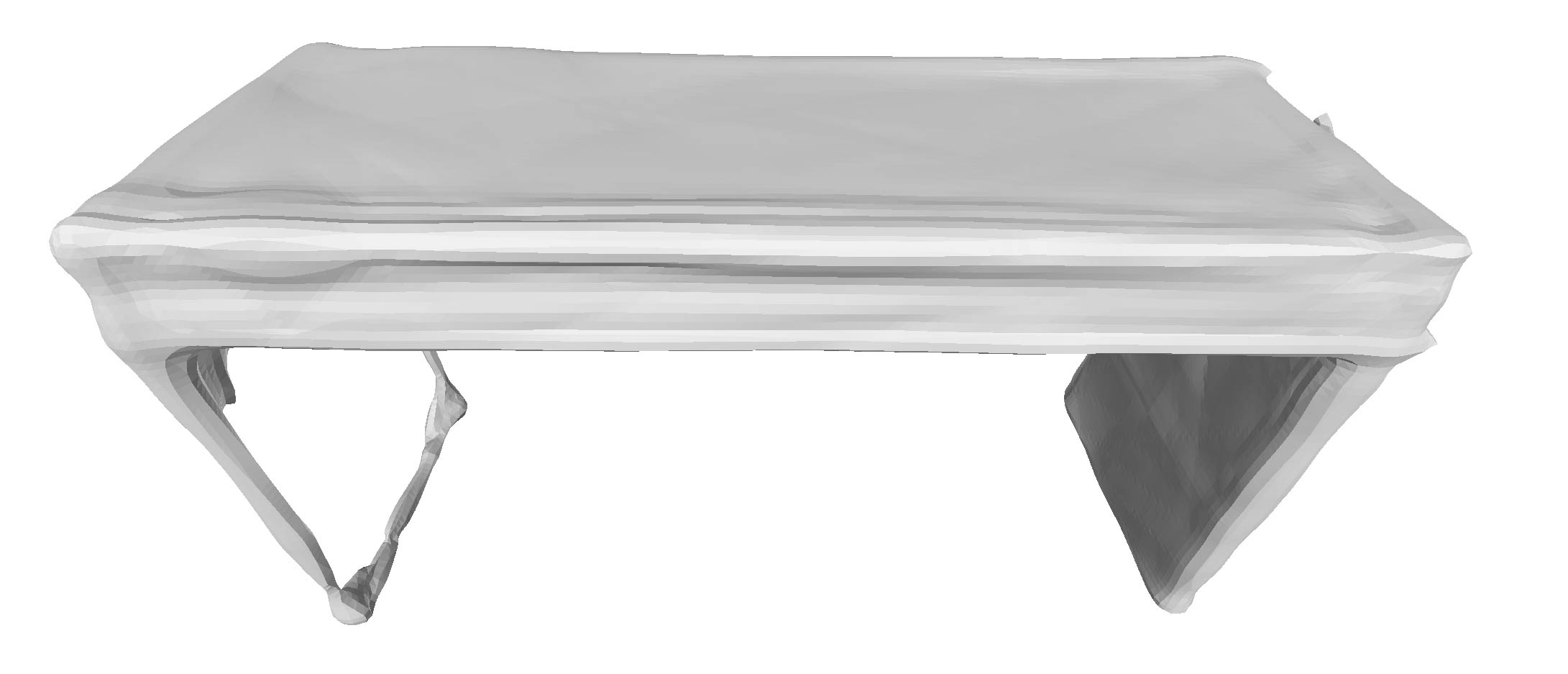} &
	    \squa{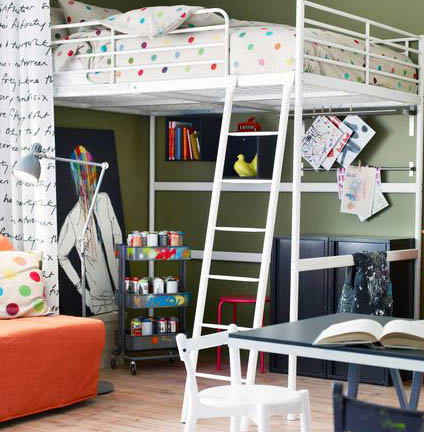} &
	    \squa{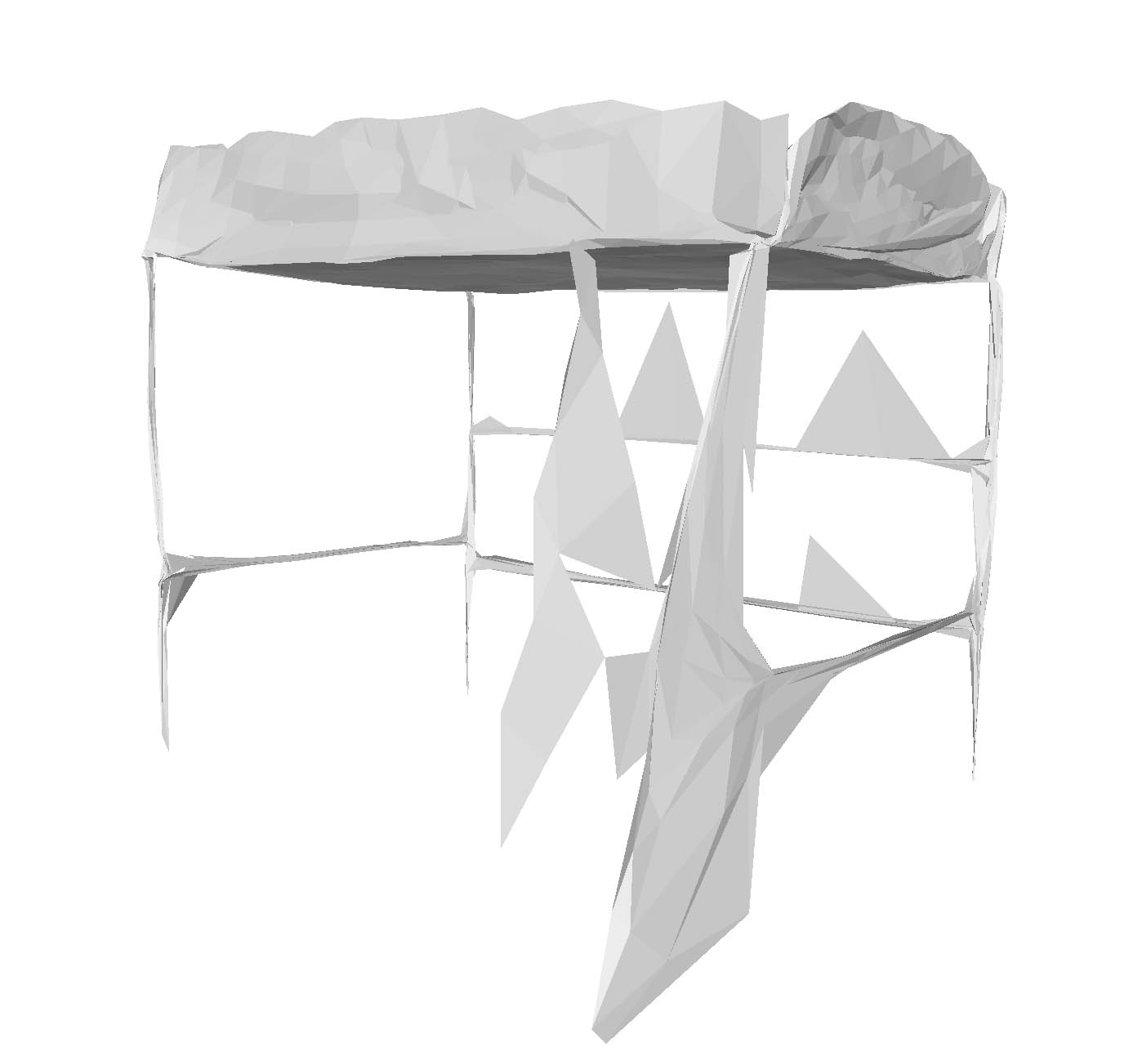} &
	    \squa{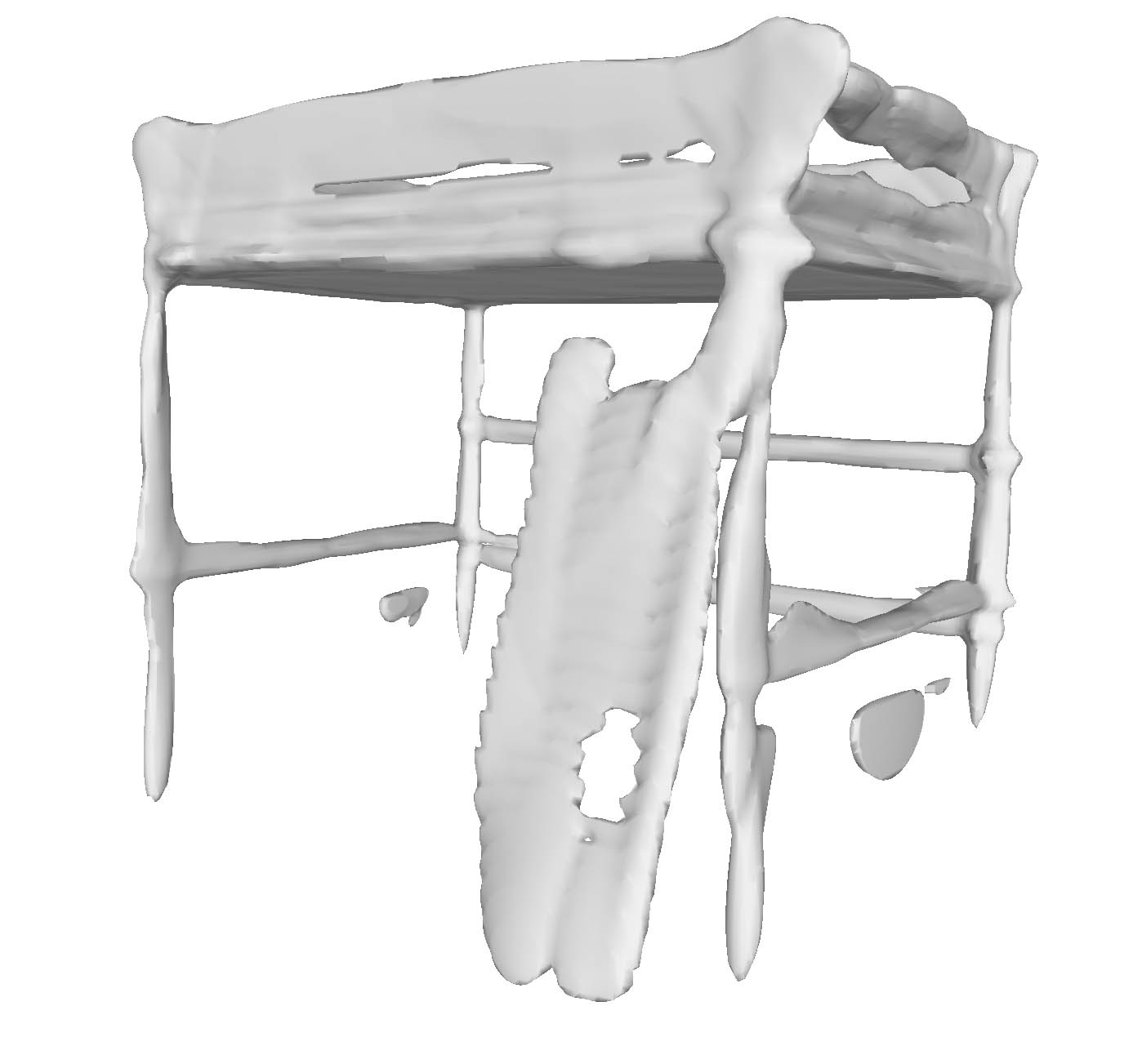} &
	    \hori{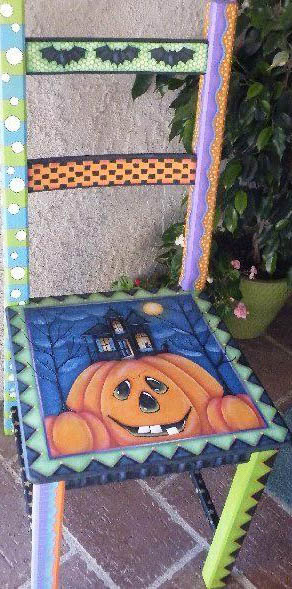} &
	    \hori{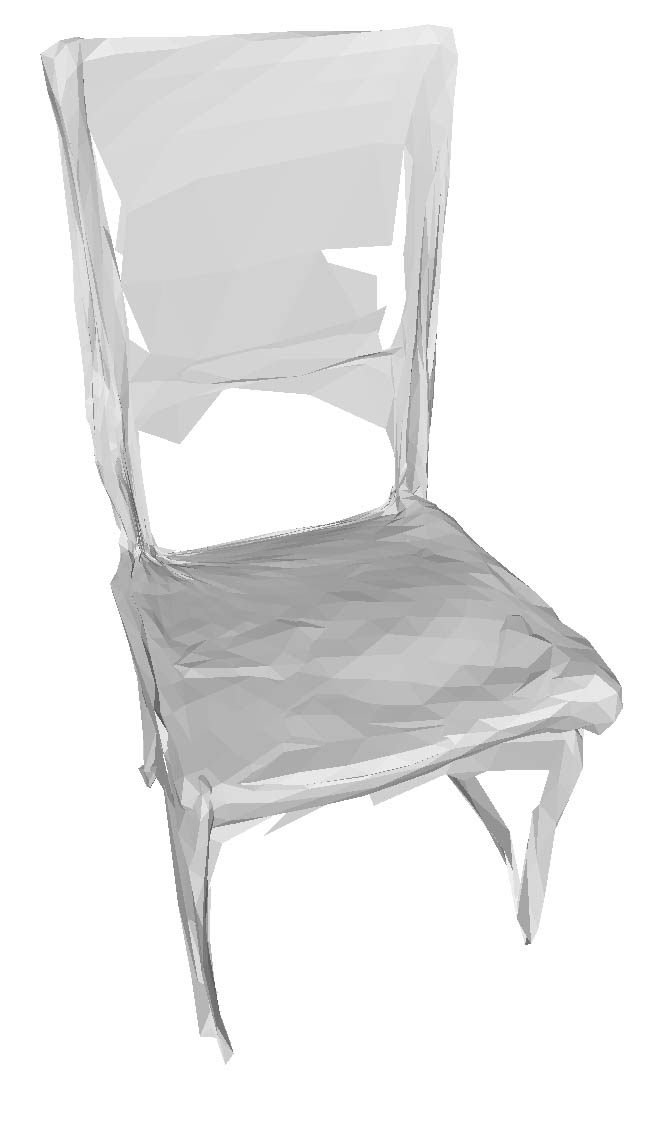} &
	    \hori{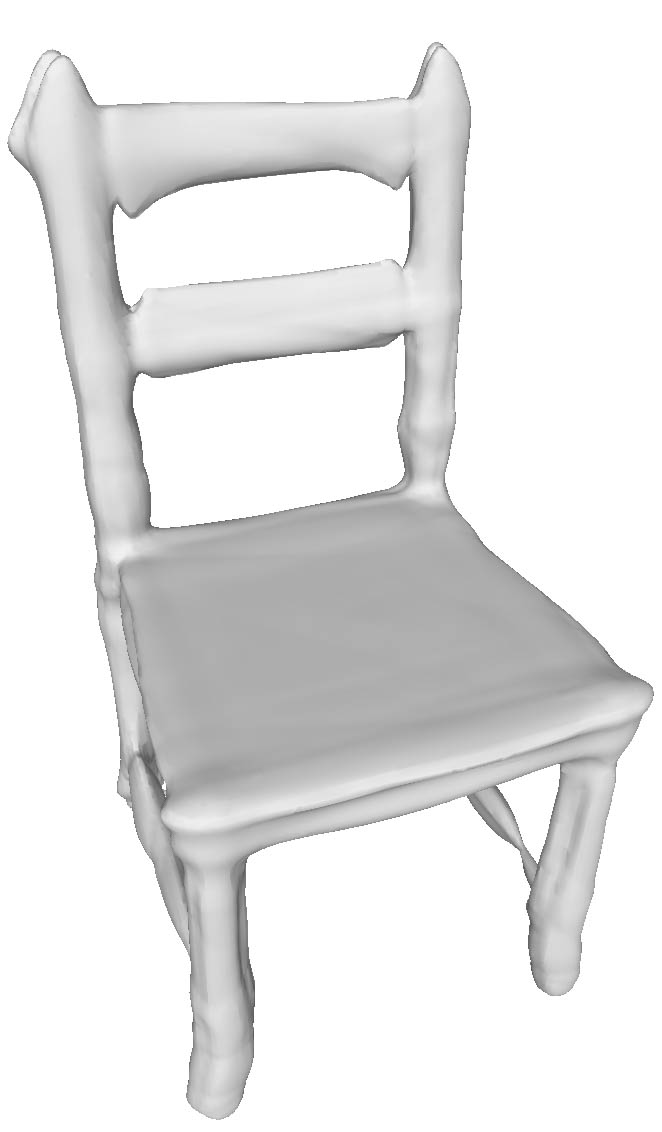} \\
	    
	    \verti{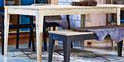} &
	    \verti{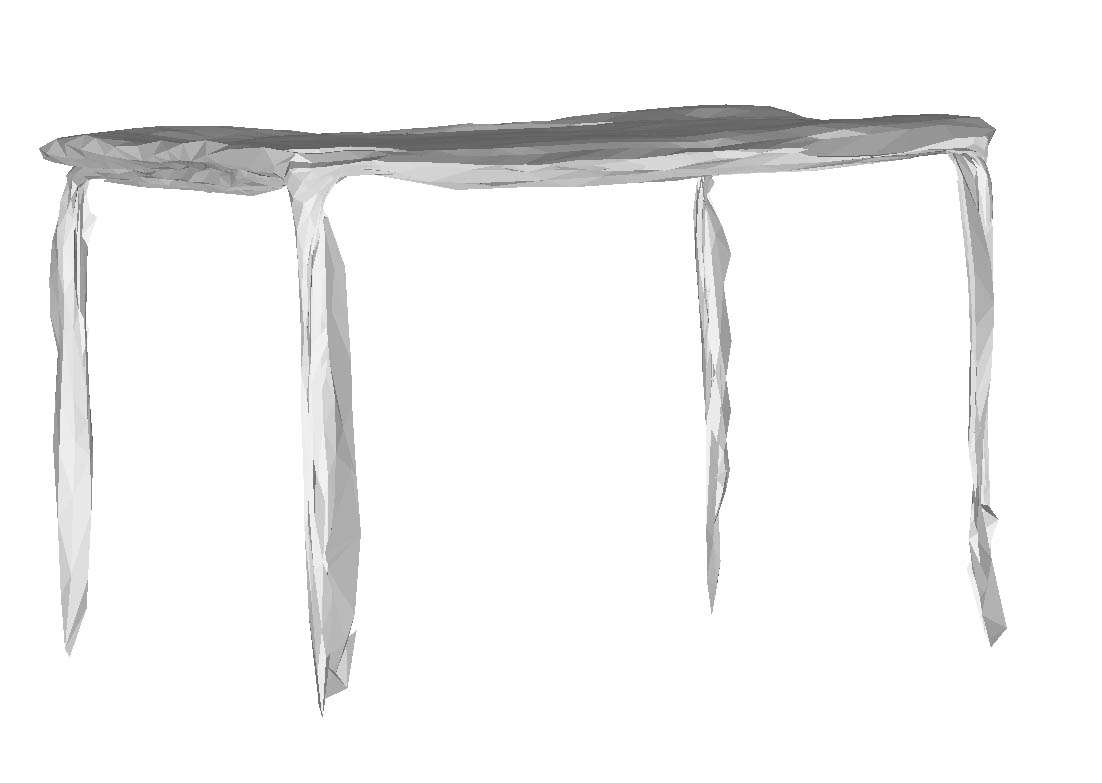} &
	    \verti{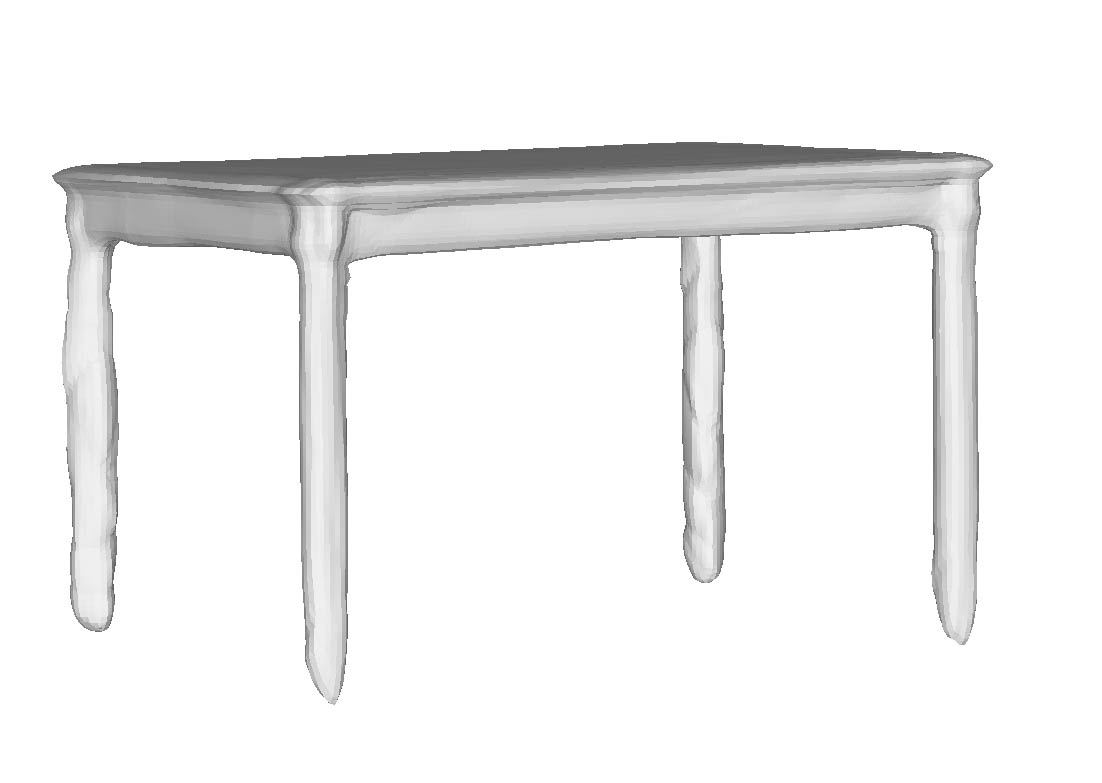} &
	    \squa{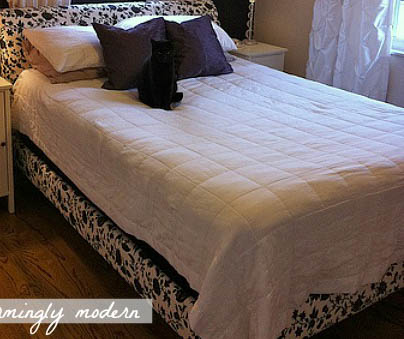} &
	    \squa{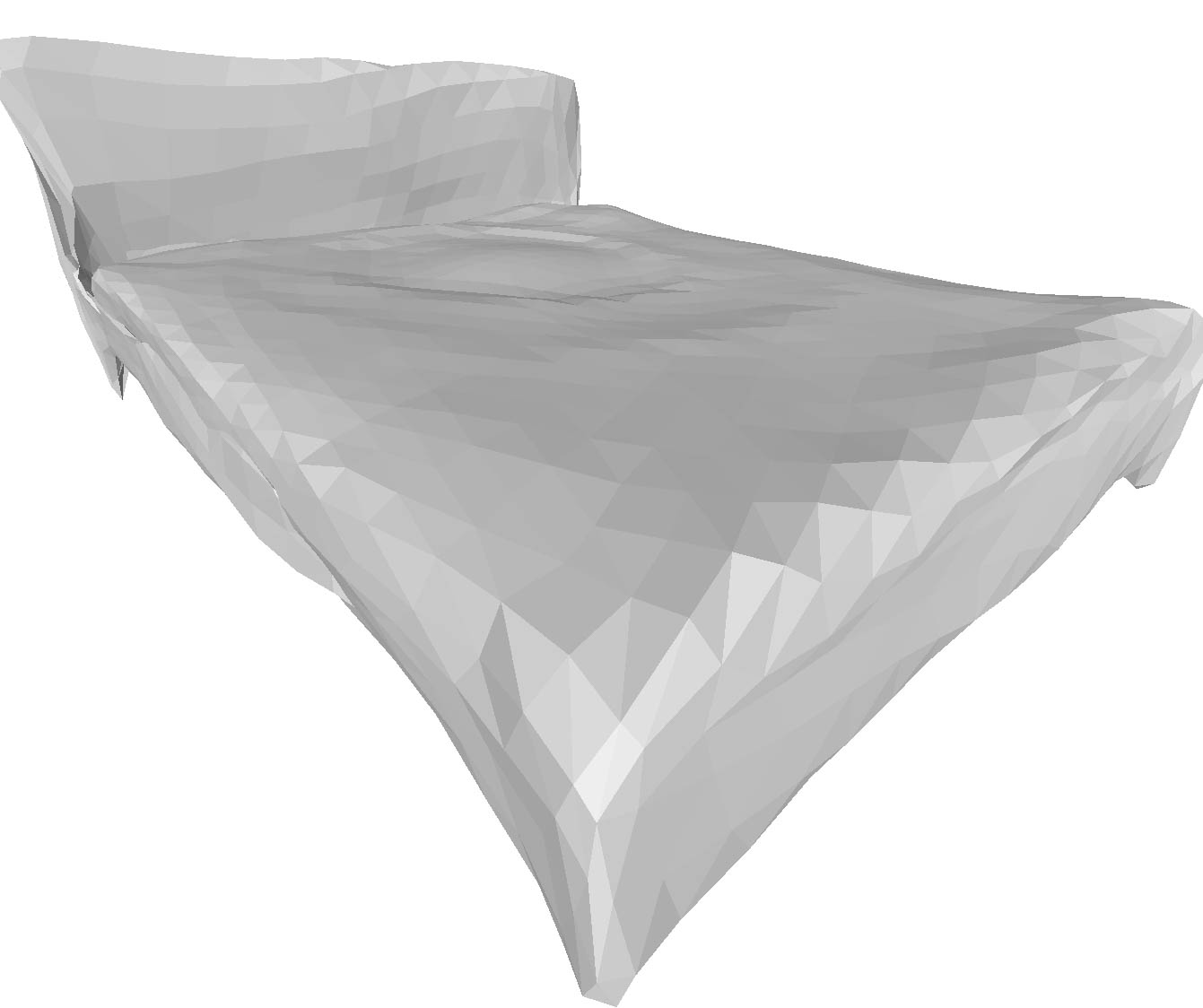} &
	    \squa{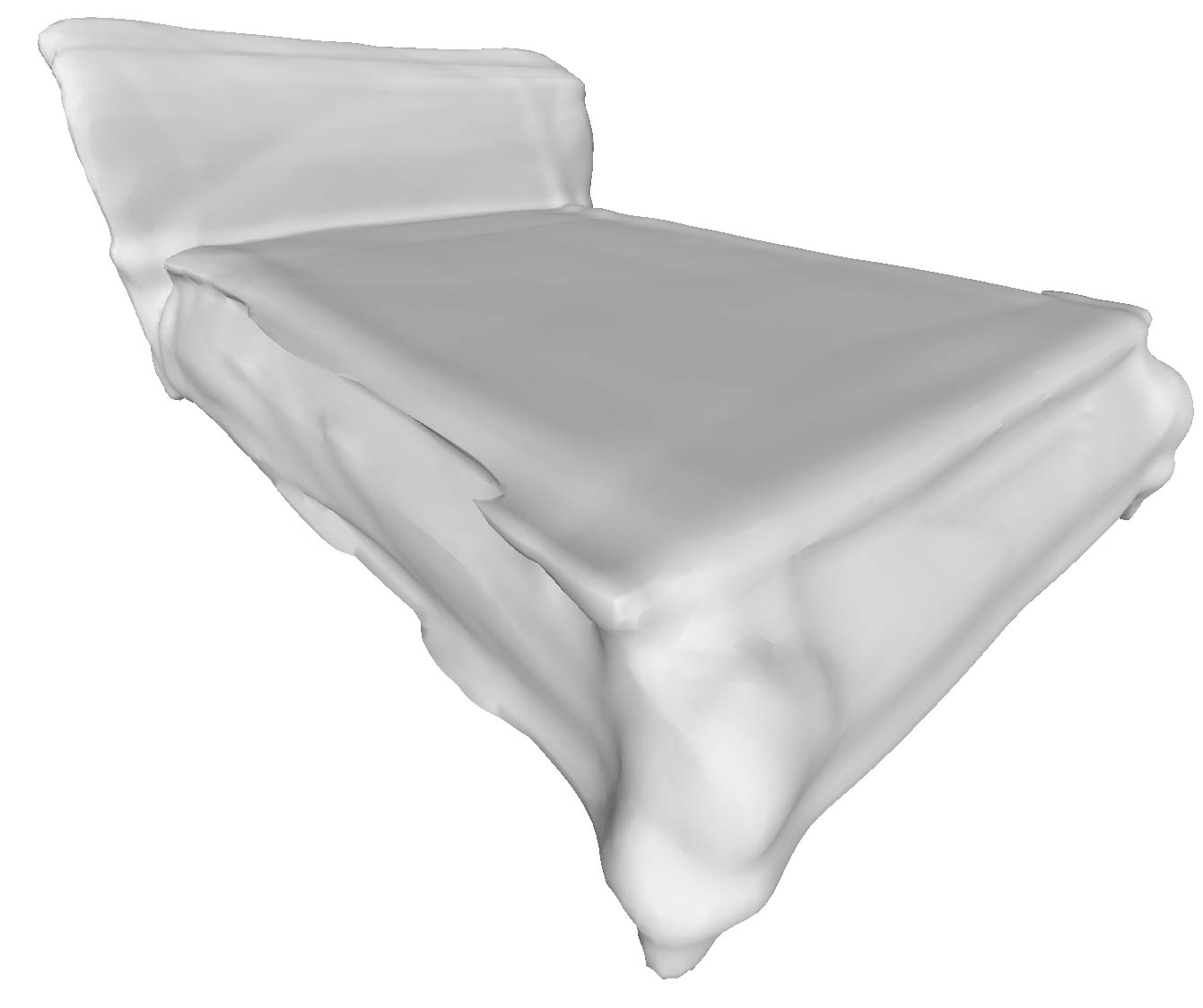} &
	    \hori{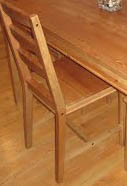} &
	    \hori{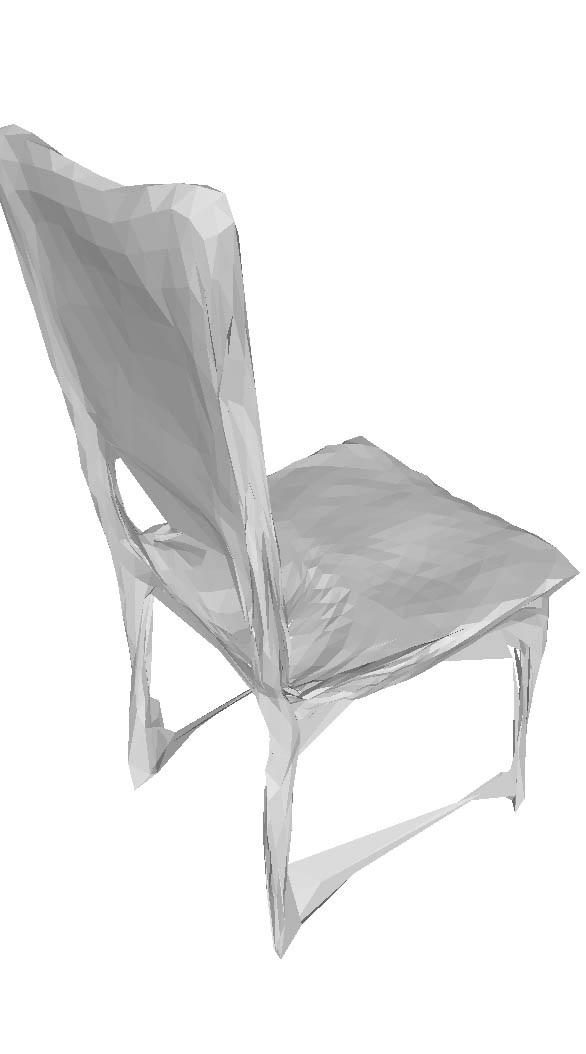} &
	    \hori{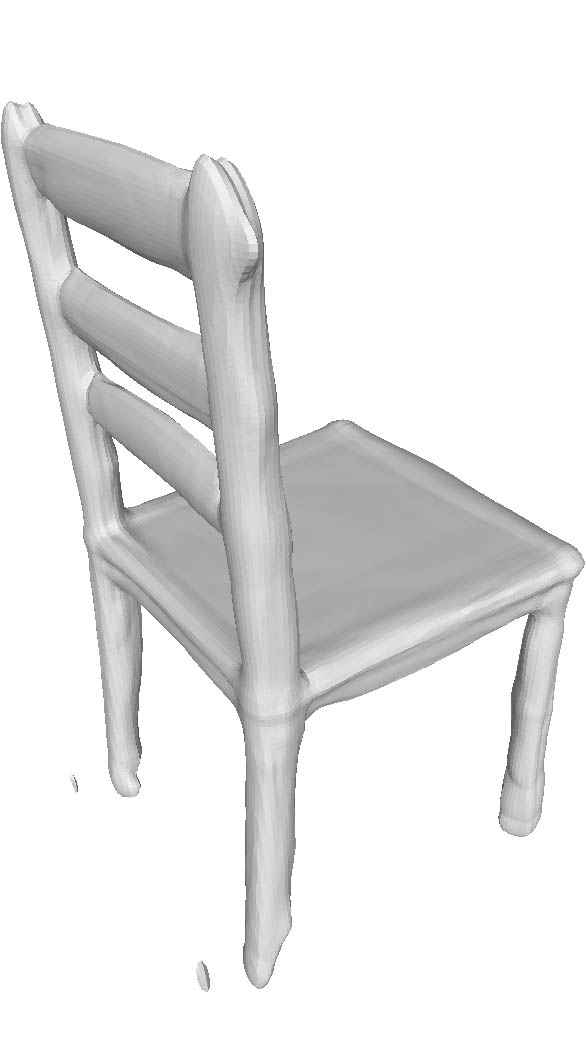} \\
	    \vspace{2px}
	    \verti{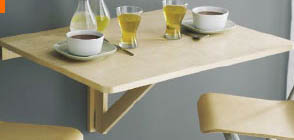} &
	    \verti{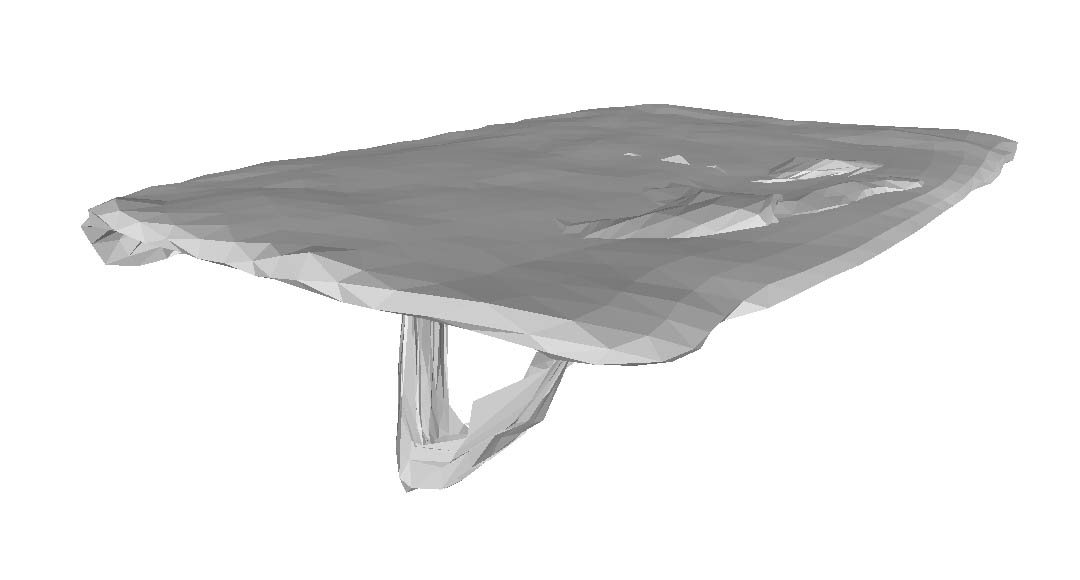} &
	    \verti{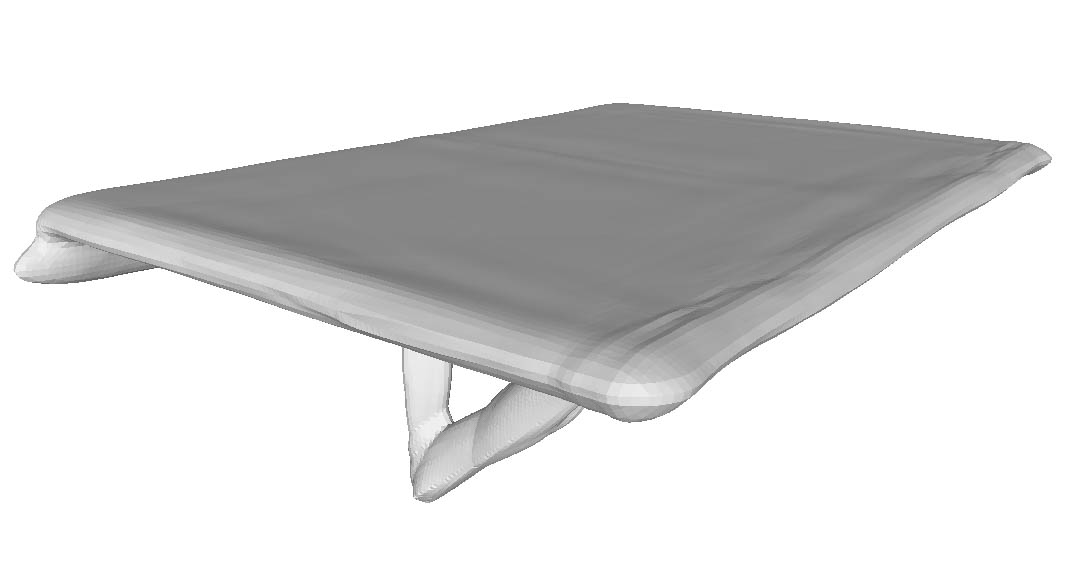} &
	    \squa{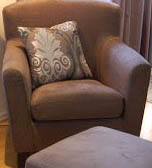} &
	    \squa{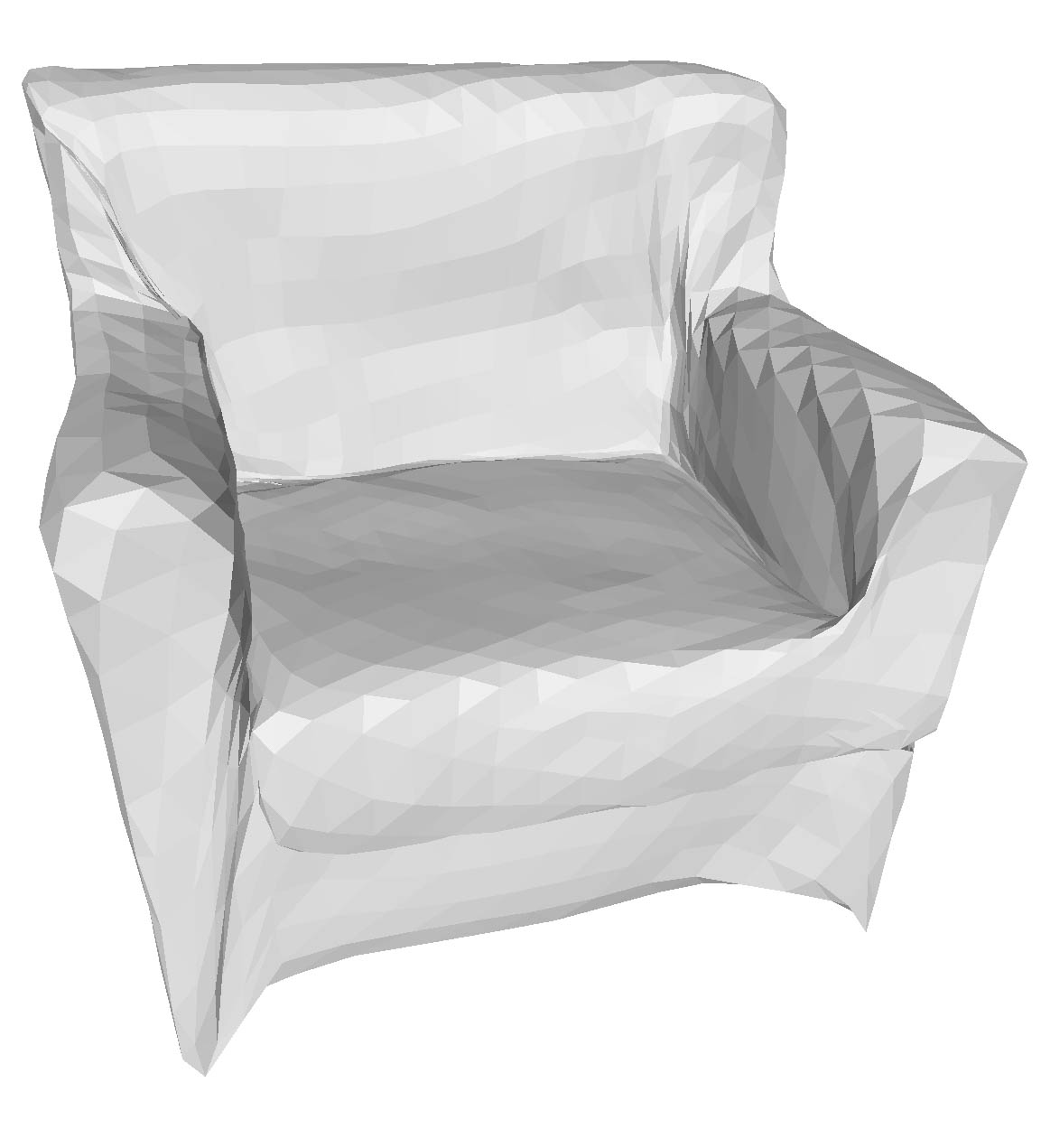} &
	    \squa{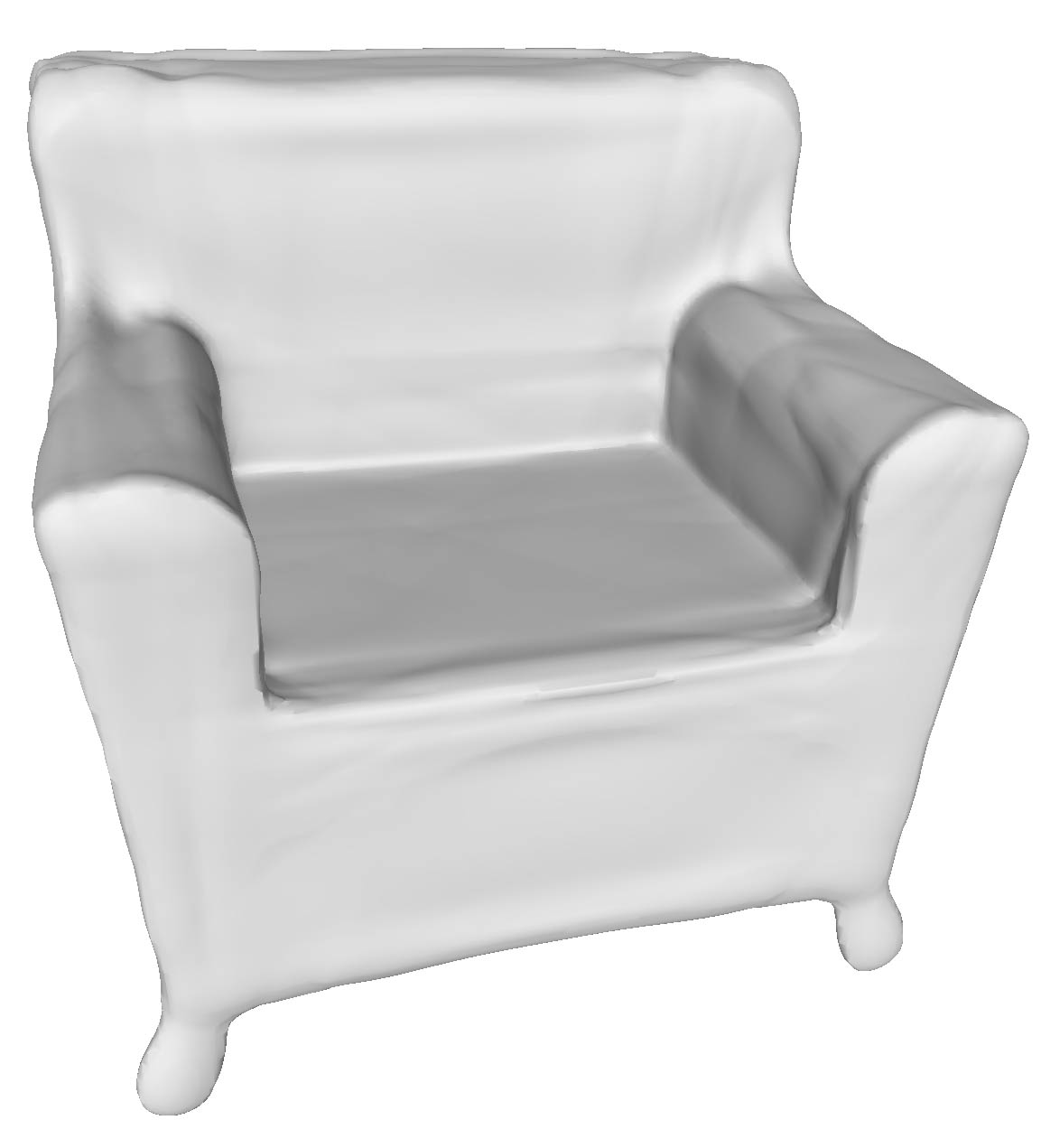} &
	    \hori{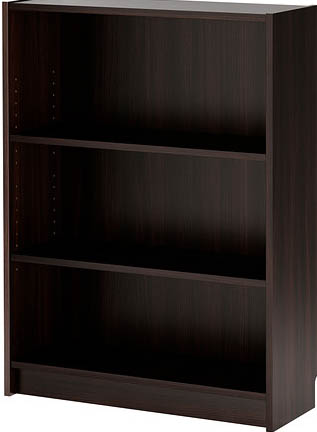} &
	    \hori{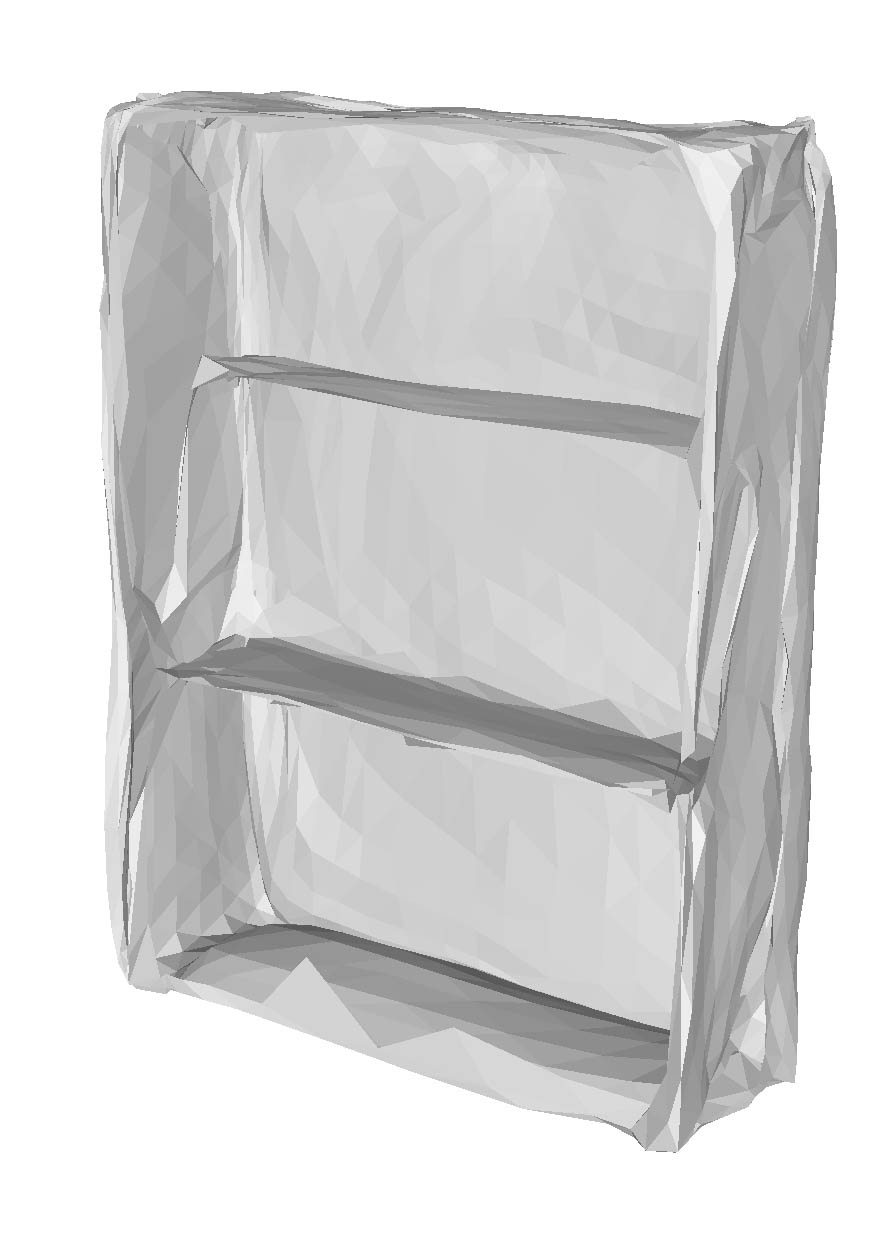} &
	    \hori{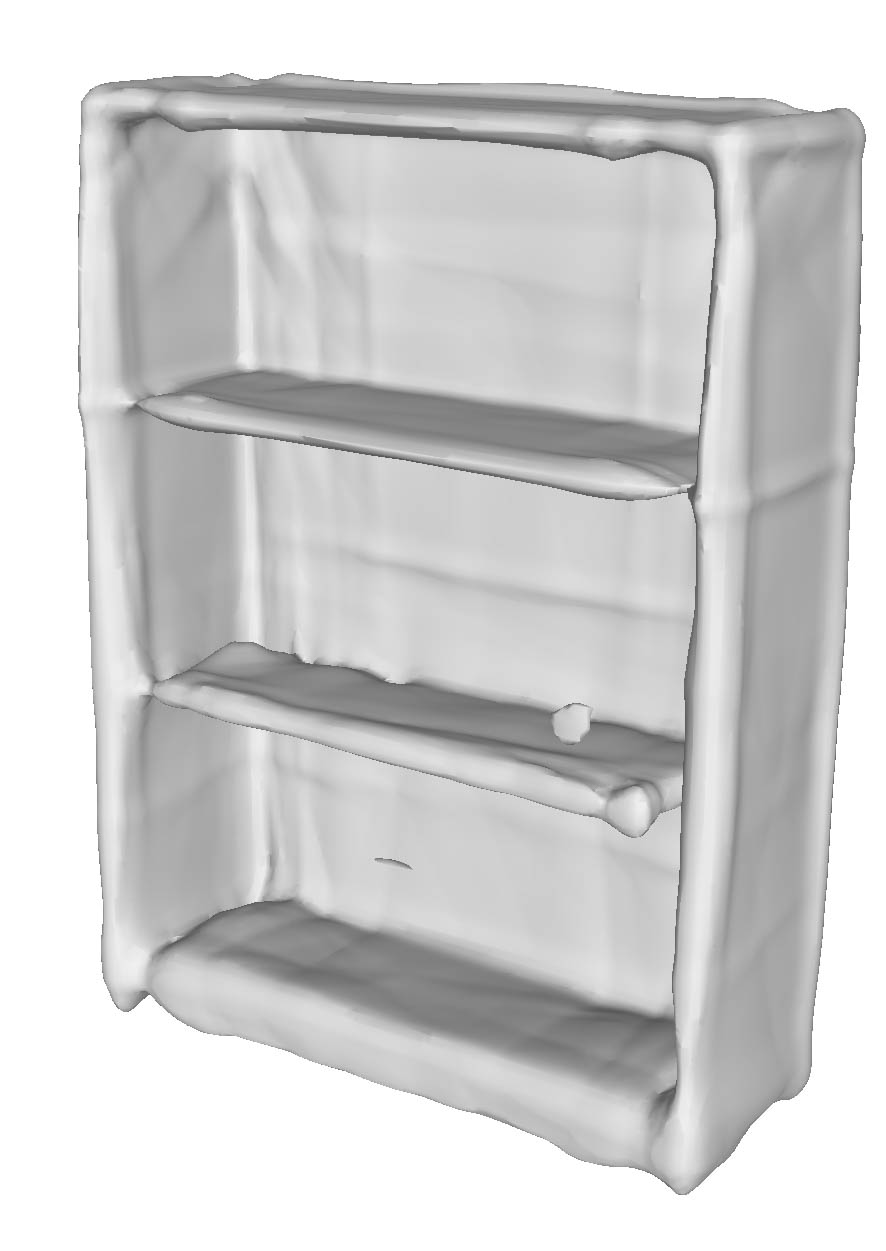} \\
	    
	    Input & MGN \cite{nie2020total3dunderstanding} & Ours & 
	    Input & MGN \cite{nie2020total3dunderstanding} & Ours & 
	    Input & MGN \cite{nie2020total3dunderstanding} & Ours \\
    \end{tabular}
    % \vspace{-1.0em}
	\caption{More qualitative comparisons on object reconstruction. We compare with MGN from Total3D \cite{nie2020total3dunderstanding}.}
	\label{fig:objreconSupp}
\end{figure*}

%In main paper Section 4.2, we show object reconstruction quality compared to \cite{groueix2018, nie2020total3dunderstanding}.
In this section, we show more results on the object reconstruction in \autoref{fig:objreconSupp}.
Compared to MGN in Total3D\cite{nie2020total3dunderstanding}, our method produces more accurate geometry preserving high-quality details especially on chairs, bookshelves, and those shapes with relatively more complex topology.

% To emphasize the more visually plausible results of using LDIF representation, we present more comparisons with MGN \cite{nie2020total3dunderstanding} in \autoref{fig:objreconSupp}.

\begin{figure}[!ht]
    % \vspace{-0.5em}
	\centering
	\begin{subfigure}[t]{0.23\textwidth}
		\includegraphics[width=\textwidth]  
		{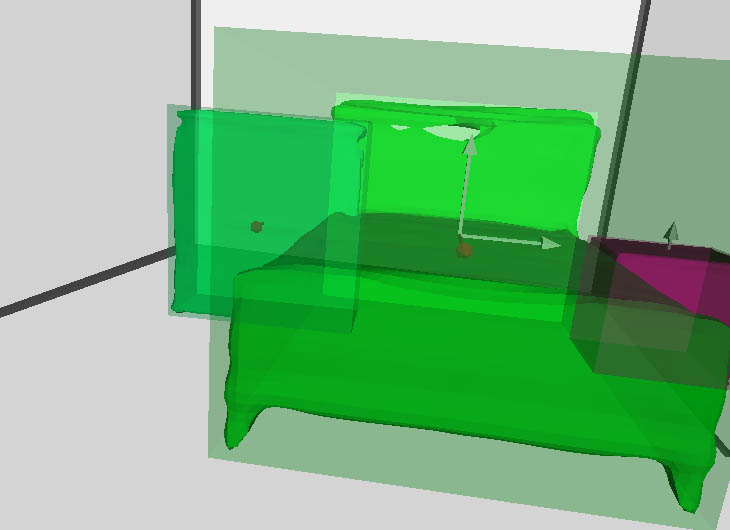}
		\includegraphics[width=\textwidth]
		{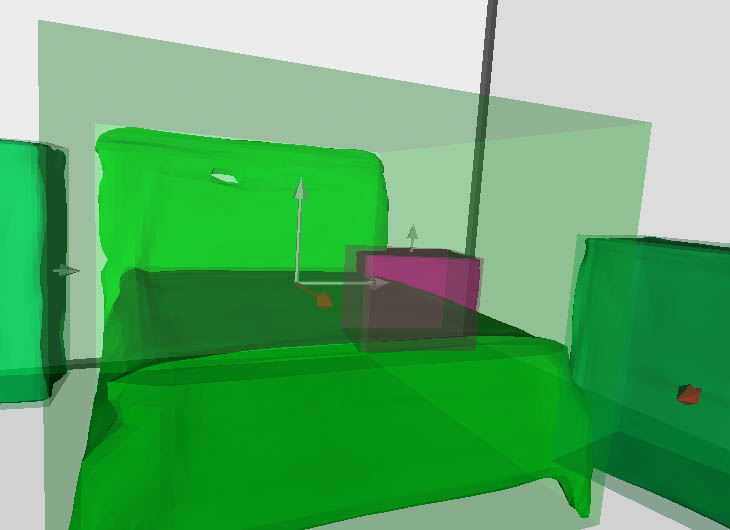}
		\caption{Ours-$\mathcal{L}_{phy}$}
	\end{subfigure}
	\begin{subfigure}[t]{0.23\textwidth}
		\includegraphics[width=\textwidth]  
		{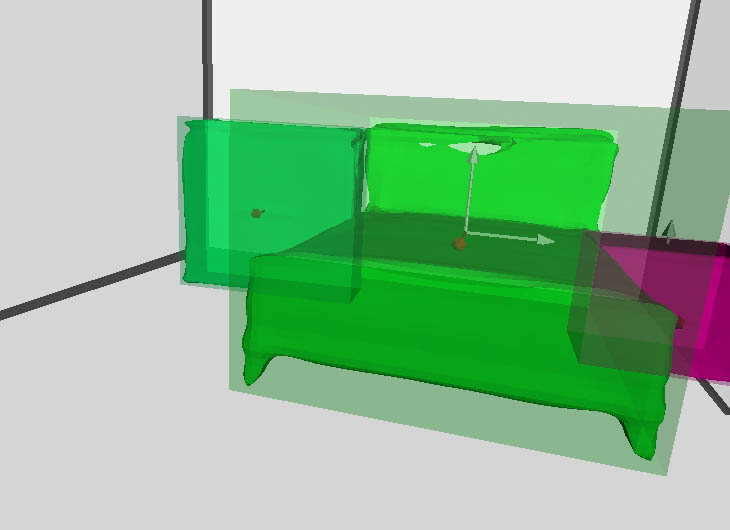}
		\includegraphics[width=\textwidth]
		{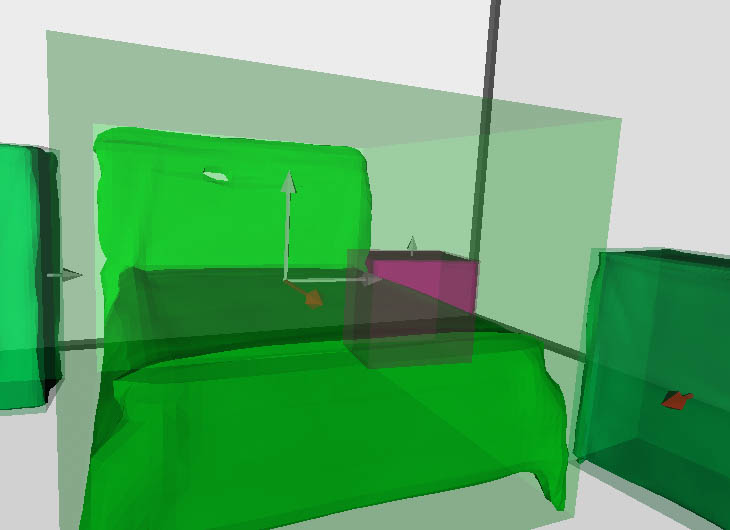}
		\caption{Full: with $\mathcal{L}_{phy}$}
	\end{subfigure}
	\vspace{-0.5em}
	\caption{\zc{Scene reconstruction samples of Ours-$\mathcal{L}_{phy}$ and Full. We observe more intersections between objects without physical violation loss in some scenes.}}
	\label{fig:lphy_qualitative}
\end{figure}

\begin{figure}[!ht]
    % \vspace{-8mm}
	\centering
	\newcommand{\front}[1]{\includegraphics[width=\textwidth,clip,trim=50 20 50 80]{#1}}
	\begin{subfigure}[t]{0.155\textwidth}
		\front{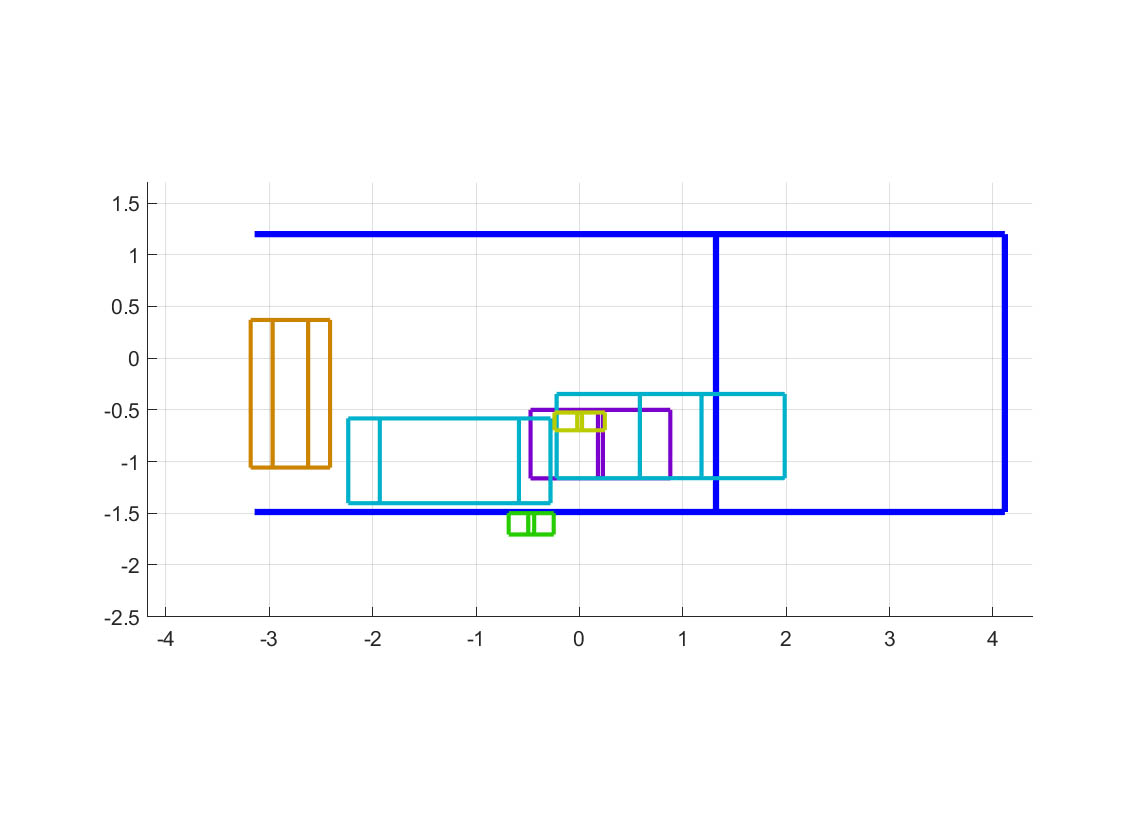}
		\vspace{-2em}
		\caption{Total3D}
	\end{subfigure}
	\begin{subfigure}[t]{0.155\textwidth}
		\front{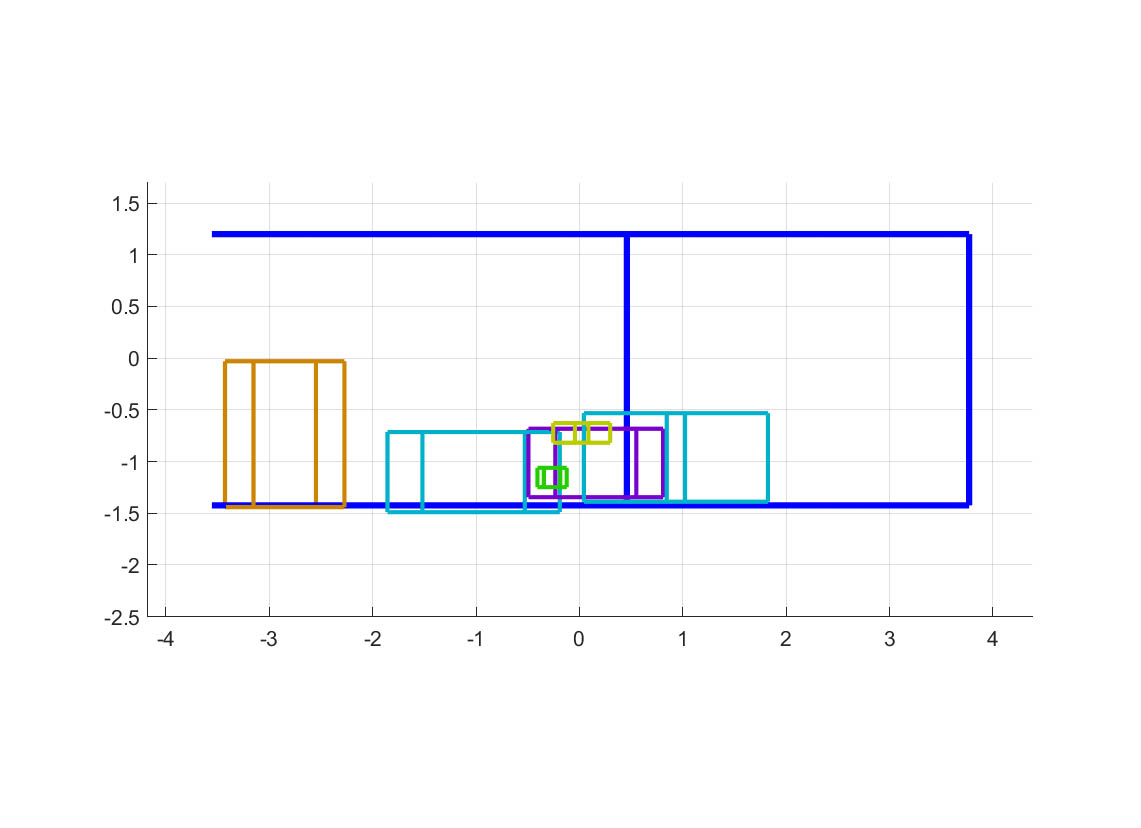}
		\vspace{-2em}
		\caption{Ours}
	\end{subfigure}
	\begin{subfigure}[t]{0.155\textwidth}
		\front{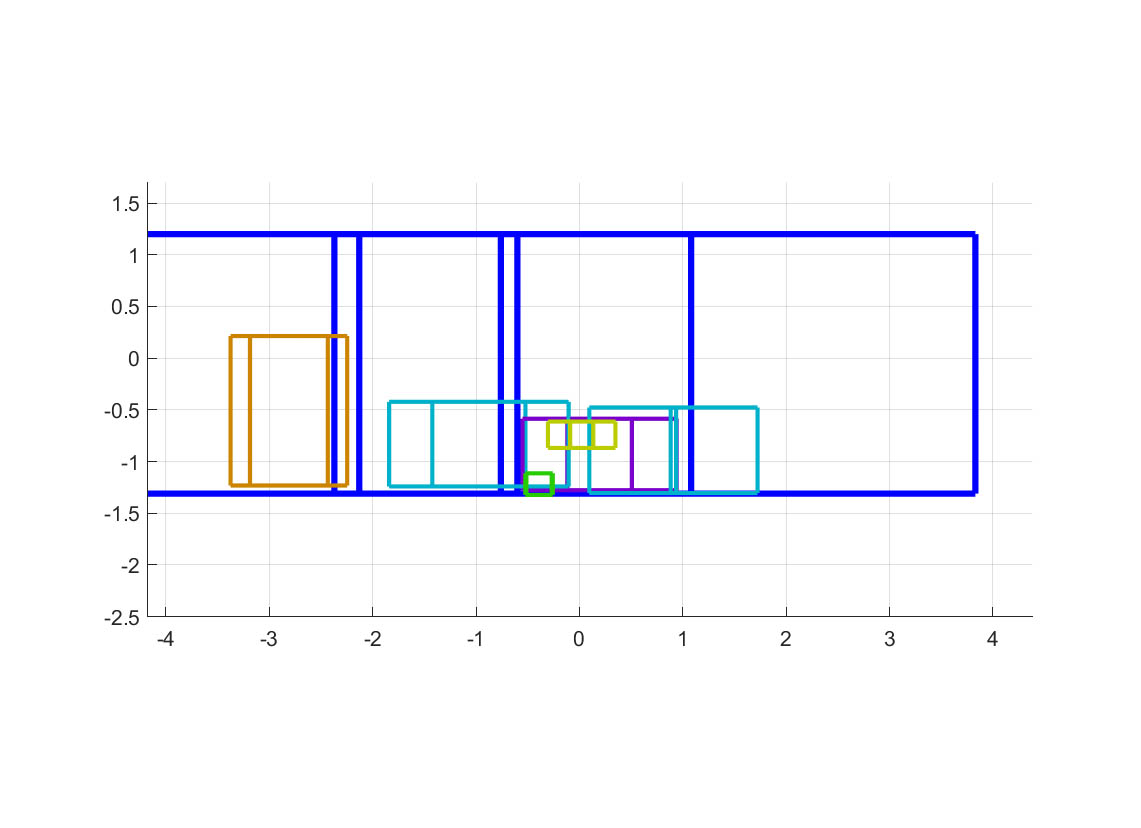}
		\vspace{-2em}
		\caption{GT}
	\end{subfigure}
	\vspace{-1em}
	\caption{\zc{Qualitative comparison of supporting relation.
	We take the front view of object 3D detection of main paper Figure 5 column 4 as an example. 
	We observe fewer flying objects in our results than Total3D, which shows a better understanding of supporting relation.
	}}
	\label{fig:sup_qualitative}
\end{figure}

\section{More Qualitatively Comparison on 3D Detection and Scene Reconstruction}
\begin{figure*}[!ht]
	\centering
	\scriptsize
	\newcommand{\rgb}[1]{\raisebox{-0.5\height}{\includegraphics[width=.175\linewidth]{#1}}}
	\newcommand{\oblique}[1]{\raisebox{-0.5\height}{\includegraphics[width=.175\linewidth,clip,trim=48 12 48 12]{#1}}}
	\newcommand{\rot}[1]{\rotatebox[origin=c]{90}{#1}}
	\def\arraystretch{0.5}%
	\begin{tabular}{c|c*{5}{c@{\hspace{1px}}}}
        &\rot{Input}& 
    	    \rgb{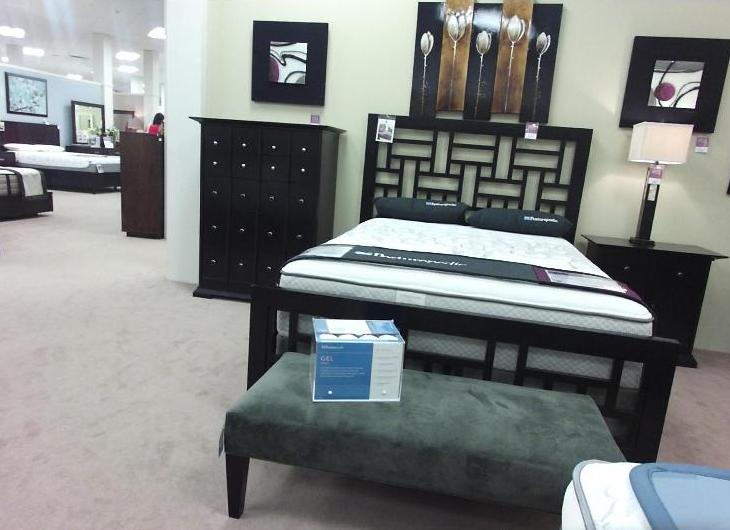} &
    	    \rgb{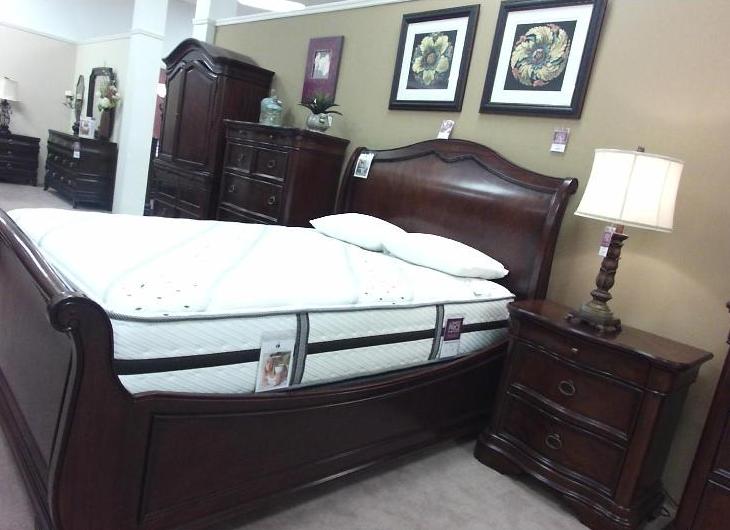} &
    	    \rgb{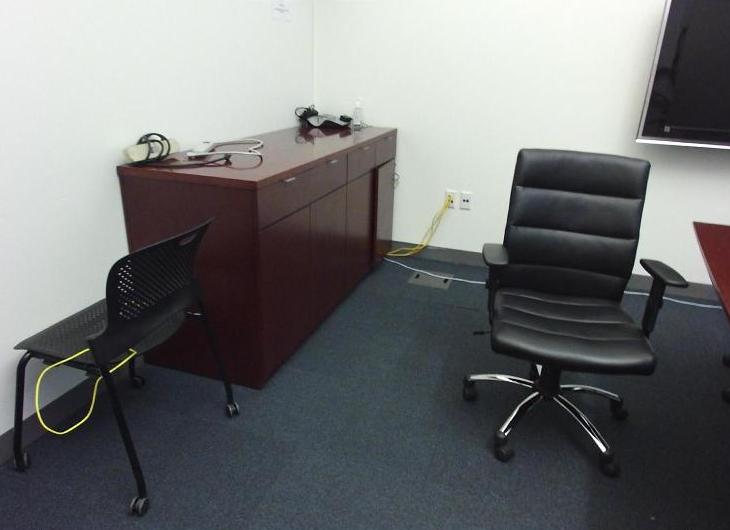} &
    	    \rgb{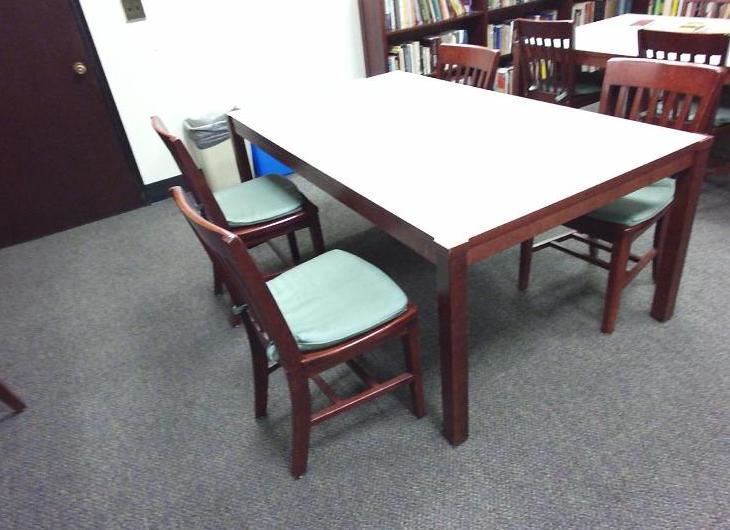} &
    	    \rgb{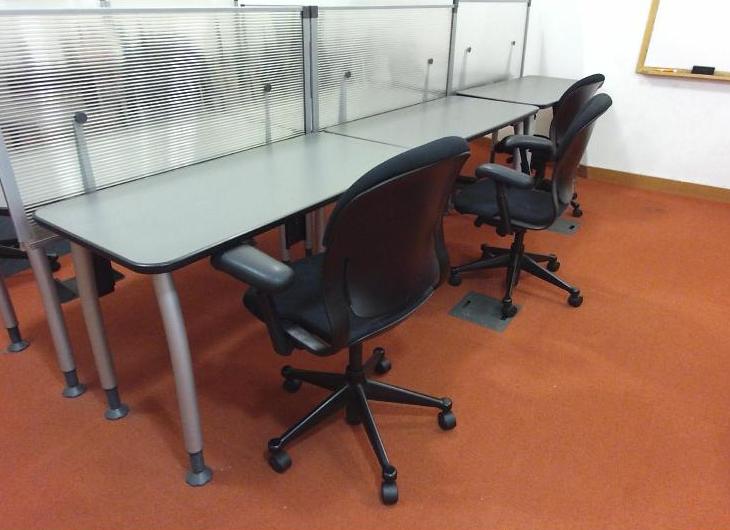} \\
    	\hline
    	    &\rot{Total3D}& 
        	    \oblique{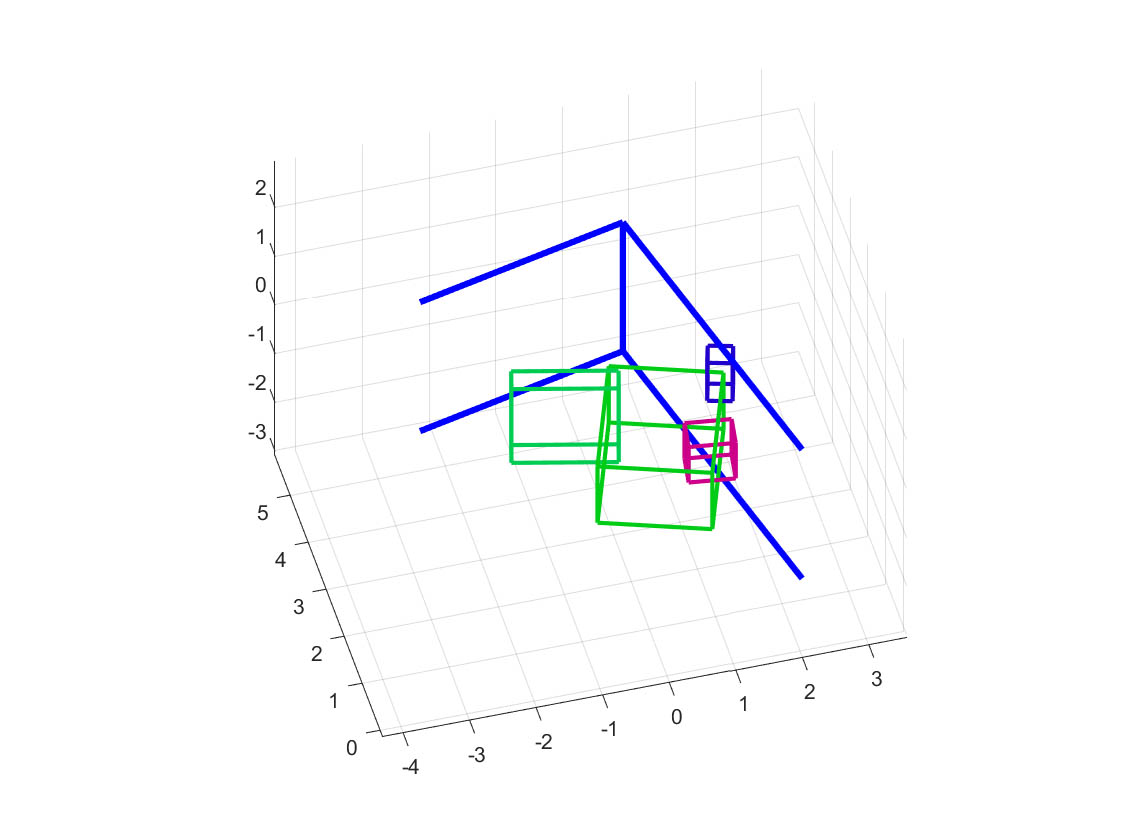} &
        	    \oblique{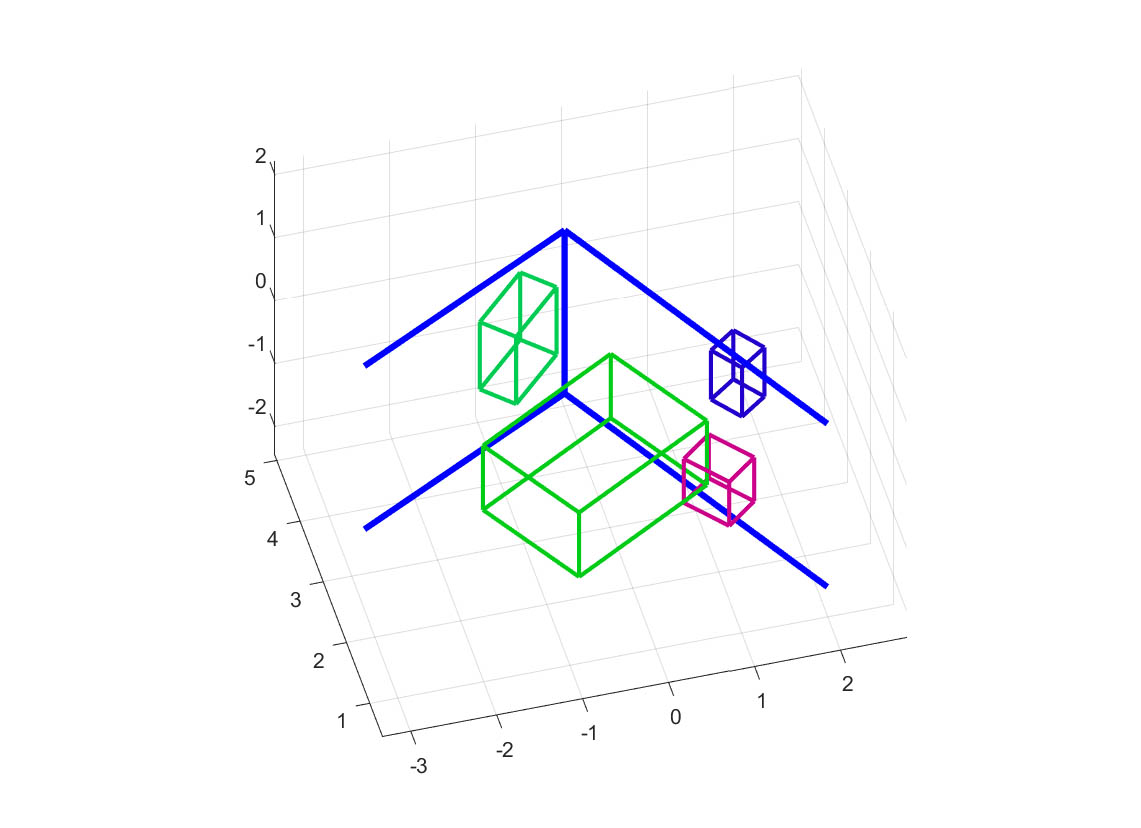} &
        	    \oblique{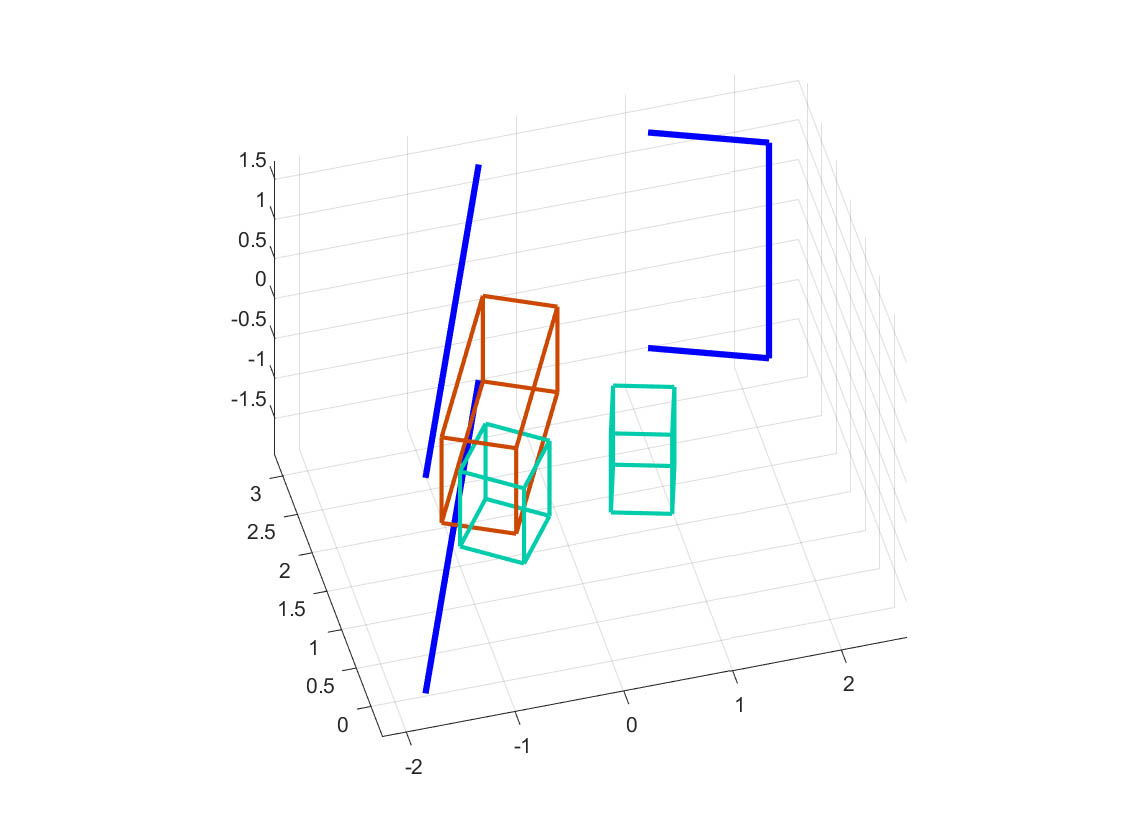} &
        	    \oblique{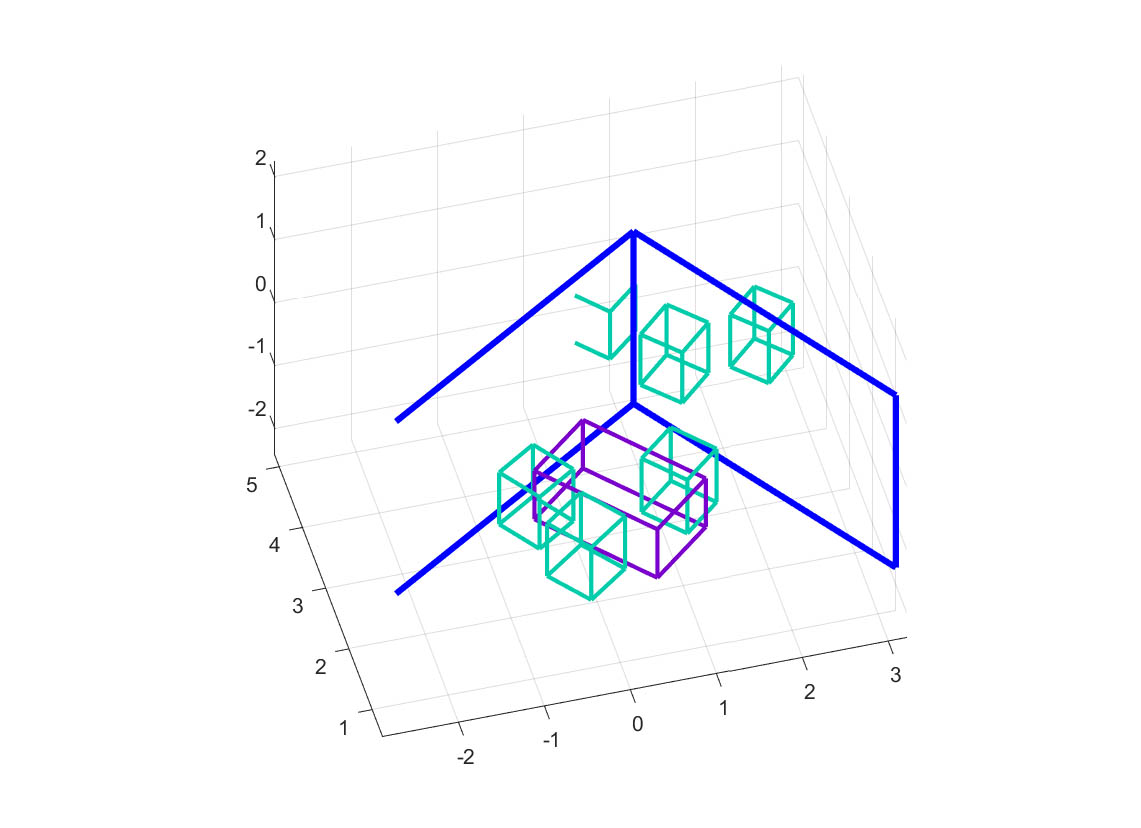} &
        	    \oblique{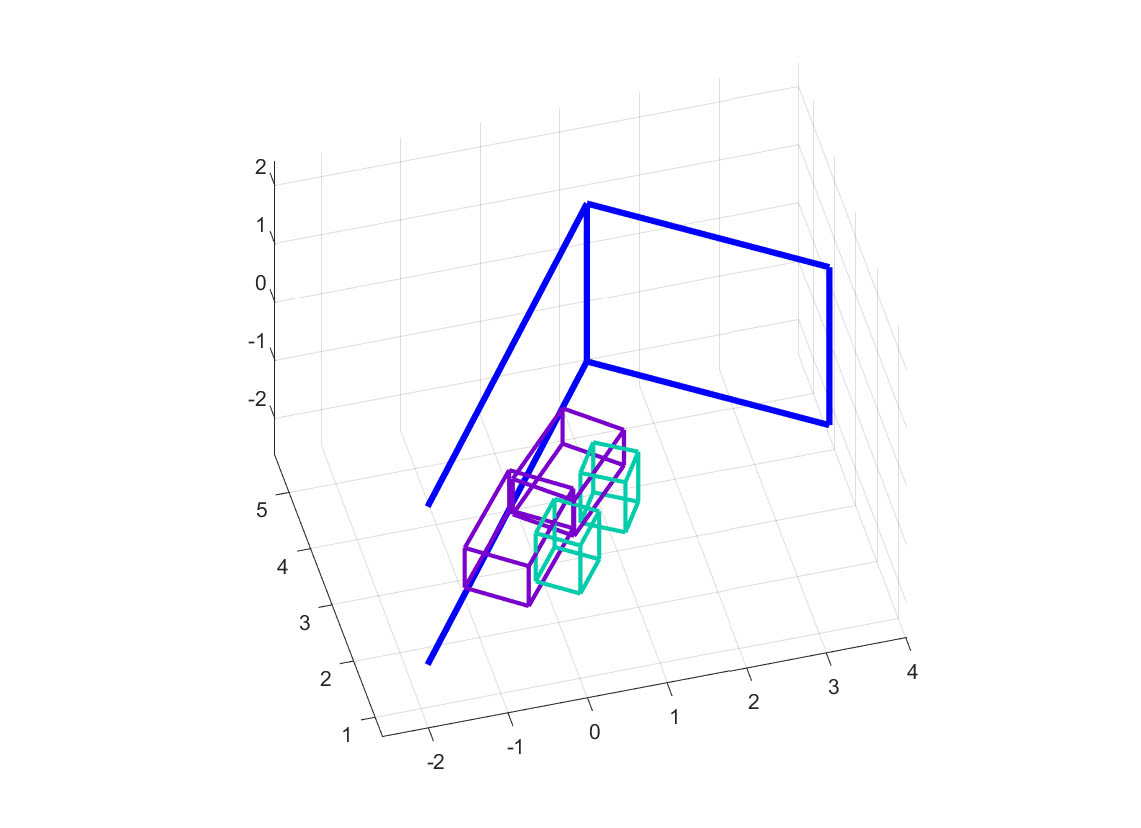} \\
        	\rot{Oblique View}& \rot{Ours}& 
        	    \oblique{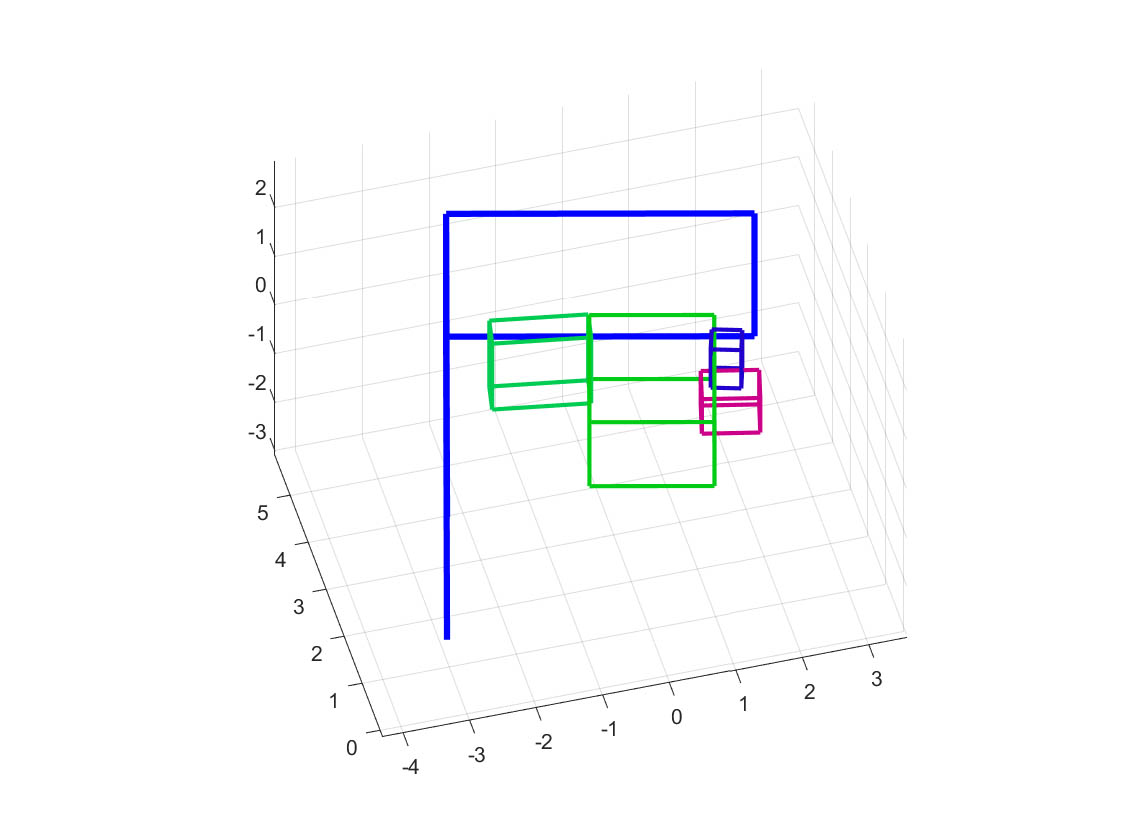} &
        	    \oblique{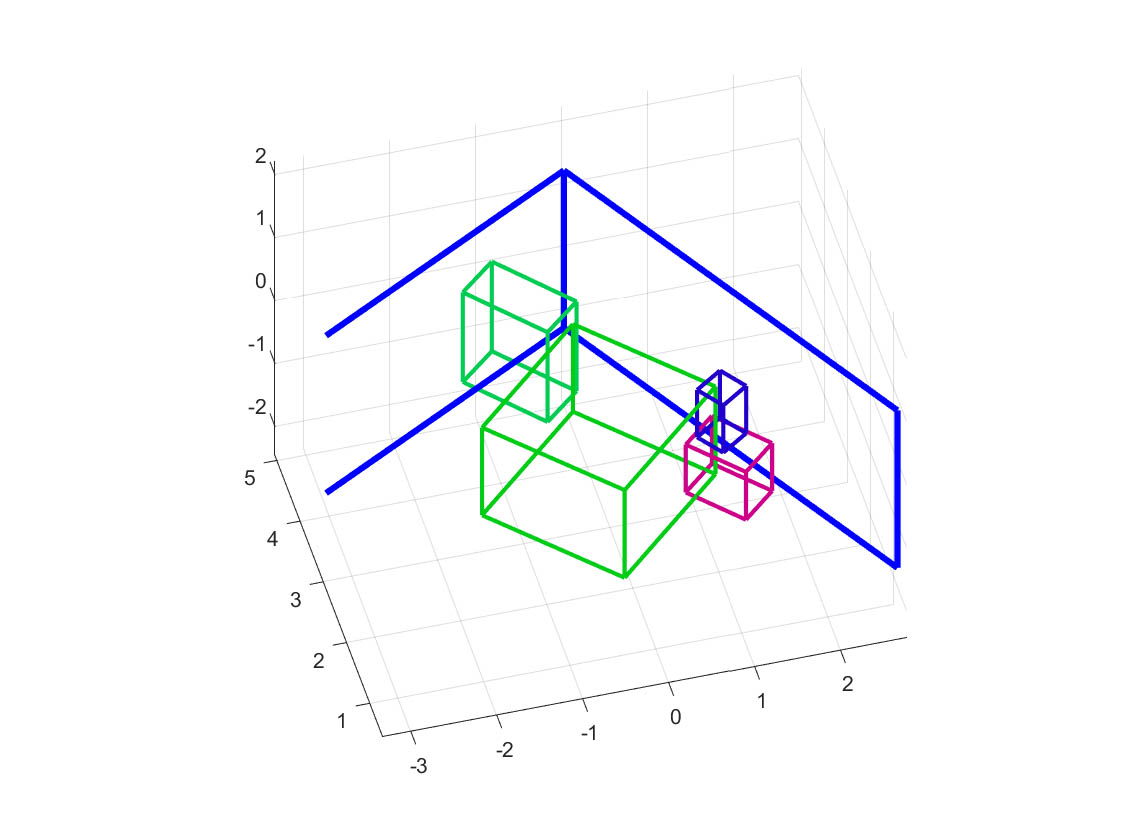} &
        	    \oblique{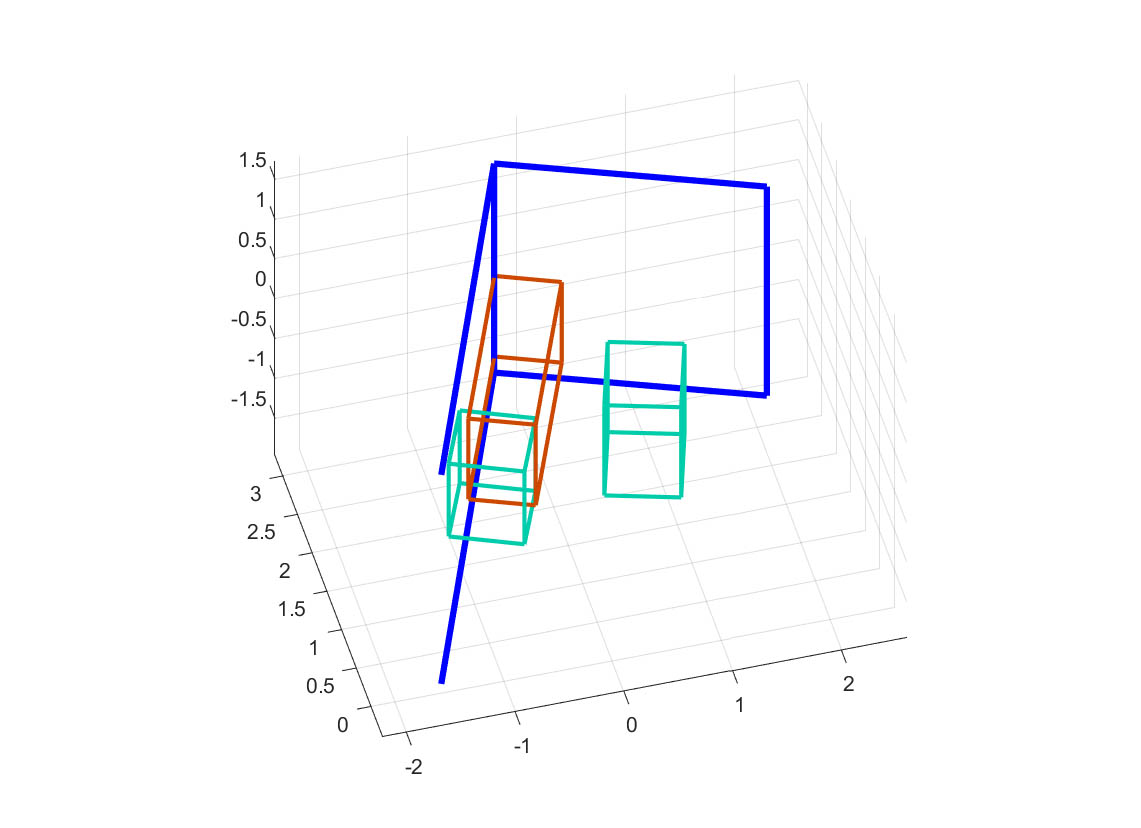} &
        	    \oblique{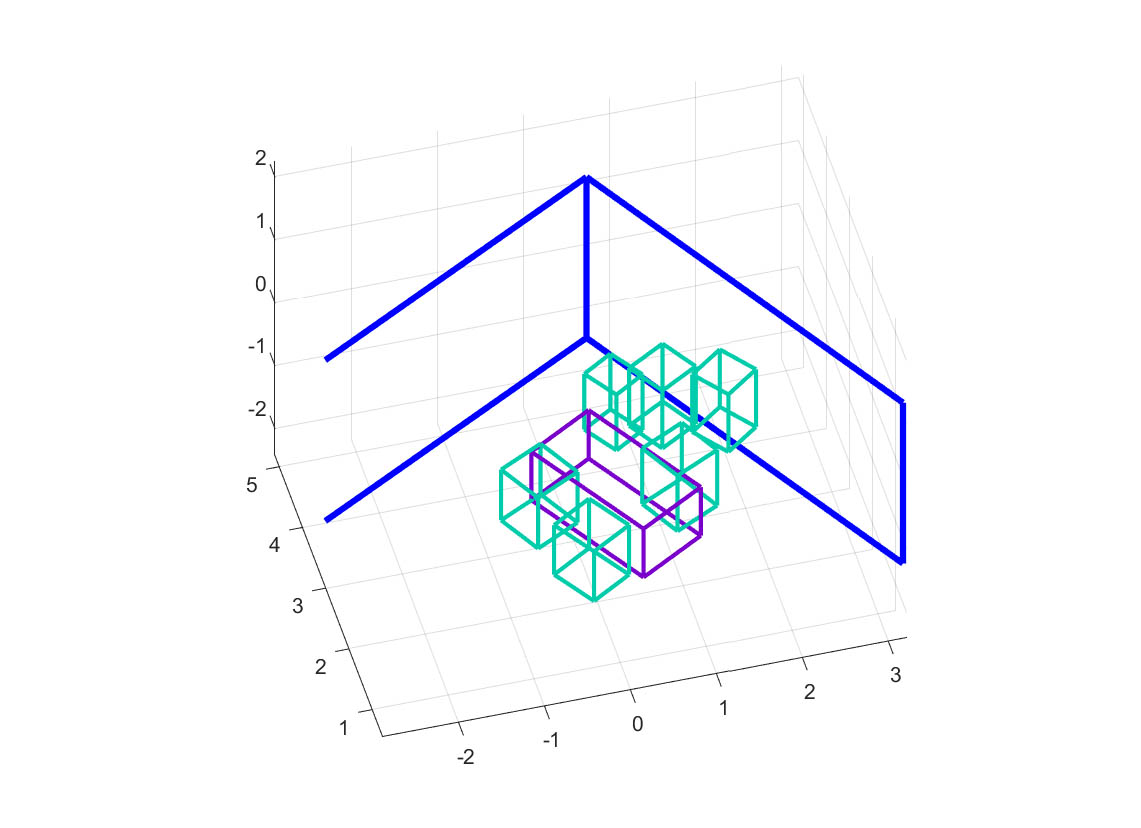} &
        	    \oblique{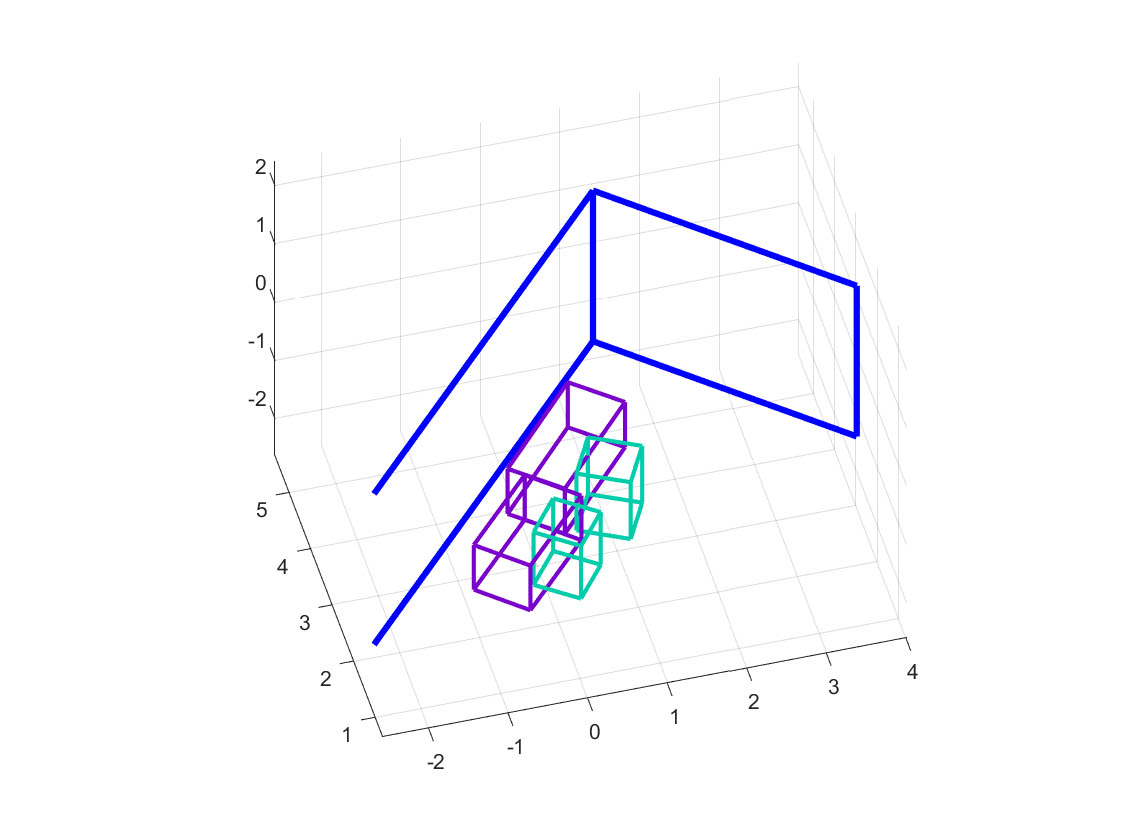} \\
        	&\rot{GT}& 
        	    \oblique{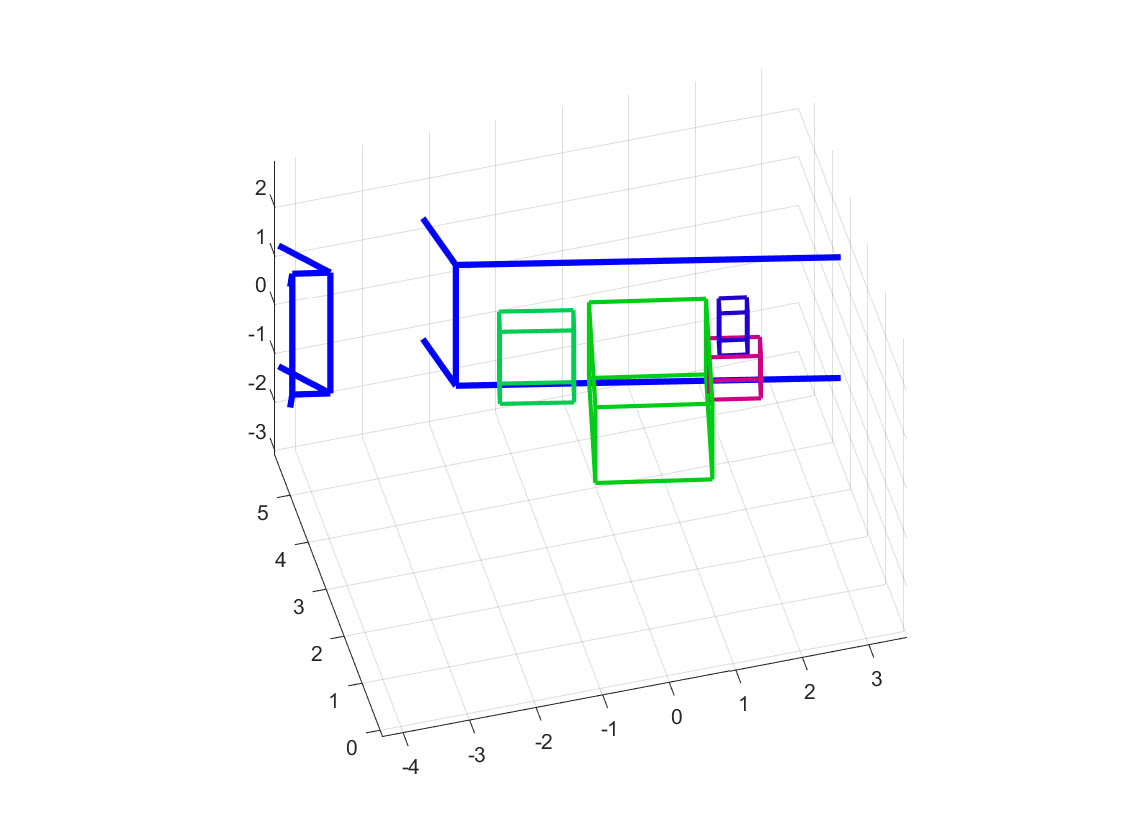} &
        	    \oblique{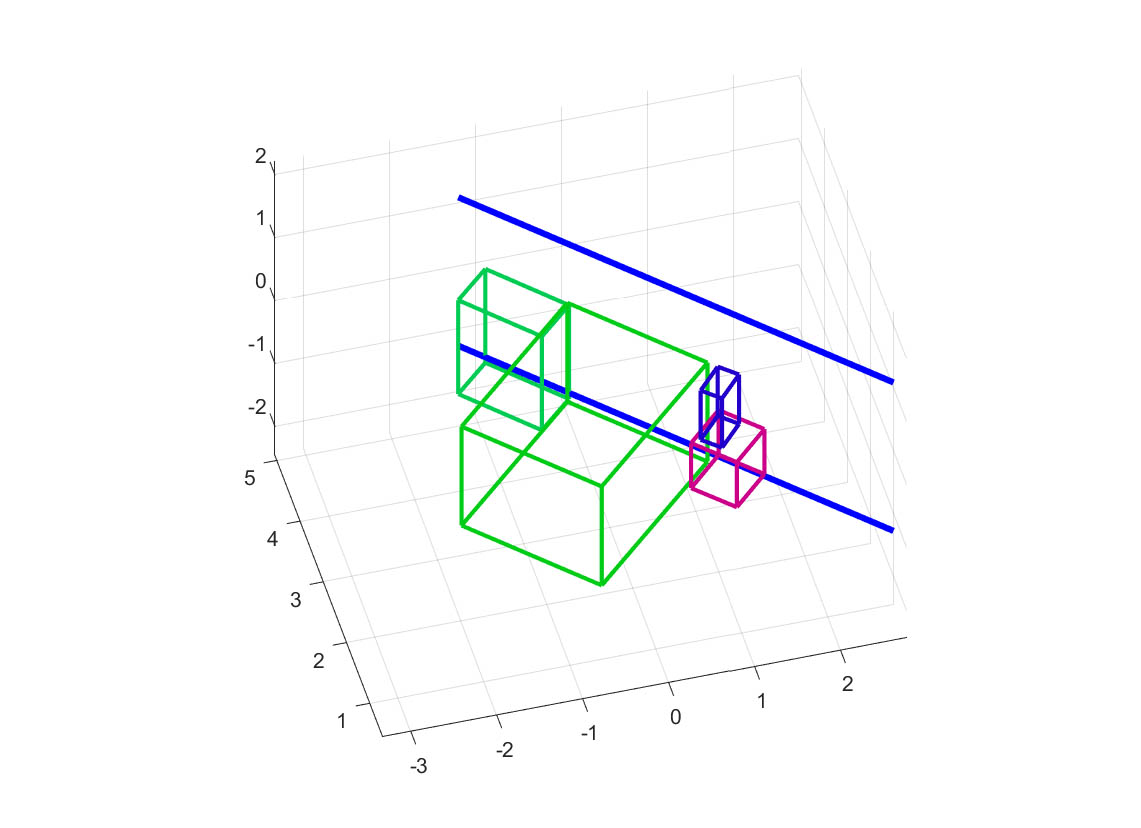} &
        	    \oblique{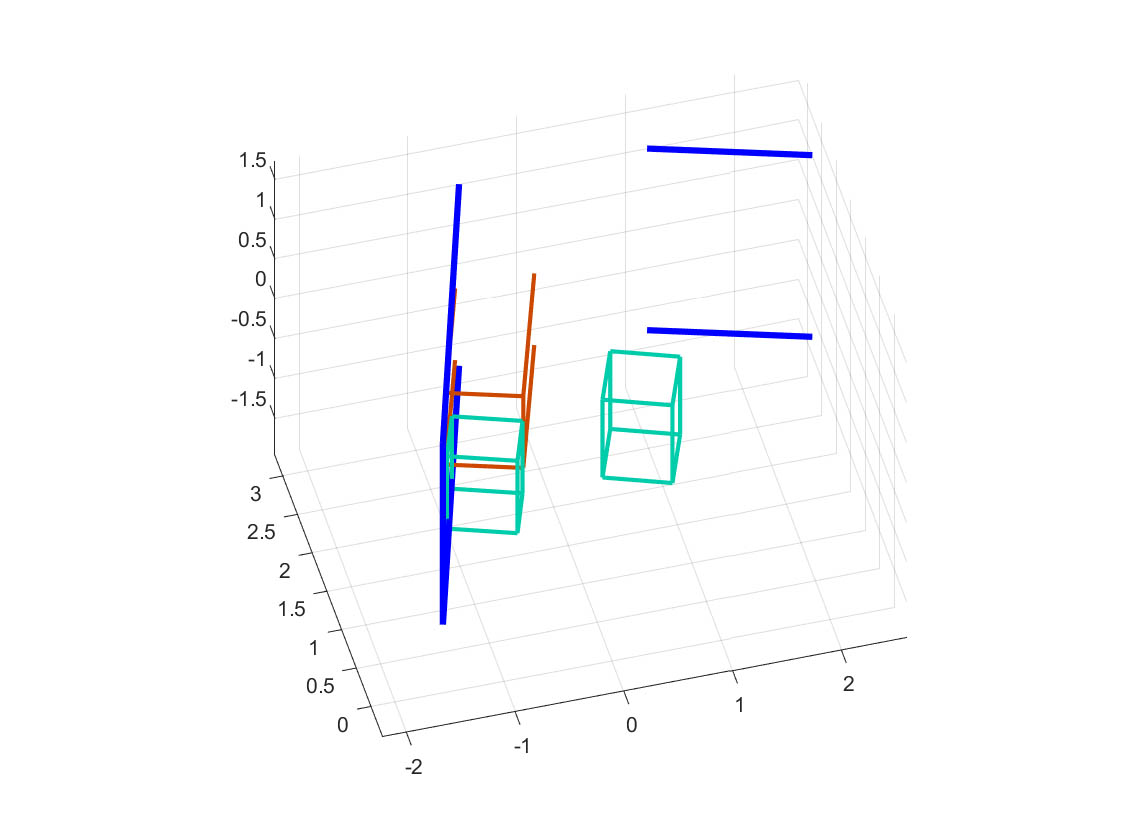} &
        	    \oblique{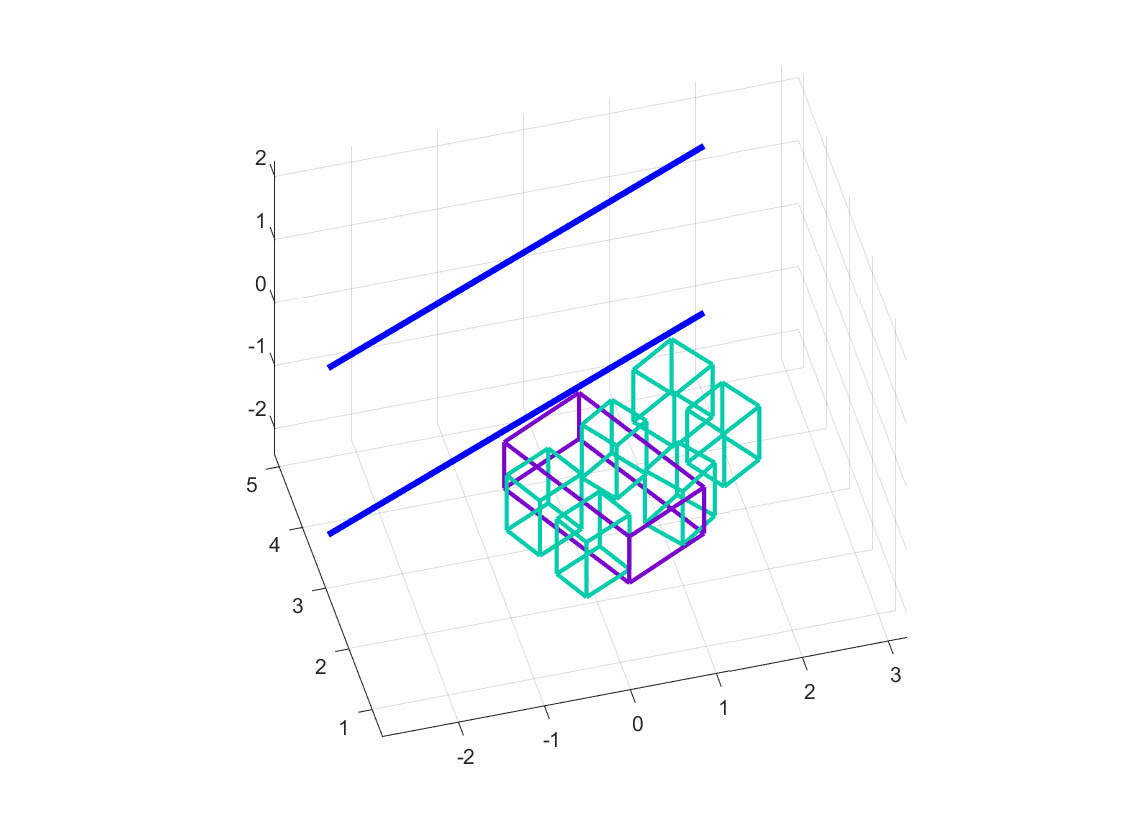} &
        	    \oblique{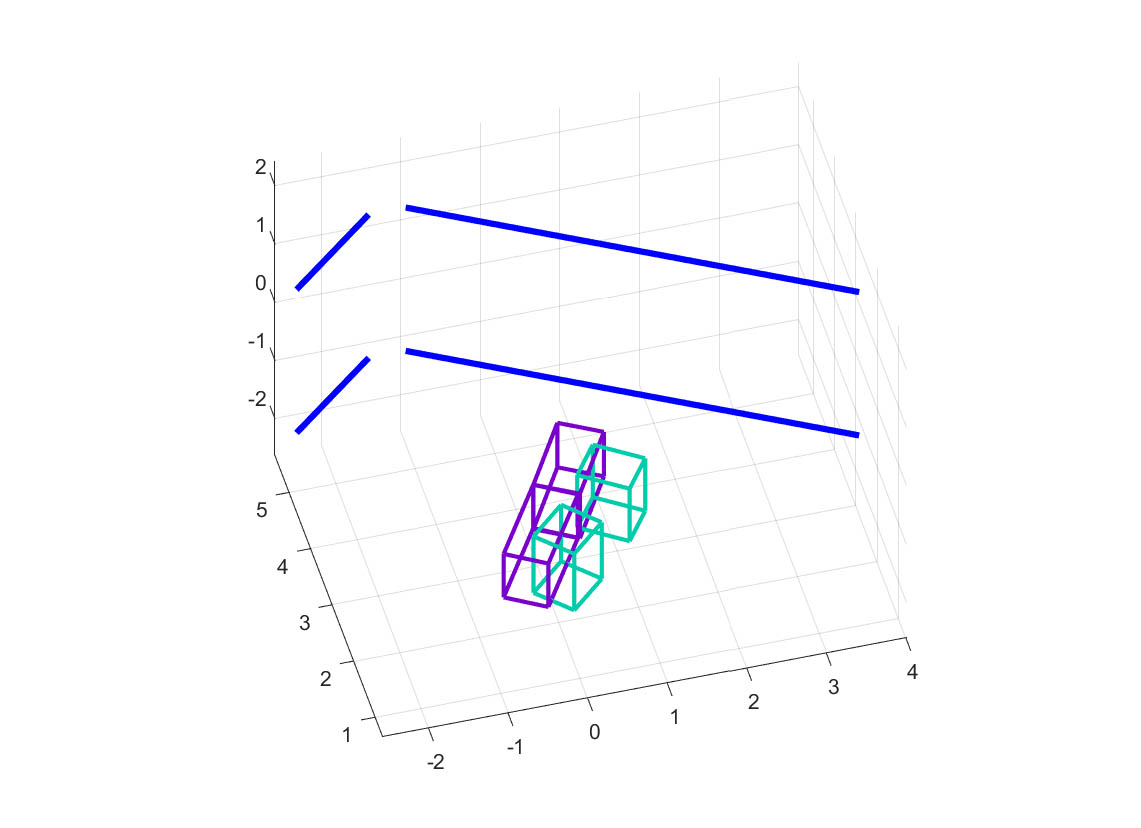} \\
    	\hline
    	    &\rot{Total3D}& 
        	    \rgb{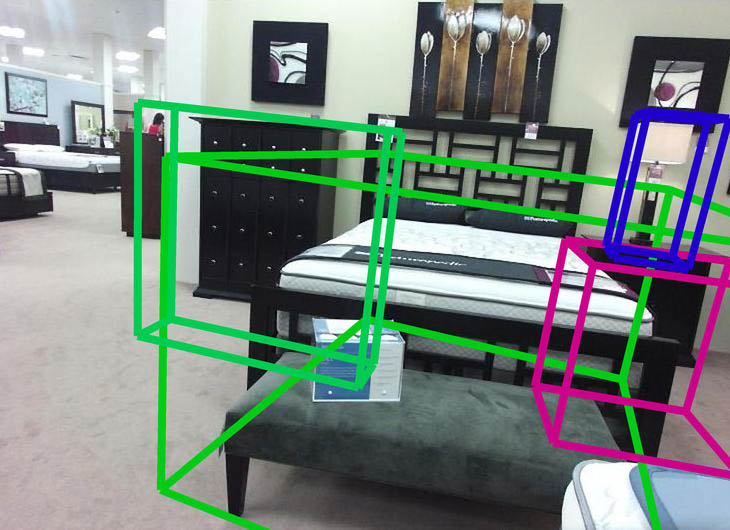} &
        	    \rgb{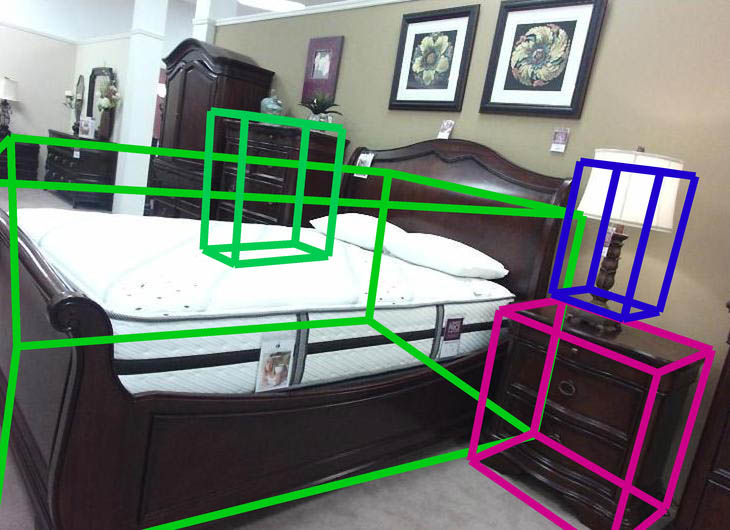} &
        	    \rgb{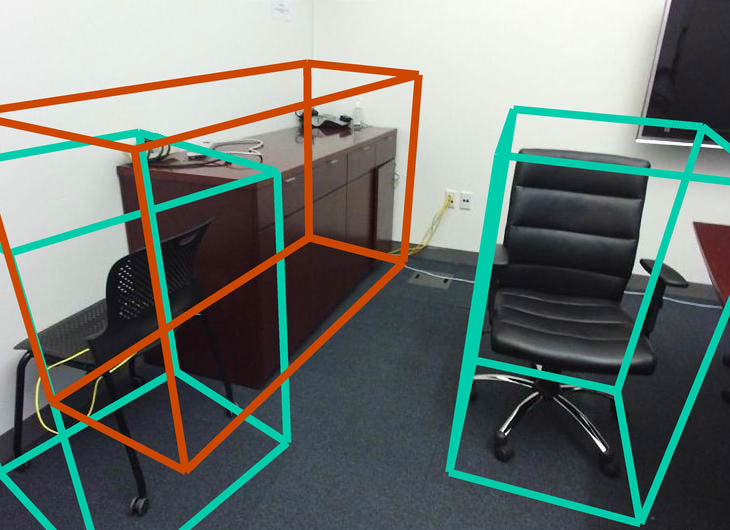} &
        	    \rgb{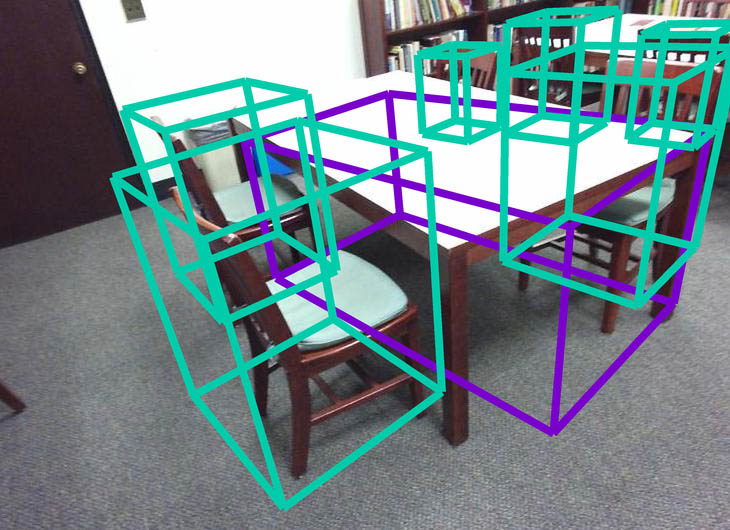} &
        	    \rgb{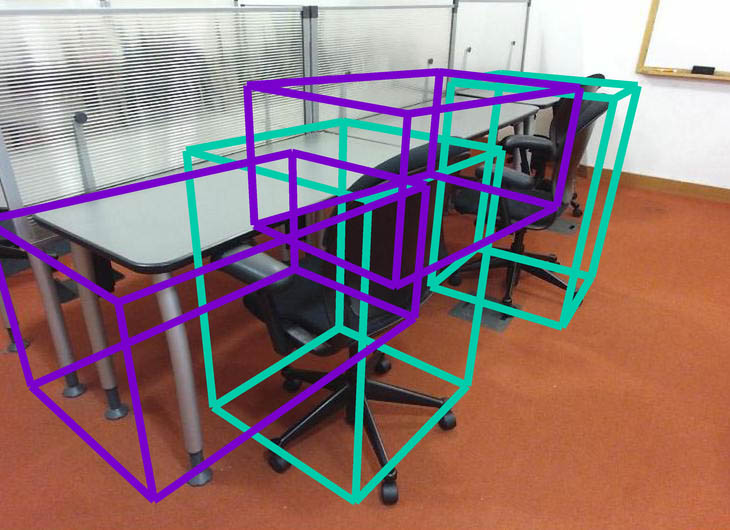} \\
            \rot{Camera View}&\rot{Ours}& 
        	    \rgb{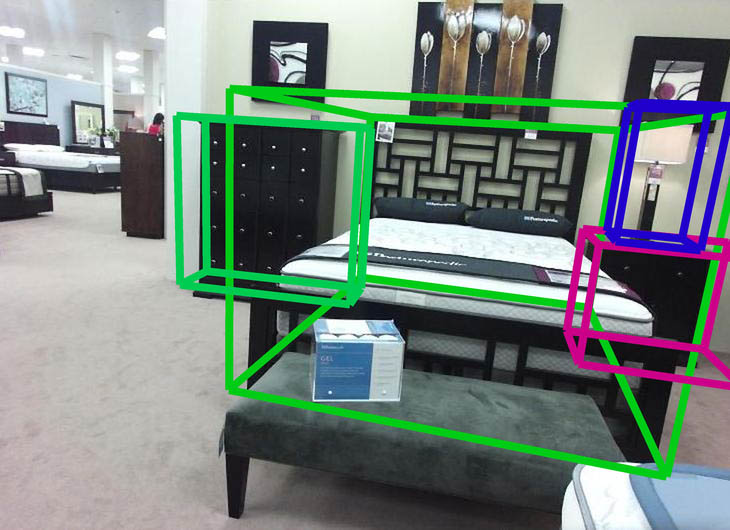} &
        	    \rgb{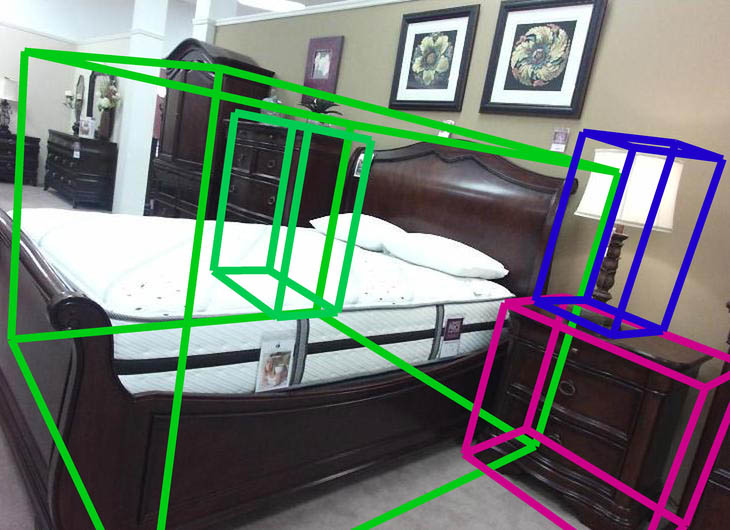} &
        	    \rgb{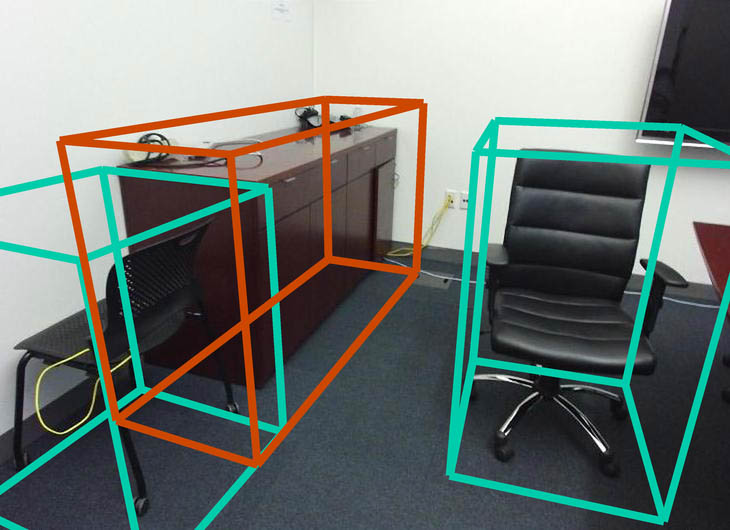} &
        	    \rgb{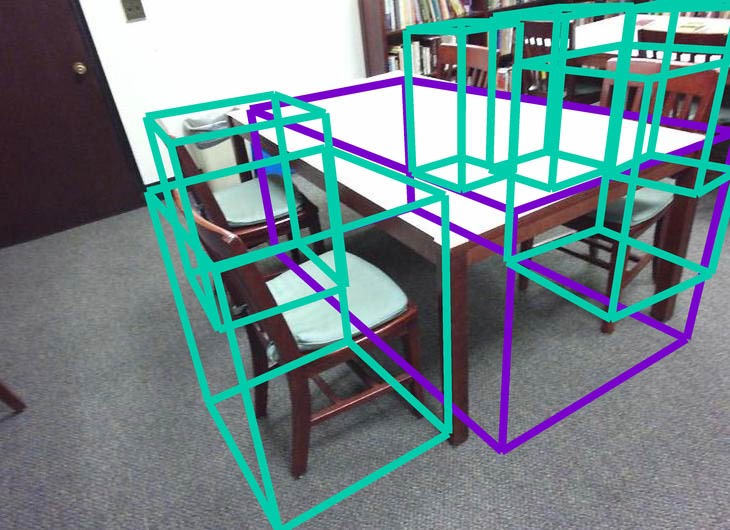} &
        	    \rgb{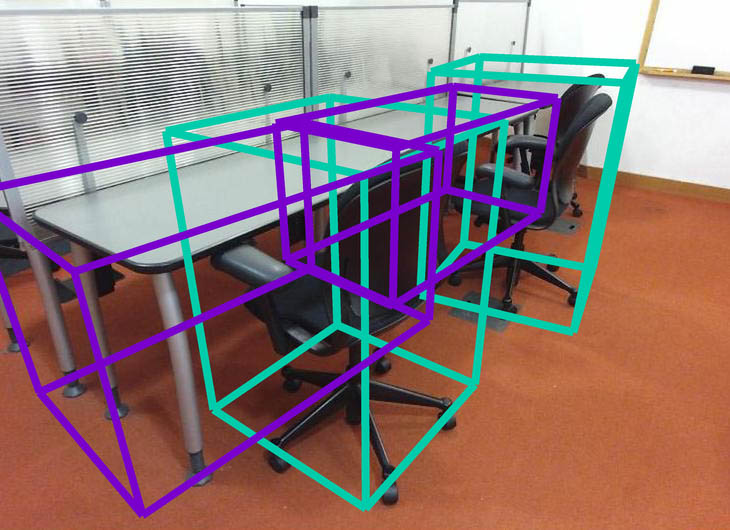} \\
        	&\rot{GT}& 
        	    \rgb{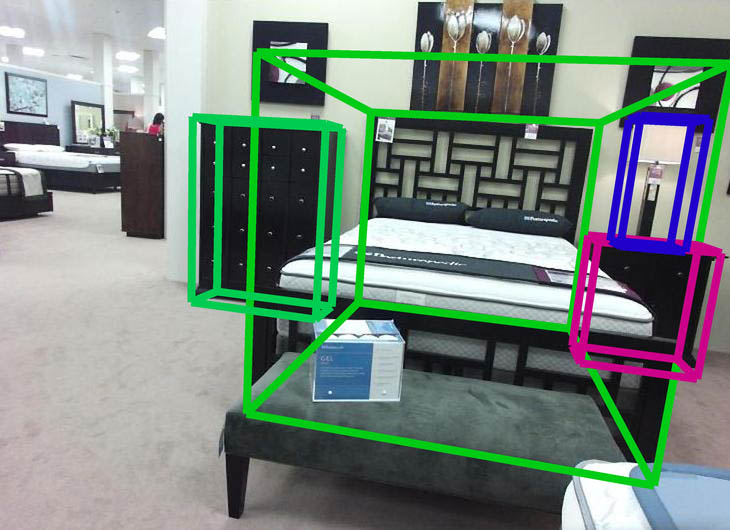} &
        	    \rgb{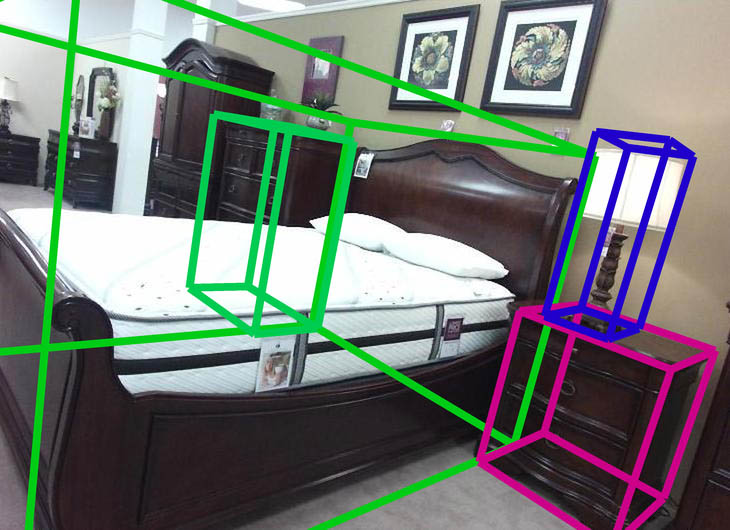} &
        	    \rgb{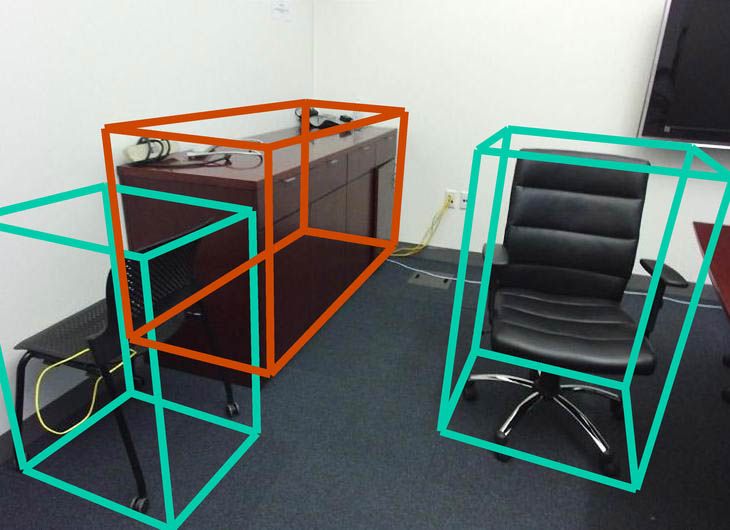} &
        	    \rgb{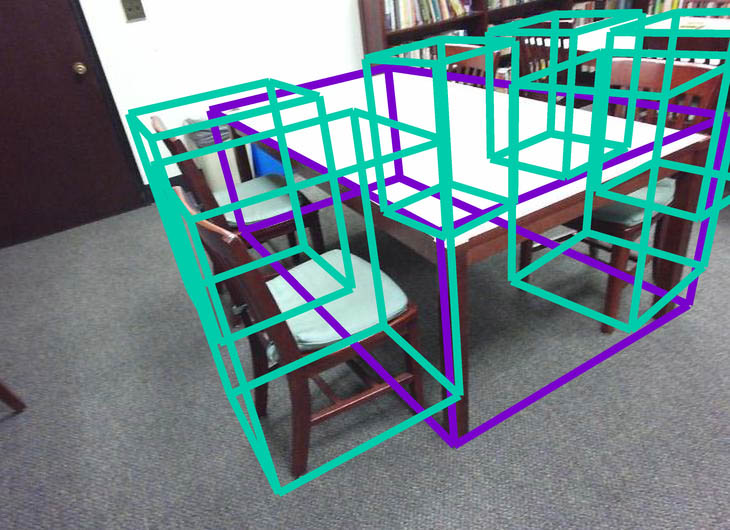} &
        	    \rgb{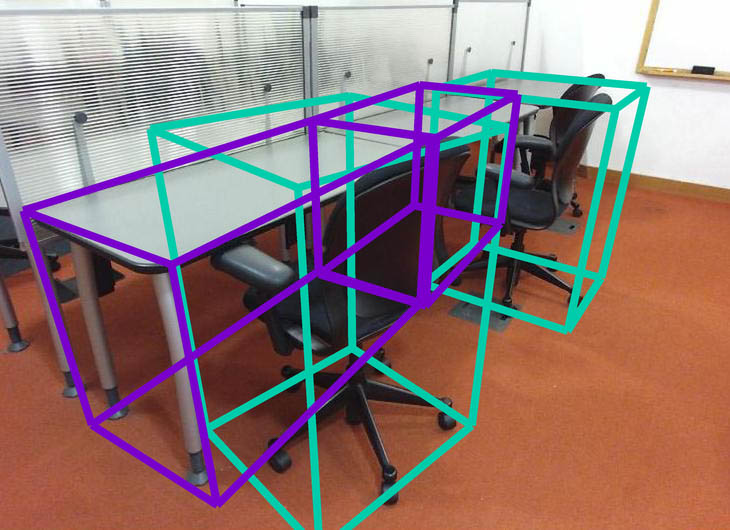} \\
    	\hline
    	\multirow{2}{*}{\rot{Scene Mesh}}
        	&\rot{Total3D}& 
        	    \rgb{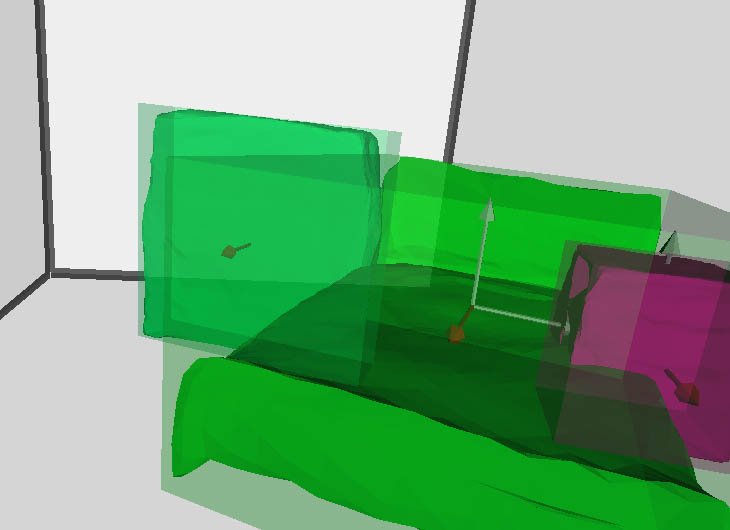} &
        	    \rgb{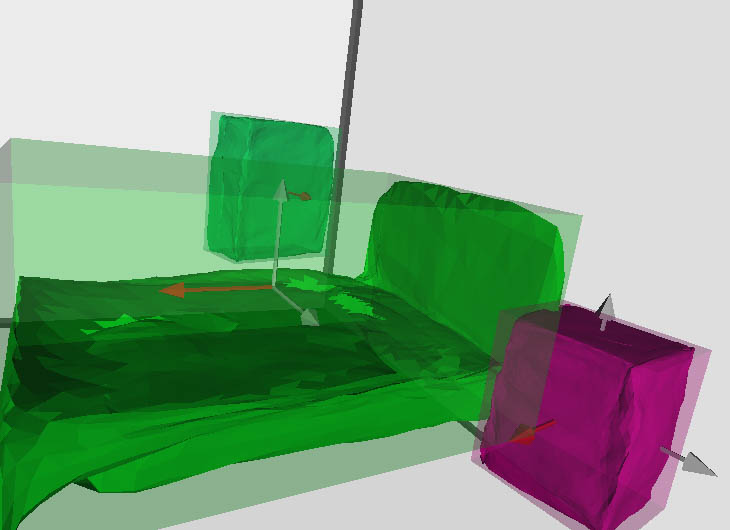} &
        	    \rgb{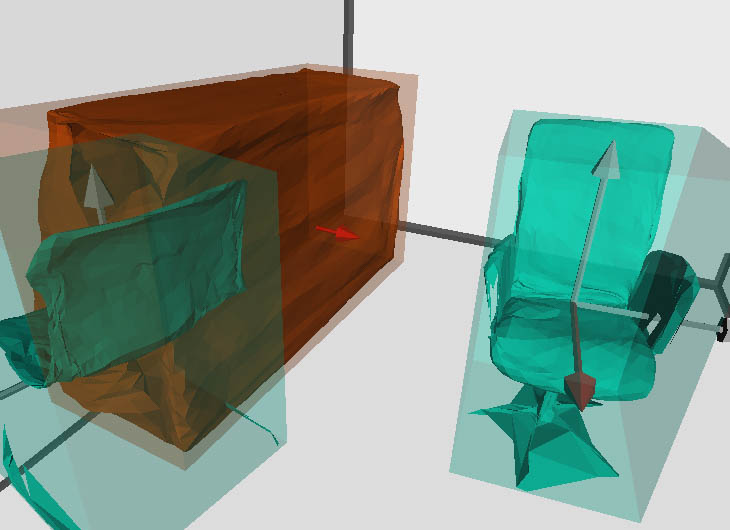} &
        	    \rgb{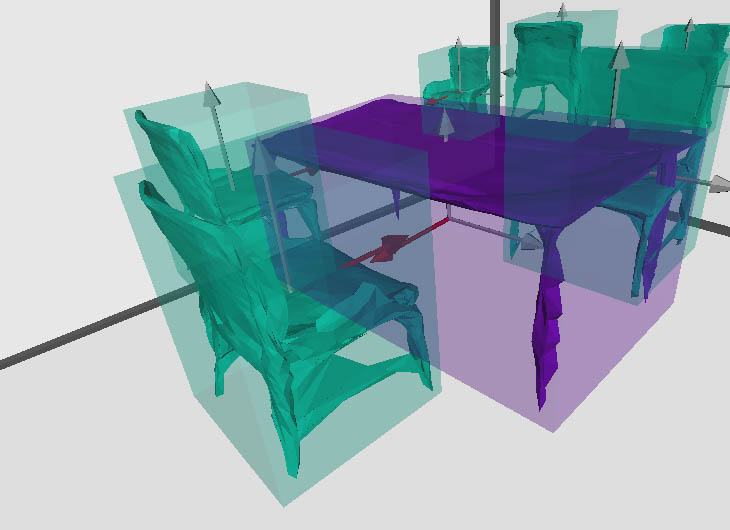} &
        	    \rgb{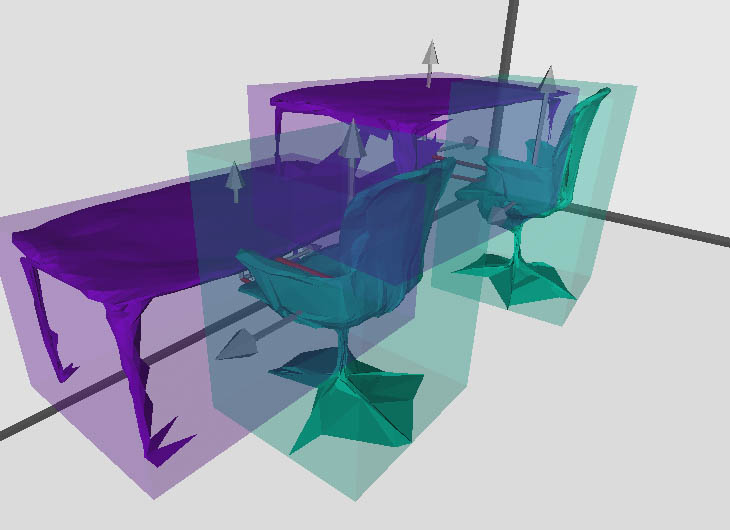} \\
        	&\rot{Ours}& 
        	    \rgb{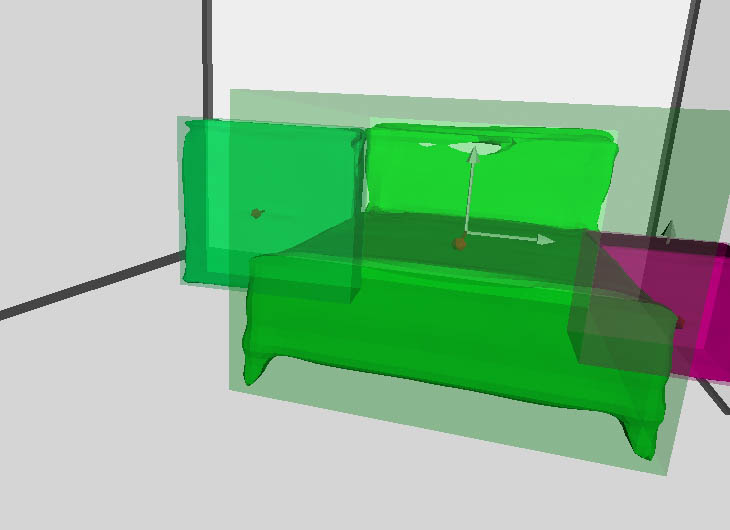} &
        	    \rgb{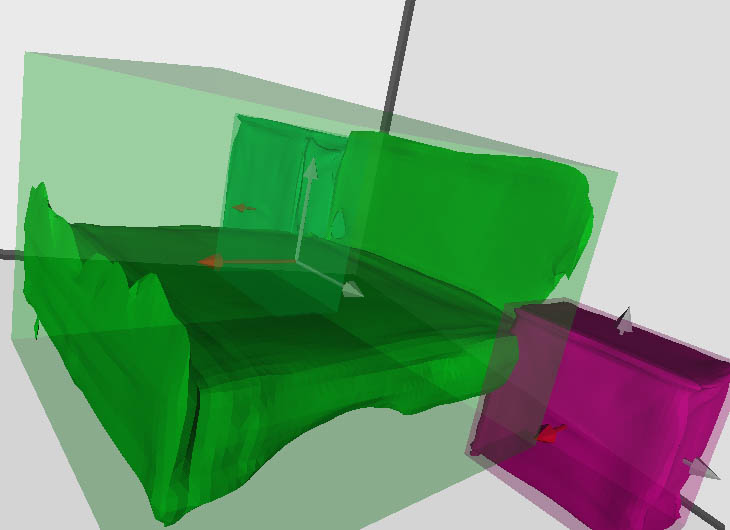} &
        	    \rgb{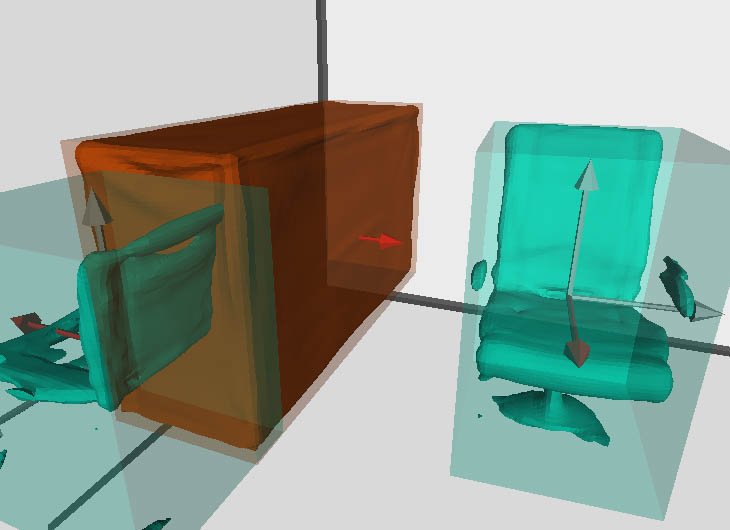} &
        	    \rgb{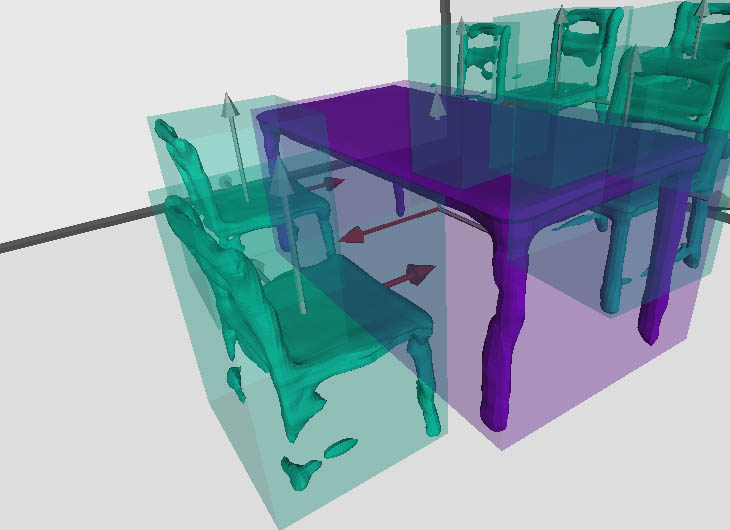} &
        	    \rgb{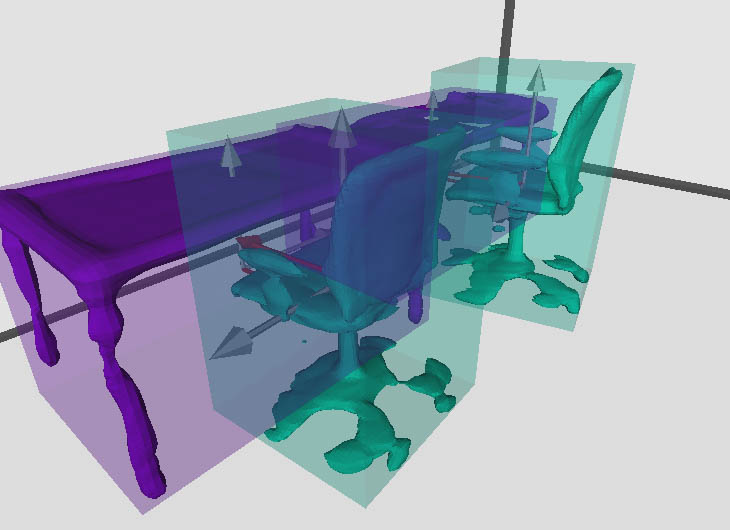} \vspace{2px}\\
        \multicolumn{2}{c}{}& (a) & (b) & (c) & (d) & (e) \\
    \end{tabular}
    % \vspace{-1.0em}
	\caption{Qualitative comparisons on object detection and scene reconstruction.}
	\label{fig:scnreconSupp1}
\end{figure*}

\begin{figure*}[!ht]
	\centering
	\scriptsize
	\newcommand{\rgb}[1]{\raisebox{-0.5\height}{\includegraphics[width=.175\linewidth]{#1}}}
	\newcommand{\oblique}[1]{\raisebox{-0.5\height}{\includegraphics[width=.175\linewidth,clip,trim=48 12 48 12]{#1}}}
	\newcommand{\rot}[1]{\rotatebox[origin=c]{90}{#1}}
	\def\arraystretch{0.5}%
	\begin{tabular}{c|c*{5}{c@{\hspace{1px}}}}
        &\rot{Input}& 
    	    \rgb{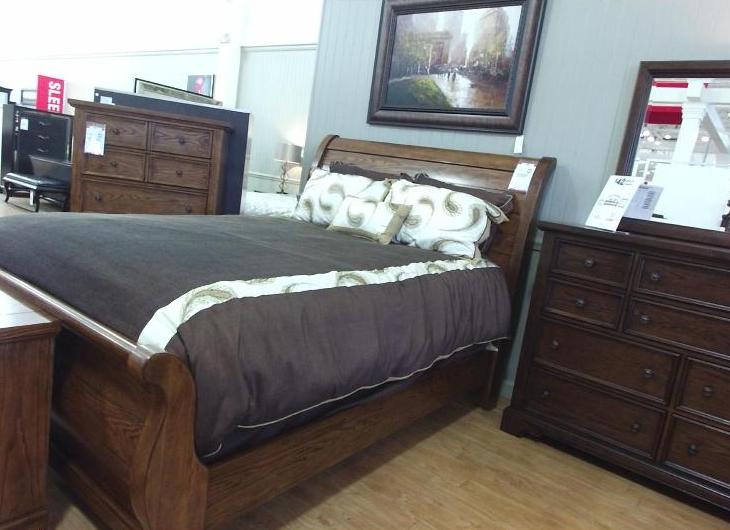} &
    	    \rgb{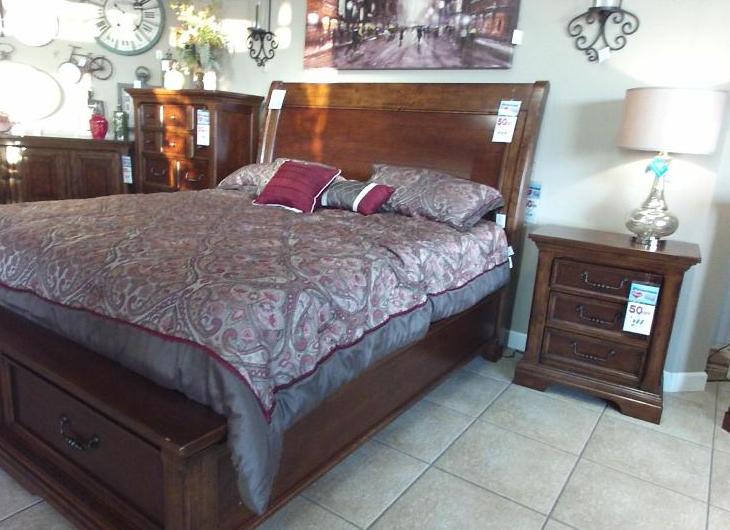} &
    	    \rgb{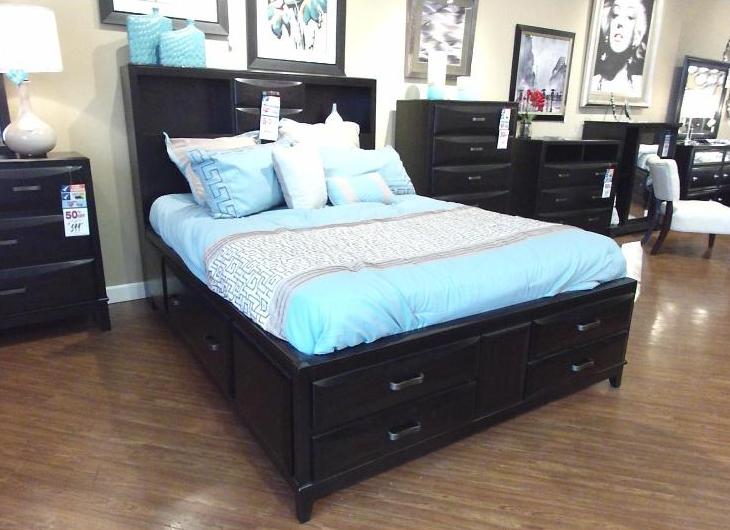} &
    	    \rgb{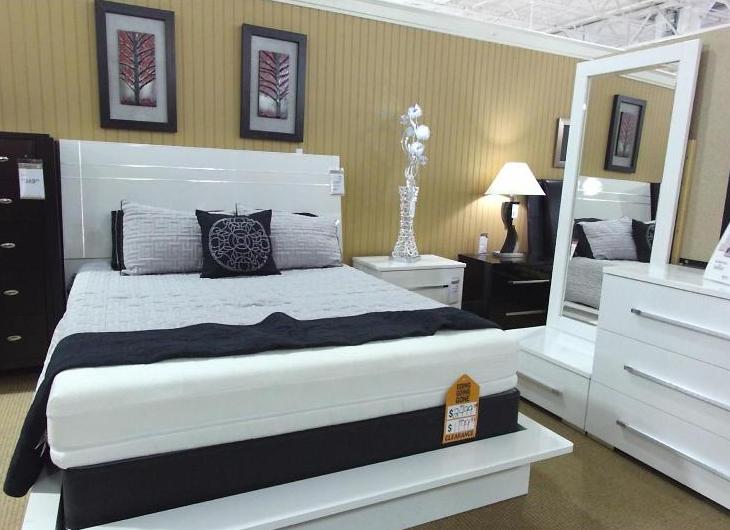} &
    	    \rgb{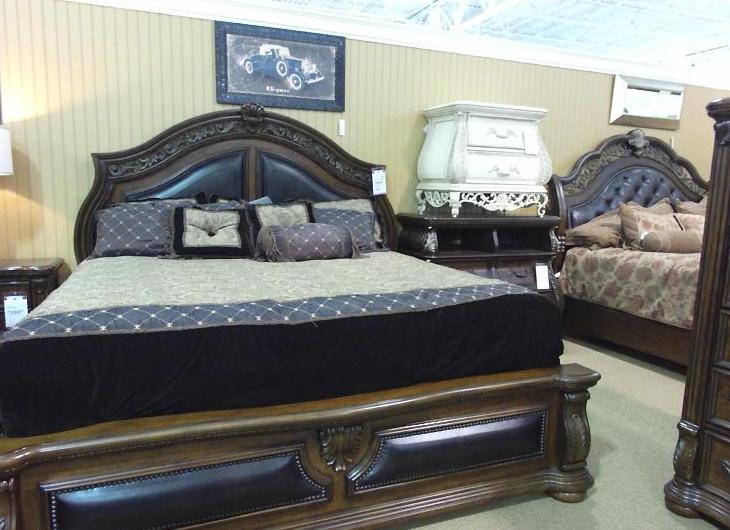} \\
    	\hline
    	    &\rot{Total3D}& 
        	    \oblique{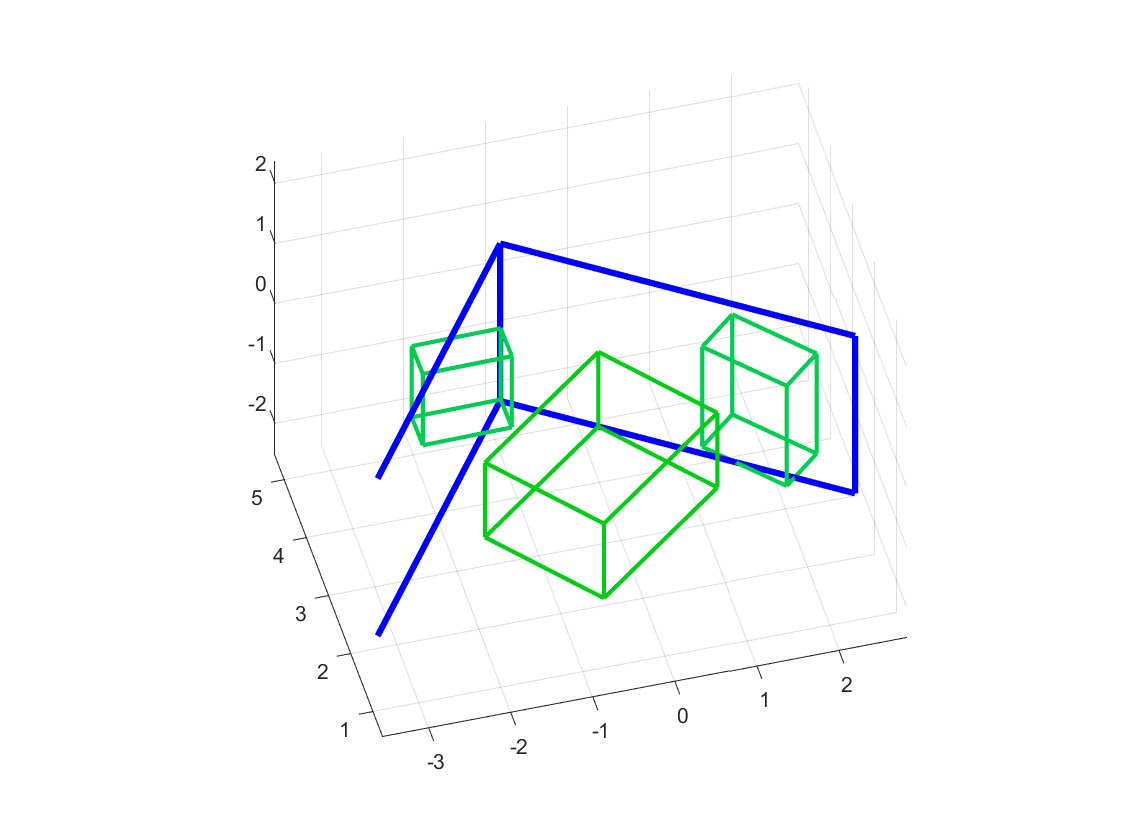} &
        	    \oblique{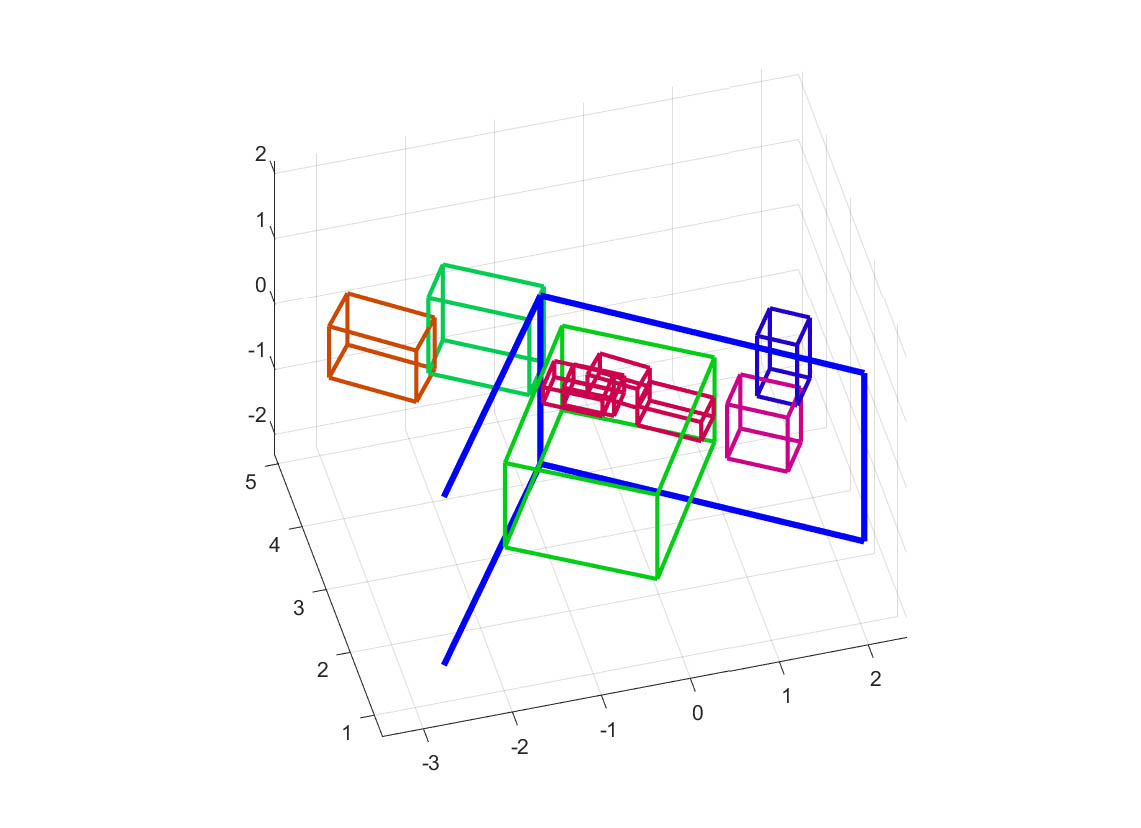} &
        	    \oblique{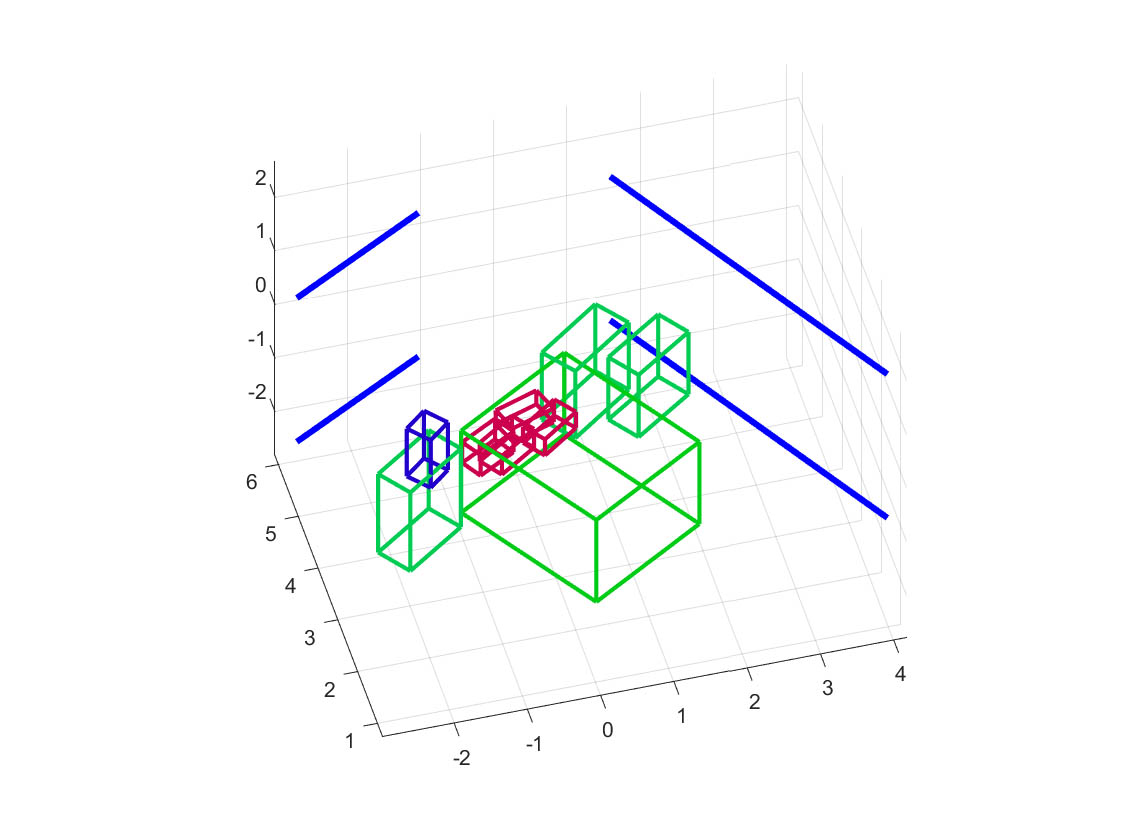} &
        	    \oblique{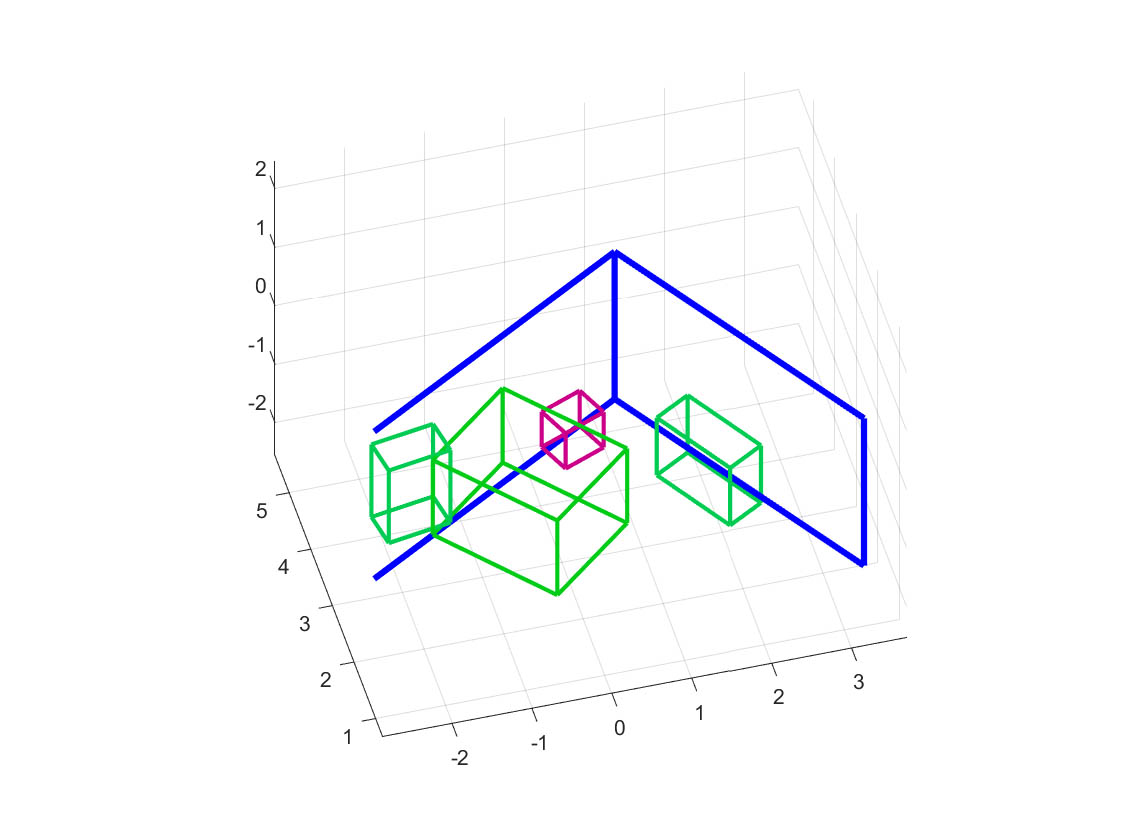} &
        	    \oblique{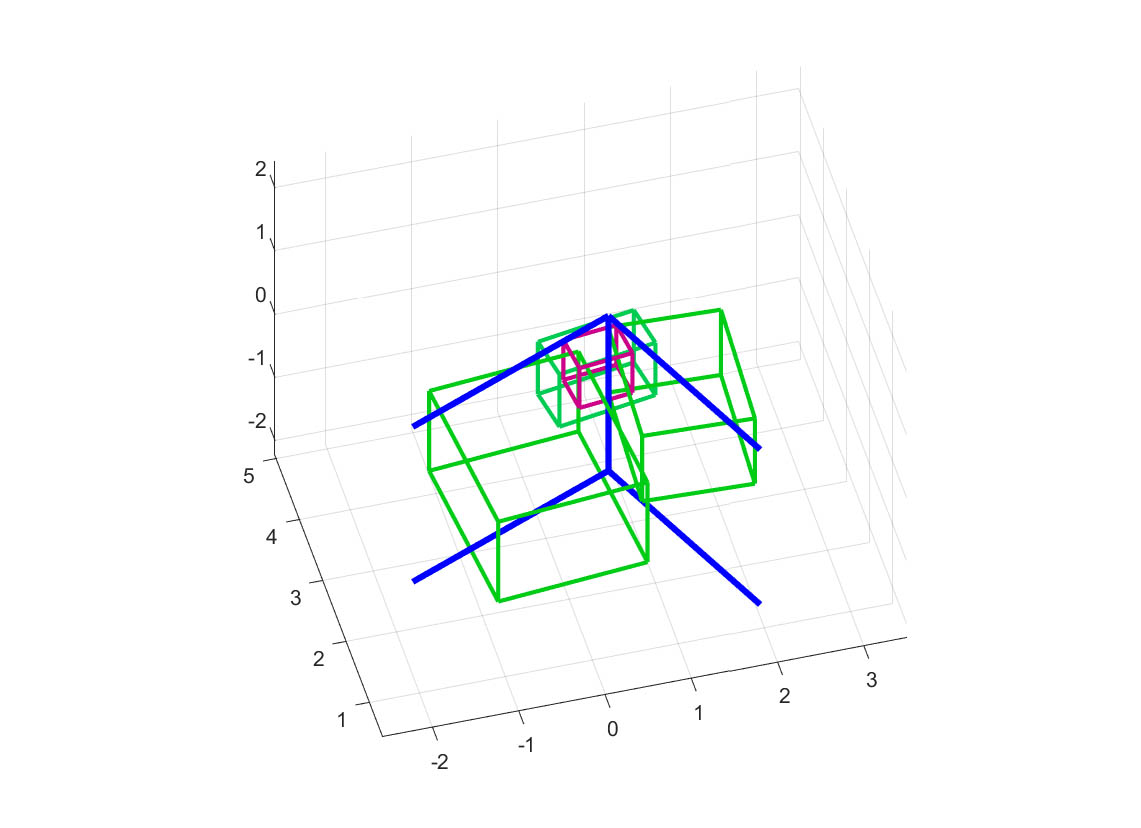} \\
        	\rot{Oblique View}& \rot{Ours}& 
        	    \oblique{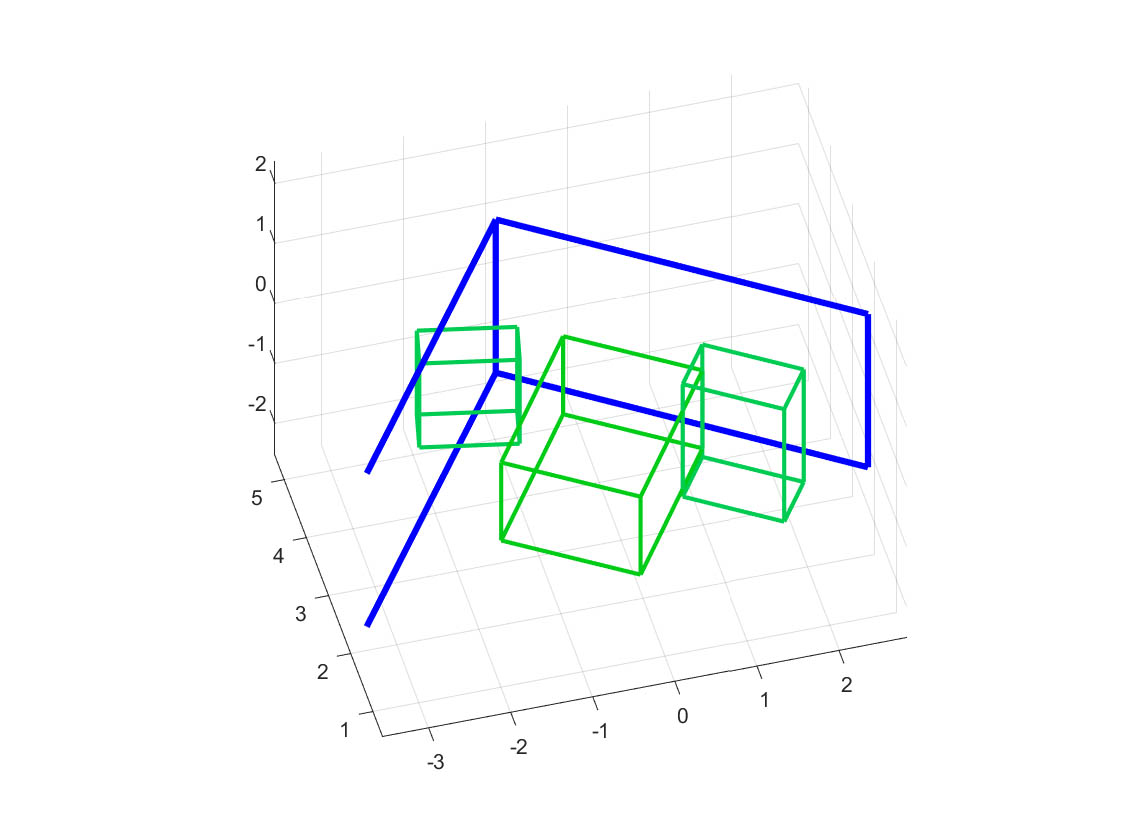} &
        	    \oblique{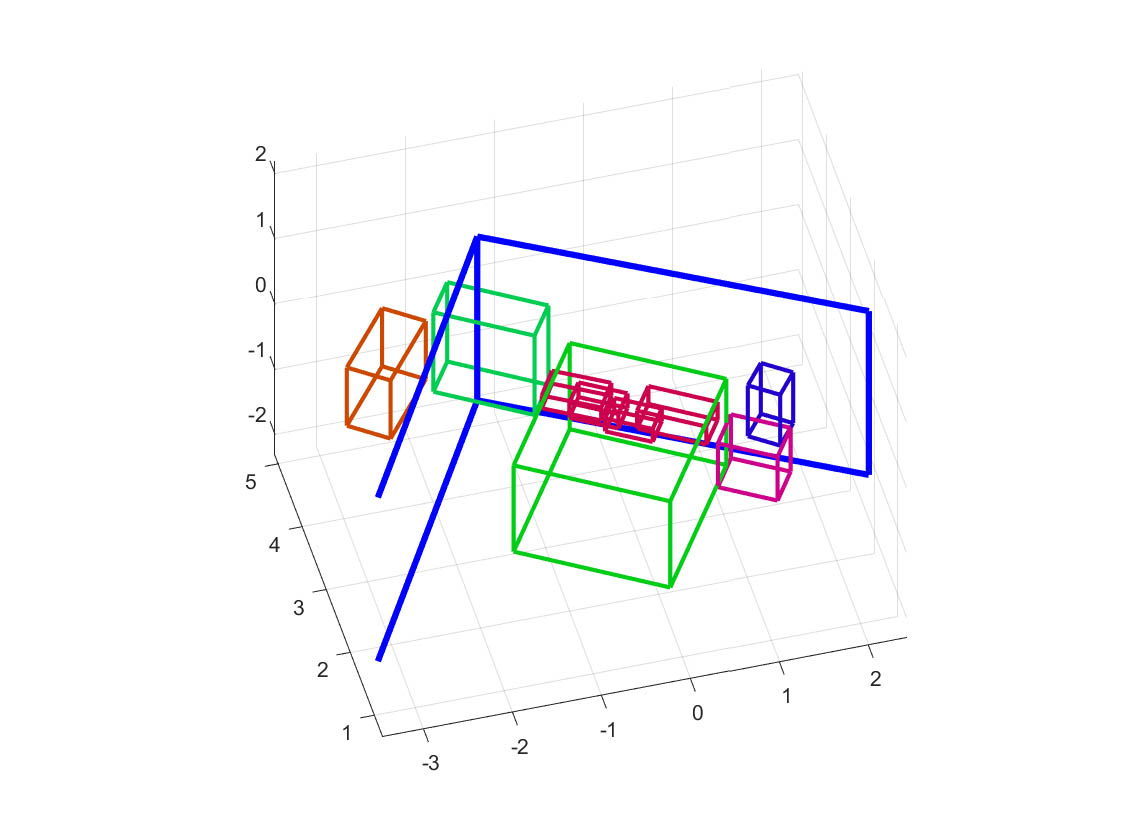} &
        	    \oblique{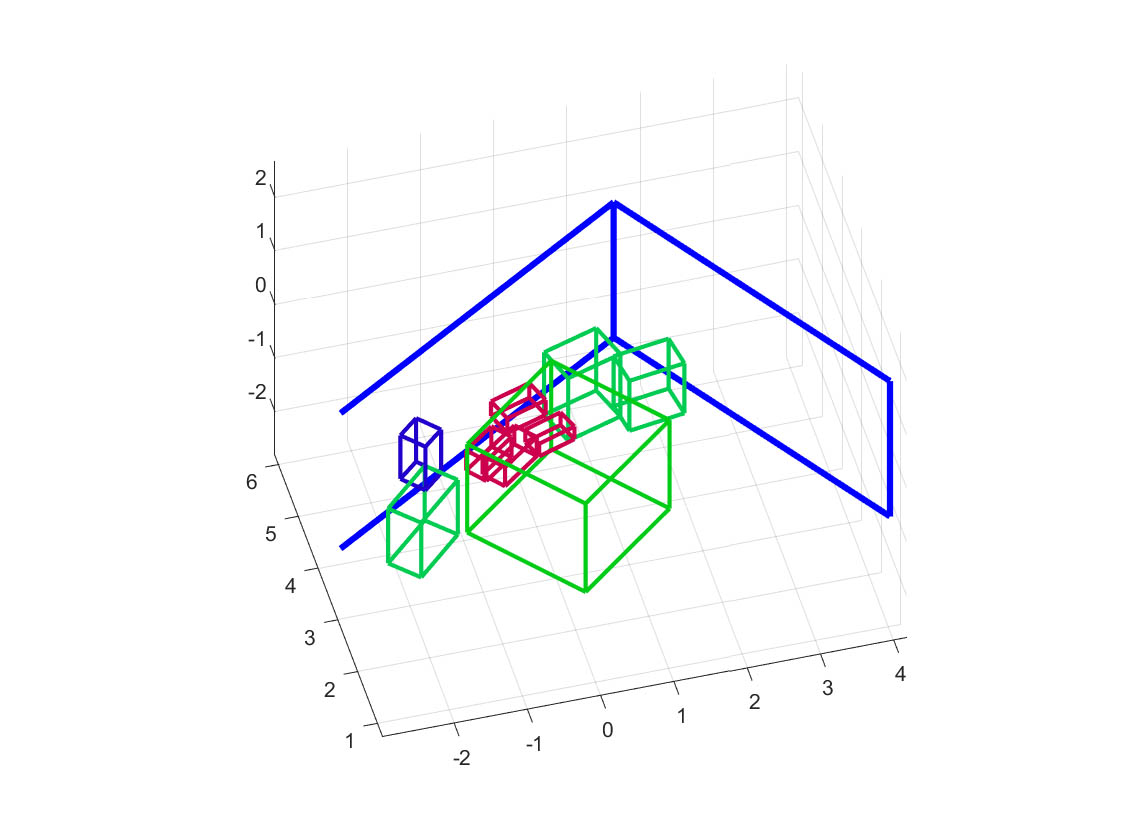} &
        	    \oblique{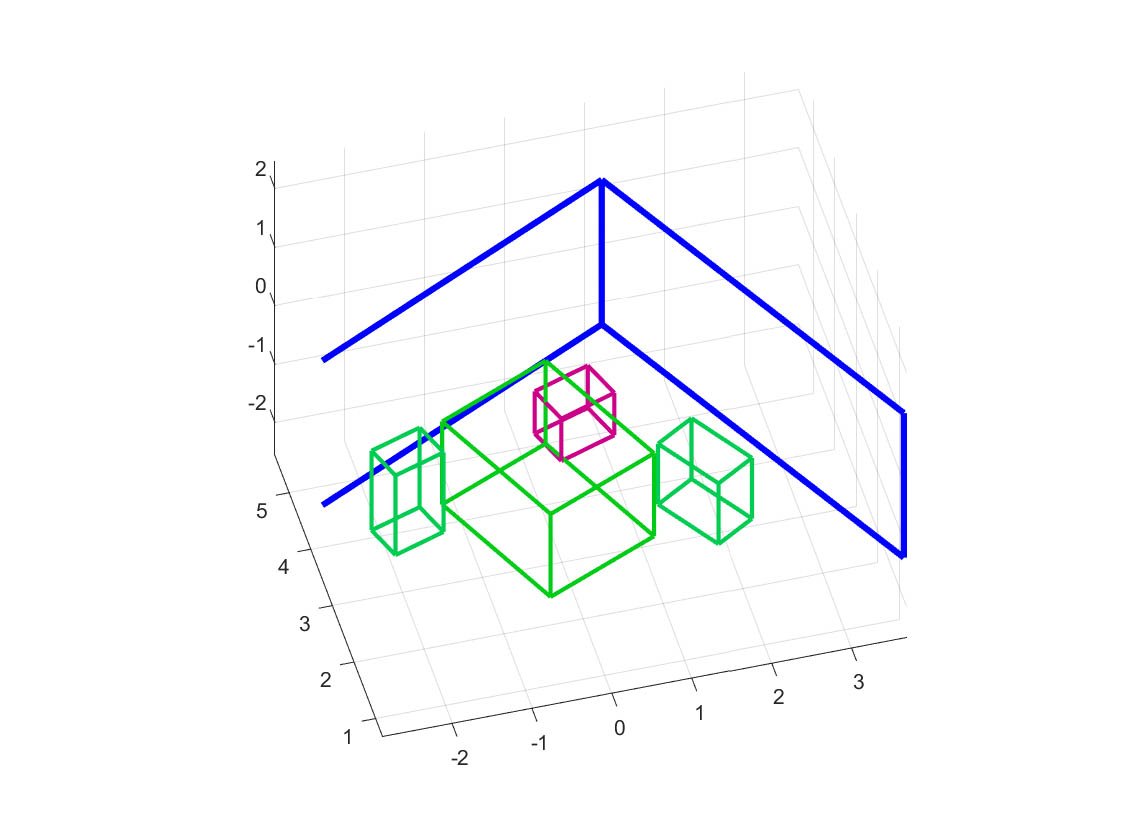} &
        	    \oblique{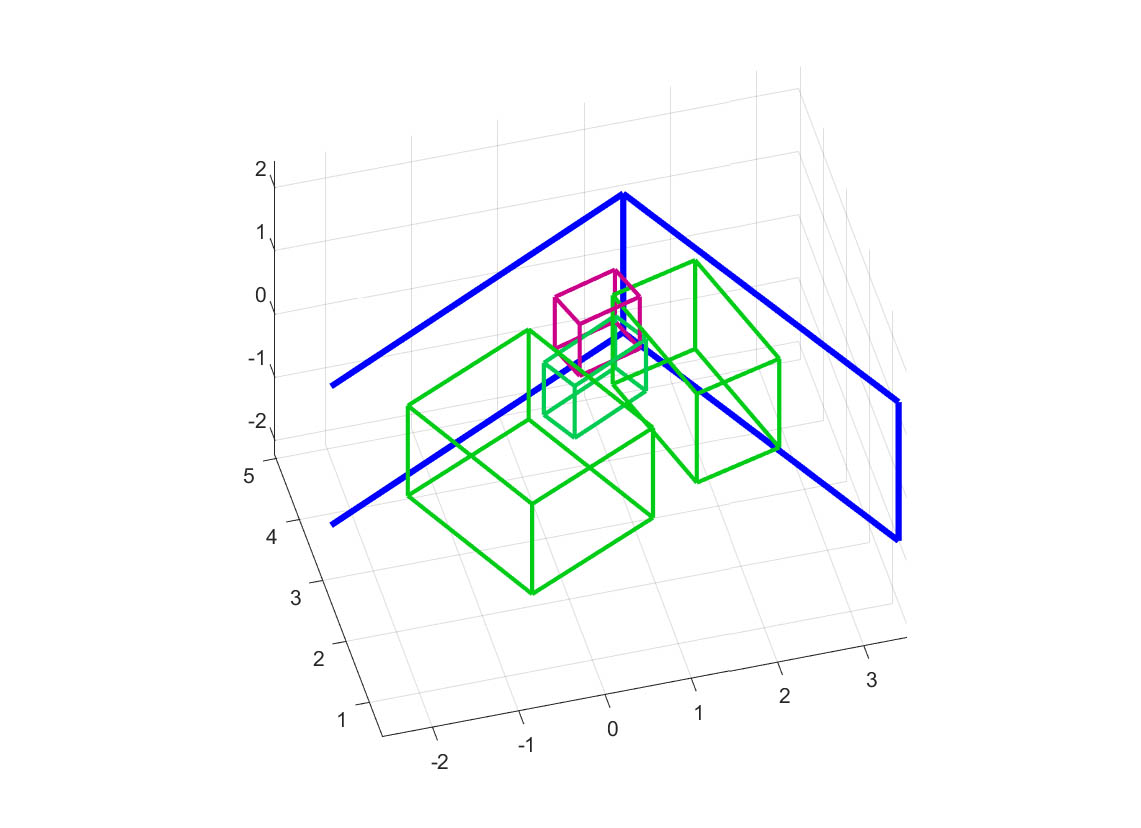} \\
        	&\rot{GT}& 
        	    \oblique{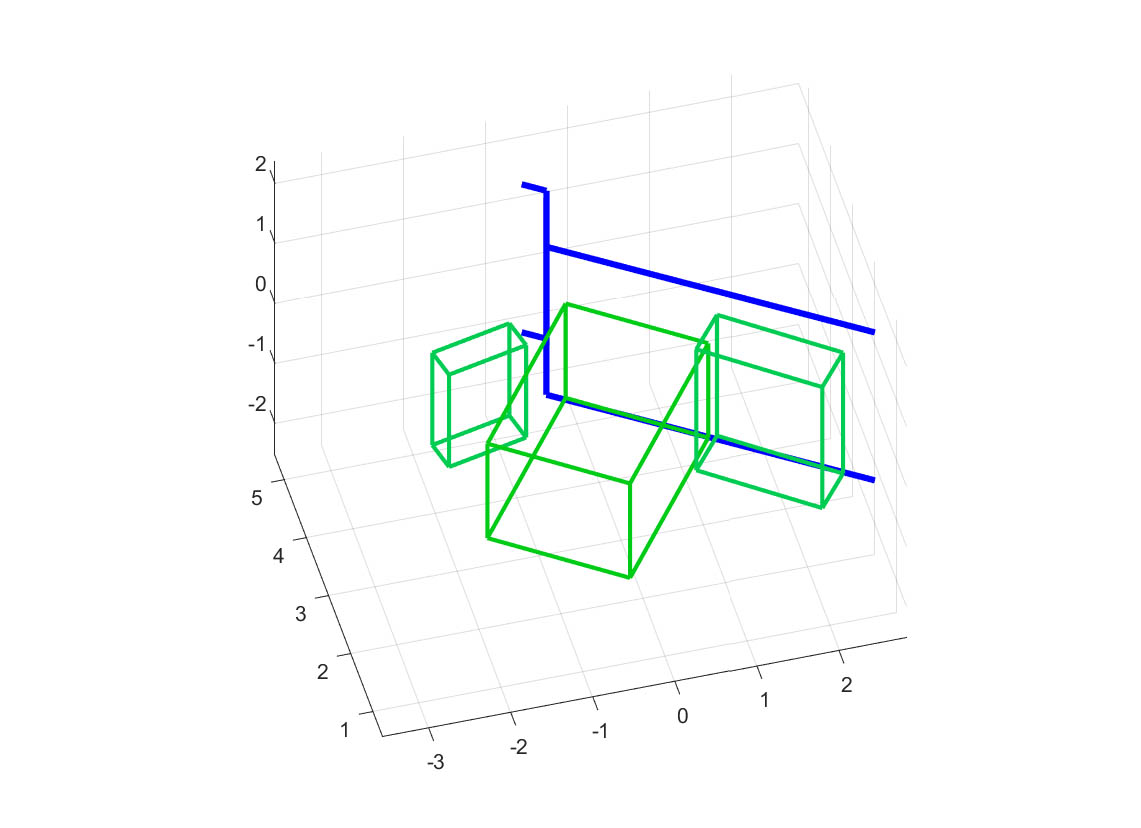} &
        	    \oblique{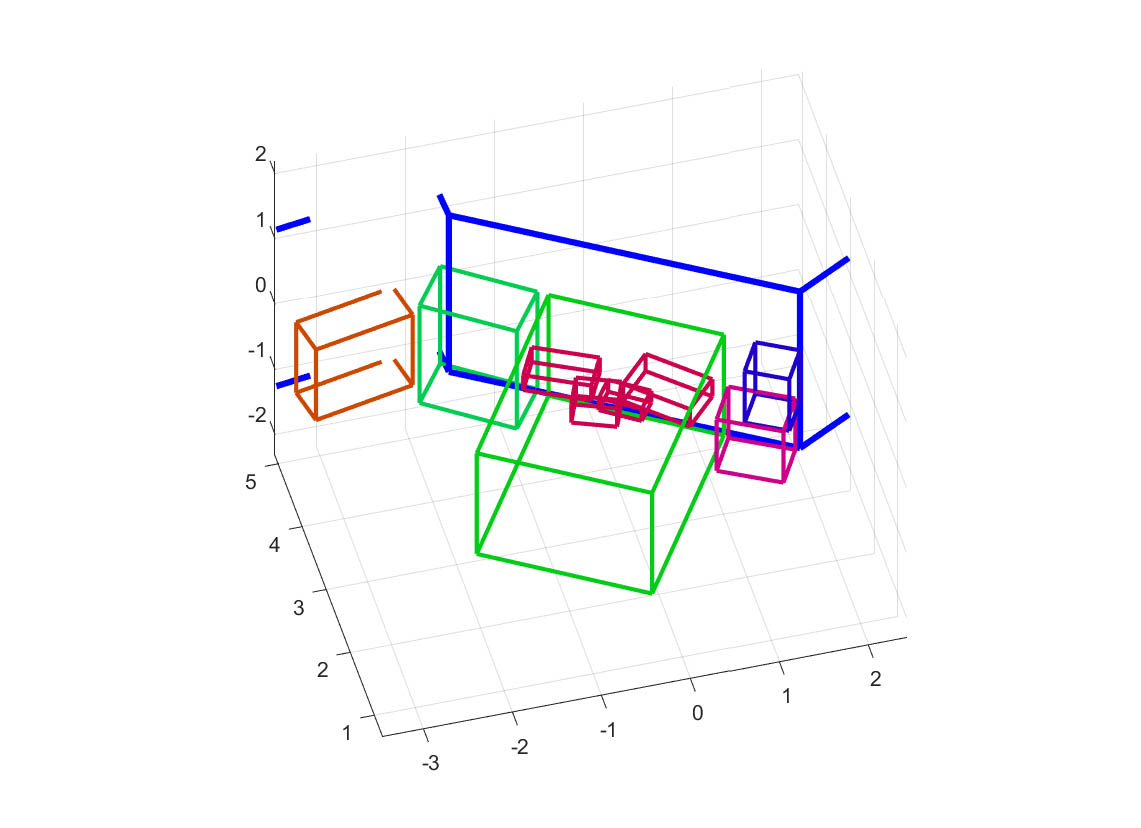} &
        	    \oblique{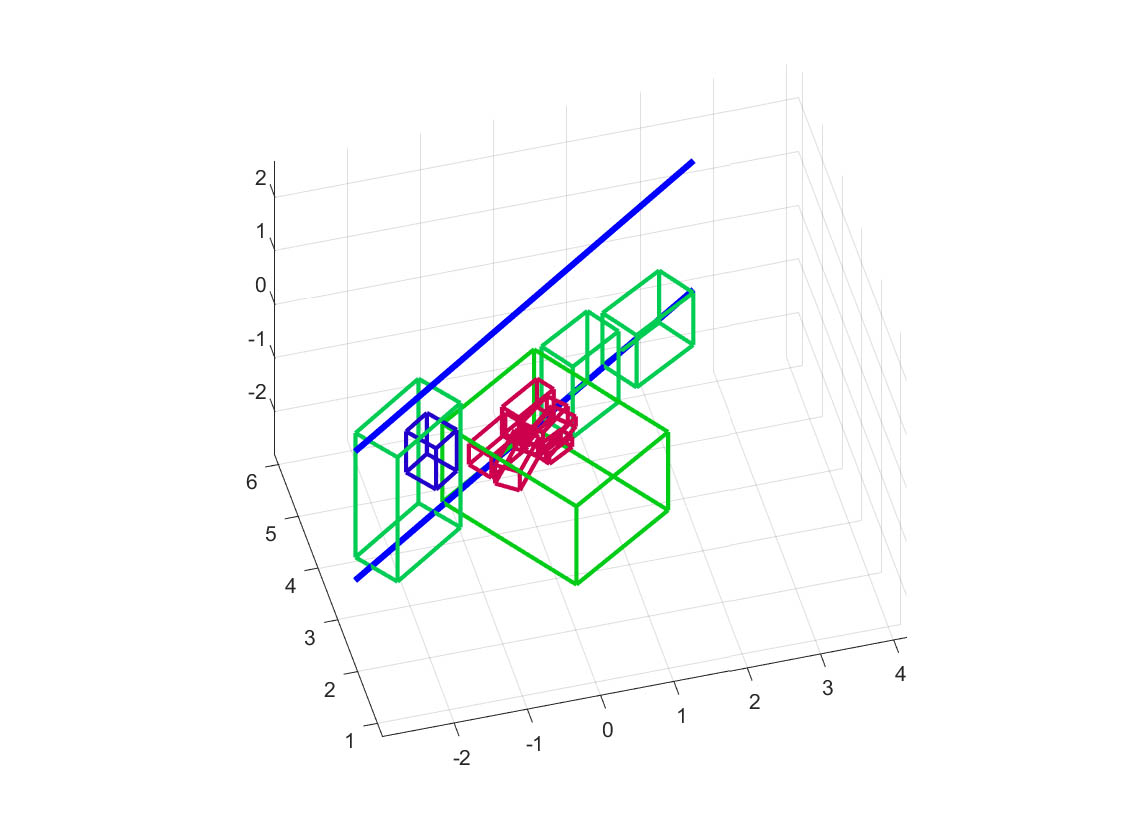} &
        	    \oblique{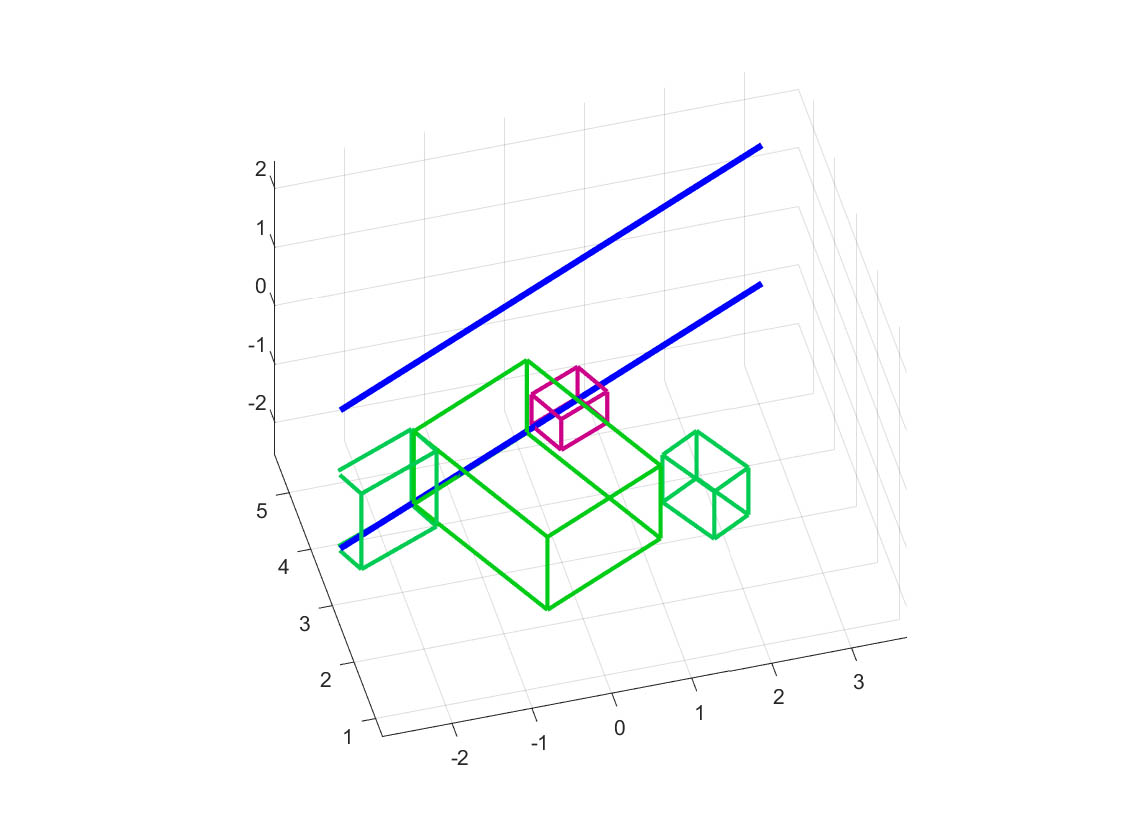} &
        	    \oblique{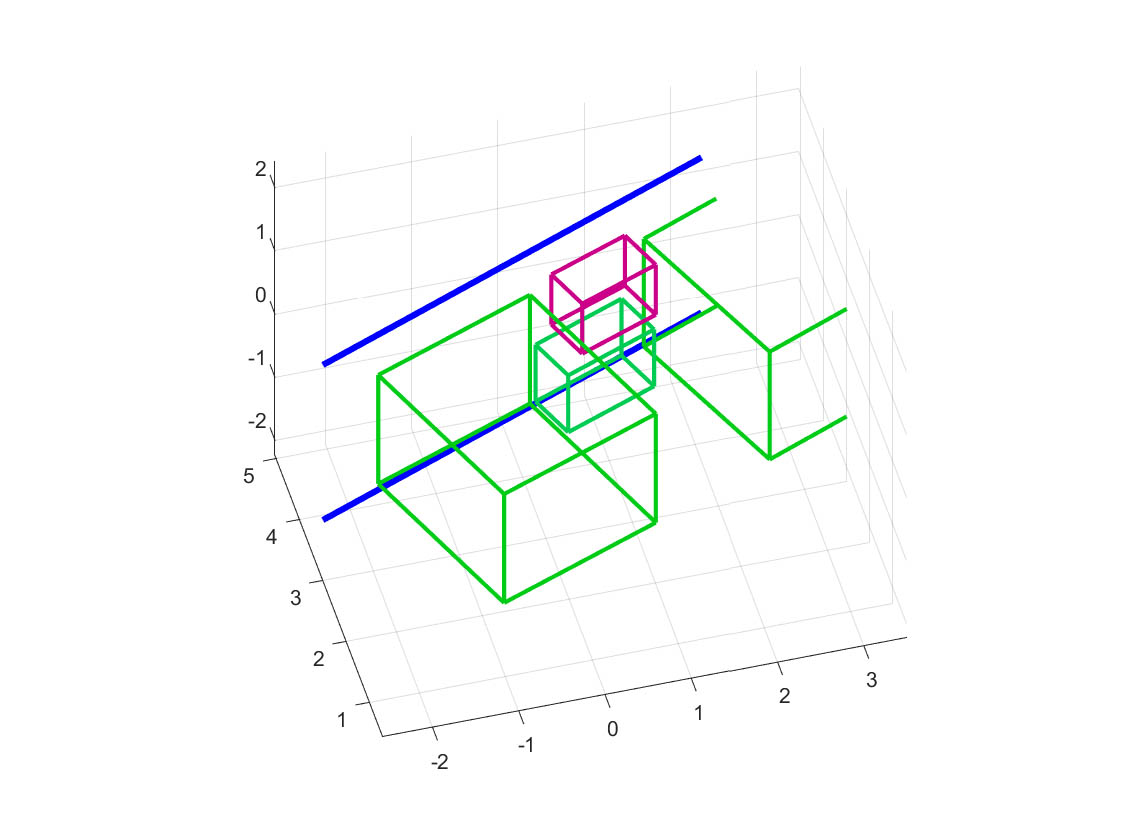} \\
    	\hline
    	    &\rot{Total3D}& 
        	    \rgb{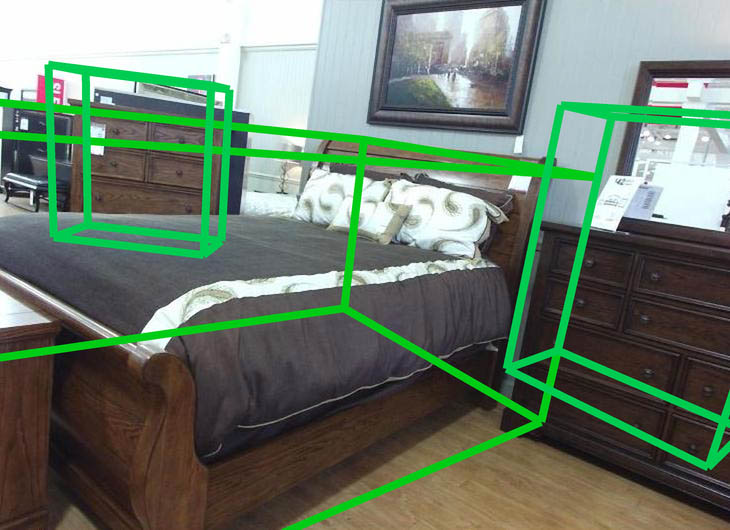} &
        	    \rgb{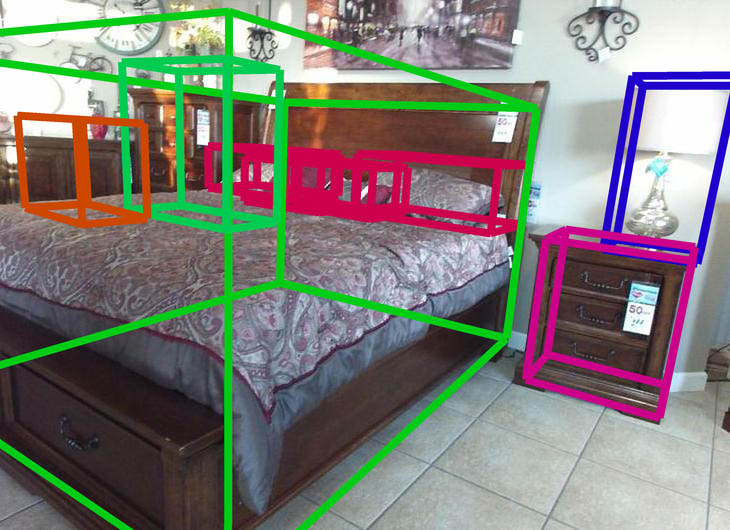} &
        	    \rgb{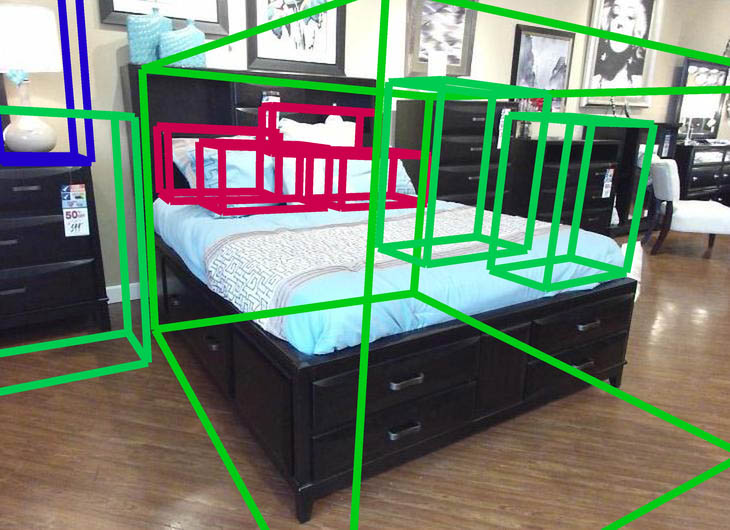} &
        	    \rgb{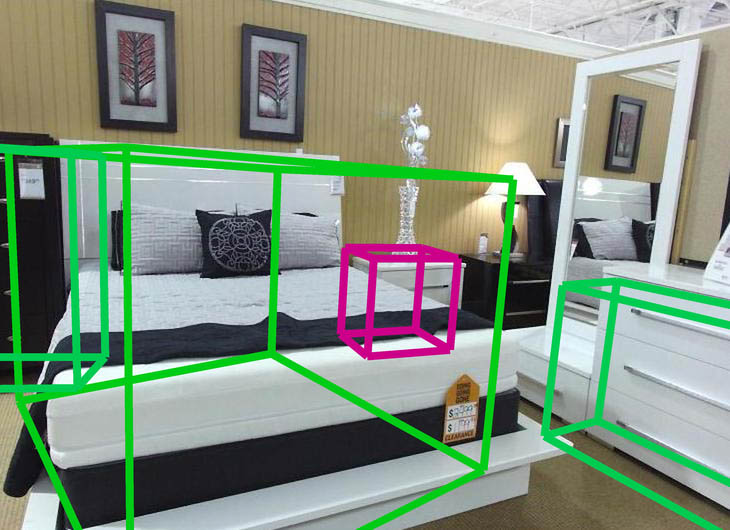} &
        	    \rgb{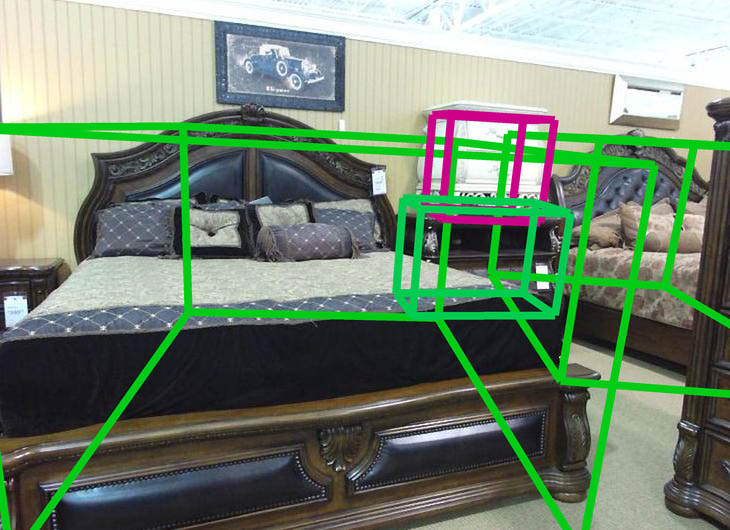} \\
            \rot{Camera View}&\rot{Ours}& 
        	    \rgb{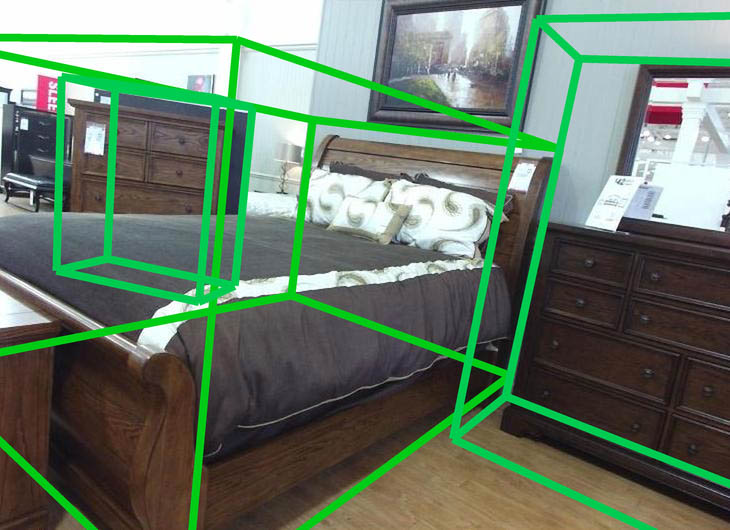} &
        	    \rgb{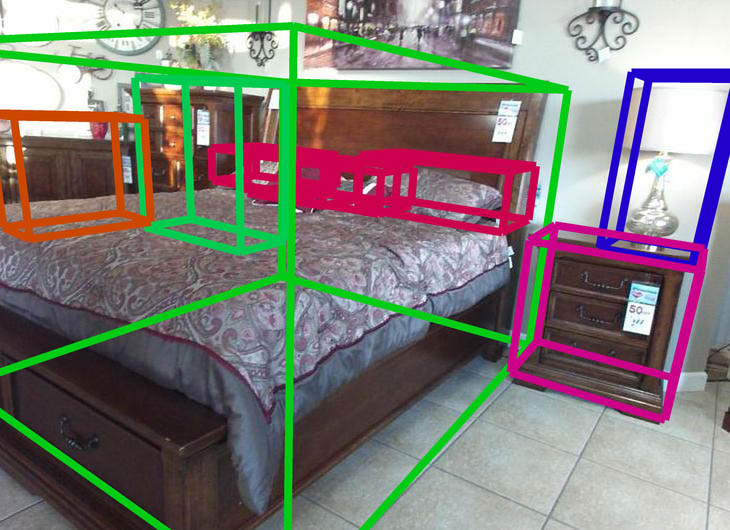} &
        	    \rgb{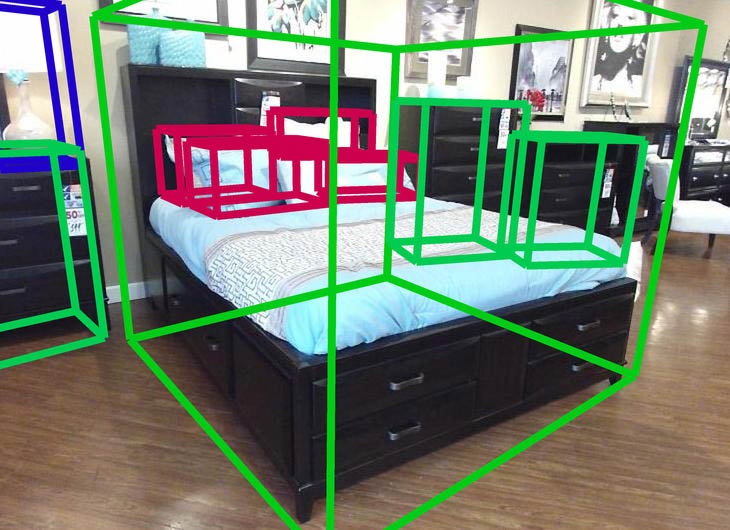} &
        	    \rgb{figure/scnrecon/ours/845_bbox} &
        	    \rgb{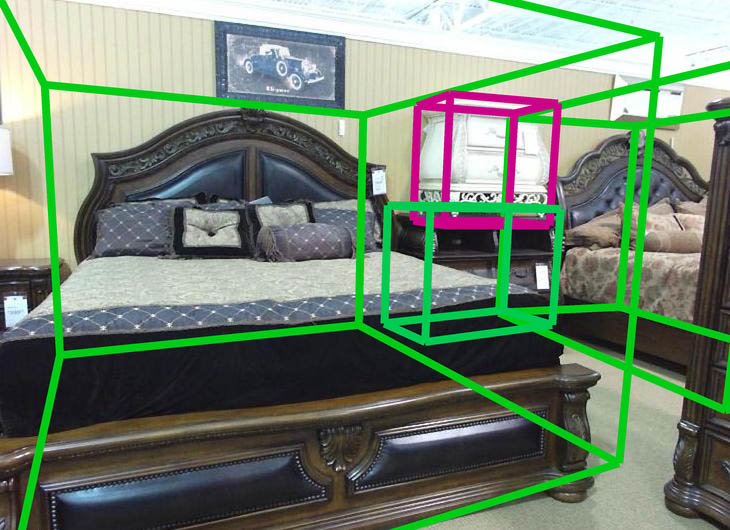} \\
        	&\rot{GT}& 
        	    \rgb{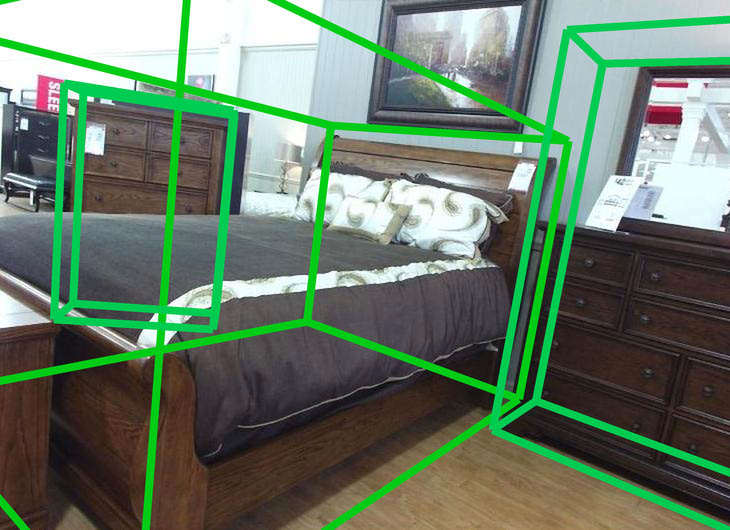} &
        	    \rgb{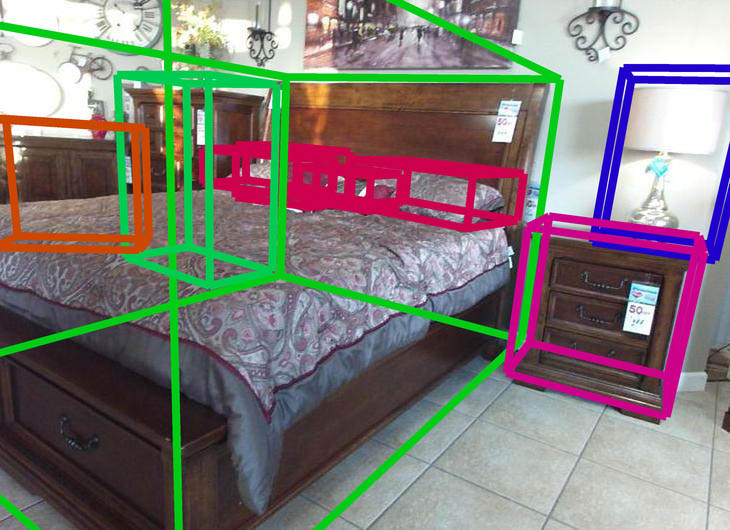} &
        	    \rgb{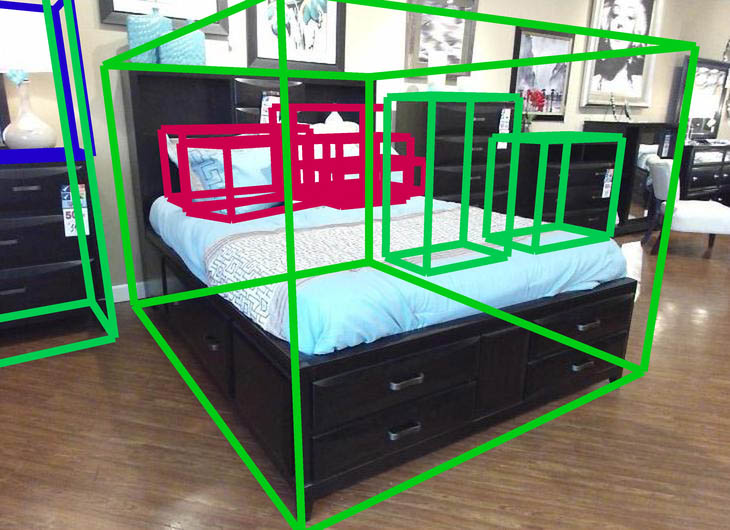} &
        	    \rgb{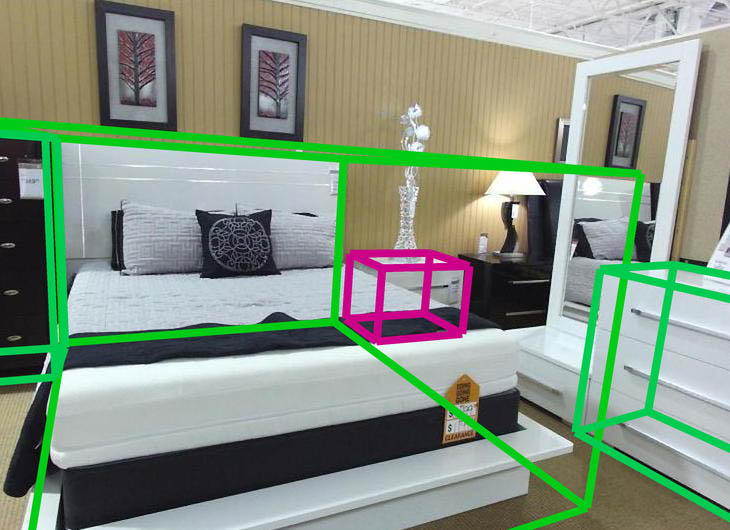} &
        	    \rgb{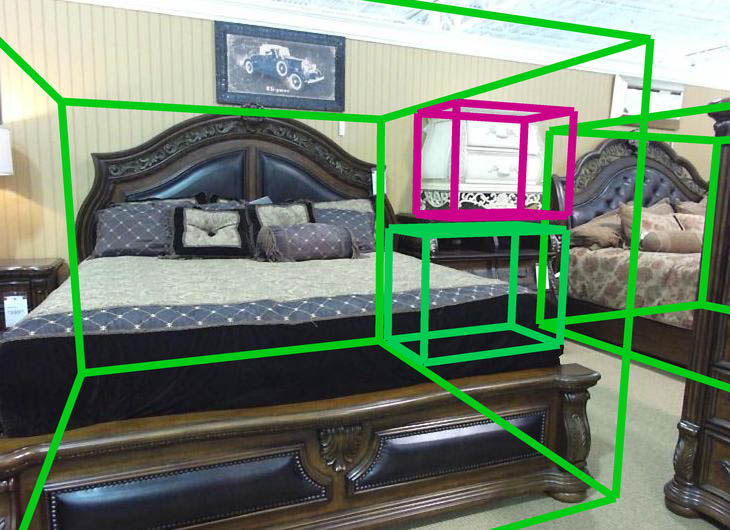} \\
    	\hline
    	\multirow{2}{*}{\rot{Scene Mesh}}
        	&\rot{Total3D}& 
        	    \rgb{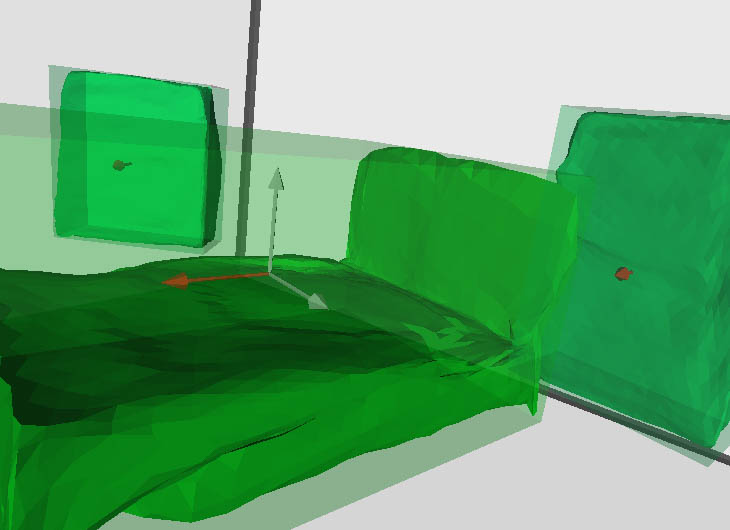} &
        	    \rgb{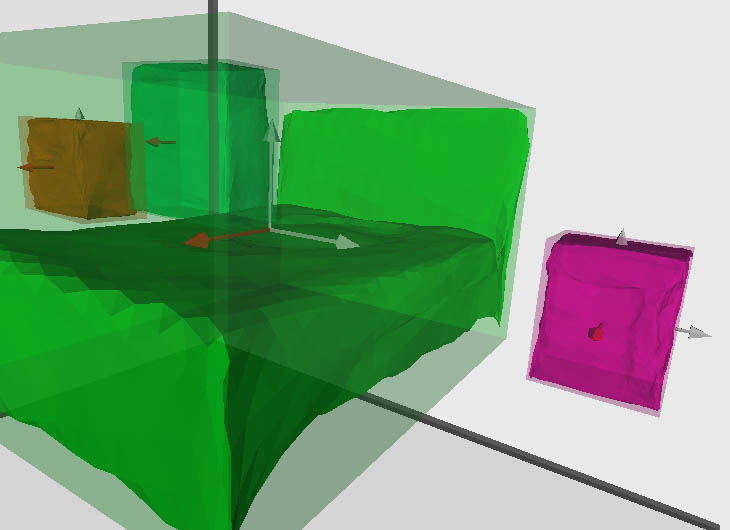} &
        	    \rgb{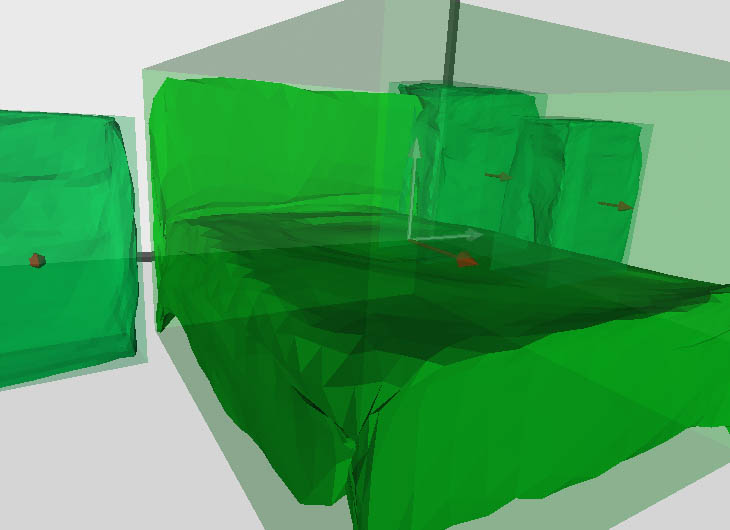} &
        	    \rgb{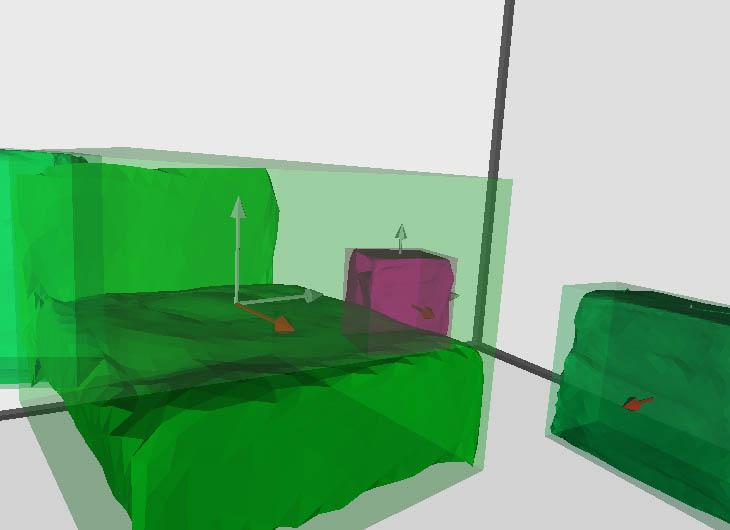} &
        	    \rgb{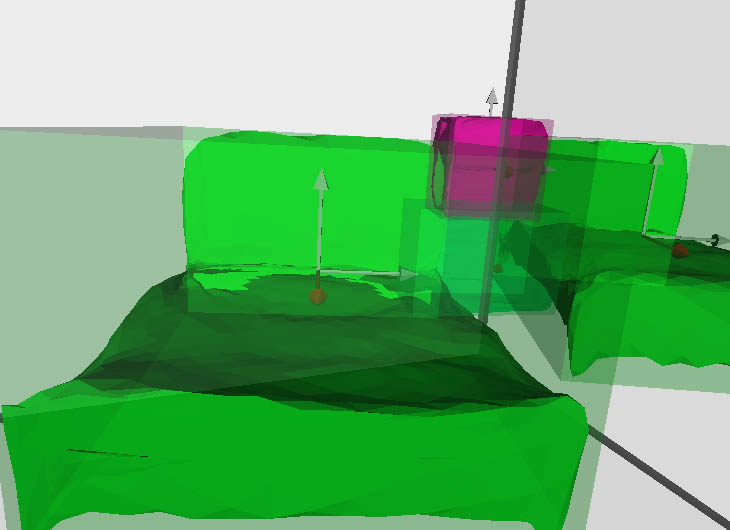} \\
        	&\rot{Ours}& 
        	    \rgb{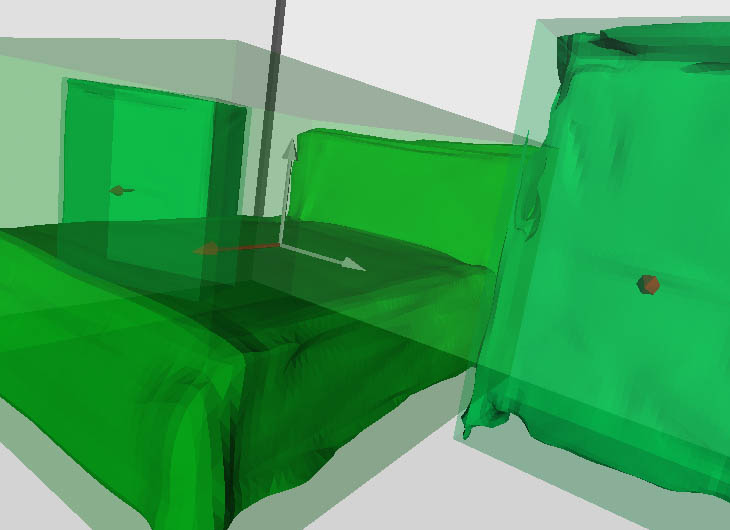} &
        	    \rgb{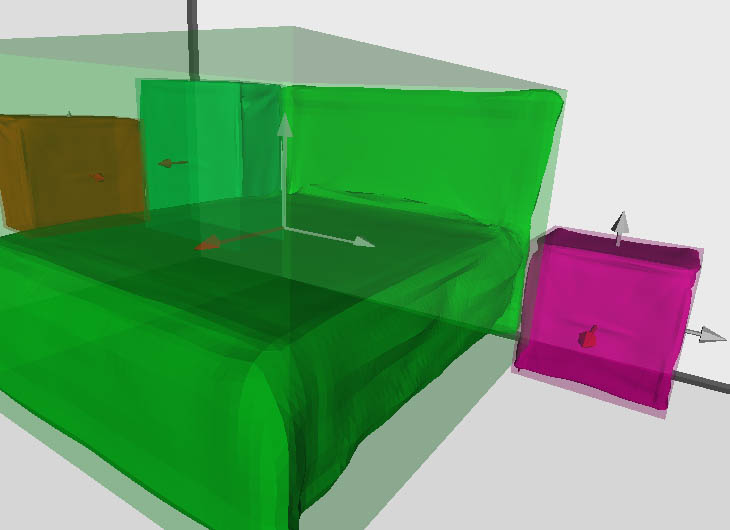} &
        	    \rgb{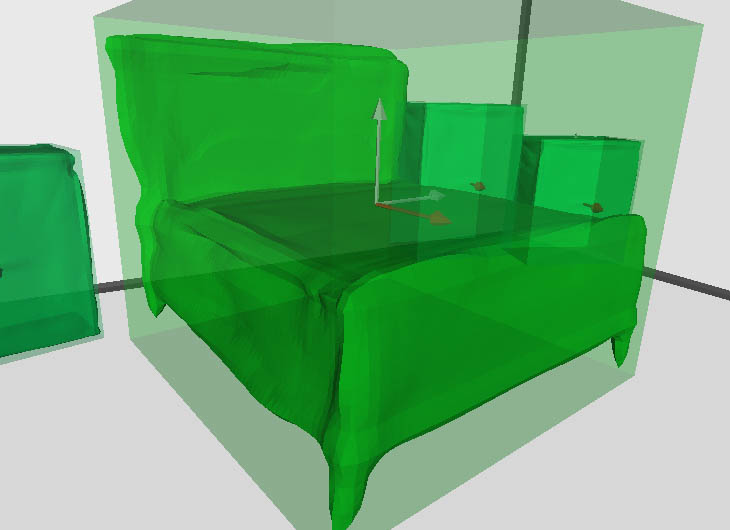} &
        	    \rgb{figure/scnrecon/ours/845_recon} &
        	    \rgb{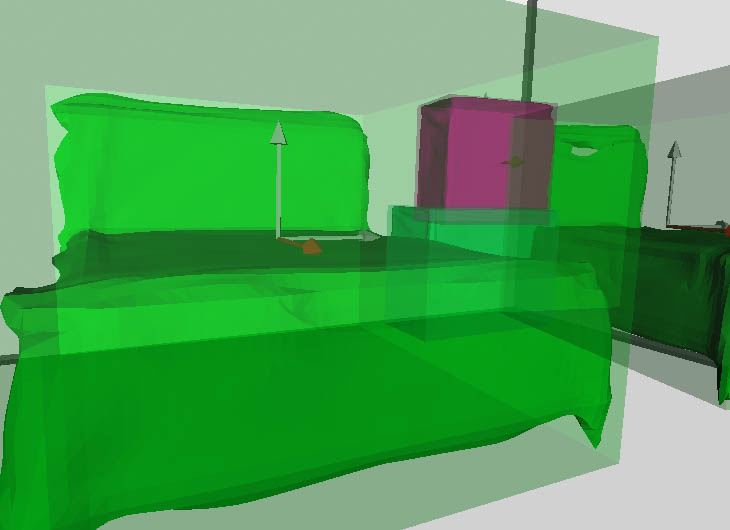} \vspace{2px}\\
        \multicolumn{2}{c}{}& (a) & (b) & (c) & (d) & (e) \\
    \end{tabular}
    % \vspace{-1.0em}
	\caption{Qualitative comparisons on object detection and scene reconstruction.}
	\label{fig:scnreconSupp2}
\end{figure*}

\begin{figure*}[!ht]
	\centering
	\scriptsize
	\newcommand{\rgb}[1]{\raisebox{-0.5\height}{\includegraphics[width=.16\linewidth]{#1}}}
	\newcommand{\oblique}[1]{\raisebox{-0.5\height}{\includegraphics[width=.16\linewidth,clip,trim=48 12 48 12]{#1}}}
	\newcommand{\rot}[1]{\rotatebox[origin=c]{90}{#1}}
	\def\arraystretch{0.5}%
	\begin{tabular}{c|c*{5}{c@{\hspace{1px}}}}
        &\rot{Input}& 
    	    \rgb{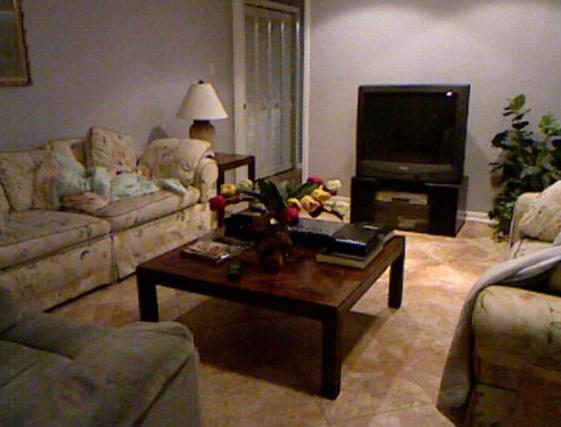} &
    	    \rgb{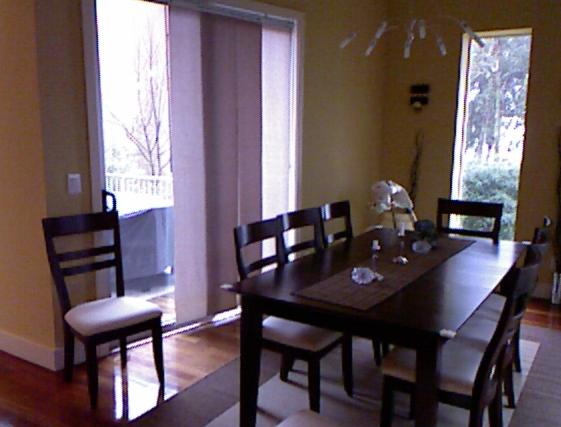} &
    	    \rgb{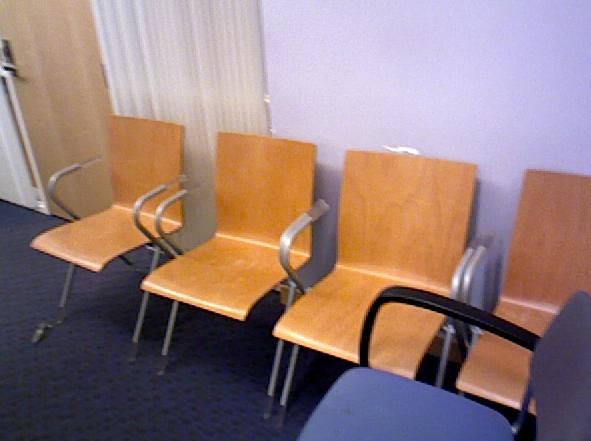} &
    	    \rgb{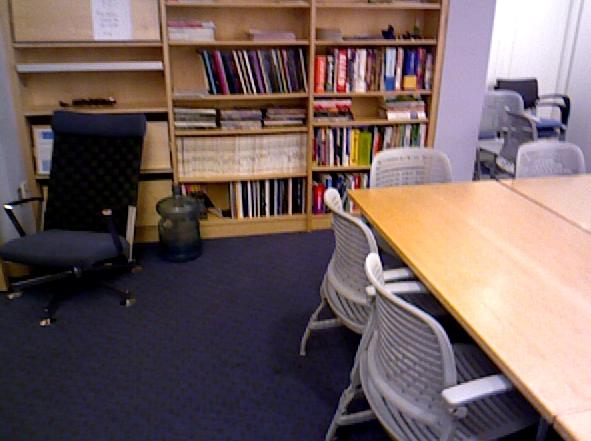} &
    	    \rgb{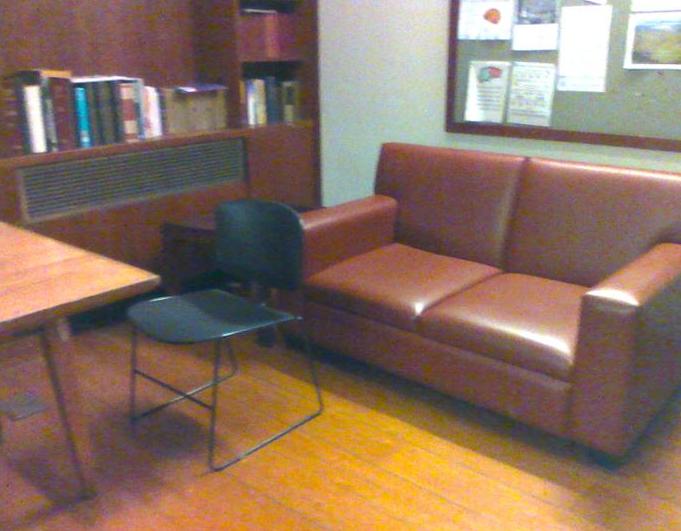} \\
    	\hline
    	    &\rot{Total3D}& 
        	    \oblique{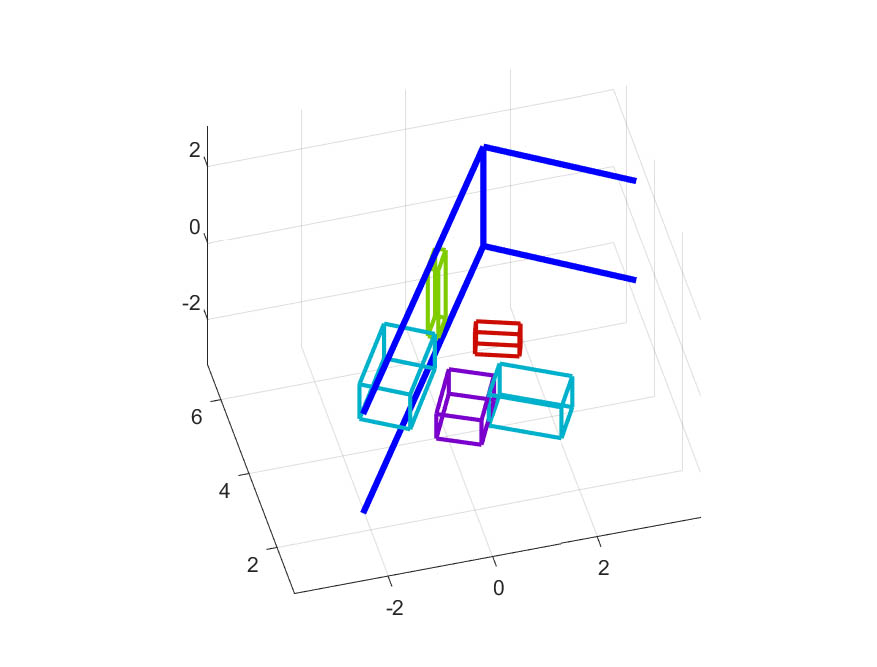} &
        	    \oblique{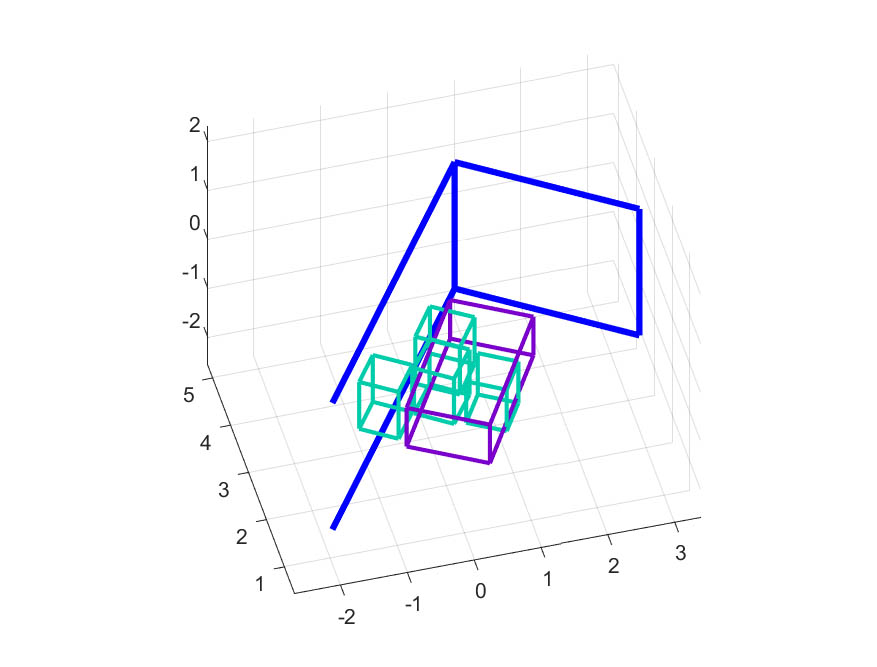} &
        	    \oblique{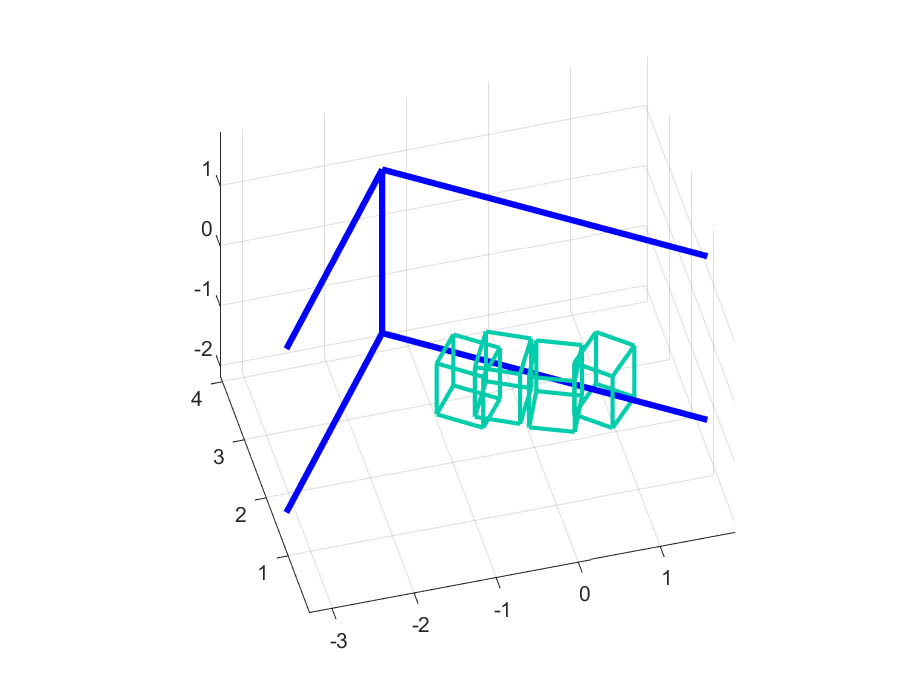} &
        	    \oblique{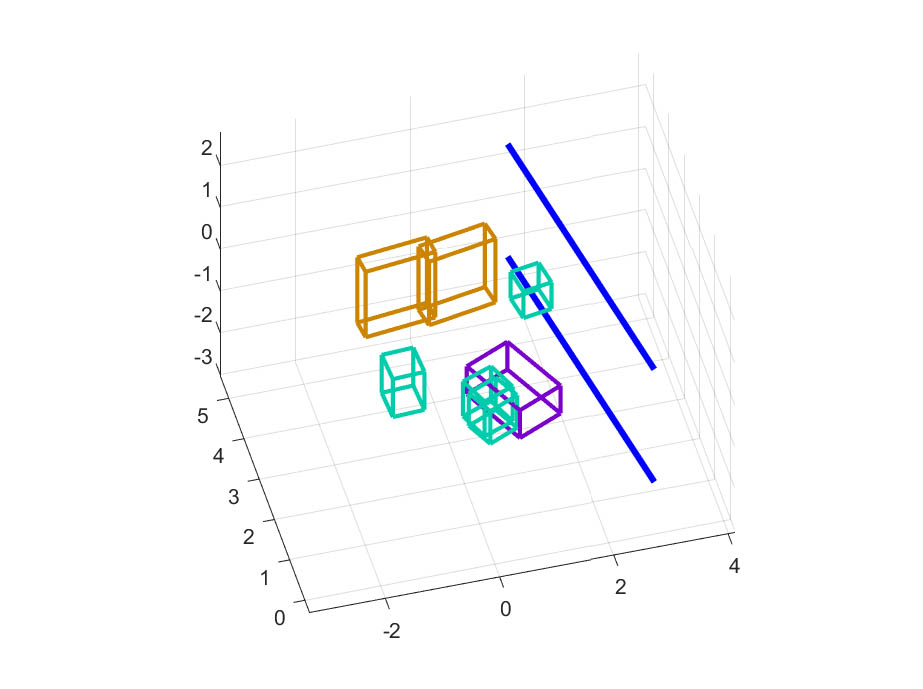} &
        	    \oblique{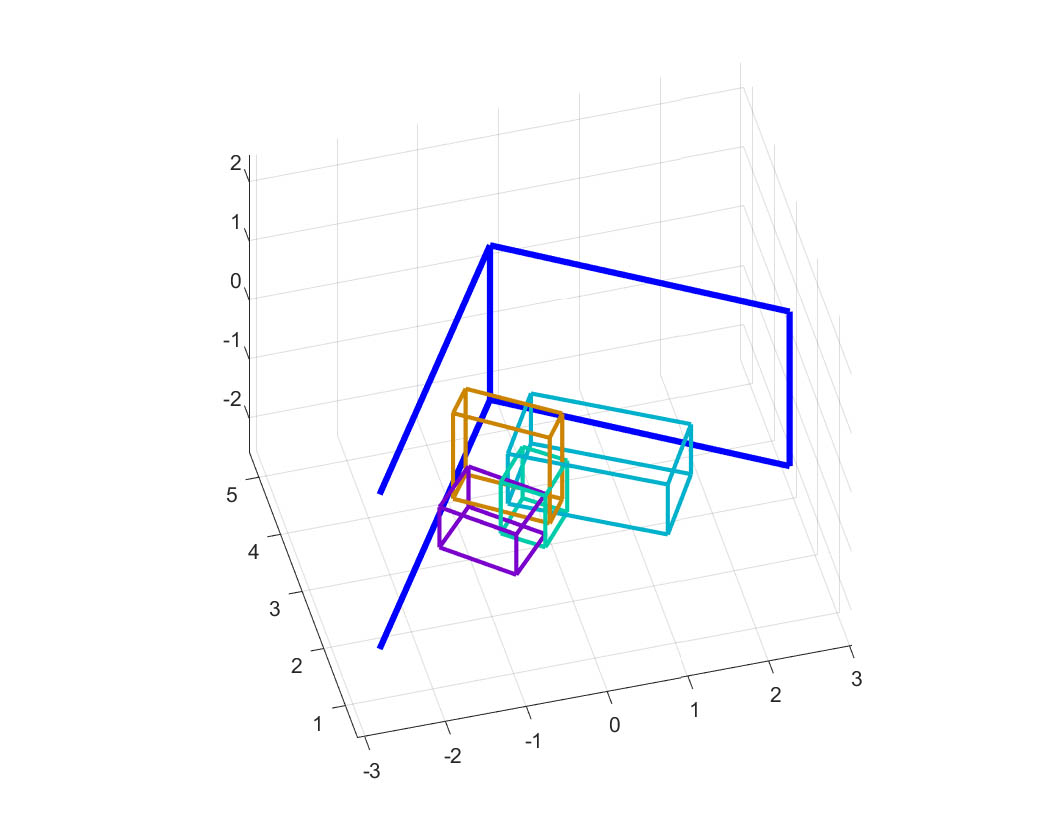} \\
        	\rot{Oblique View}& \rot{Ours}& 
        	    \oblique{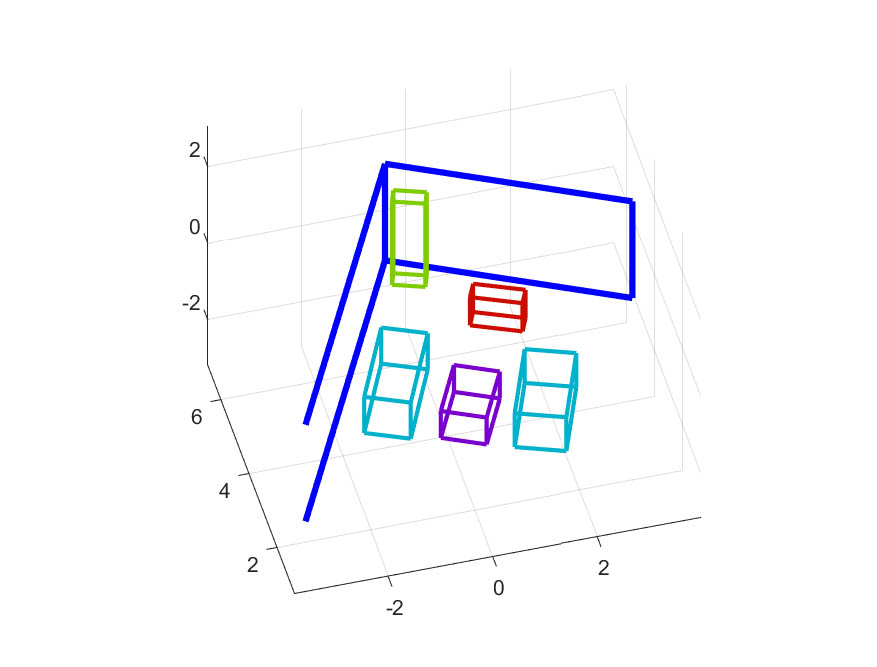} &
        	    \oblique{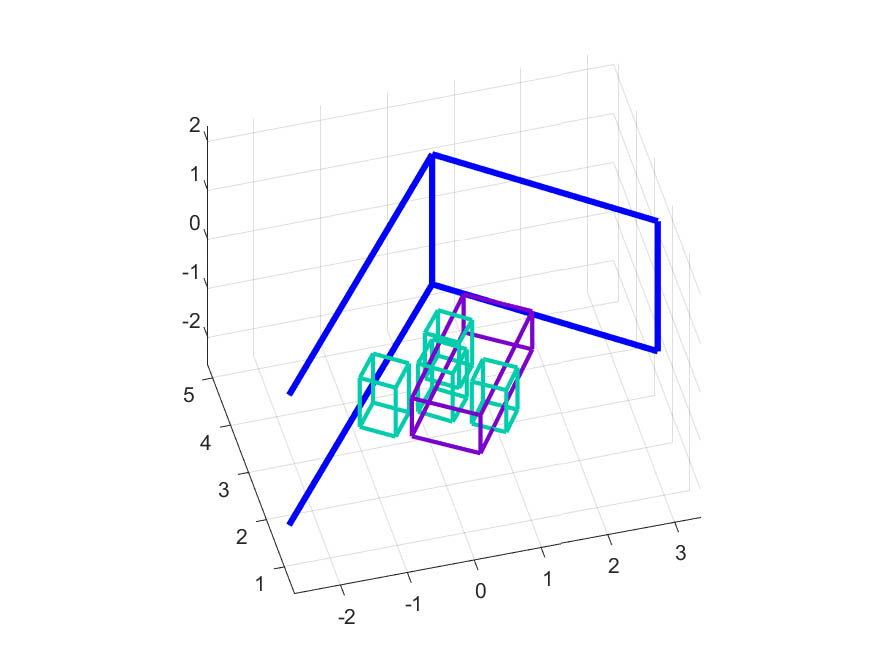} &
        	    \oblique{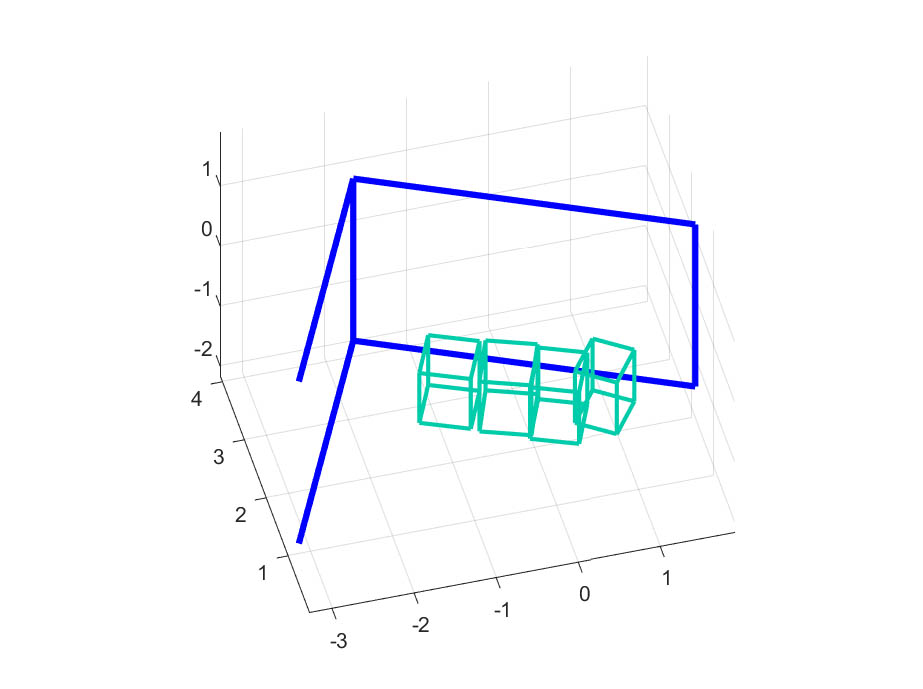} &
        	    \oblique{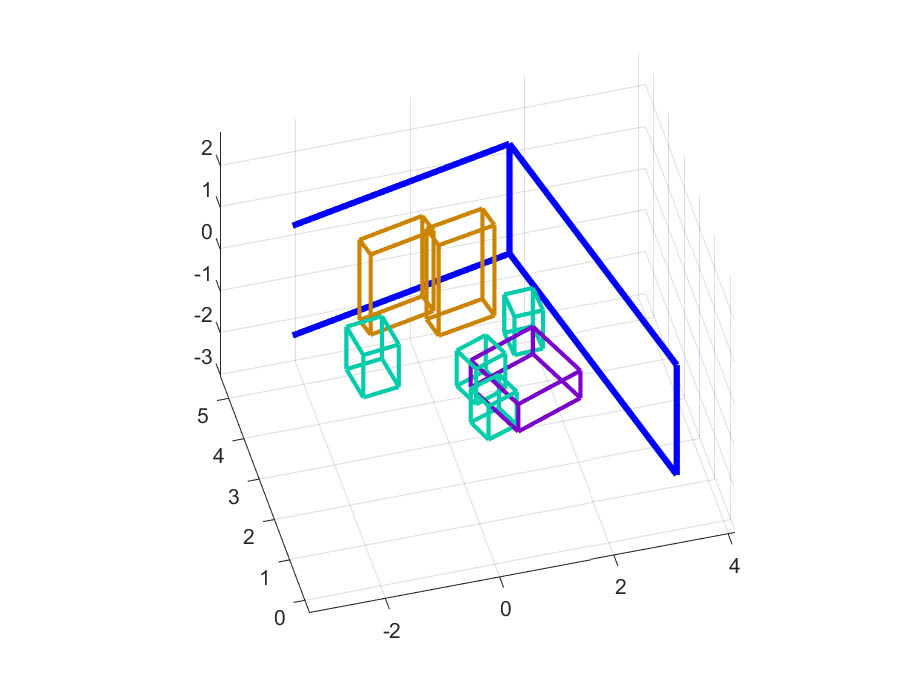} &
        	    \oblique{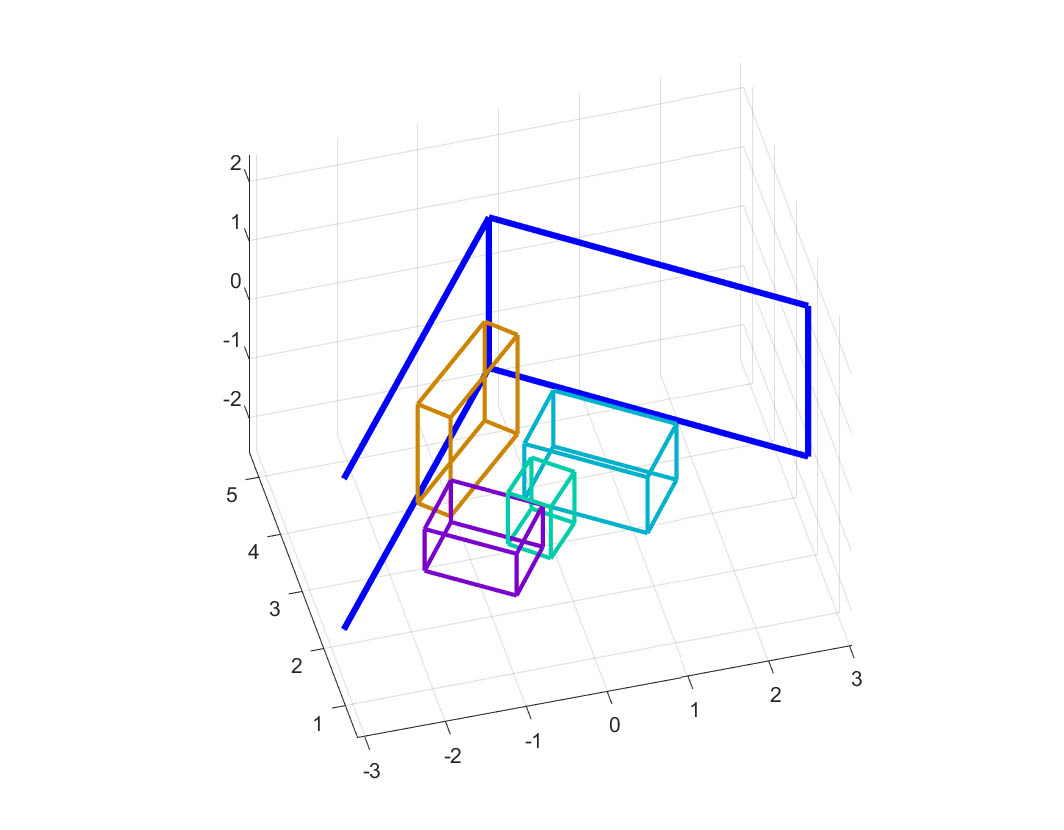} \\
        	&\rot{GT}& 
        	    \oblique{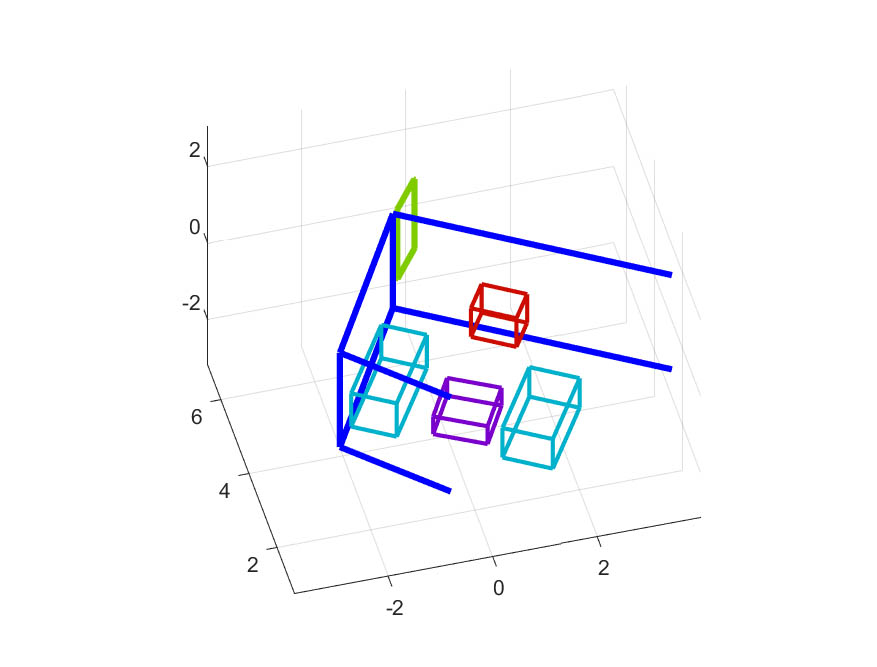} &
        	    \oblique{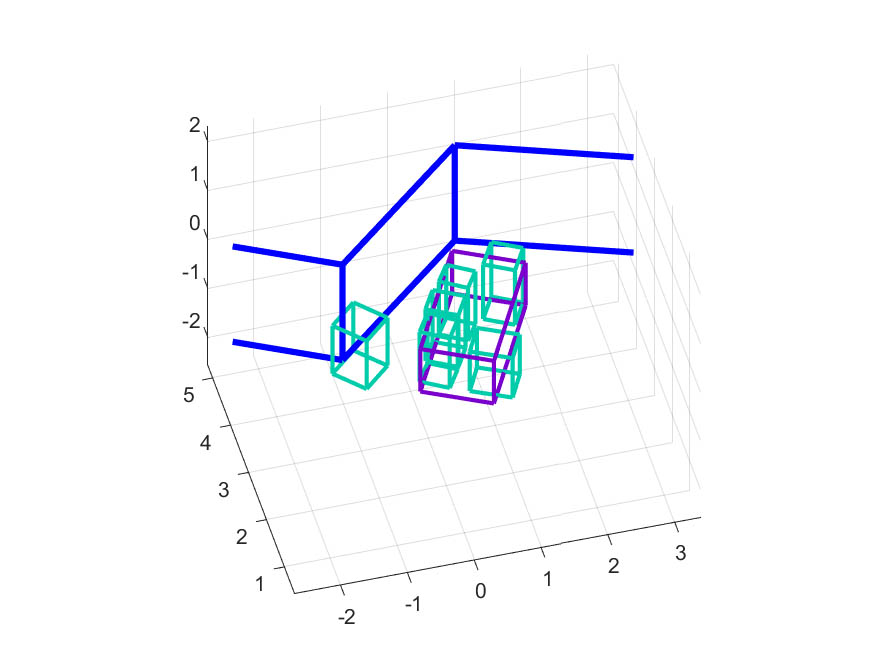} &
        	    \oblique{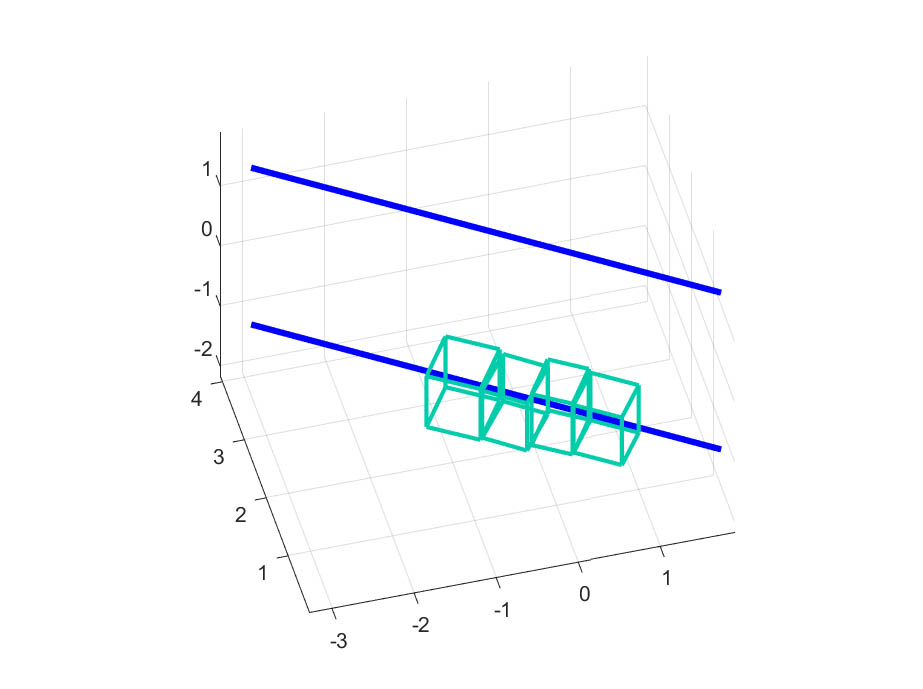} &
        	    \oblique{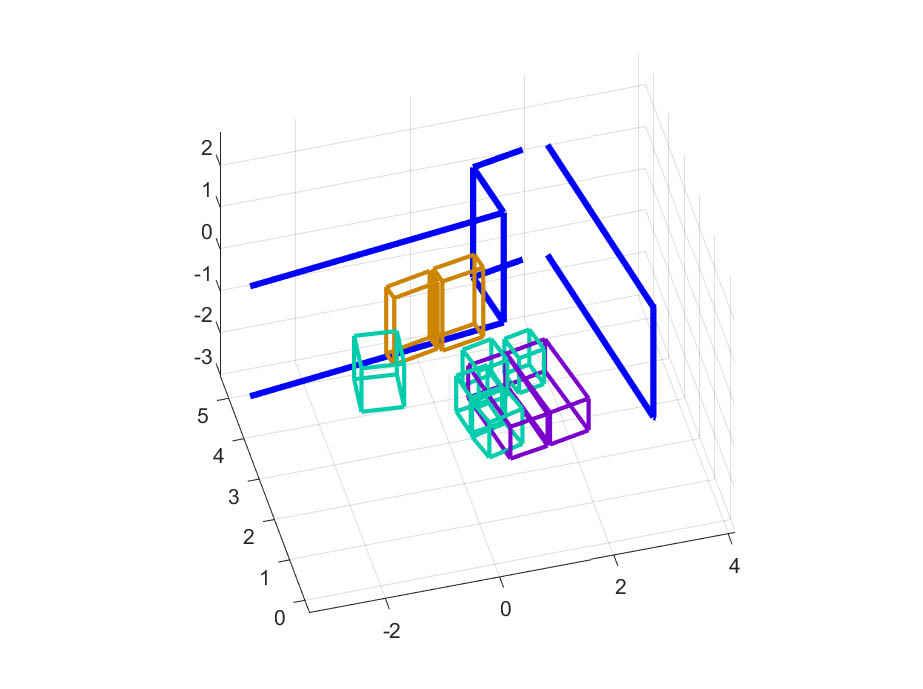} &
        	    \oblique{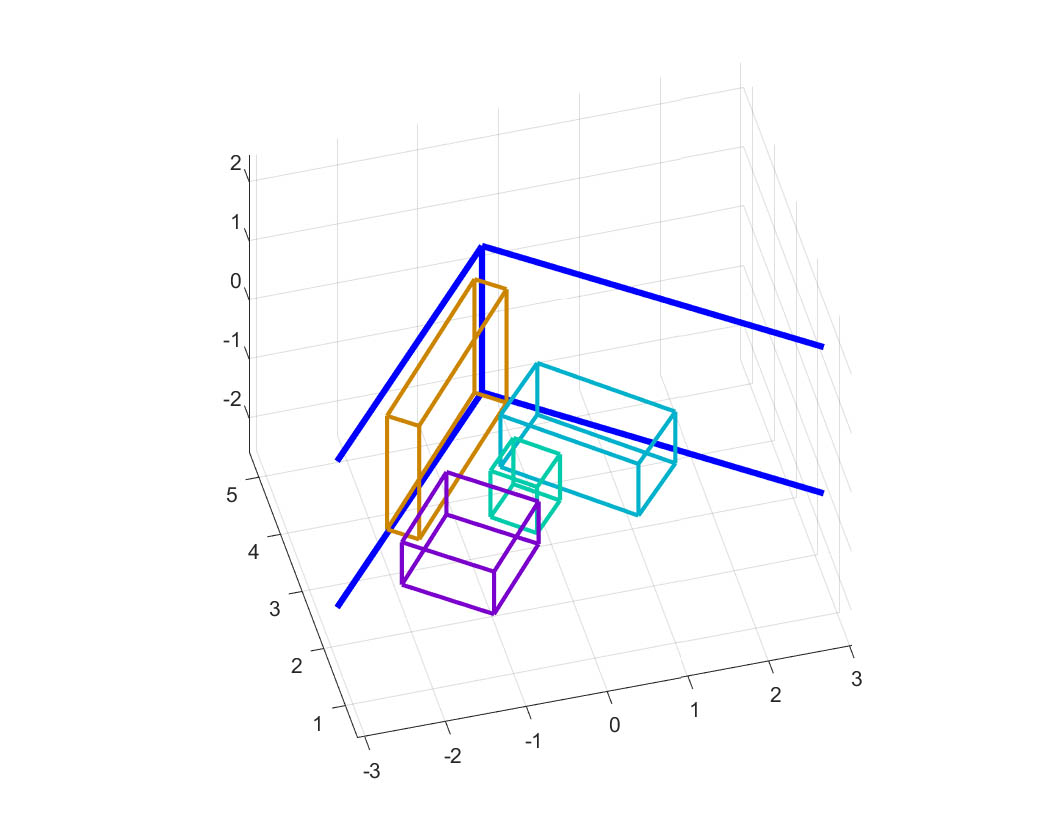} \\
    	\hline
    	    &\rot{Total3D}& 
        	    \rgb{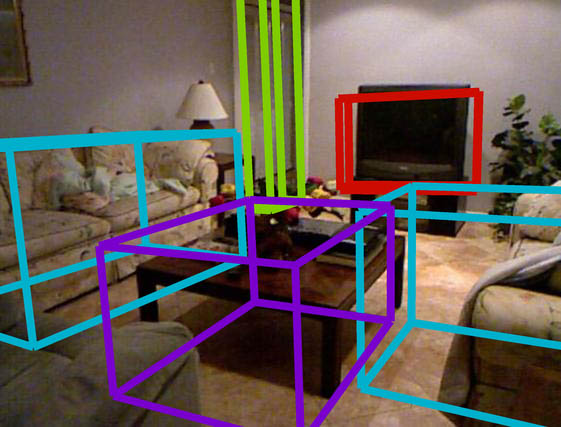} &
        	    \rgb{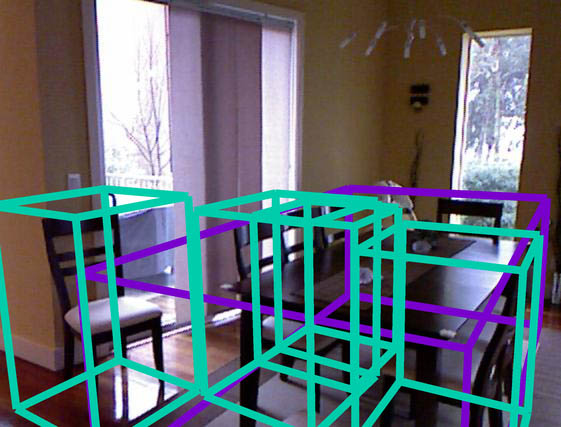} &
        	    \rgb{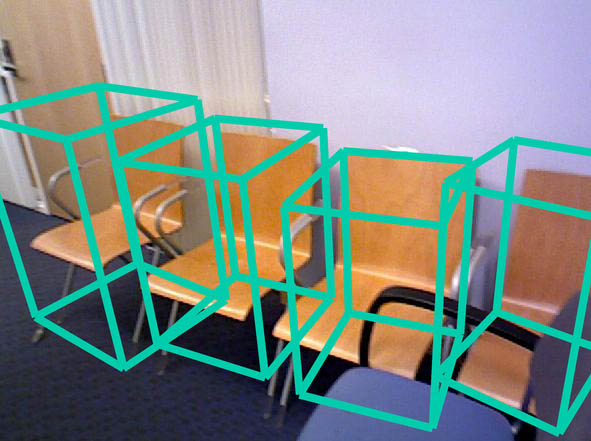} &
        	    \rgb{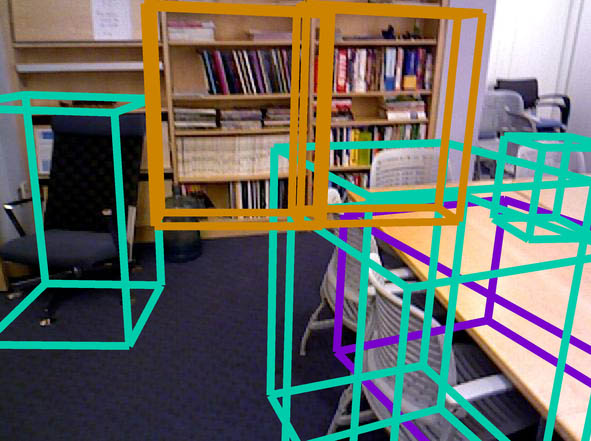} &
        	    \rgb{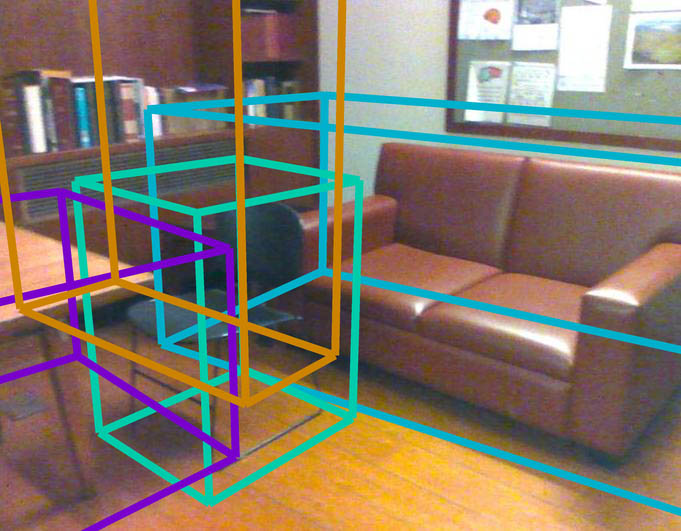} \\
            \rot{Camera View}&\rot{Ours}& 
        	    \rgb{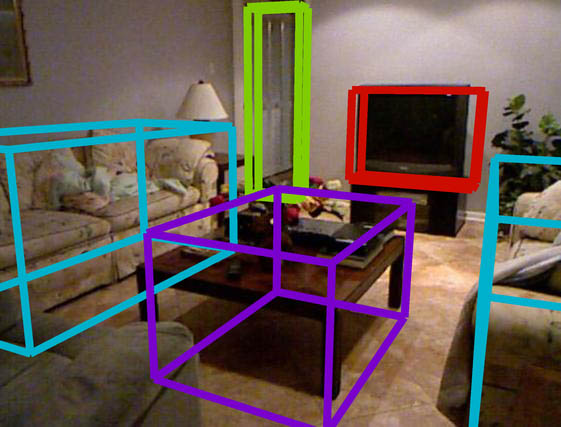} &
        	    \rgb{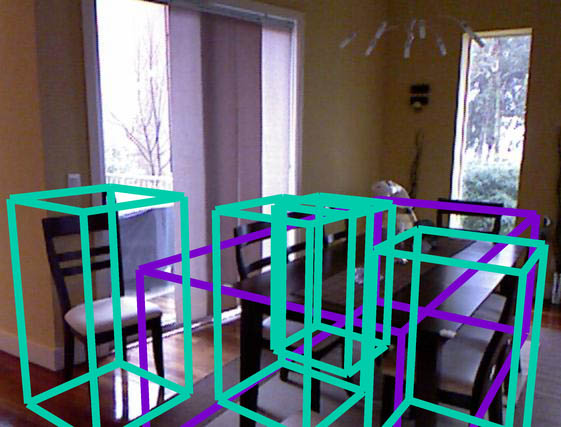} &
        	    \rgb{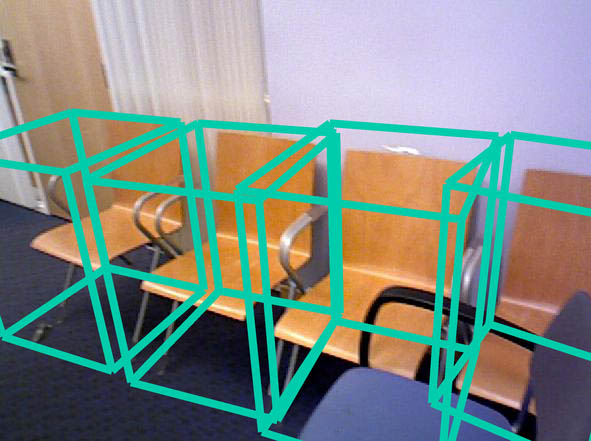} &
        	    \rgb{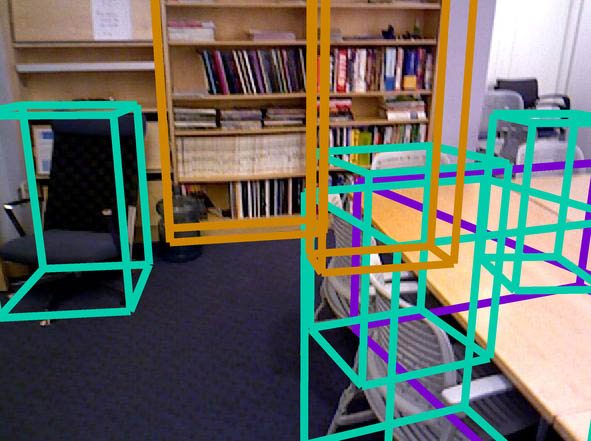} &
        	    \rgb{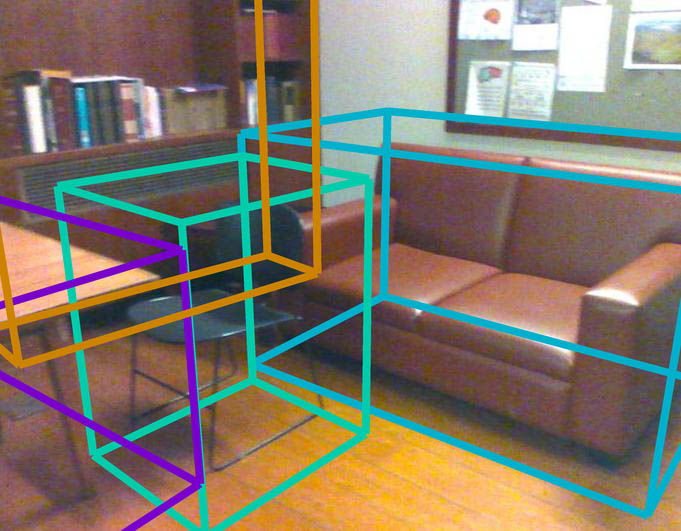} \\
        	&\rot{GT}& 
        	    \rgb{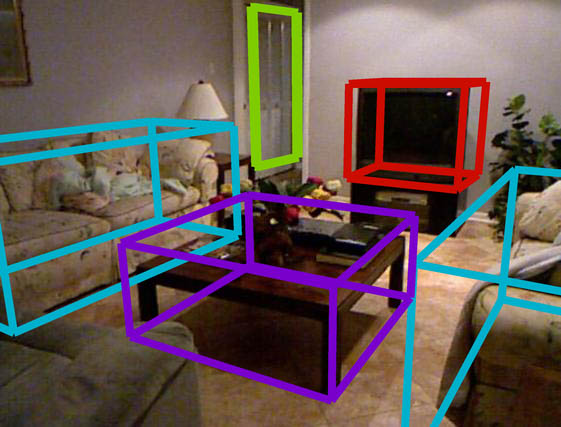} &
        	    \rgb{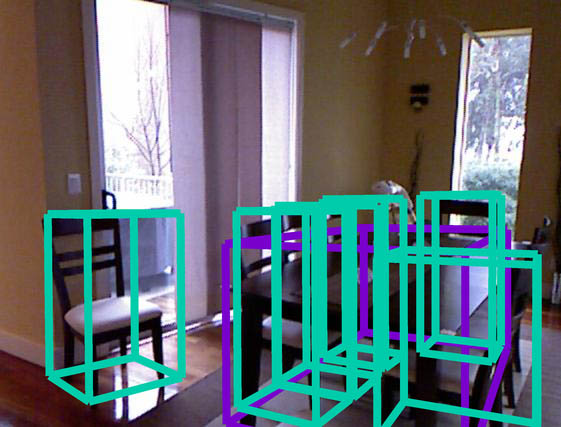} &
        	    \rgb{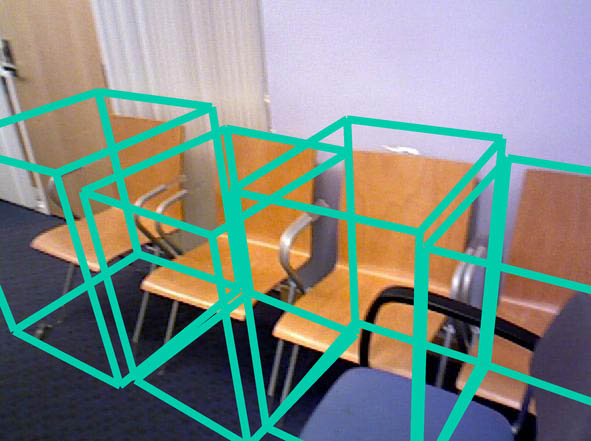} &
        	    \rgb{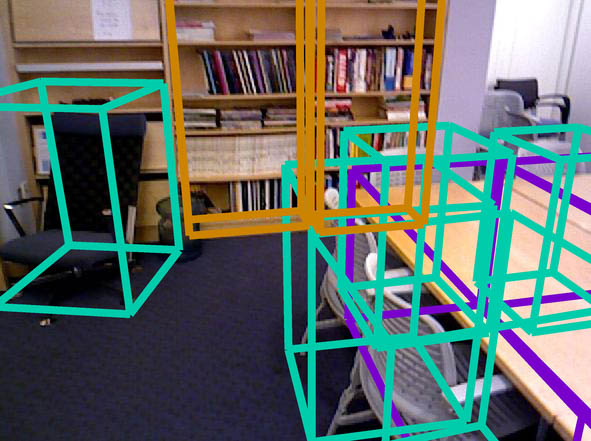} &
        	    \rgb{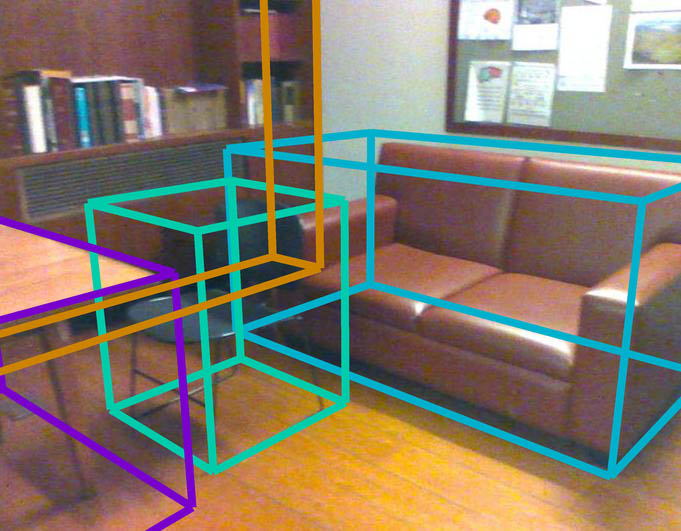} \\
    	\hline
    	\multirow{2}{*}{\rot{Scene Mesh}}
        	&\rot{Total3D}& 
        	    \rgb{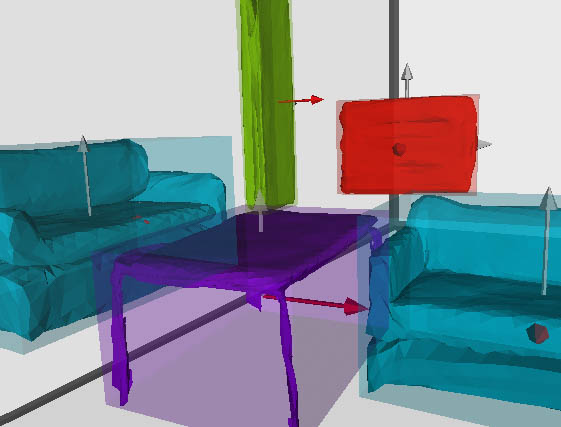} &
        	    \rgb{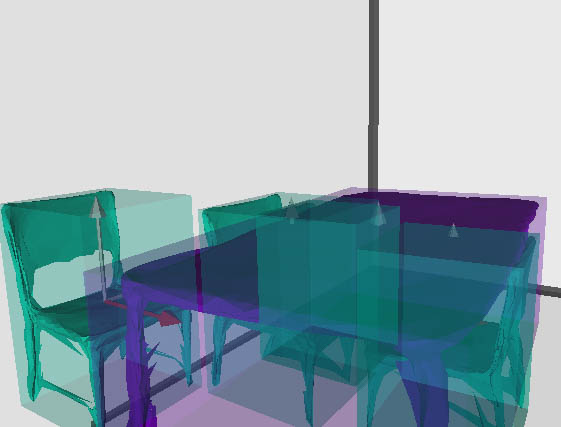} &
        	    \rgb{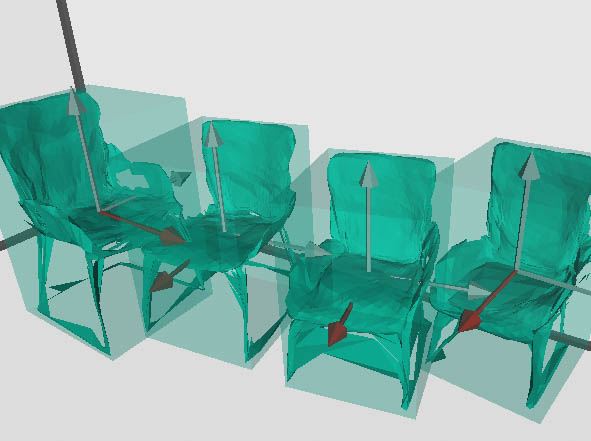} &
        	    \rgb{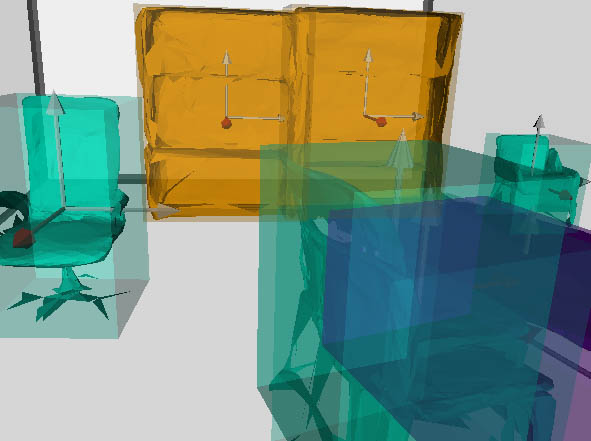} &
        	    \rgb{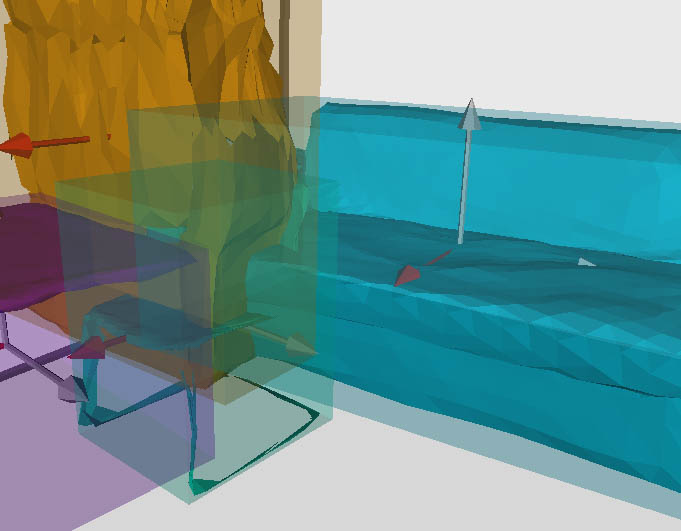} \\
        	&\rot{Ours}& 
        	    \rgb{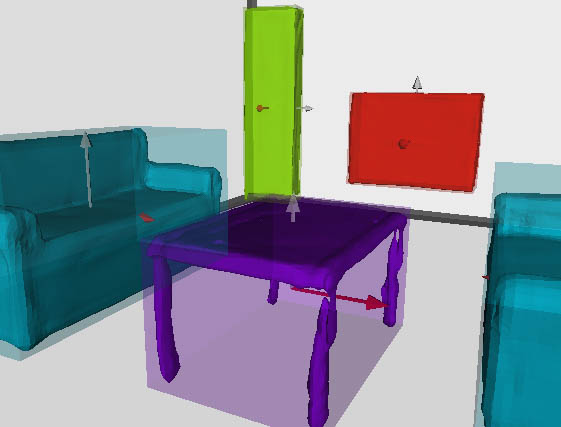} &
        	    \rgb{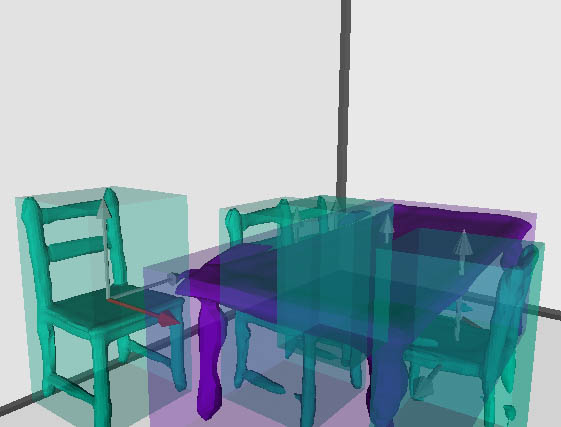} &
        	    \rgb{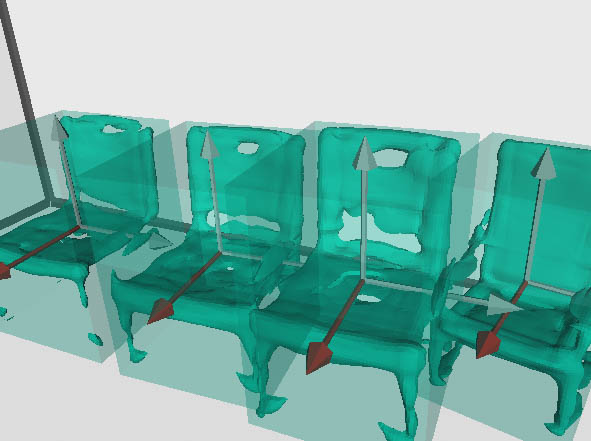} &
        	    \rgb{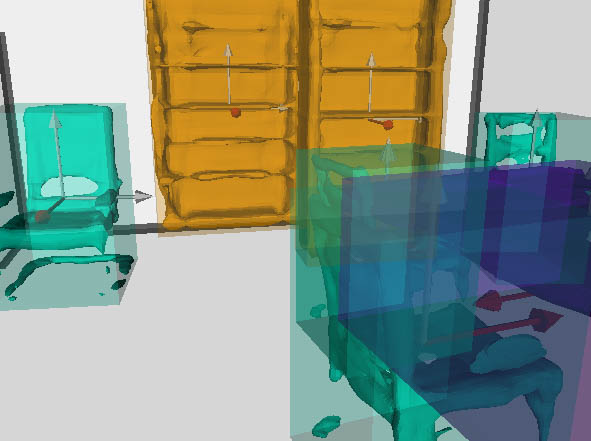} &
        	    \rgb{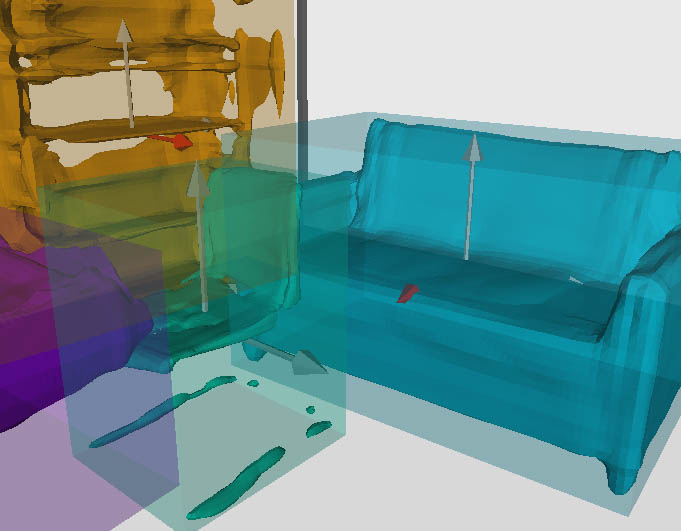} \vspace{2px}\\
        \multicolumn{2}{c}{}& (a) & (b) & (c) & (d) & (e) \\
    \end{tabular}
    % \vspace{-1.0em}
	\caption{Qualitative comparisons on object detection and scene reconstruction.}
	\label{fig:scnreconSupp3}
\end{figure*}

In main paper Section 4.2, we show qualitative results of the 3D object detection and scene reconstruction.
Here, we show more results in \autoref{fig:scnreconSupp1}, \autoref{fig:scnreconSupp2}, and \autoref{fig:scnreconSupp3}.
We can observe that compared to the state-of-the-art method \cite{nie2020total3dunderstanding}, our method produces significantly more accurate object pose estimation with fewer flying objects (\autoref{fig:scnreconSupp1}e, \autoref{fig:scnreconSupp2}a), fewer objects intersected with each other (\autoref{fig:scnreconSupp1}a, \autoref{fig:scnreconSupp2}d, \autoref{fig:scnreconSupp3}e), and more accurate object orientation estimation (\autoref{fig:scnreconSupp1}c, \autoref{fig:scnreconSupp2}e, \autoref{fig:scnreconSupp3}c). We also observe fewer objects intersected with the layout box (\autoref{fig:scnreconSupp1}d, \autoref{fig:scnreconSupp2}e, \autoref{fig:scnreconSupp3}a). 
%In general, our method produces accurate scene layout and object arrangement.

% More testing results of 3D detection and scene reconstruction are shown in \autoref{fig:scnreconSupp1}, \autoref{fig:scnreconSupp2}, and \autoref{fig:scnreconSupp3}

\section{Qualitative Comparison of Ablation Study}

\zc{
In main paper Section 4.3, we quantitatively compare the improvement of our proposed $\mathcal{L}_{phy}$.
While exhibiting a small gap from the metric, we show in qualitative results (\autoref{fig:lphy_qualitative}) that the visual difference is relatively large.
Objects are more likely to intersect with each other when trained without $\mathcal{L}_{phy}$, which disobeys physical context severely.
On the contrary, training with $\mathcal{L}_{phy}$ effectively prevents these errors in the results.
}

\zc{
We also quantitatively compare the supporting relation, in main paper Section 4.3.
Here in \autoref{fig:sup_qualitative}, we qualitatively compare the understanding of supporting relation in the front view of object 3D detection.
}

\section{Failure Cases}
\begin{figure*}[!ht]
	\centering
	\scriptsize
	\newcommand{\rgb}[1]{\raisebox{-0.5\height}{\includegraphics[width=.14\linewidth]{#1}}}
	\newcommand{\oblique}[1]{\raisebox{-0.5\height}{\includegraphics[width=.14\linewidth,clip,trim=48 12 48 12]{#1}}}
	\newcommand{\rot}[1]{\rotatebox[origin=c]{90}{#1}}
	\begin{tabular}{c|c*{6}{c@{\hspace{1px}}}}
        &\rot{Input}& 
    	    \rgb{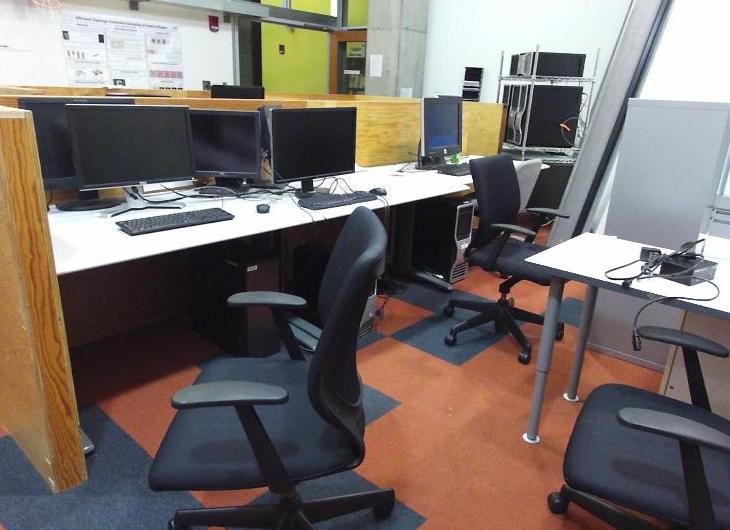} &
    	    \rgb{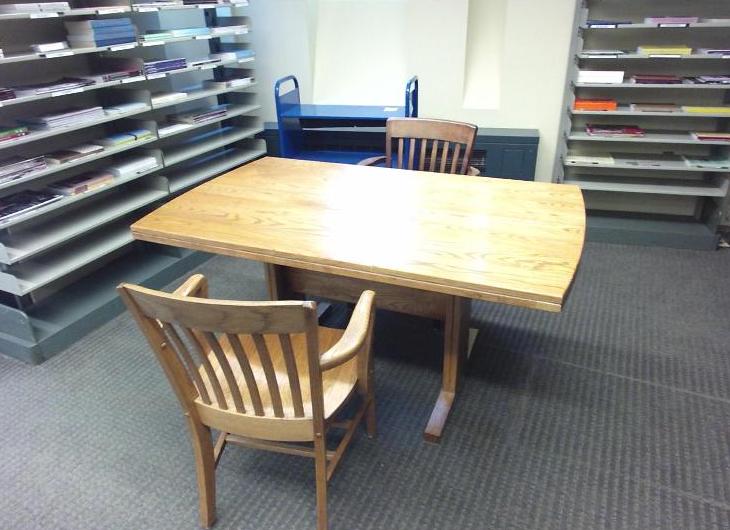} &
    	    \rgb{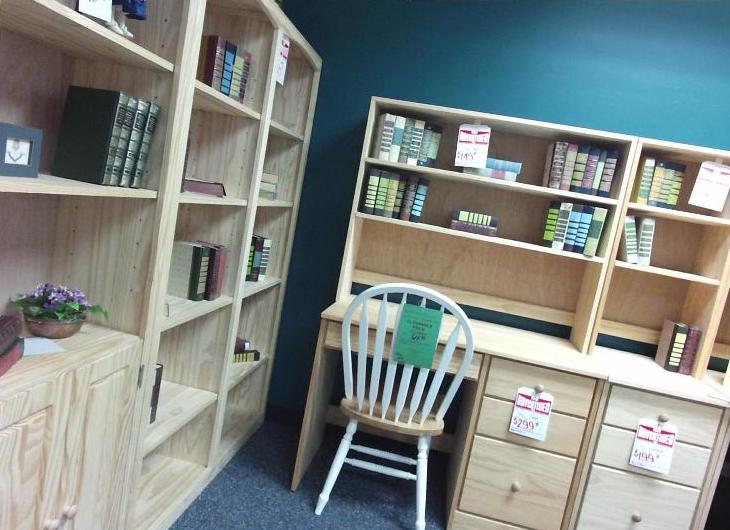} &
    	    \rgb{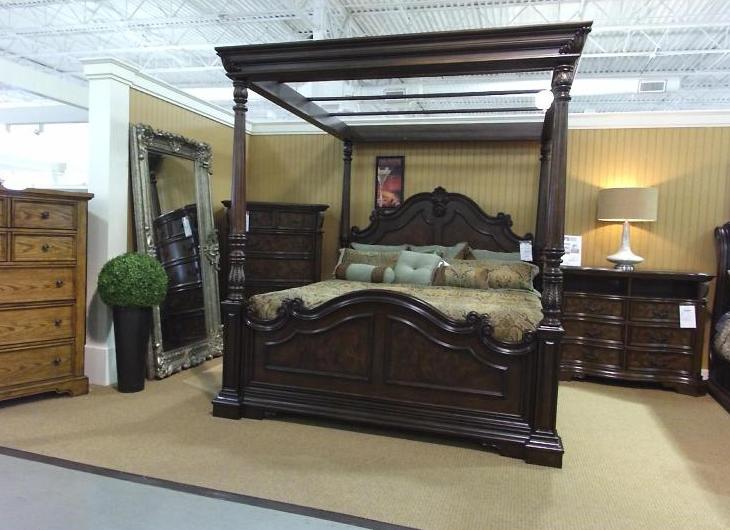} &
    	    \rgb{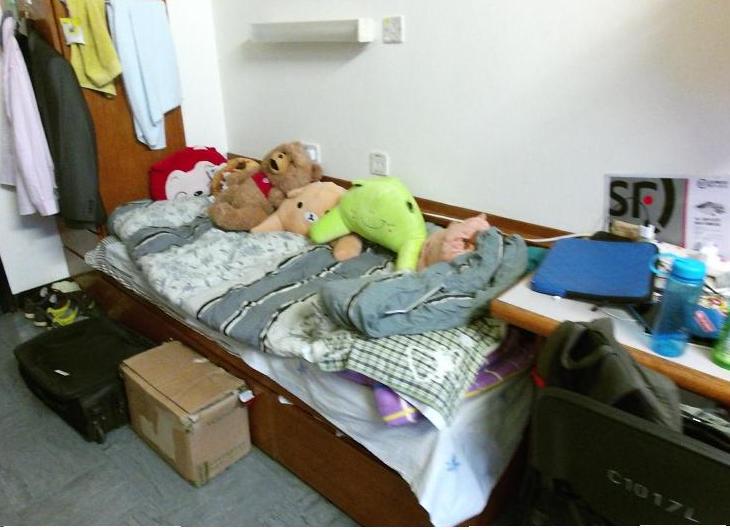} &
    	    \rgb{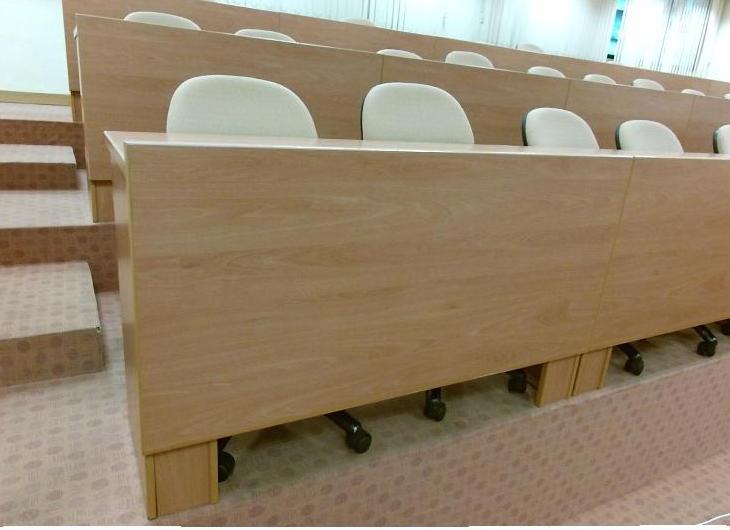} \\
    	\hline
        	\multirow{2}{*}{\rot{Oblique View}}& 
        	\rot{Ours}& 
        	    \oblique{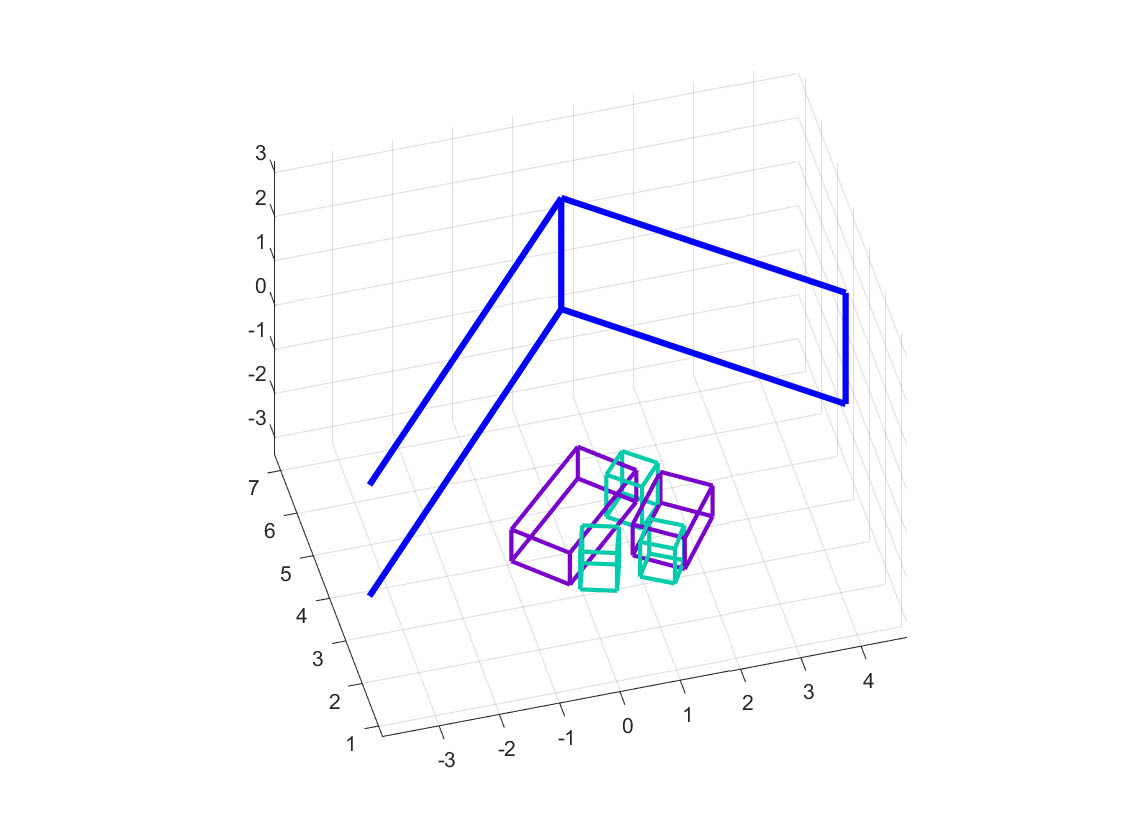} &
        	    \oblique{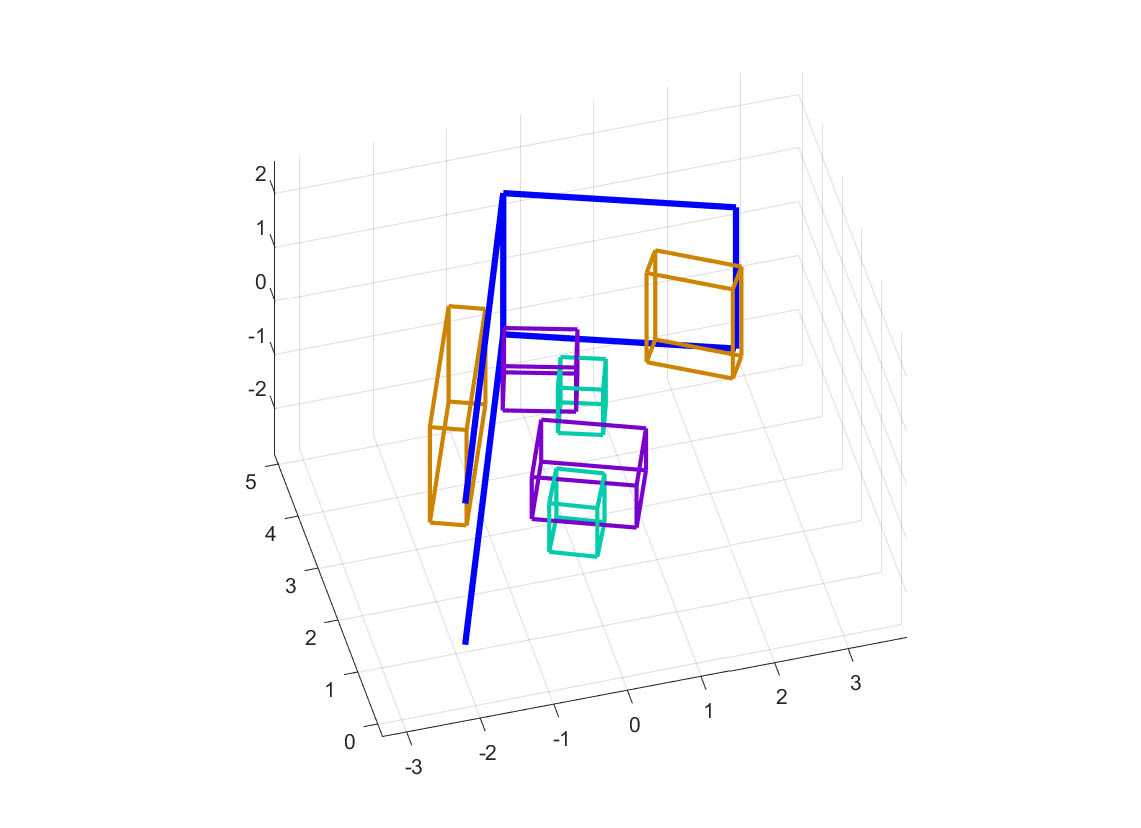} &
        	    \oblique{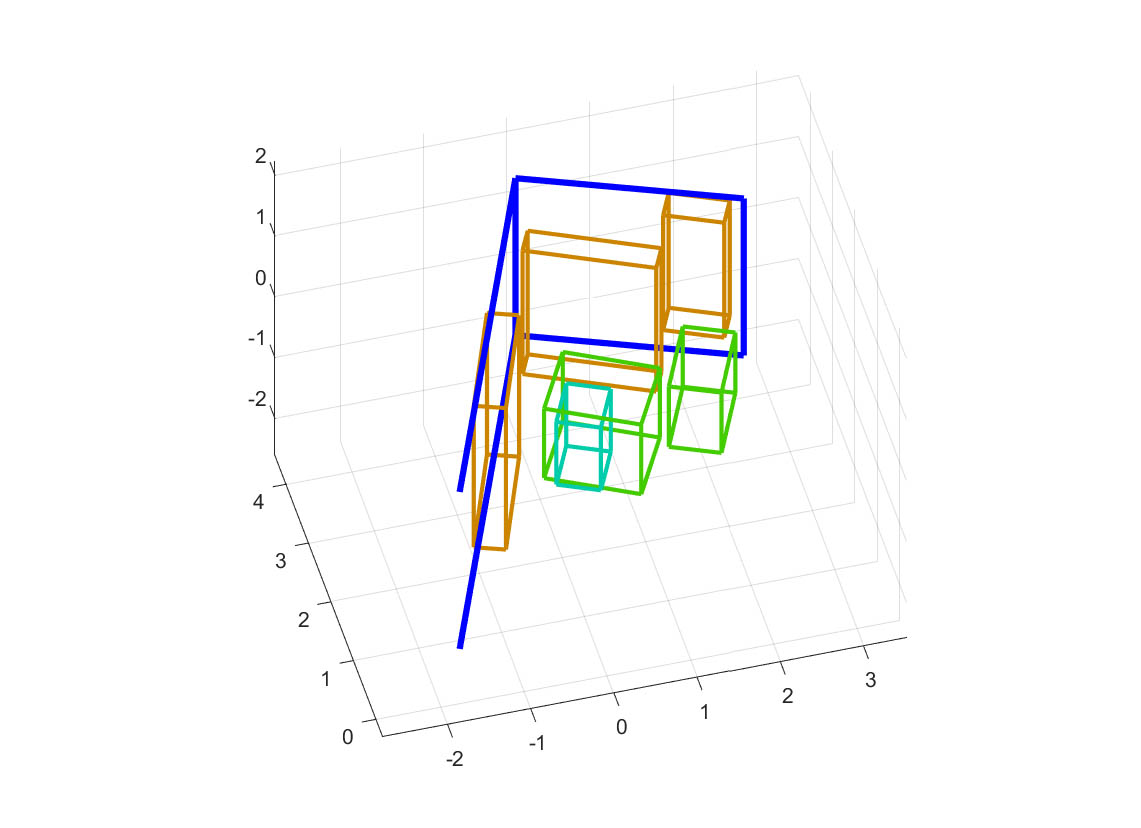} &
        	    \oblique{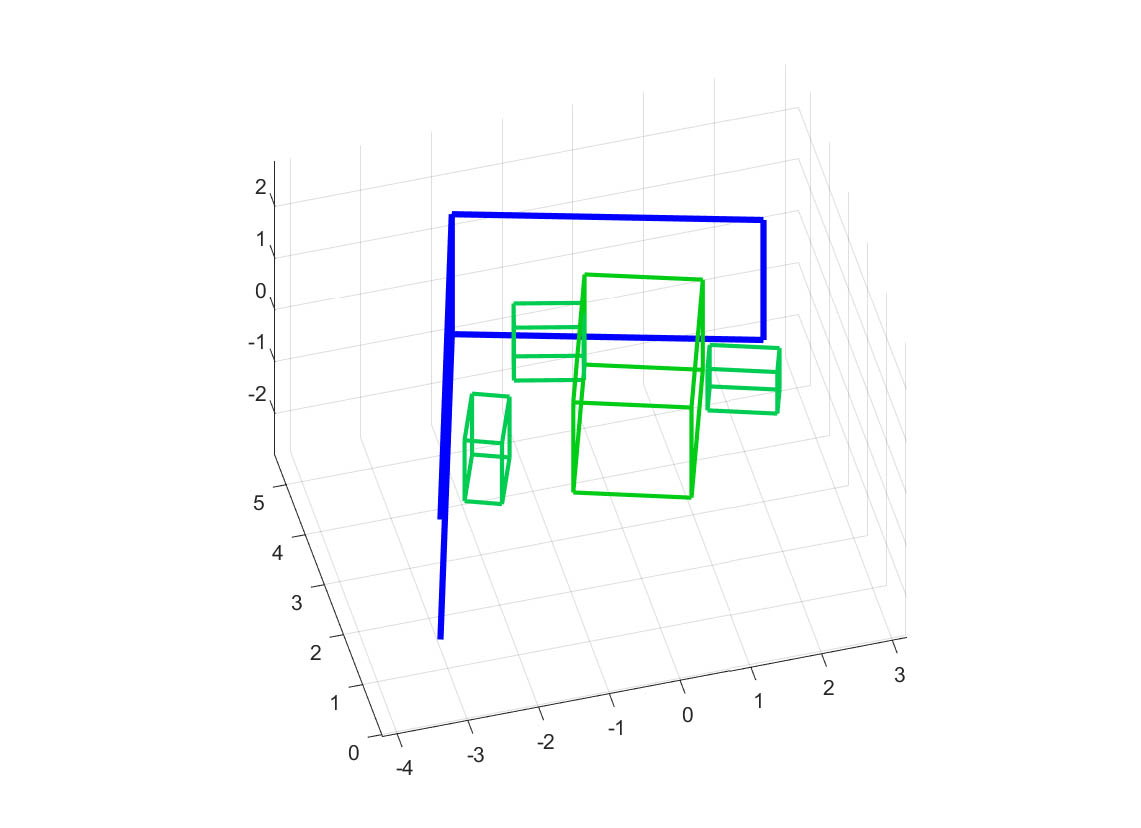} &
        	    \oblique{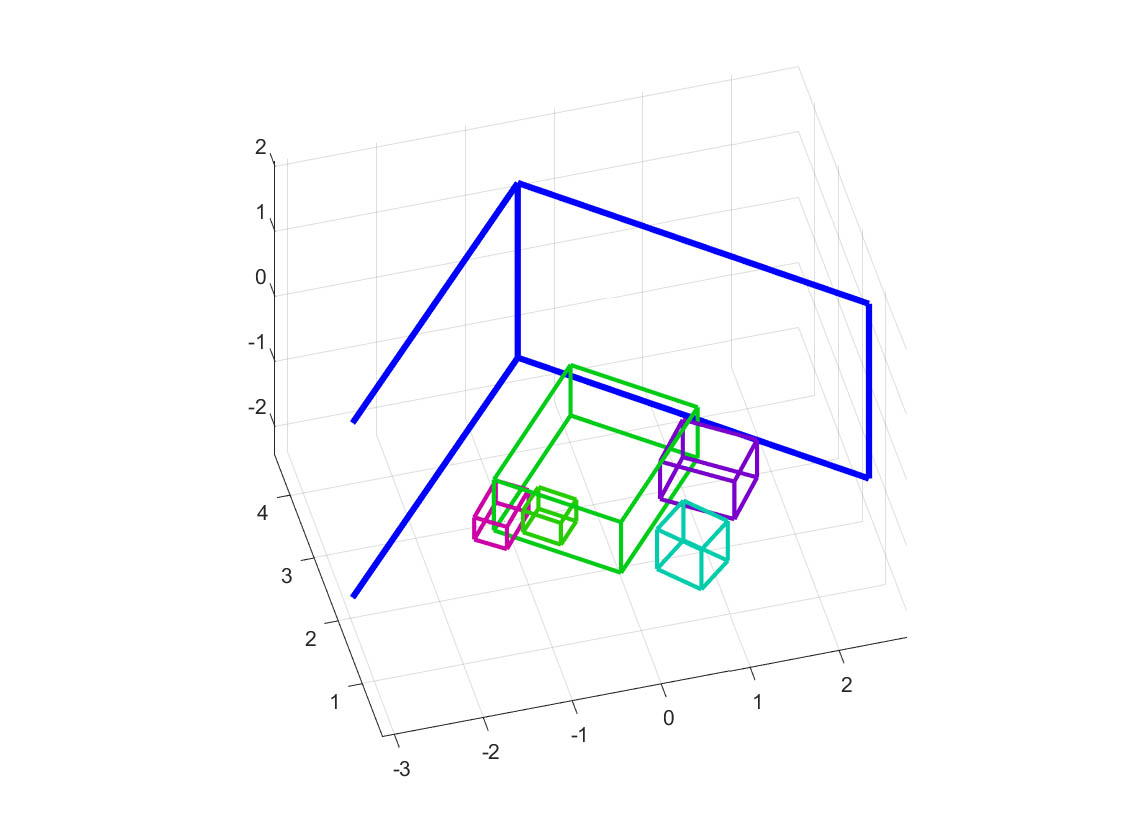} &
        	    \oblique{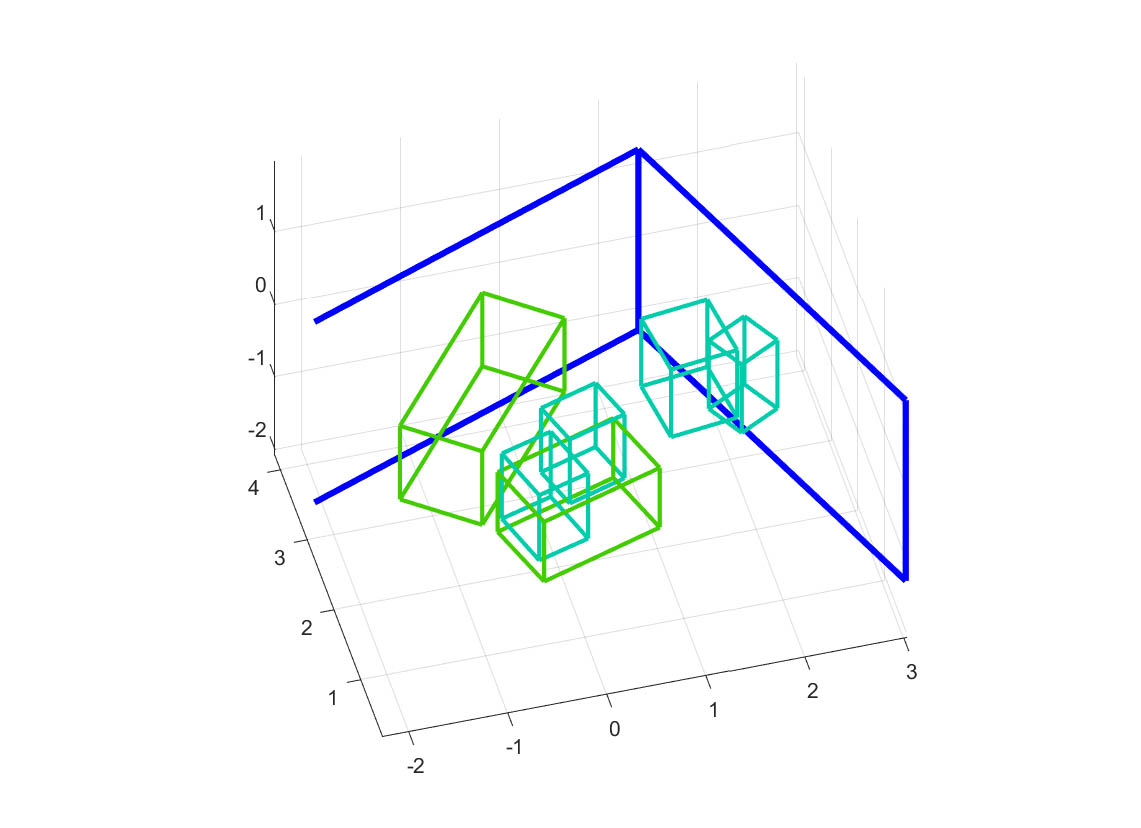} \\
        	&\rot{GT}& 
        	    \oblique{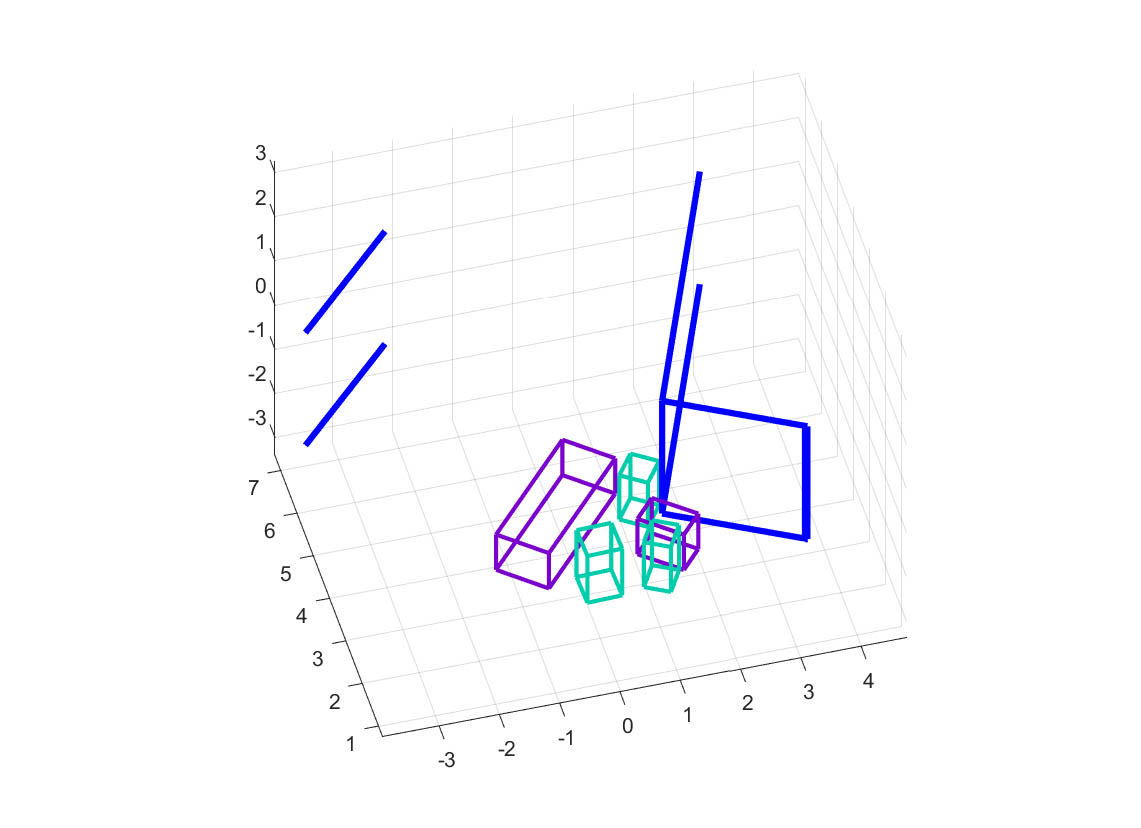} &
        	    \oblique{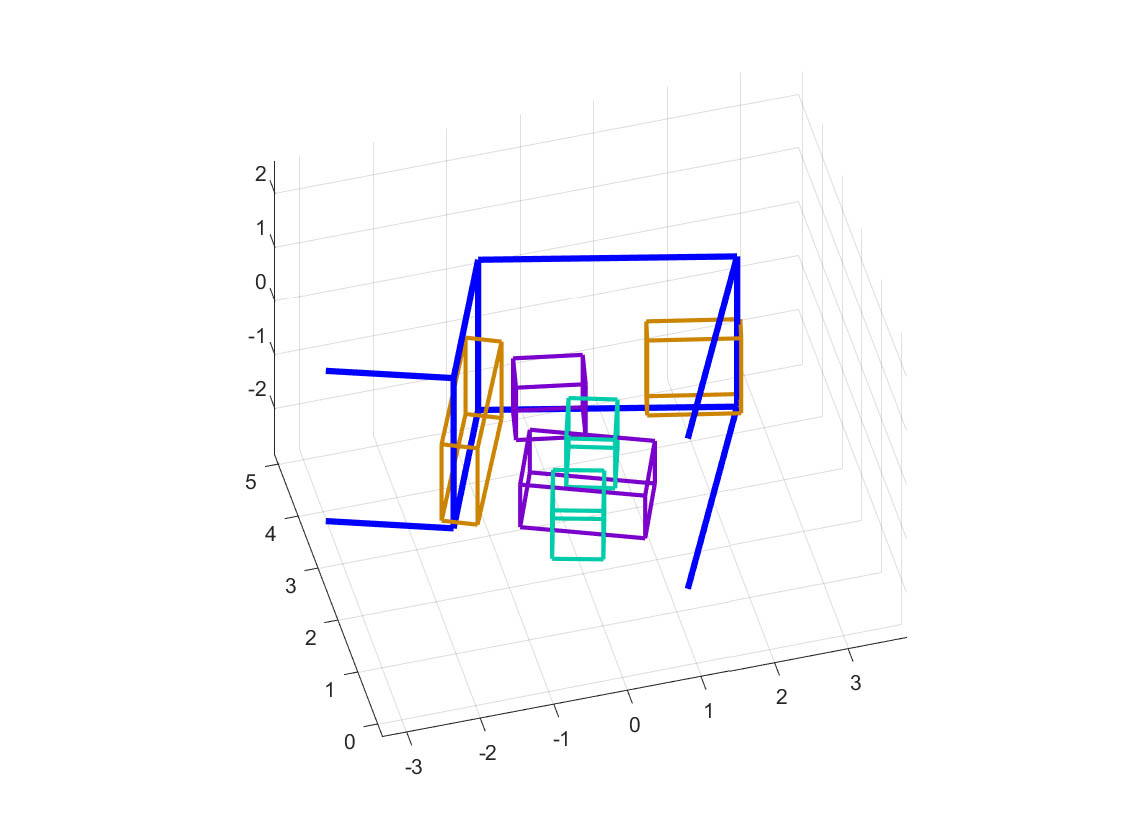} &
        	    \oblique{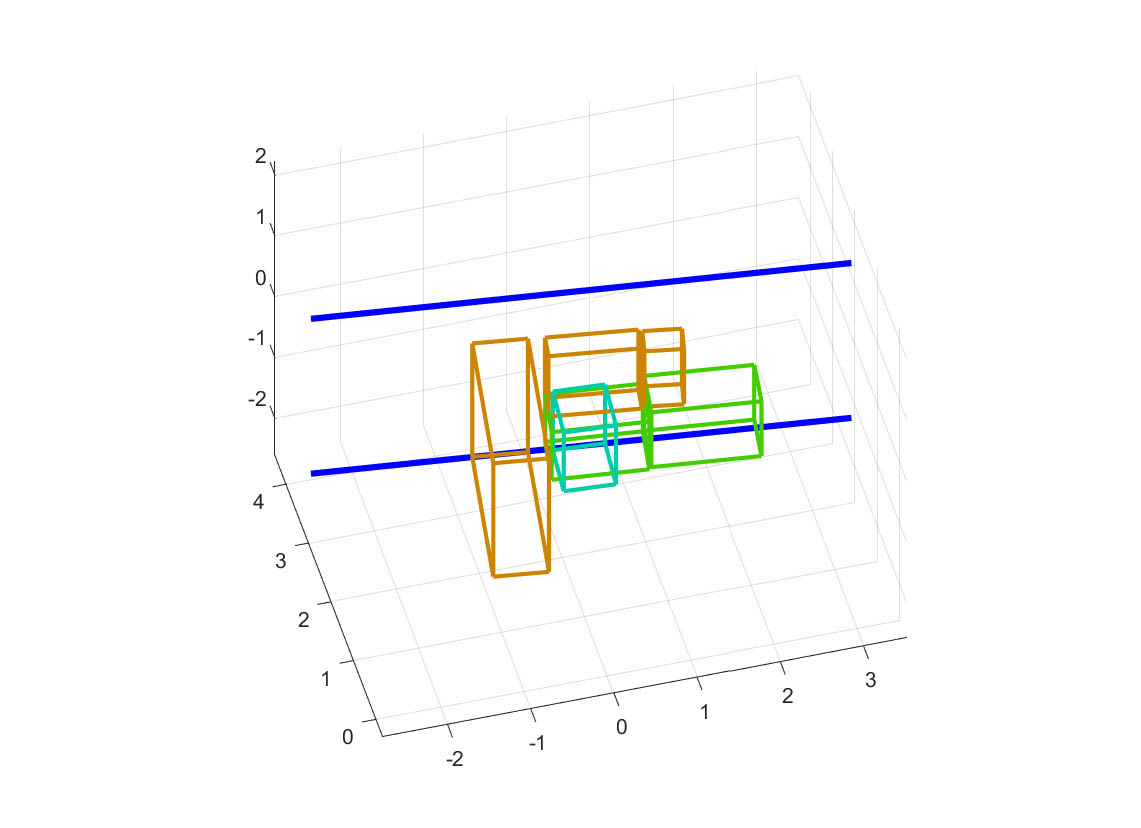} &
        	    \oblique{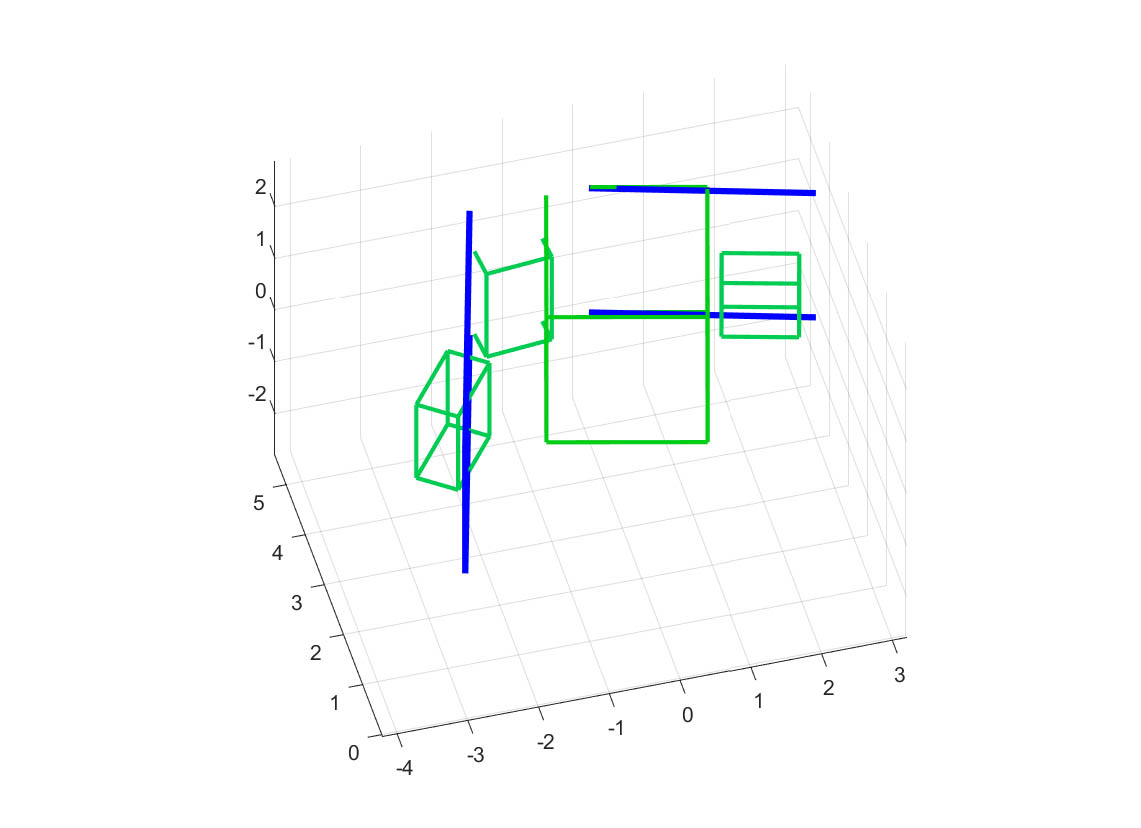} &
        	    \oblique{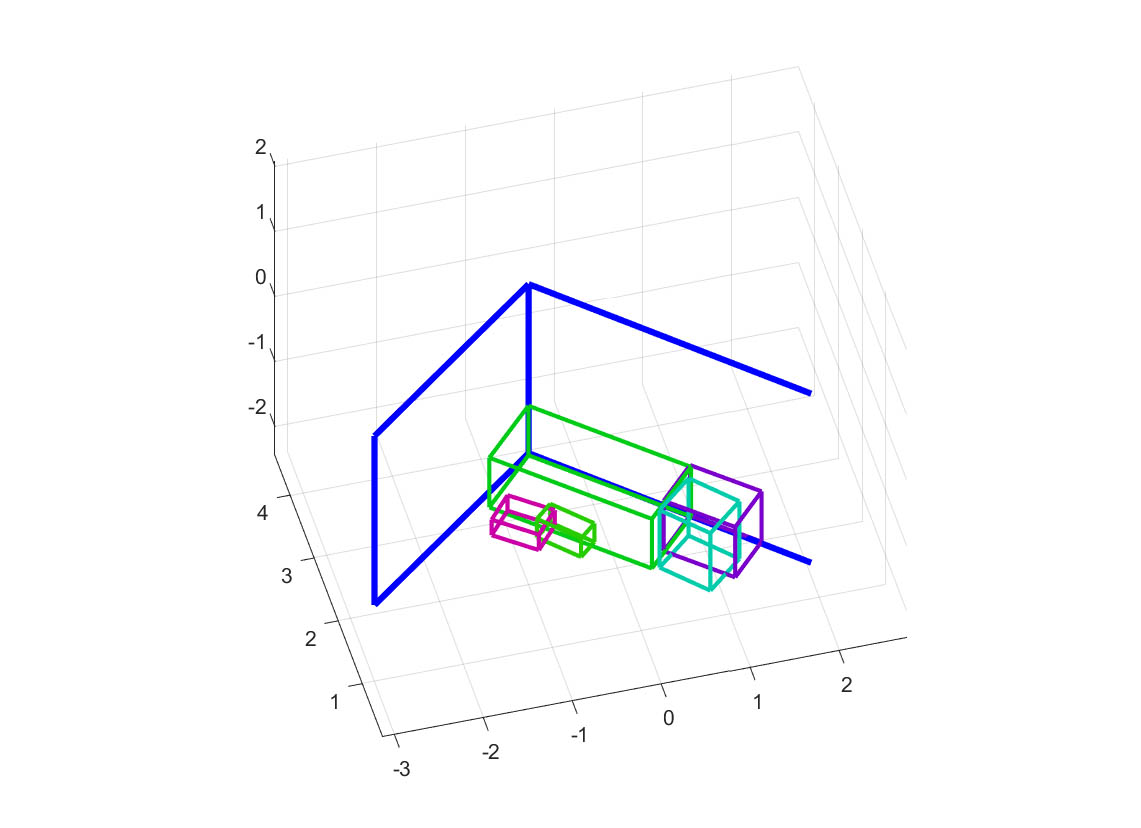} &
        	    \oblique{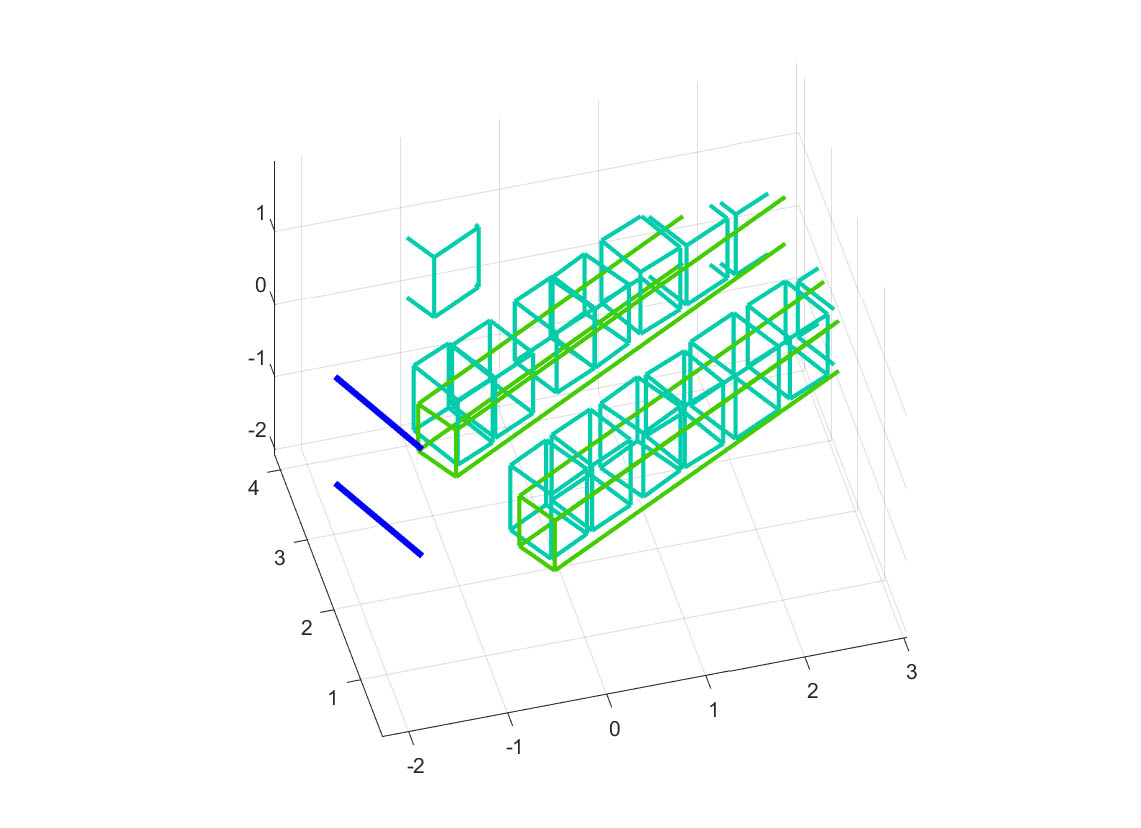} \\
    	\hline
            \multirow{2}{*}{\rot{Camera View}}&
            \rot{Ours}& 
        	    \rgb{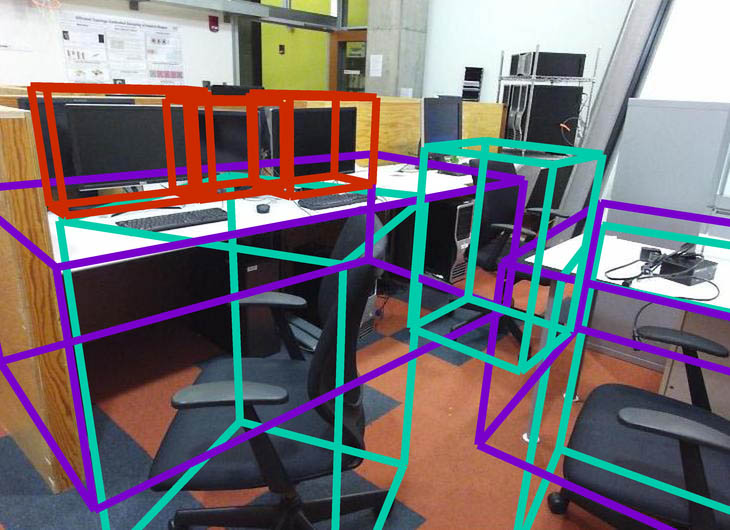} &
        	    \rgb{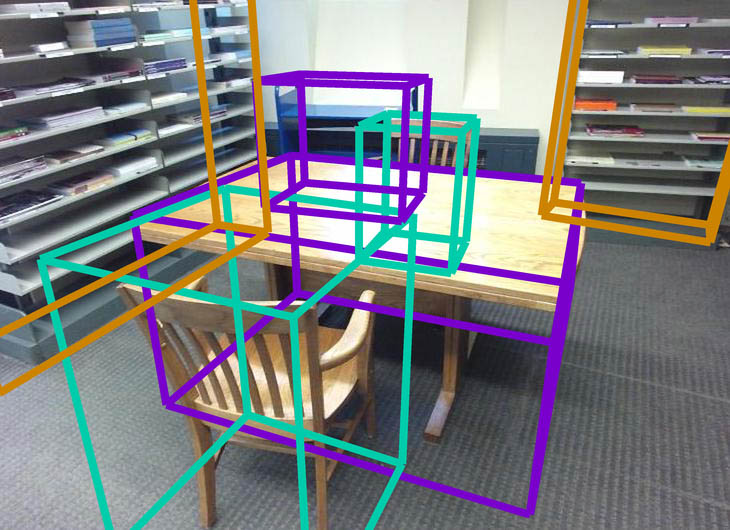} &
        	    \rgb{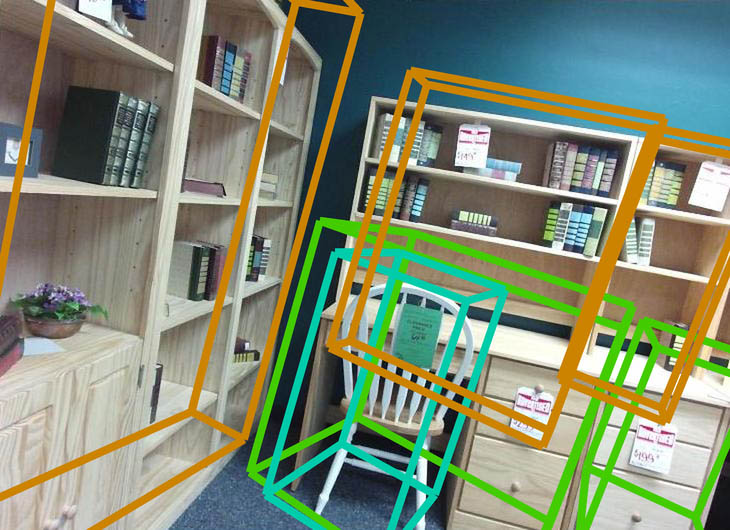} &
        	    \rgb{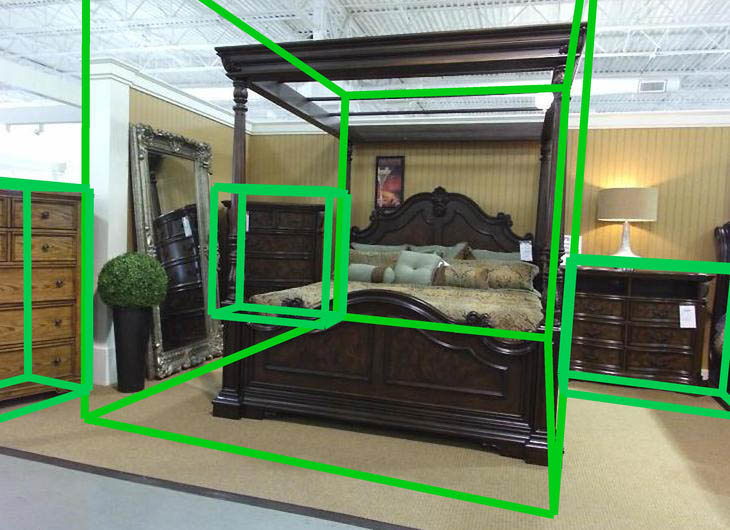} &
        	    \rgb{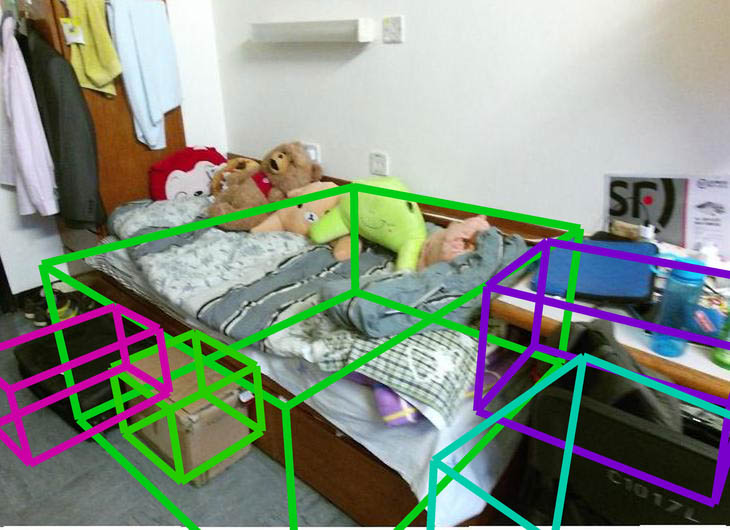} &
        	    \rgb{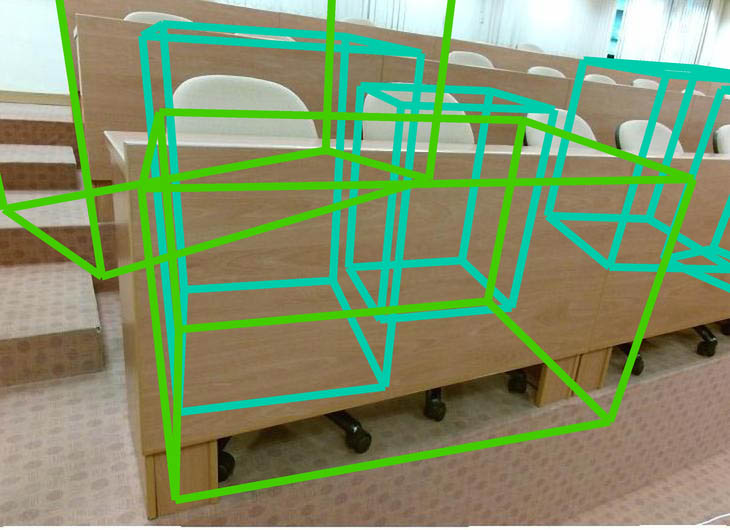} \\
        	&\rot{GT}& 
        	    \rgb{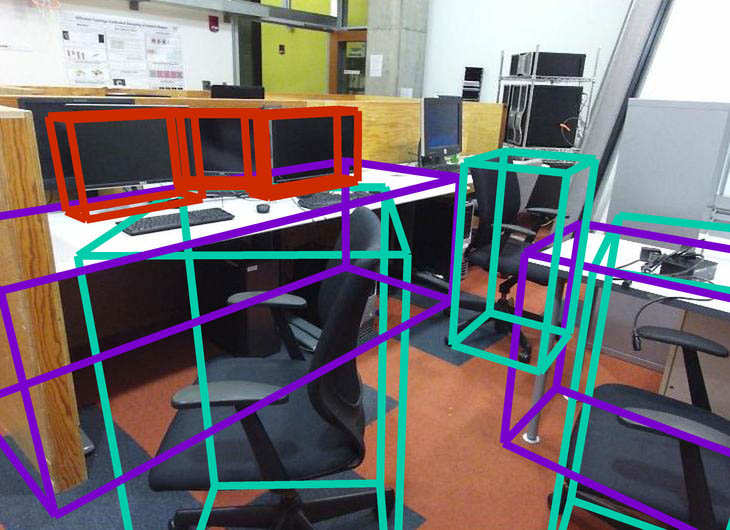} &
        	    \rgb{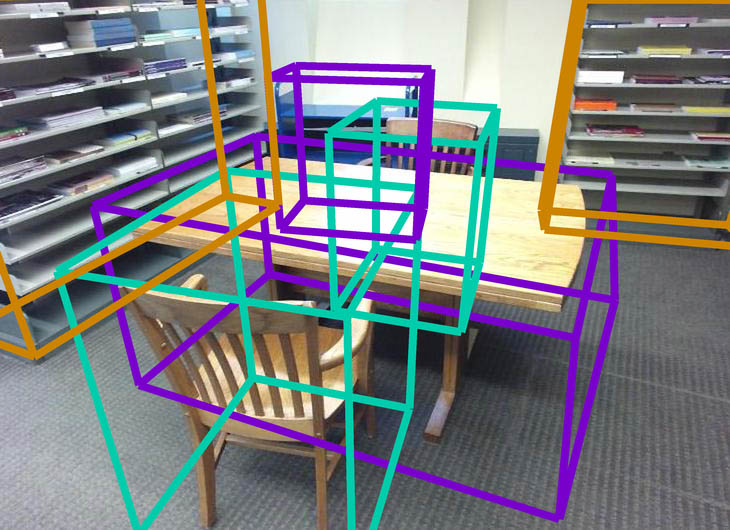} &
        	    \rgb{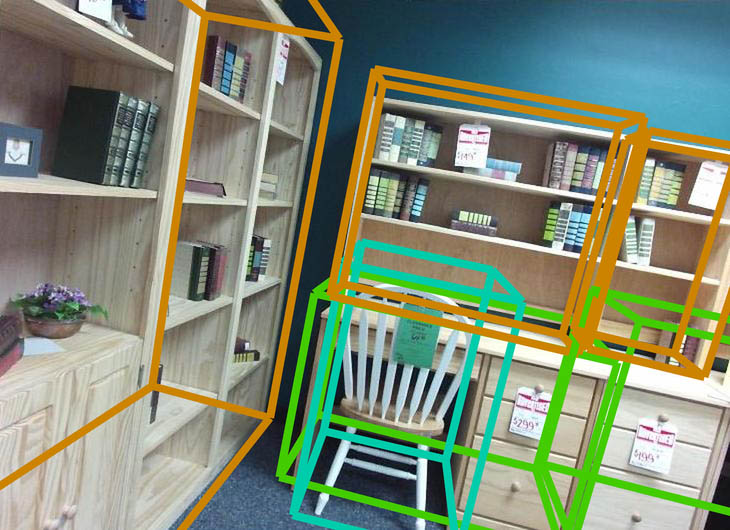} &
        	    \rgb{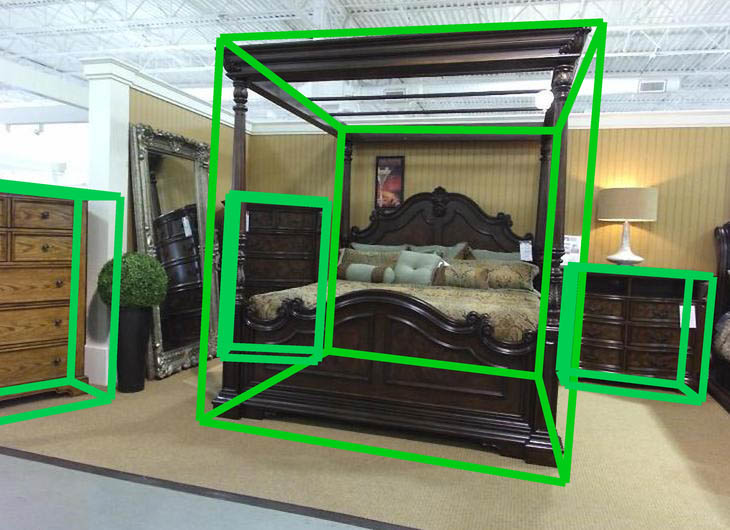} &
        	    \rgb{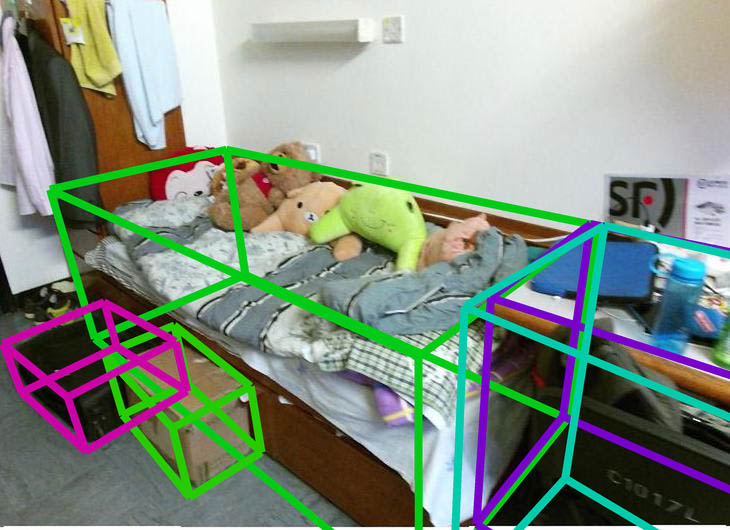} &
        	    \rgb{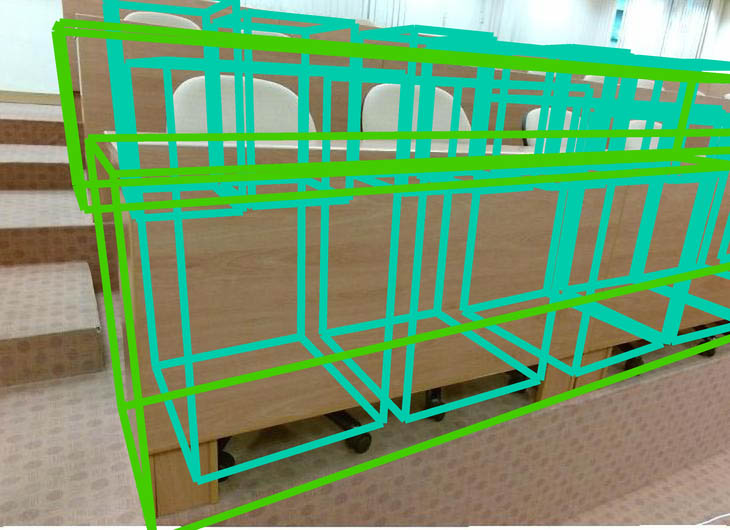} \\
    	\hline
    	    \rot{Scene Mesh}&
    	    \rot{Ours}& 
        	    \rgb{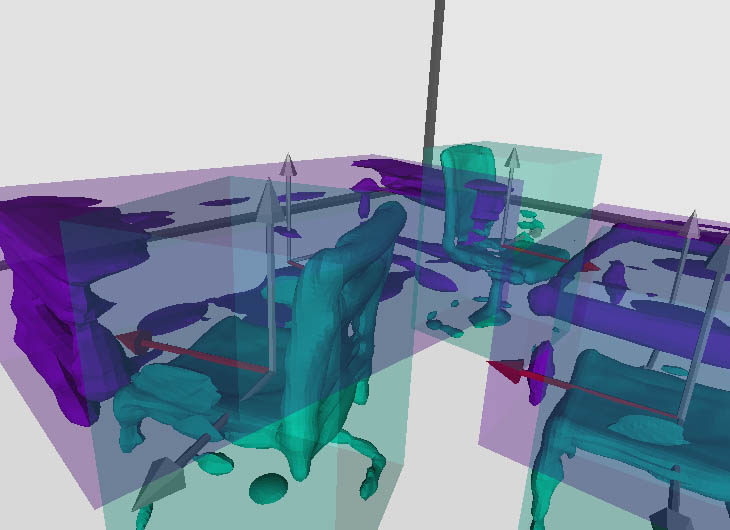} &
        	    \rgb{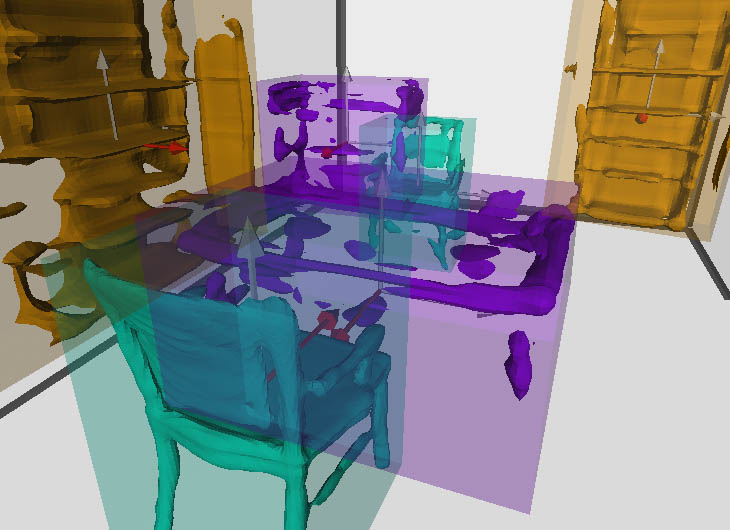} &
        	    \rgb{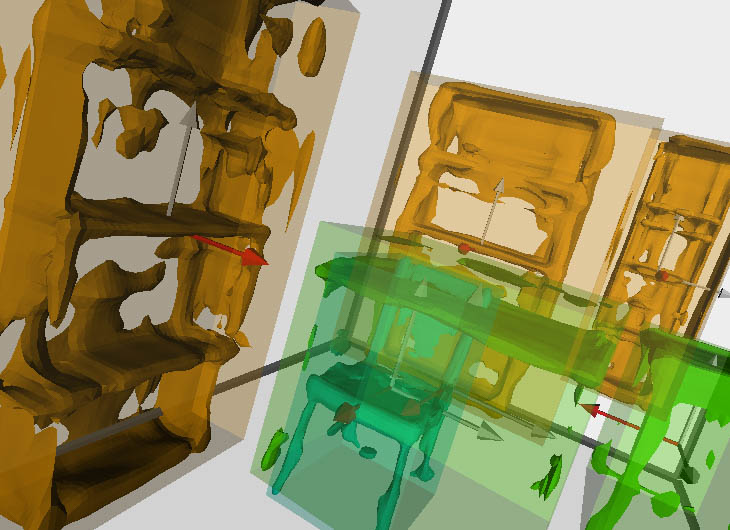} &
        	    \rgb{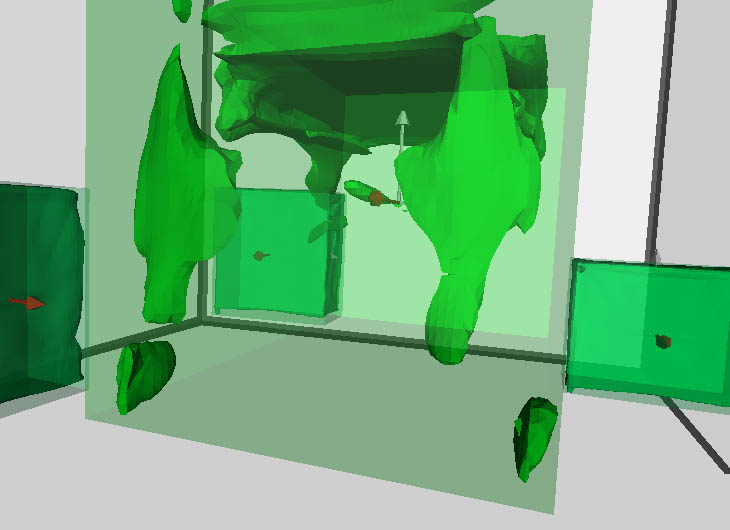} &
        	    \rgb{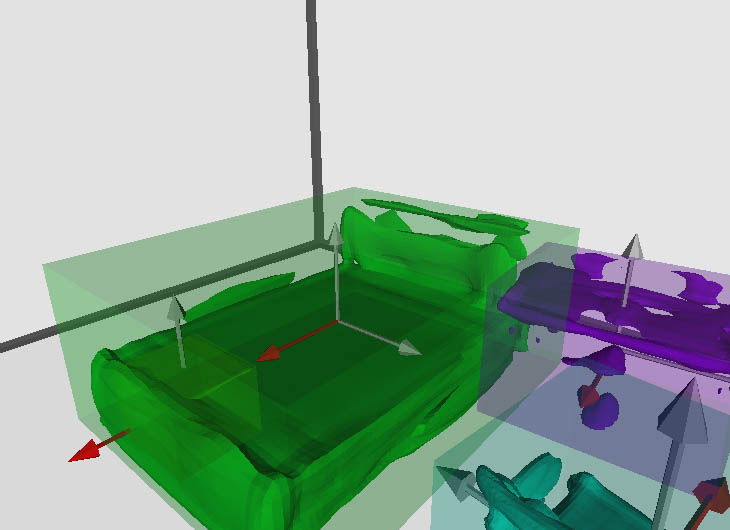} &
        	    \rgb{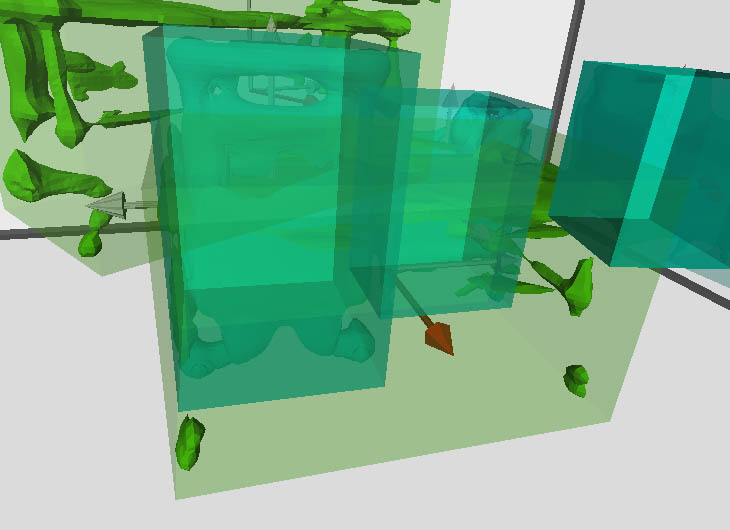} \vspace{2px}\\
        \multicolumn{2}{c}{}& (a) & (b) & (c) & (d) & (e) & (f) \\
    \end{tabular}
    % \vspace{-1.0em}
	\caption{Failure Cases. Possible reasons might be unseen object shapes (a, b, c, d), heavy occlusion (f), cluttered scene (e).}
	\label{fig:scnreconFail}
\end{figure*}

We also show some failure cases in \autoref{fig:scnreconFail}. We observe that although our LIEN performs well on Pix3D and is generalized to SUN RGB-D, it still cannot make plausible reconstruction for some objects in rarely seen shapes (i.e. the desks of (a) and (b), the bookshelves of (b) and (c), the bed of (d)). For object detection, our pipeline fails to correctly estimate the pose of the bed in (e), which might result from the clustered scenes. Also, in some extreme cases, heavy occlusion might cause our pipeline to fail like in (f).

% \newpage

\end{document}